\documentclass{esann}

\usepackage[dvips]{graphicx}
\usepackage[utf8]{inputenc}

\usepackage{microtype}
\usepackage{graphicx}
\usepackage{booktabs} 

\usepackage{wrapfig}

\usepackage{hyperref}

\usepackage{amsmath,amssymb,amsfonts,array}
\usepackage{algorithmic}
\usepackage[noline,ruled,algo2e,linesnumbered]{algorithm2e}
\usepackage{caption}
\usepackage{subcaption}

\SetKwIF{If}{ElseIf}{Else}{if}{then}{else if}{else}{end if}%

\usepackage{todonotes}

\definecolor{blue}{HTML}{1f77b4}
\definecolor{orange}{HTML}{ff7f0e}
\definecolor{green}{HTML}{2ca02c}
\definecolor{red}{HTML}{d62728}
\definecolor{purple}{HTML}{9467bd}
\definecolor{darkblue}{HTML}{000055}
\definecolor{darkred}{HTML}{880000}
\begin{document}
\title{Is One Epoch All You Need For Multi-Fidelity Hyperparameter Optimization?}

\author{Romain Egel\'e$^{1,3}$, Isabelle Guyon$^{1,2}$, Yixuan Sun$^3$, Prasanna Balaprakash$^4$
%
%
\vspace{.3cm}\\
%
1- LISN, U. Paris-Saclay, France;
2- Google, USA \& ChaLearn, USA; \\ 3- MCS, Argonne Ntl Lab, USA; 
4- CCSD, Oak Ridge Ntl Lab, USA\\
}

\maketitle

\begin{abstract}

Hyperparameter optimization (HPO) is crucial for fine-tuning machine learning models, but it can be computationally expensive. To reduce costs, Multi-fidelity HPO (MF-HPO) leverages intermediate accuracy levels in the learning process and discards low-performing models early on.
We conducted a comparison of various representative MF-HPO methods against a simple baseline on classical benchmark data. The baseline involved discarding all models except the Top-$K$ after training for only one epoch, followed by further training to select the best model. Surprisingly, this baseline achieved similar results to its counterparts, while requiring an order of magnitude less computation.
Upon analyzing the learning curves of the benchmark data, we observed a few dominant learning curves, which explained the success of our baseline. This suggests that researchers should (1) always use the suggested baseline in benchmarks and (2) broaden the diversity of MF-HPO benchmarks to include more complex cases.
\end{abstract}

\section{Introduction}

Hyperparameter optimization (HPO) is a key component of AutoML systems. It aims to find the best configuration of a machine learning (ML) pipeline, which consists of data and model components. Hyperparameters (HP) are parameters that cannot be learned directly with the primarly learning algorithm (e.g., gradient descent), but affect the learning process and the performance of the pipeline. Those may include number of layers and units per layer in deep networks, learning rates, etc. HPO is usually formulated as a black-box optimization problem\cite{yu_hyper-parameter_2020}, where a function maps a HP configuration to a performance score. However, evaluating this function can be costly and time-consuming. Therefore, Multi-fidelity HPO (MF-HPO) methods have been proposed~\cite{jamieson2016non,li2017hyperband}, which use intermediate learning machine performance estimates to reduce the computational cost and speed up the optimization process.
In multi-fidelity algorithms, fidelity refers to the level of accuracy or resolution at which a machine learning model is trained and evaluated. Different levels of fidelity can be used to balance the trade-off between computational cost and accuracy. For example, a low fidelity level could involve training and evaluating a model on a small subset of the data, while a high fidelity level would involve using the entire dataset. Multi-fidelity algorithms use this concept to optimize hyperparameters and other model parameters efficiently by exploring the model space using a combination of different fidelity levels. The accuracy obtained at each fidelity level is used to guide the optimization process towards better performing models while minimizing the computational cost.

\begin{wrapfigure}{R}{0.55\textwidth}
    \centering
    \includegraphics[width=0.54\textwidth]{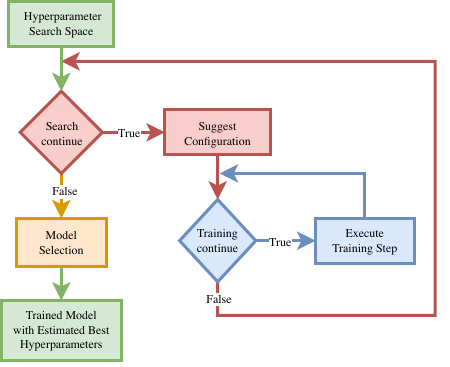}
    \caption{\small Hyperparameter optimization and its components including {\color{green}input/output}, {\color{red} outer optimization loop exploring new HP configurations}, {\color{blue}inner optimization loop incrementally allocating new training ``steps"} and {\color{orange} model selection}.}
    \label{fig:components-hpo}
    \vspace{-0.1in}
\end{wrapfigure}

Fig.~\ref{fig:components-hpo} presents the generic components of a muti-fidelity HPO. The input to HPO is the HP search space. The output is a trained model with the corresponding hyperparameters. HPO is a bi-level optimization problem: the upper level ({\color{red}outer loop}) optimizes the HP of the ML pipeline while the lower level ({\color{blue}inner loop}) optimizes the parameters of the learning machine, given the HP configuration of the upper level. After completing the search process, a final step of {\color{orange}model selection} is performed before returning the trained model.
In regular HPO, the inner loop is halted by a given stopping criterion. A natural improvement is to consider the multi-fidelity setting, where intermediate performance estimates are obtained at different stages of the inner optimization process (as a function of the number of training examples or training epochs). This involves monitoring a "learning curve."

This study evaluates the effectiveness of early discarding policies in the {\color{blue}inner loop}, while random search is used for the {\color{red}outer loop}. We use widely-used approaches for early discarding, performing denovo search without prior knowledge of learning curve typology or meta-learning involvement.

We consider two baselines at each extreme. The first evaluates each HP configuration candidate at maximum fidelity, i.e., after being fully trained within the overall time budget (100-Epochs). The second evaluates them at minimum fidelity, after only one epoch of training. We demonstrate the effectiveness of the 1-Epoch approach and explain why it is successful. Our findings provide a new perspective on previous studies that omitted this baseline. Only one study~\cite{bohdal2023pasha} included a similar baseline and reported results similar to ours. Therefore, we advocate for including the 1-Epoch baseline in further MF-HPO benchmarks.


\section{Related Work}~\label{sec:related-work}


Our study focuses on methods, which train only a single model at a time, but keep all check-points for further reference. Early discarding means switching to training a model with another HP configuration, before attaining the maximum number of epochs allowed. A typical example of such strategy, sometimes referred to as ``vertical'' model selection~\cite{mohr_learning_2022} is Asynchronous Successive Halving~\cite{li_system_2020} (SHA). Several methods can be adapted to this setting, including Hyperband~\cite{li2017hyperband}, which can explore different levels of fidelity to differentiate noisy learning curves; 
Learning Curve Extrapolation~\cite{domhan_speeding_2015} (LCE), which can observe early performances and extrapolate future performances to decide whether training should continue; FABOLAS~\cite{klein2017fast}, which learns correlations in the candidates' ranking between different levels of fidelity; Bayesian Optimization Hyperband~\cite{falkner_bohb_2018}, which embeds Bayesian optimization in Hyperband to sample candidates more efficiently; Learning Curve with Support Vector Regression~\cite{baker_accelerating_2017}, which predicts final performance based on the configuration and early observations; Learning Curve with Bayesian neural network~\cite{klein_learning_2017}, which is similar to the previous method but switches the model with a Bayesian neural network; and Trace Aware Knowledge-Gradient~\cite{wu2020practical}, which leverages an observed curve to update the posterior distribution of a Gaussian process more efficiently.




The previous works had some limitations, which were primarily due to the assumptions made about the learning curve. For example, methods based on SHA assume that the discarded learning curves will not cross in the future, since only the Top-$K$ models are allowed to continue at any given step. In this context, models that start slowly are often discarded.
This phenomenon is known as ``short-horizon bias''~\cite{wu2018understanding}. For methods based on LCE, they either assume a "biased" parametric model, or they are more generic but require a more expensive initialization\cite{white_how_2021}. The second limitation is that the configuration returned is often the lowest validation error observed during the search. However, this approach is quite limiting because it does not consider more sophisticated model selection schemes such as cross-validation or Hyperband. Therefore, we added the {\color{orange} model selection} block in Fig.~\ref{fig:components-hpo}.

Benchmarks play a critical role in the design and development of HPO methods. We have surveyed several recent benchmarks for learning curves that are continuously evolving, such as HPOBench~\cite{eggensperger_hpobench_2021,klein_tabular_2019}, LCBench~\cite{zimmer2021auto}, JAHS-Bench-201~\cite{bansal2022jahsbench}, and YAHPO-Gym~\cite{pfisterer2022yahpo}.

\section{Method}

Our experiments include representative state-of-the-art methods in multi-fidelity HPO research.
The first method considered is {\bf Successive Halving} (SHA)~\cite{jamieson2016non} which we consider in its asynchronous variant~\cite{li_system_2020} to be able to handle a stream of learners. SHA decides to continue the training loop at exponentially increasing steps called rungs (based on learning curves~\cite{hutter_learning_2021} theory). At each rung, it continues training only if the current observed score is among the Top-$K$ scores and discards the model otherwise. This method is error-prone when learning curves are noisy. To make SHA more robust {\bf Hyperband} (HB)~\cite{li2017hyperband} enforces the exploration of more levels of fidelity on top of SHA to reduce the impact of noise in decisions. However this is at the expense of consuming more training steps than SHA. Finally, the last approach we consider consists in using a surrogate model to predict the future of the learning curve, called {\bf Learning Curve Extrapolation} (LCE)~\cite{domhan_speeding_2015}. In theory, this approach is less likely to suffer from ``short-horizon'' bias unless the learning curve surrogate is biased. 
The method also suffers from instabilities in its original implementation \cite{domhan_speeding_2015}. 
To address this shortcoming, we are using our own implementation of this method (See the RoBER method described in supplemental material$^1$).

We compare these methods with two baselines: the maximum fidelity baseline ({\bf 100-Epochs}), which trains all methods for 100 Epochs and selects the best one and a mimimal fidelity baseline ({\bf 1-Epoch}), which comprises in fact two phases: (i) exploration search phase pre-selecting the Top-$K$ after only 1 Epoch (minimum fidelity blue loop in Fig.~\ref{fig:components-hpo}); (ii) model selection phase evaluating the Top-$K$ at maximum fidelity and returing the best model. 1-Epoch is expected to perform poorly if there is a strong ``short-horizon'' bias (see previous secton for a definiton) and noisy learning curves.

\section{Experimental Results}

\begin{wrapfigure}{R}{0.45\textwidth}
    \vspace{-0.2in}
    \centering
    \includegraphics[width=0.45\textwidth]{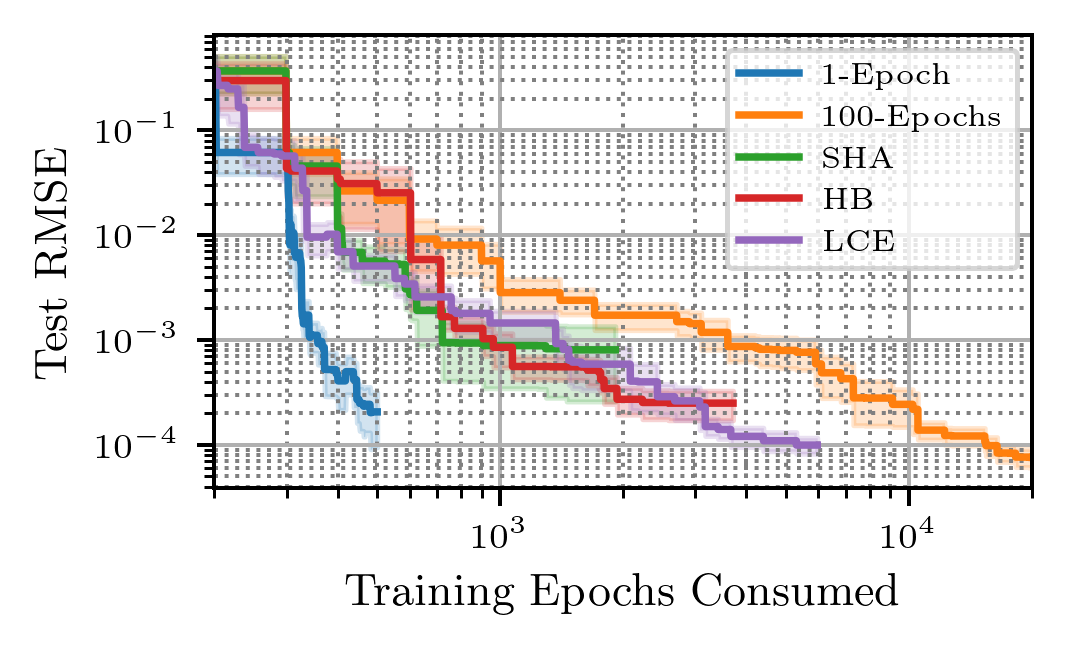}
    \vspace{-0.15in}
    \caption{\small Comparing the performance (mean and standard error) of various early discarding strategies (all combined with with random search) on the Naval Propulsion problem from the HPOBench benchmark. 
    }
    \label{fig:model-based-pruning-with-random-search-on-hpobench}
    \vspace{-0.15in}
\end{wrapfigure}

In our analysis we looked at four benchmarks: HPOBench, LCBench, YAHPO-Gym and JAHS-Bench. But, since they all yielded similar results, for brevity we only expose the Naval Propulsion problem from HPOBench. We provide complementary materials \footnote{\url{https://github.com/deephyper/scalable-bo/blob/main/esann-23/One_Epoch_Is_Often_All_You_Need_Extended.pdf}} with our full set of results.

To visualize the algorithm progress, we display the test error as a function of training epochs \cite{pfisterer2022yahpo} (Fig.\ref{fig:components-hpo}). This type of learning curve weighs training iterations (epochs) equally for all HP configurations, which may be deceiving since they can vary in computational cost. Still, this is a convenient simple method abstrating from details of implementation.
 We configured the number of iterations for the outer-loop optimization to 200 (red loop).
As a result, the maximum fidelity policy (100-Epochs),  consumes the maximum number of 20,000 training epochs. In contrast, its counterpart at minimum fidelity (1-Epoch, performed with a Top-3 model selection ($K=3$) at maximum fidelity after the search), consumes a fixed number of $200 \times 1 + 3 \times 100 = 500$ training epochs. 

Figure~\ref{fig:model-based-pruning-with-random-search-on-hpobench} shows the evolution of the test RMSE as a function of the number of training epochs used. The key observation is that the test RMSE for the 1-Epoch (blue curve) and 100-Epochs (orange curve) strategies is similar at the final point. However, the 1-Epoch policy uses 40 times fewer training epochs than its counterpart, the 100-Epoch policy (20,000/500=40), as per our experimental setup, which is evident from the last point of the blue and orange curves. Additionally, we note that the more complex agents (SHA, HB, and LCE) do not differ significantly from each other and all result in similar final test RMSE, but they consume significantly more epochs. Hyperband consumes slightly more resources than SHA, which is consistent with its design. SHA consumes the least training epochs, about 10 times fewer epochs than 100-Epochs, but still, four times more than 1-Epoch.

\begin{figure}
    \centering
    \begin{subfigure}[b]{0.45\columnwidth}
        \centering
        \includegraphics[width=\textwidth]{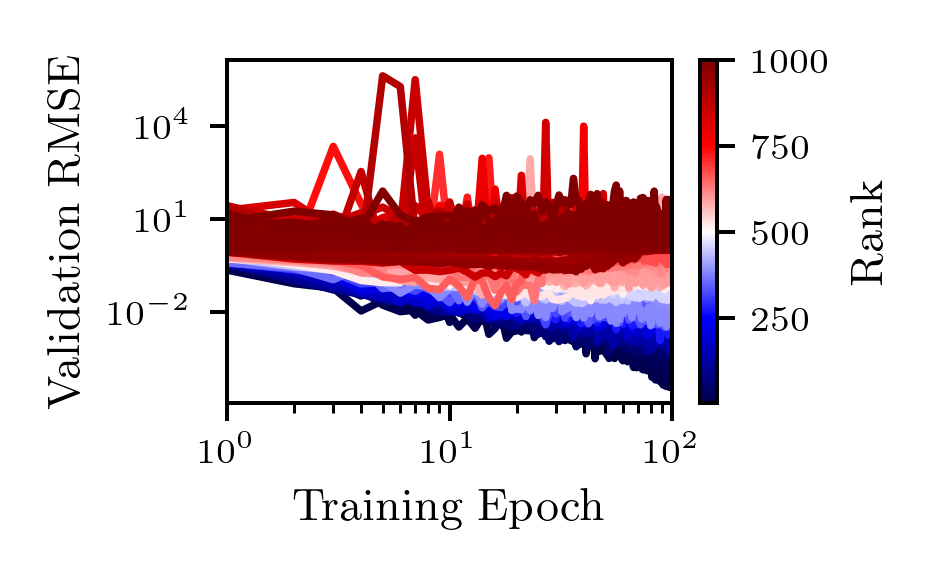}
        \vspace{-0.25in}
        \caption{Learning curves colored by final rank.}
        \label{fig:viz-learning-curves-by-ranking-navalpropulsion}
    \end{subfigure}
    \begin{subfigure}[b]{0.45\columnwidth}
        \centering
        \includegraphics[width=\textwidth]{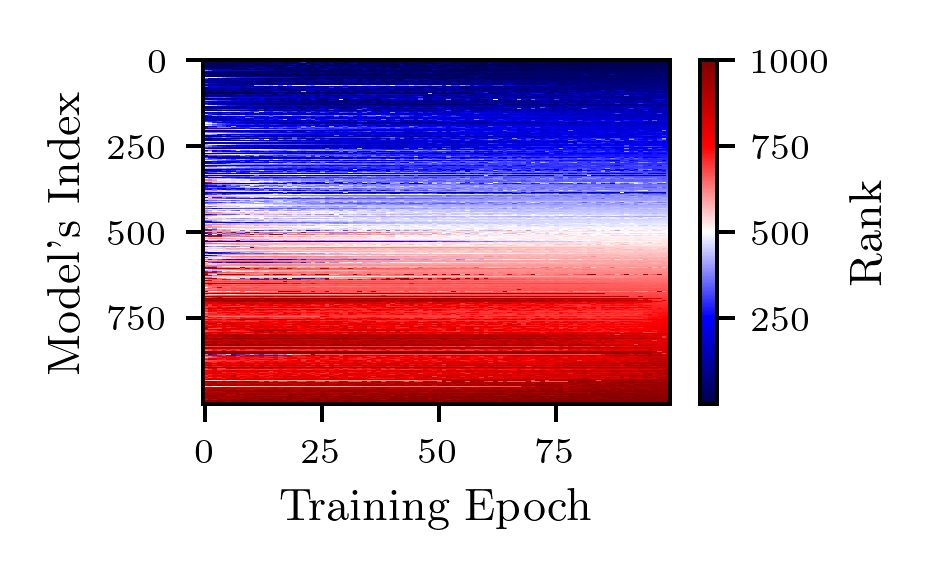}
        \vspace{-0.25in}
        \caption{Evolution of the ranking with epochs.}
        \label{fig:viz-heatmap-raking-navalpropulsion}
    \end{subfigure}
    \vspace{-0.1in}
    \caption{\small Visualizing the ranking for {\color{darkblue}good (blue)} and {\color{darkred}bad (red)} models for 1,000 randomly sampled learning curves from the Naval Propulsion benchmark. (a) Models can be selected from the first epoch. (b) The ranking stabilizes faster among the best models.}
    \label{fig:viz-ranking-navalpropulsion}
    \vspace{-0.2in}
\end{figure}

After observing the performance of training for only one epoch during the search, we will explain why. In Fig.~\ref{fig:viz-learning-curves-by-ranking-navalpropulsion}, we display 1,000 randomly sampled learning curves from the same benchmark, colored by their ranking at maximum fidelity. Low ranks, colored in blue, correspond to good models, while high ranks, colored in red, correspond to bad models. This visualization reveals that the groups of bad and good models can be identified in the first epoch of training, making selection with one epoch effective. Some noise exists between models, as seen in the blue curves, necessitating the top-$K$ tournament selection to differentiate among different models. The noise in the ranking for the same benchmark can be visualized in Fig.\ref{fig:viz-heatmap-raking-navalpropulsion}, where the ranking for the same models at each training epoch is displayed. The first half (before 50 epochs) has more noise than the second half, and the ranking of good models stabilizes faster, already after 5 epochs, compared to the ranking of bad models, which remains relatively noisy until the end.

\section{Conclusion}
\label{sec:conclusion}

In this study, we investigated the optimization of hyperparameters for machine learning models with minimal training steps using MF-HPO agents. We compared various state-of-the-art multi-fidelity policies and discovered that a new simple baseline method, not considered in prior becnhmarks, namely the 1-Epoch strategy assessing candidate configuration at lowest fidelity, performs surprisingly well. This was traced to the fact that good and bad models can be distinguished early in the training process. This phenomenon occurs frequently in benchmarks commonly used in the literature. Therefore, this point to the need to include 1-Epoch in forthcoming benchmarks and eventually designing new harder (possibly more realistic) benchmarks that will defeat it. Our work is limitated to using ``epoch" as a unit of fidelity. While this is convenient and appealing to conduct studies independent of hardware implementation considerations, practical application settings may require considering wall time or other options as units of fidelity.

\section{Acknowledgment}
This material is based upon work supported by the U.S.\ Department of Energy 
(DOE), Office of Science, Office of Advanced Scientific Computing Research, under
Contract DE-AC02-06CH11357. This research used resources of the Argonne 
Leadership Computing Facility, which is a DOE Office of Science User Facility. 
This material is based upon work supported by ANR Chair of Artificial Intelligence HUMANIA ANR-19-CHIA-0022 and TAILOR EU Horizon 2020 grant 952215.


\begin{footnotesize}


\bibliographystyle{unsrt}
\bibliography{esann-short}

\begin{thebibliography}{10}

\bibitem{yu_hyper-parameter_2020}
T.~Yu et~al.
\newblock Hyper-parameter optimization: A review of algorithms and
  applications.
\newblock {\em arXiv preprint arXiv:2003.05689}, 2020.

\bibitem{jamieson2016non}
K.~Jamieson et~al.
\newblock Non-stochastic best arm identification and hyperparameter
  optimization.
\newblock In {\em Artificial intelligence and statistics}, pages 240--248.
  PMLR, 2016.

\bibitem{li2017hyperband}
L.~Li et~al.
\newblock Hyperband: A novel bandit-based approach to hyperparameter
  optimization.
\newblock {\em The Journal of Machine Learning Research}, 18(1):6765--6816,
  2017.

\bibitem{bohdal2023pasha}
O.~Bohdal et~al.
\newblock {PASHA}: Efficient {HPO} and {NAS} with progressive resource
  allocation.
\newblock In {\em The Eleventh International Conference on Learning
  Representations}, 2023.

\bibitem{mohr_learning_2022}
F.~Mohr et~al.
\newblock Learning curves for decision making in supervised machine learning--a
  survey.
\newblock {\em arXiv preprint arXiv:2201.12150}, 2022.

\bibitem{li_system_2020}
L.~Li et~al.
\newblock A system for massively parallel hyperparameter tuning.
\newblock {\em Proceedings of Machine Learning and Systems}, 2:230--246, 2020.

\bibitem{domhan_speeding_2015}
T.~Domhan et~al.
\newblock Speeding up automatic hyperparameter optimization of deep neural
  networks by extrapolation of learning curves.
\newblock In {\em Twenty-fourth international joint conference on artificial
  intelligence}, 2015.

\bibitem{klein2017fast}
A.~Klein et~al.
\newblock Fast bayesian optimization of machine learning hyperparameters on
  large datasets.
\newblock In {\em Artificial intelligence and statistics}, pages 528--536.
  PMLR, 2017.

\bibitem{falkner_bohb_2018}
S.~Falkner et~al.
\newblock Bohb: Robust and efficient hyperparameter optimization at scale.
\newblock In {\em International Conference on Machine Learning}, pages
  1437--1446. PMLR, 2018.

\bibitem{baker_accelerating_2017}
B.~Baker et~al.
\newblock Accelerating neural architecture search using performance prediction.
\newblock {\em arXiv preprint arXiv:1705.10823}, 2017.

\bibitem{klein_learning_2017}
A.~Klein et~al.
\newblock Learning curve prediction with bayesian neural networks.
\newblock In {\em International Conference on Learning Representations}, 2017.

\bibitem{wu2020practical}
J.~Wu et~al.
\newblock Practical multi-fidelity bayesian optimization for hyperparameter
  tuning.
\newblock In {\em Uncertainty in Artificial Intelligence}, pages 788--798.
  PMLR, 2020.

\bibitem{wu2018understanding}
Y.~Wu et~al.
\newblock Understanding short-horizon bias in stochastic meta-optimization.
\newblock In {\em International Conference on Learning Representations}, 2018.

\bibitem{white_how_2021}
C.~White et~al.
\newblock How powerful are performance predictors in neural architecture
  search?
\newblock {\em Advances in Neural Information Processing Systems},
  34:28454--28469, 2021.

\bibitem{eggensperger_hpobench_2021}
K.~Eggensperger et~al.
\newblock Hpobench: A collection of reproducible multi-fidelity benchmark
  problems for hpo.
\newblock {\em arXiv preprint arXiv:2109.06716}, 2021.

\bibitem{klein_tabular_2019}
A.~Klein et~al.
\newblock Tabular benchmarks for joint architecture and hyperparameter
  optimization.
\newblock {\em arXiv preprint arXiv:1905.04970}, 2019.

\bibitem{zimmer2021auto}
L.~Zimmer et~al.
\newblock Auto-pytorch: Multi-fidelity metalearning for efficient and robust
  autodl.
\newblock {\em IEEE Transactions on Pattern Analysis and Machine Intelligence},
  43(9):3079--3090, 2021.

\bibitem{bansal2022jahsbench}
A.~Bansal et~al.
\newblock Jahs-bench-201: A foundation for research on joint architecture and
  hyperparameter search.
\newblock In {\em Thirty-sixth Conference on Neural Information Processing
  Systems Datasets and Benchmarks Track}, 2022.

\bibitem{pfisterer2022yahpo}
F.~Pfisterer et~al.
\newblock Yahpo gym-an efficient multi-objective multi-fidelity benchmark for
  hyperparameter optimization.
\newblock In {\em International Conference on Automated Machine Learning},
  pages 3--1. PMLR, 2022.

\bibitem{hutter_learning_2021}
M.~Hutter.
\newblock Learning curve theory.
\newblock {\em arXiv preprint arXiv:2102.04074}, 2021.

\bibitem{10.5555/645940.671380}
B.~Gu et~al.
\newblock Modelling classification performance for large data sets.
\newblock In {\em Proceedings of the Second International Conference on
  Advances in Web-Age Information Management}, WAIM '01, page 317–328,
  Berlin, Heidelberg, 2001. Springer-Verlag.

\bibitem{mohr2023lcdb}
F.~Mohr et~al.
\newblock Lcdb 1.0: An extensive learning curves database for classification
  tasks.
\newblock In {\em Machine Learning and Knowledge Discovery in Databases:
  European Conference, ECML PKDD 2022, Grenoble, France, September 19--23,
  2022, Proceedings, Part V}, pages 3--19. Springer, 2023.

\bibitem{gavin2019levenberg}
HP. Gavin.
\newblock The levenberg-marquardt algorithm for nonlinear least squares
  curve-fitting problems.
\newblock {\em Department of Civil and Environmental Engineering, Duke
  University}, 19, 2019.

\bibitem{jax2018github}
J.~Bradbury et~al.
\newblock {JAX}: composable transformations of {P}ython+{N}um{P}y programs,
  2018.

\bibitem{bergstra_algorithms_2011}
J.~Bergstra et~al.
\newblock Algorithms for hyper-parameter optimization.
\newblock {\em Advances in neural information processing systems}, 24, 2011.

\bibitem{bergstra_making_2013}
J.~Bergstra et~al.
\newblock Making a science of model search: Hyperparameter optimization in
  hundreds of dimensions for vision architectures.
\newblock In {\em International conference on machine learning}, pages
  115--123. PMLR, 2013.

\bibitem{hutter2014algorithm}
F.~Hutter et~al.
\newblock Algorithm runtime prediction: Methods \& evaluation.
\newblock {\em Artificial Intelligence}, 206:79--111, 2014.

\bibitem{scikit-learn}
F.~Pedregosa et~al.
\newblock Scikit-learn: Machine learning in {P}ython.
\newblock {\em Journal of Machine Learning Research}, 12:2825--2830, 2011.

\end{thebibliography}

\end{footnotesize}


\appendix
\section{Visualization of Learning Curves from HPOBench}~\label{sec:appendix-viz-learning-curves}

\begin{figure}[b]
    \centering
    \begin{subfigure}[b]{0.45\textwidth}
        \centering
        \includegraphics[width=\textwidth]{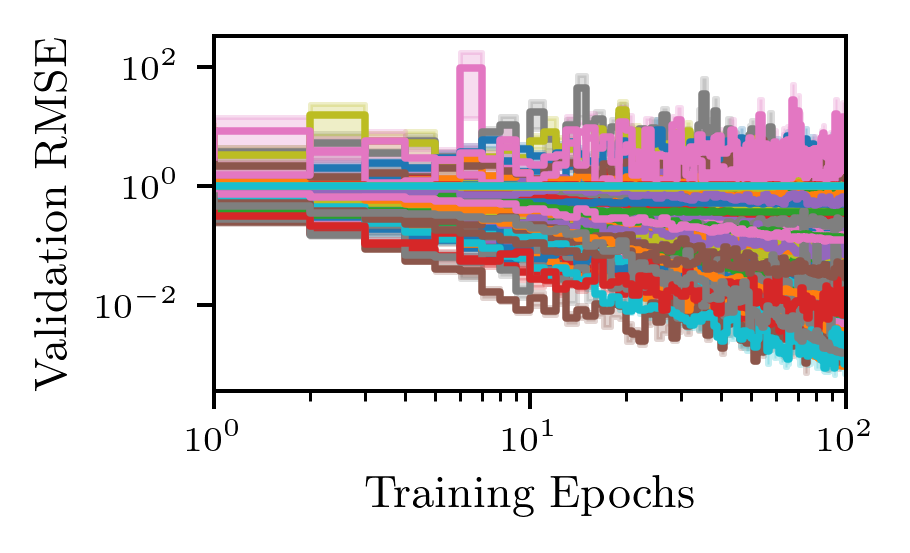}
        \caption{Naval Propulsion}
        \label{fig:viz-learning-curves-hpobench-log-scales-navalpropulsion}
    \end{subfigure}
    \begin{subfigure}[b]{0.45\textwidth}
        \centering
        \includegraphics[width=\textwidth]{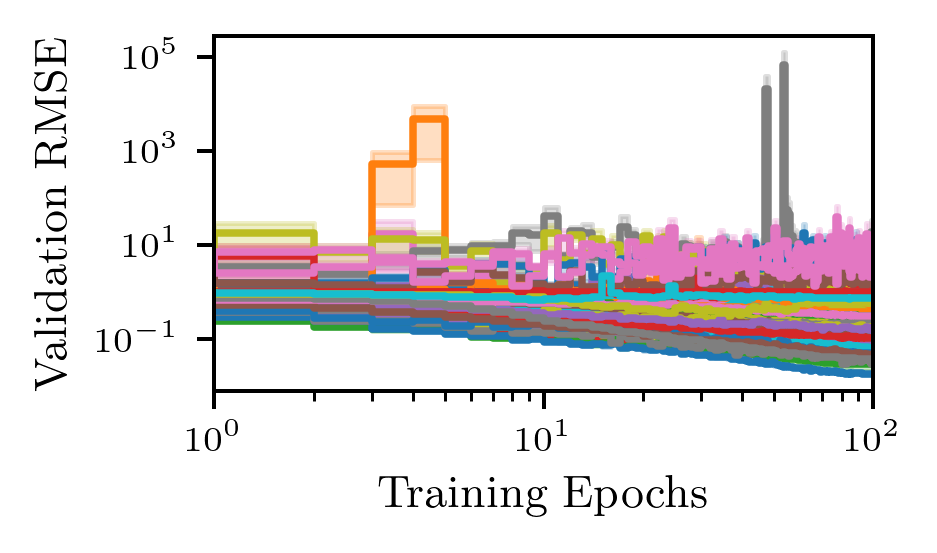}
        \caption{Parkinsons Telemonitoring}
        \label{fig:viz-learning-curves-hpobench-log-scales-parkinsonstelemonitoring}
    \end{subfigure}
    \begin{subfigure}[b]{0.45\textwidth}
        \centering
        \includegraphics[width=\textwidth]{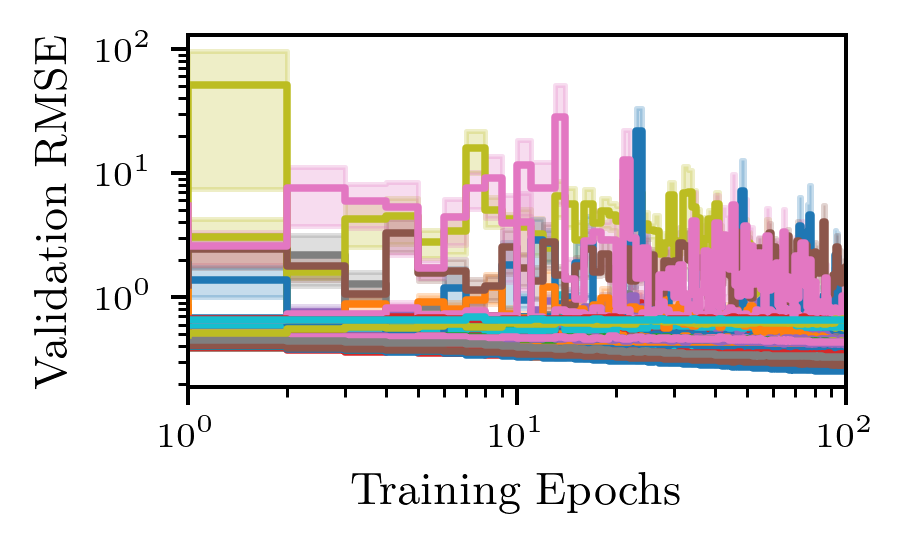}
        \caption{Protein Structure}
        \label{fig:viz-learning-curves-hpobench-log-scales-proteinstructure}
    \end{subfigure}
    \begin{subfigure}[b]{0.45\textwidth}
        \centering
        \includegraphics[width=\textwidth]{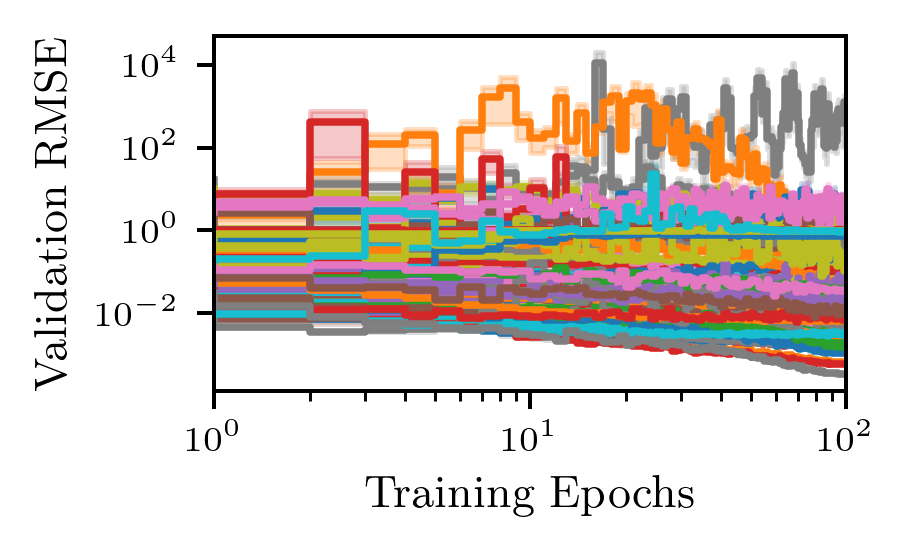}
        \caption{Slice Localization}
        \label{fig:viz-learning-curves-hpobench-log-scales-slicelocalization}
    \end{subfigure}
    \caption{A set of 200 randomly sampled curves from the different benchmarks in HPOBench. The mean (plain-line) and standard error of the validation objective (RMSE) for each curve are displayed as different random-seeds are provided in HPOBench.}
    \label{fig:viz-learning-curves-hpobench-log-scales}
\end{figure}

\begin{figure}
    \centering
    \begin{subfigure}[b]{0.45\textwidth}
        \centering
        \includegraphics[width=\textwidth]{figures/hpobench/navalpropulsion/ranking_curves_naval_propulsion.png}
        \caption{Naval Propulsion}
        \label{fig:ranking-curves-top10-navalpropulsion}
    \end{subfigure}
    \begin{subfigure}[b]{0.45\textwidth}
        \centering
        \includegraphics[width=\textwidth]{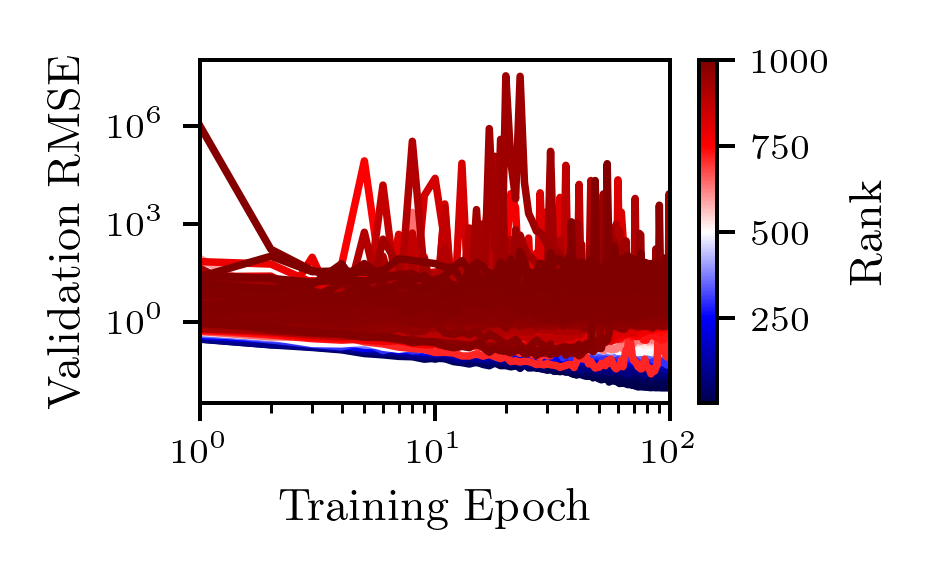}
        \caption{Parkinsons Telemonitoring}
        \label{fig:ranking-curves-top10-parkinsonstelemonitoring}
    \end{subfigure}
    \begin{subfigure}[b]{0.45\textwidth}
        \centering
        \includegraphics[width=\textwidth]{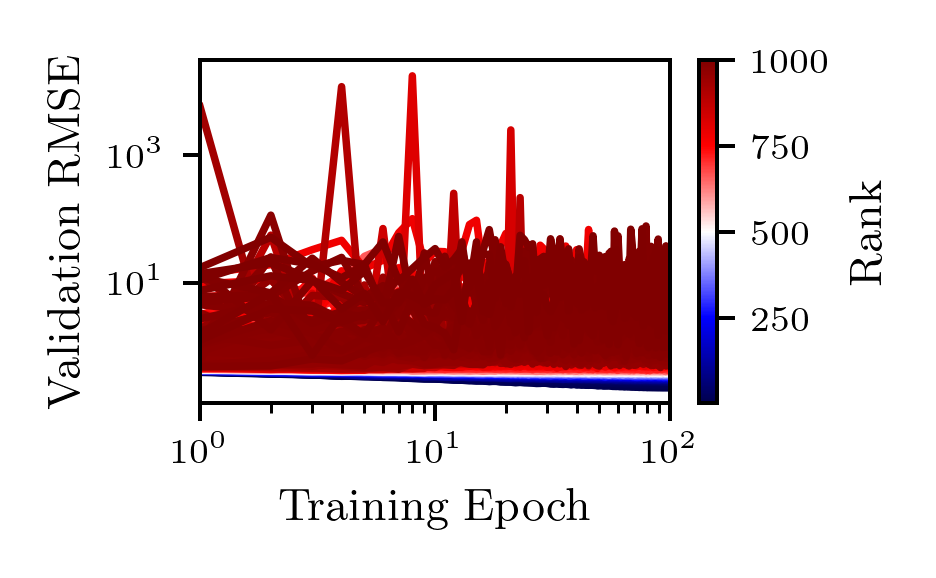}
        \caption{Protein Structure}
        \label{fig:ranking-curves-top10-proteinstructure}
    \end{subfigure}
    \begin{subfigure}[b]{0.45\textwidth}
        \centering
        \includegraphics[width=\textwidth]{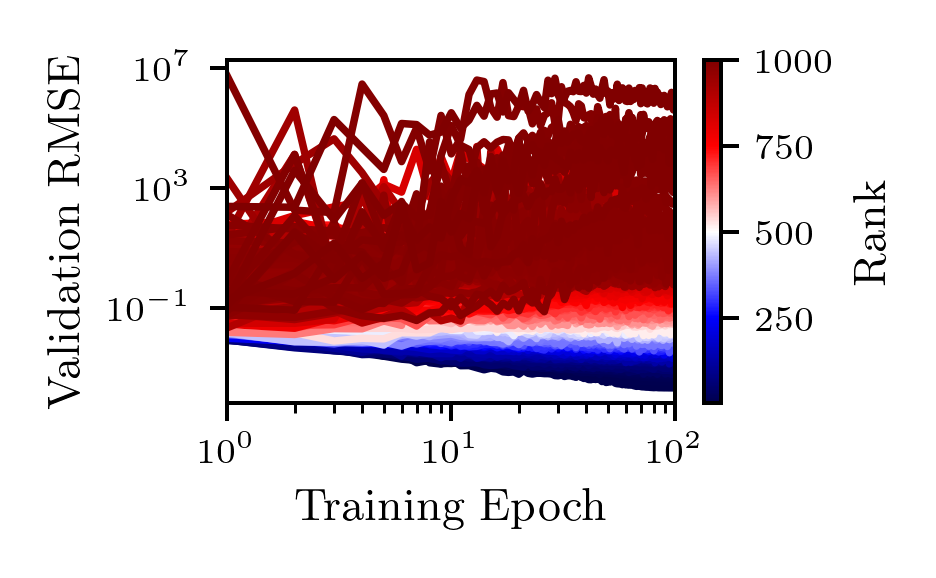}
        \caption{Slice Localization}
        \label{fig:ranking-curves-top10-slicelocalization}
    \end{subfigure}
    \caption{Learning curves of 1000 randomly sampled models from HPOBench colored by the rank at maximum fidelity.}
    \label{fig:all-ranking-curves-hpobench}
\end{figure}

\begin{figure}
    \centering
    \begin{subfigure}[b]{0.45\textwidth}
        \centering
        \includegraphics[width=\textwidth]{figures/hpobench/navalpropulsion/heatmap_ranks_naval_propulsion.png}
        \caption{NavalPropulsion}
        \label{fig:heatmap-top10-navalpropulsion}
    \end{subfigure}
    \begin{subfigure}[b]{0.45\textwidth}
        \centering
        \includegraphics[width=\textwidth]{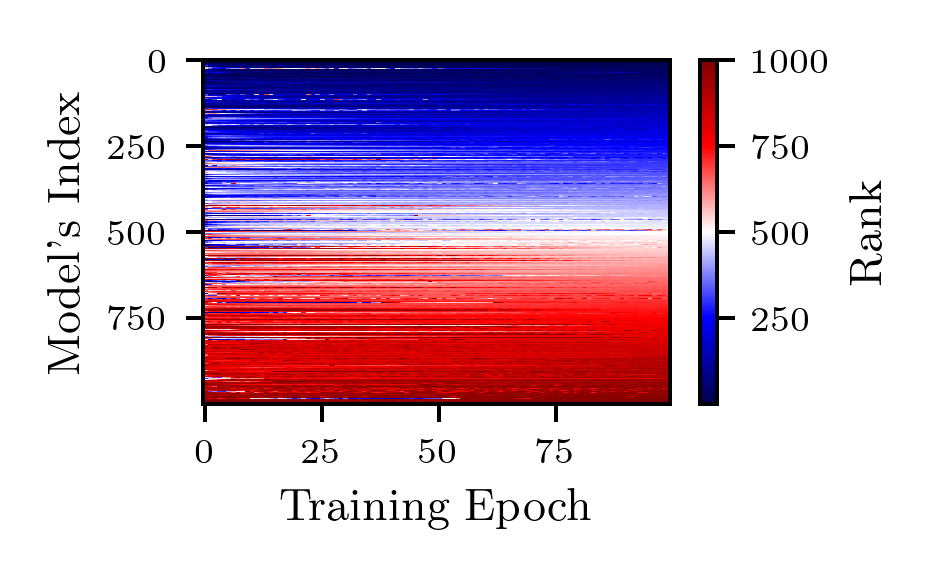}
        \caption{ParkinsonsTelemonitoring}
        \label{fig:heatmap-top10-parkinsonstelemonitoring}
    \end{subfigure}
    \begin{subfigure}[b]{0.45\textwidth}
        \centering
        \includegraphics[width=\textwidth]{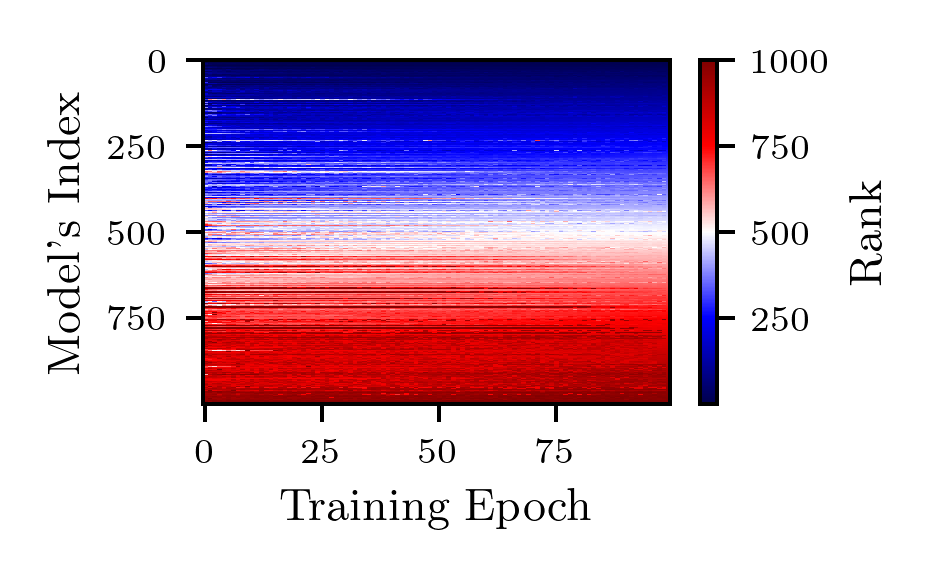}
        \caption{ProteinStructure}
        \label{fig:heatmap-top10-proteinstructure}
    \end{subfigure}
    \begin{subfigure}[b]{0.45\textwidth}
        \centering
        \includegraphics[width=\textwidth]{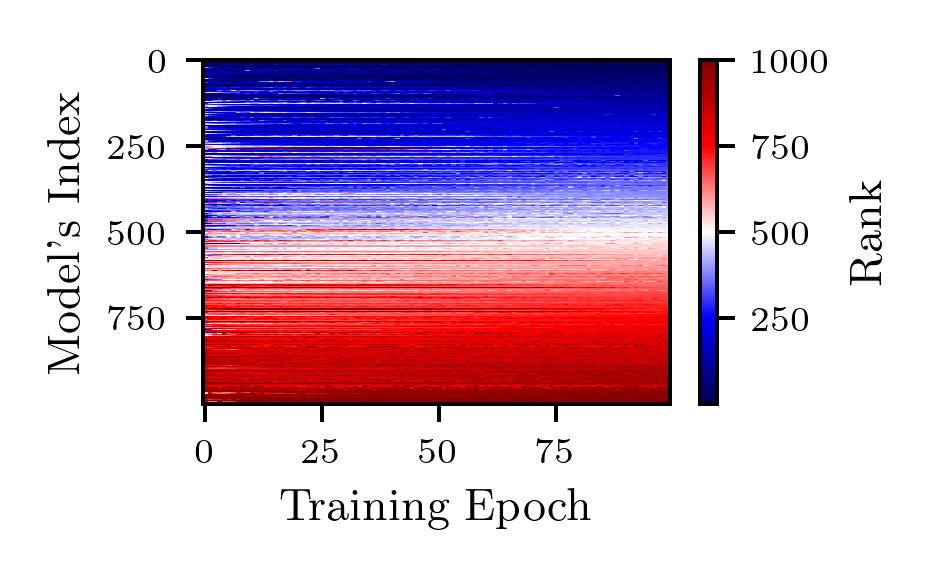}
        \caption{SliceLocalization}
        \label{fig:heatmap-top10-slicelocalization}
    \end{subfigure}
    \caption{Evolution of the ranking from minimal to maximal fidelity on 1,000 randomly sampled models from the search space of HPOBench.}
    \label{fig:all-heatmap-hpobench}
\end{figure}

In this section, we display in Figure~\ref{fig:viz-learning-curves-hpobench-log-scales} the 200 randomly picked learning curves for the four benchmarks considered. It is interesting to observe these curves to notice their diversity. Also, many outliers are visible which makes regular and well behaving learning curves harder to distinguish (and impossible in a linear scale).
Then, we present 1,000 randomly selected learning curves colored by their final ranking in Figure~\ref{fig:all-ranking-curves-hpobench} and the corresponding heatmaps in Figure~\ref{fig:all-heatmap-hpobench}. In such visualization, we can identify clearly two groups among learning curves (good and bad candidates). Also we can notice that it is possible to identify good candidates only based on the first epoch.

\section{Visualization of Learning Curves from LCBench}~\label{sec:appendix-viz-learning-curves-lcbench}

\begin{figure}[!h]
    \centering

    \begin{subfigure}[b]{0.24\textwidth}
        \centering
        \includegraphics[width=\textwidth]{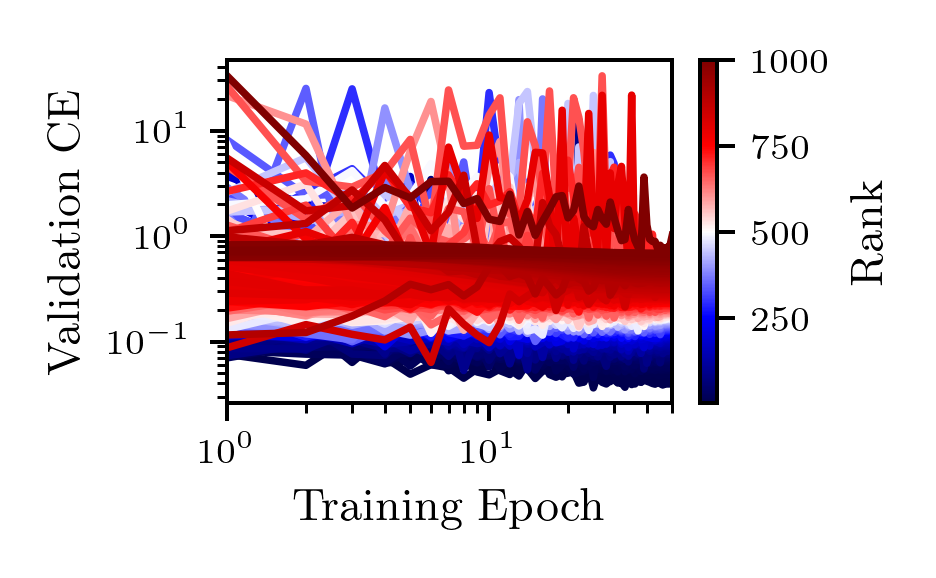}
        \caption{APSFailure}
        \label{fig:ranking-curves-top10-APSFailure}
    \end{subfigure}
    \begin{subfigure}[b]{0.24\textwidth}
        \centering
        \includegraphics[width=\textwidth]{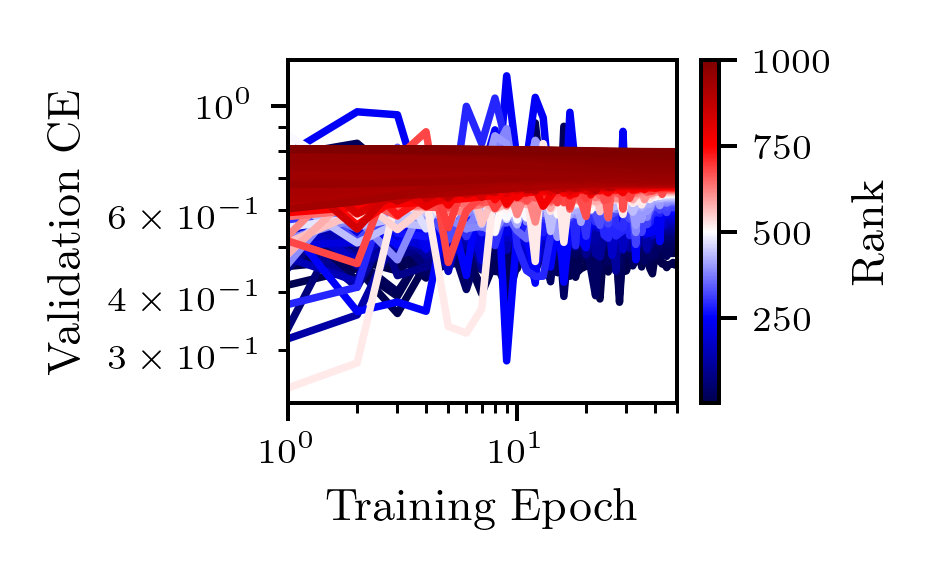}
        \caption{Amazon employee access}
        \label{fig:ranking-curves-top10-Amazon_employee_access}
    \end{subfigure}
    \begin{subfigure}[b]{0.24\textwidth}
        \centering
        \includegraphics[width=\textwidth]{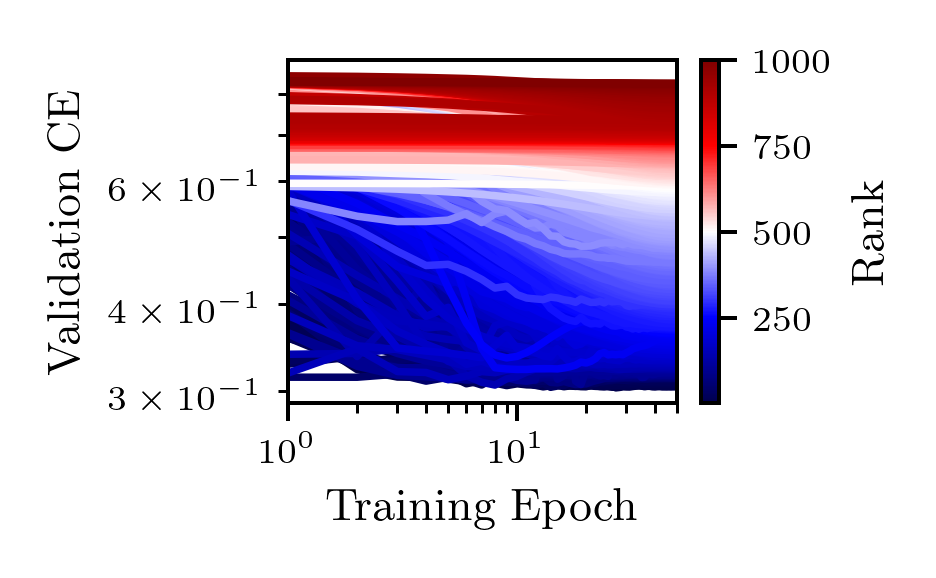}
        \caption{Australian}
        \label{fig:ranking-curves-top10-Australian}
    \end{subfigure}
    \begin{subfigure}[b]{0.24\textwidth}
        \centering
        \includegraphics[width=\textwidth]{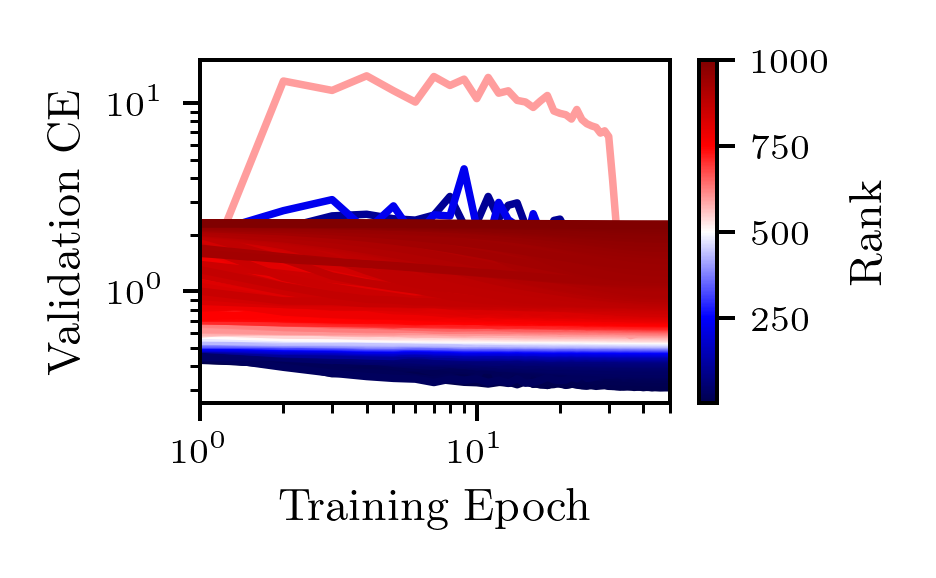}
        \caption{Fashion MNIST}
        \label{fig:ranking-curves-top10-Fashion-MNIST}
    \end{subfigure}
    \begin{subfigure}[b]{0.24\textwidth}
        \centering
        \includegraphics[width=\textwidth]{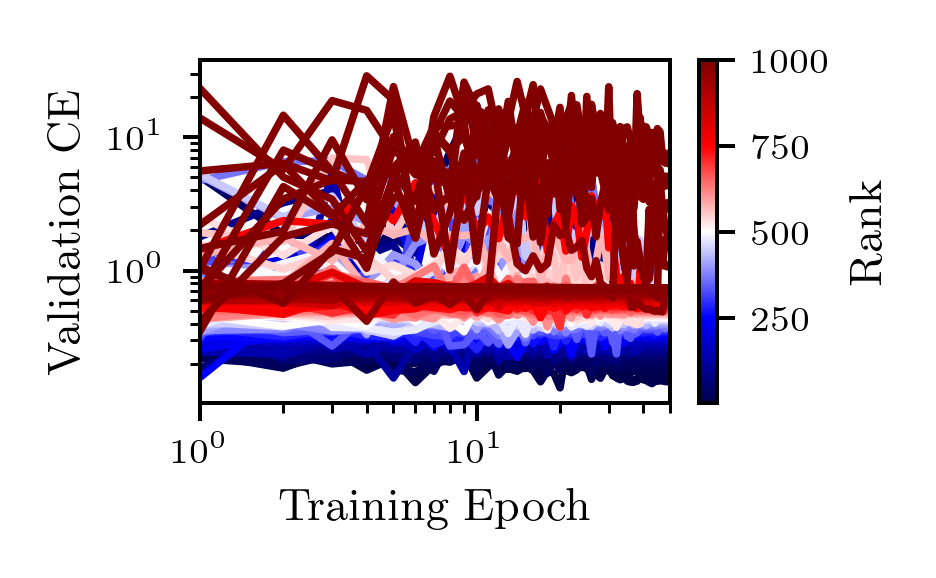}
        \caption{KDDCup09 appetency}
        \label{fig:ranking-curves-top10-KDDCup09_appetency}
    \end{subfigure}
    \begin{subfigure}[b]{0.24\textwidth}
        \centering
        \includegraphics[width=\textwidth]{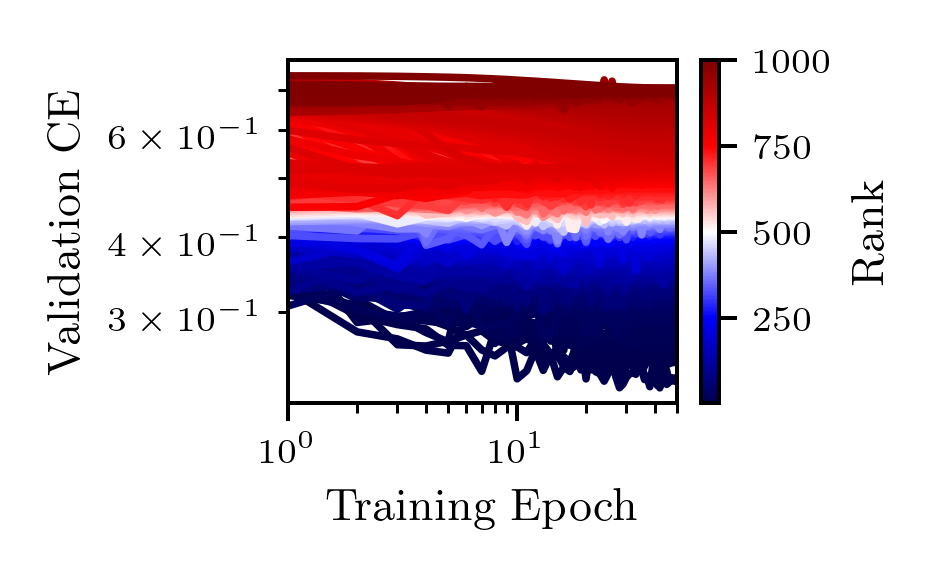}
        \caption{MiniBooNE}
        \label{fig:ranking-curves-top10-MiniBooNE}
    \end{subfigure}
    \begin{subfigure}[b]{0.24\textwidth}
        \centering
        \includegraphics[width=\textwidth]{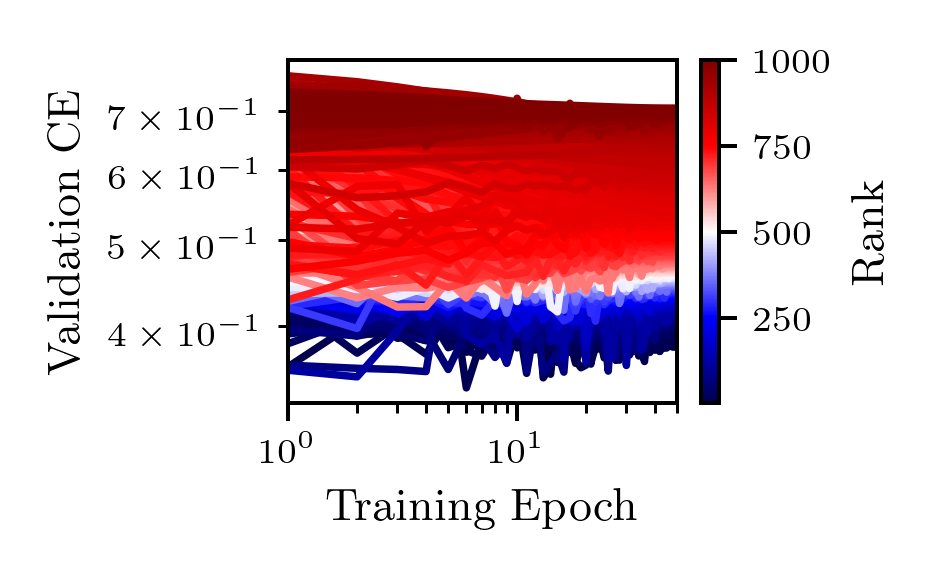}
        \caption{adult}
        \label{fig:ranking-curves-top10-adult}
    \end{subfigure}
    \begin{subfigure}[b]{0.24\textwidth}
        \centering
        \includegraphics[width=\textwidth]{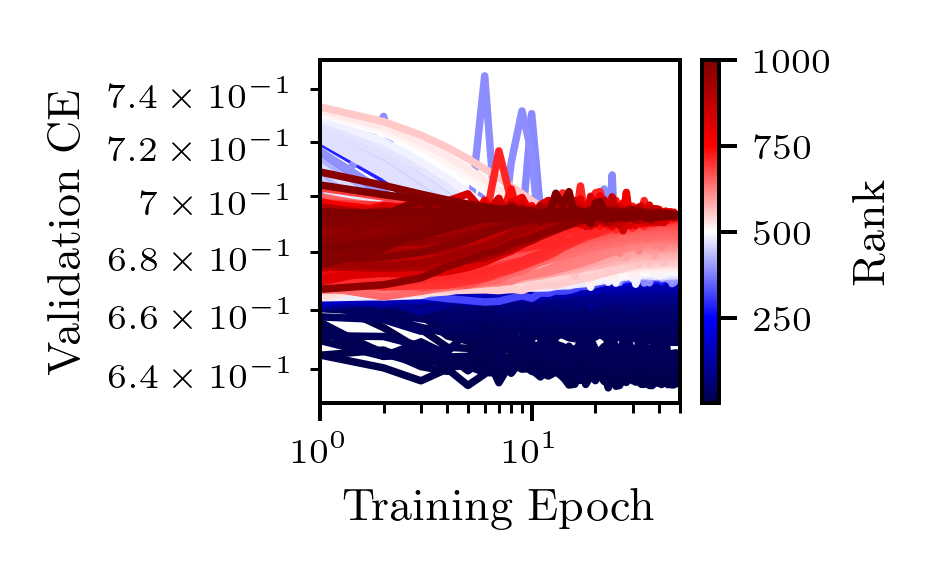}
        \caption{airlines}
        \label{fig:ranking-curves-top10-airlines}
    \end{subfigure}
    \begin{subfigure}[b]{0.24\textwidth}
        \centering
        \includegraphics[width=\textwidth]{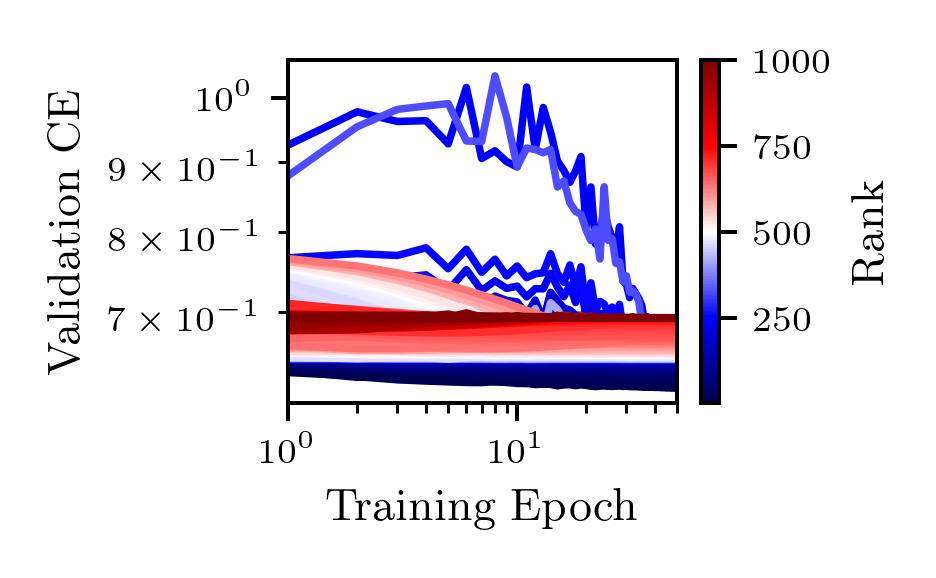}
        \caption{albert}
        \label{fig:ranking-curves-top10-albert}
    \end{subfigure}
    \begin{subfigure}[b]{0.24\textwidth}
        \centering
        \includegraphics[width=\textwidth]{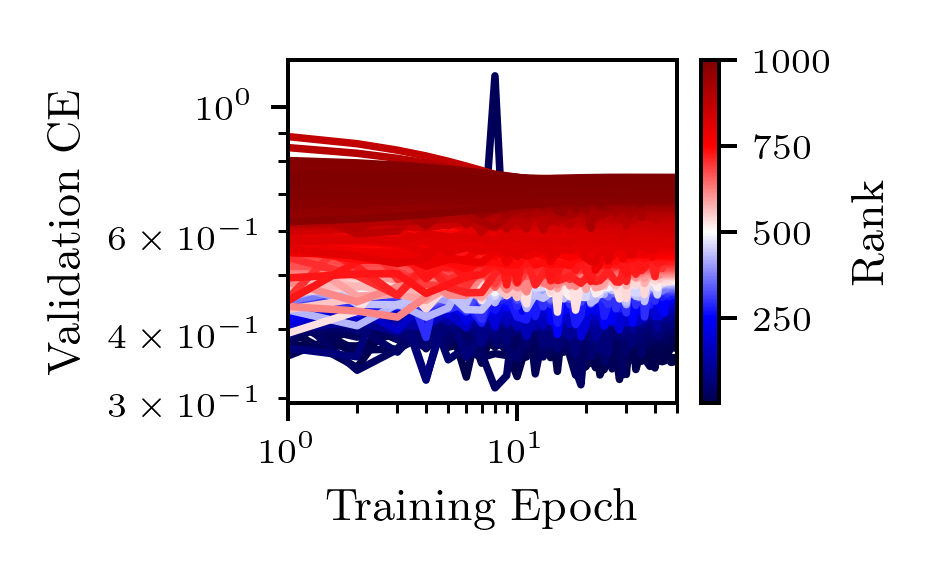}
        \caption{bank marketing}
        \label{fig:ranking-curves-top10-bank-marketing}
    \end{subfigure}
    \begin{subfigure}[b]{0.24\textwidth}
        \centering
        \includegraphics[width=\textwidth]{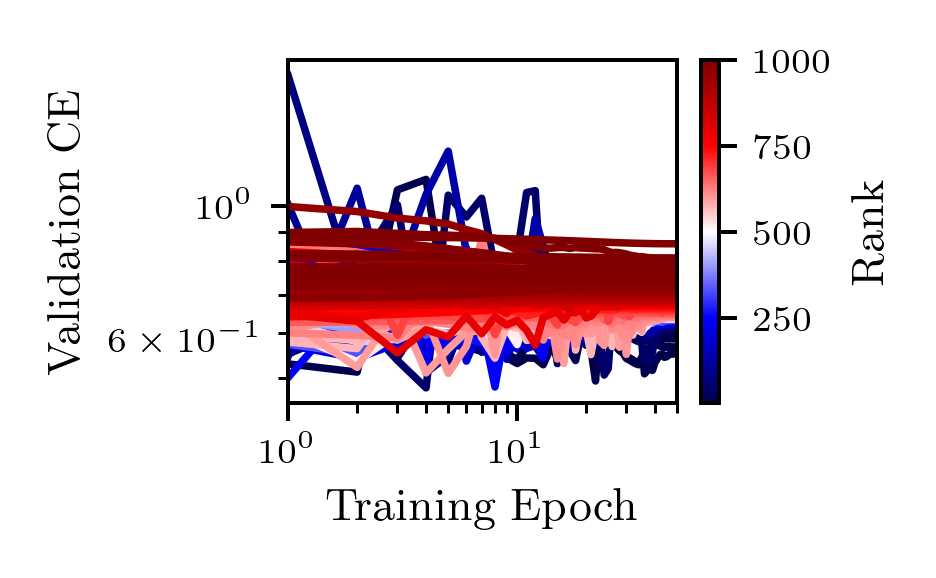}
        \caption{blood transfusion}
        \label{fig:ranking-curves-top10-blood-transfusion-service-center}
    \end{subfigure}
    \begin{subfigure}[b]{0.24\textwidth}
        \centering
        \includegraphics[width=\textwidth]{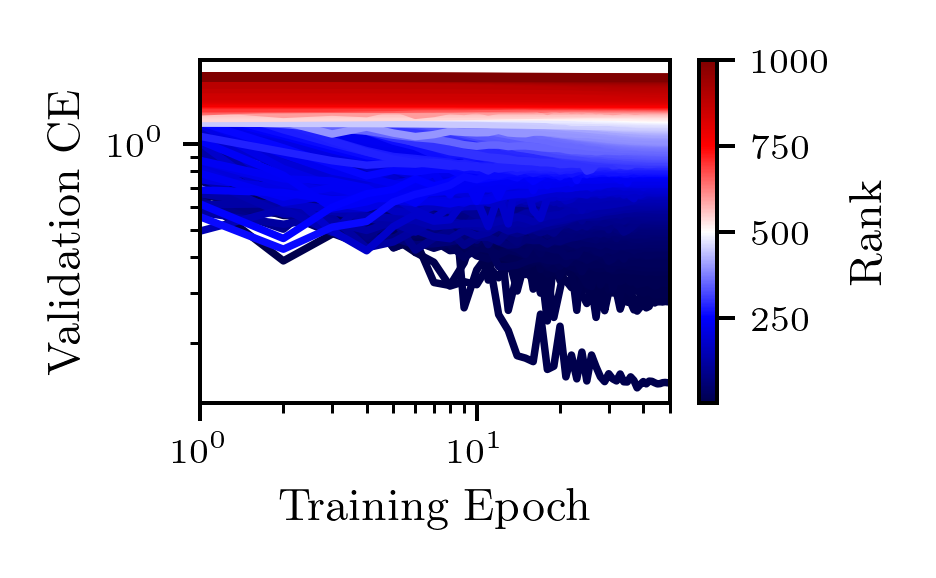}
        \caption{car}
        \label{fig:ranking-curves-top10-car}
    \end{subfigure}
    \begin{subfigure}[b]{0.24\textwidth}
        \centering
        \includegraphics[width=\textwidth]{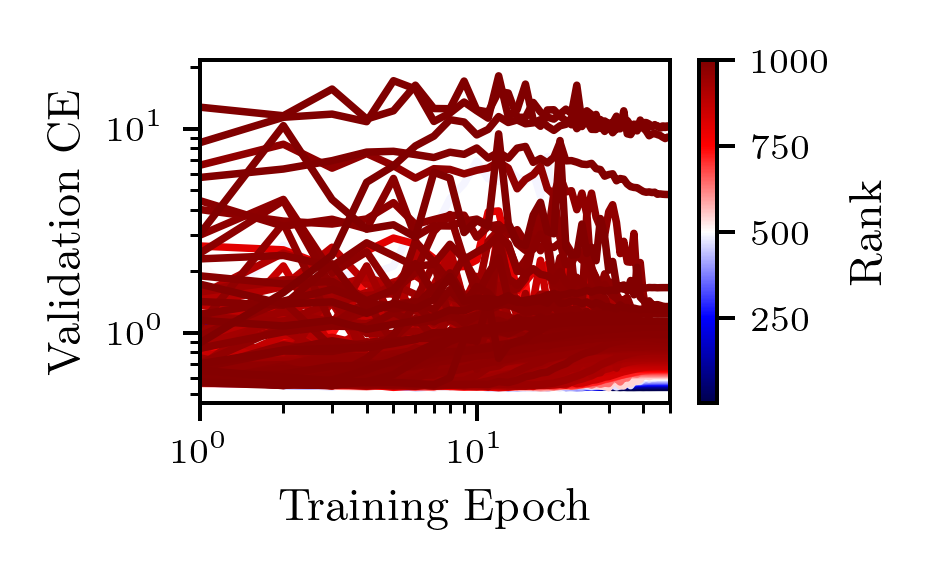}
        \caption{christine}
        \label{fig:ranking-curves-top10-christine}
    \end{subfigure}
    \begin{subfigure}[b]{0.24\textwidth}
        \centering
        \includegraphics[width=\textwidth]{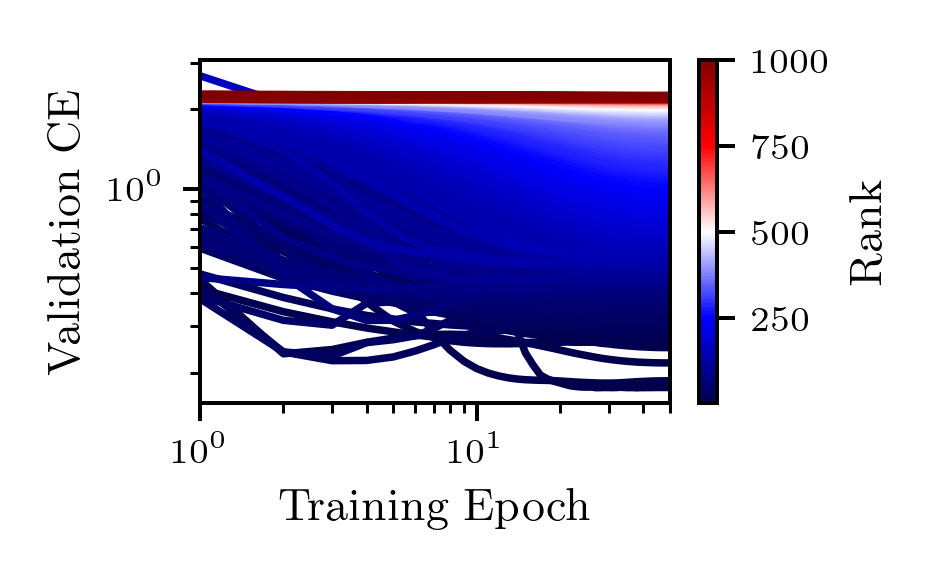}
        \caption{cnae 9}
        \label{fig:ranking-curves-top10-cnae-9}
    \end{subfigure}
    \begin{subfigure}[b]{0.24\textwidth}
        \centering
        \includegraphics[width=\textwidth]{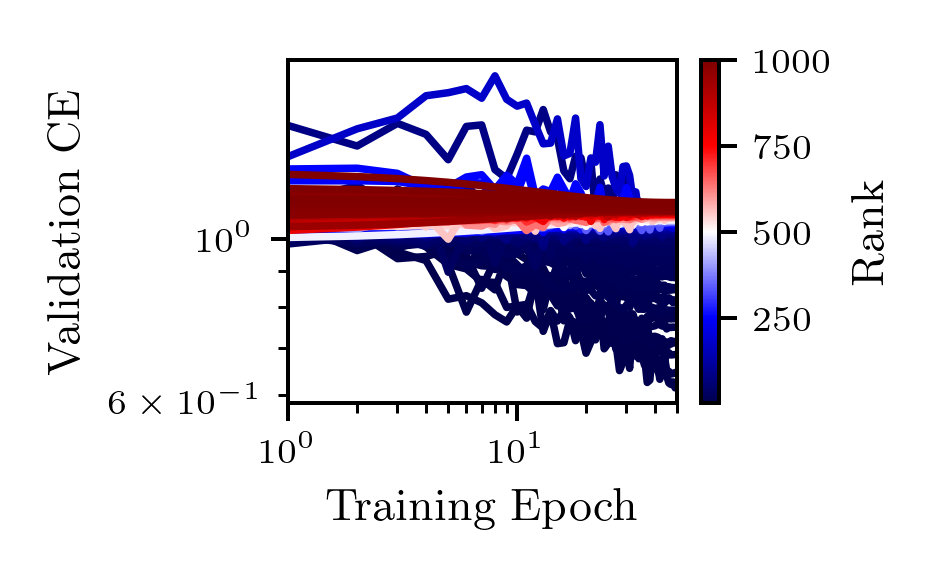}
        \caption{connect 4}
        \label{fig:ranking-curves-top10-connect-4}
    \end{subfigure}
    \begin{subfigure}[b]{0.24\textwidth}
        \centering
        \includegraphics[width=\textwidth]{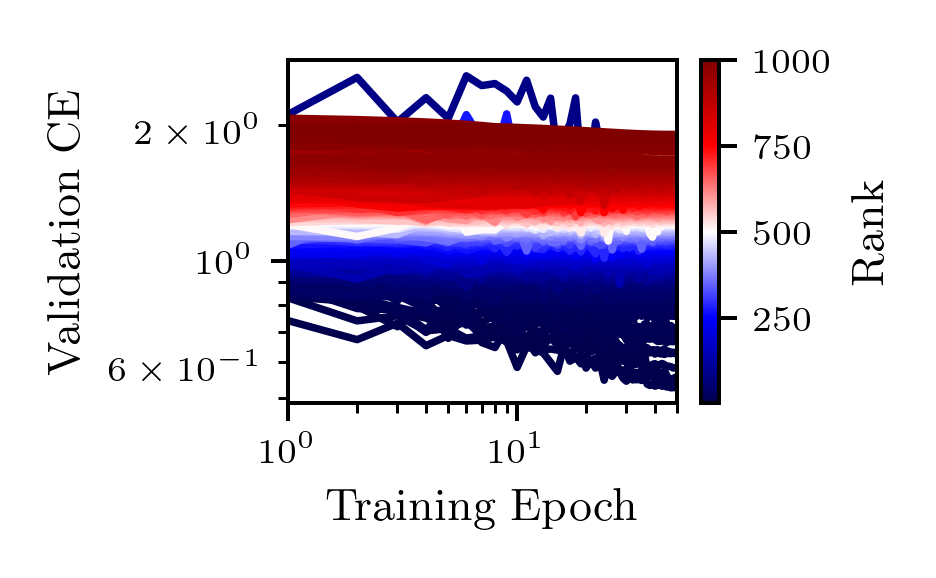}
        \caption{covertype}
        \label{fig:ranking-curves-top10-covertype}
    \end{subfigure}
    \begin{subfigure}[b]{0.24\textwidth}
        \centering
        \includegraphics[width=\textwidth]{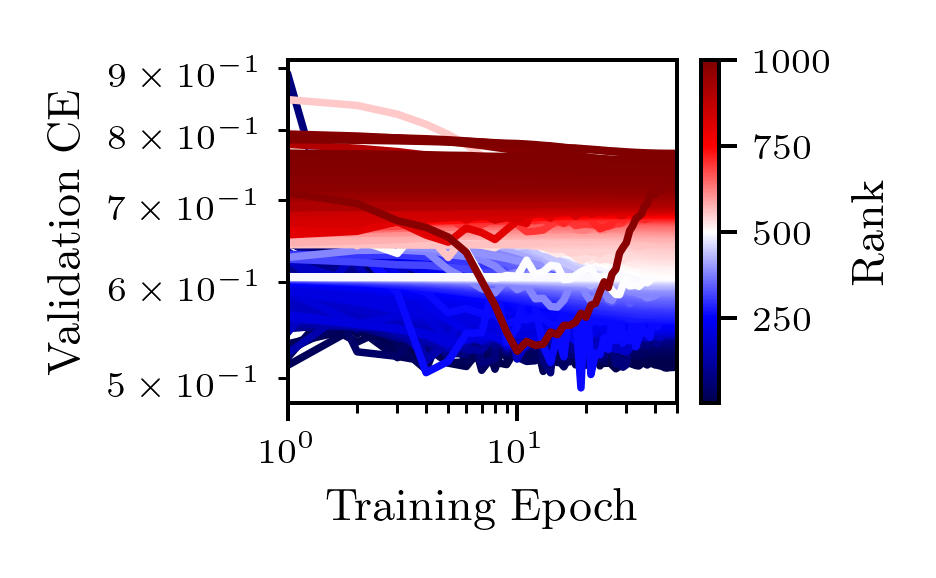}
        \caption{credit g}
        \label{fig:ranking-curves-top10-credit-g}
    \end{subfigure}
    \begin{subfigure}[b]{0.24\textwidth}
        \centering
        \includegraphics[width=\textwidth]{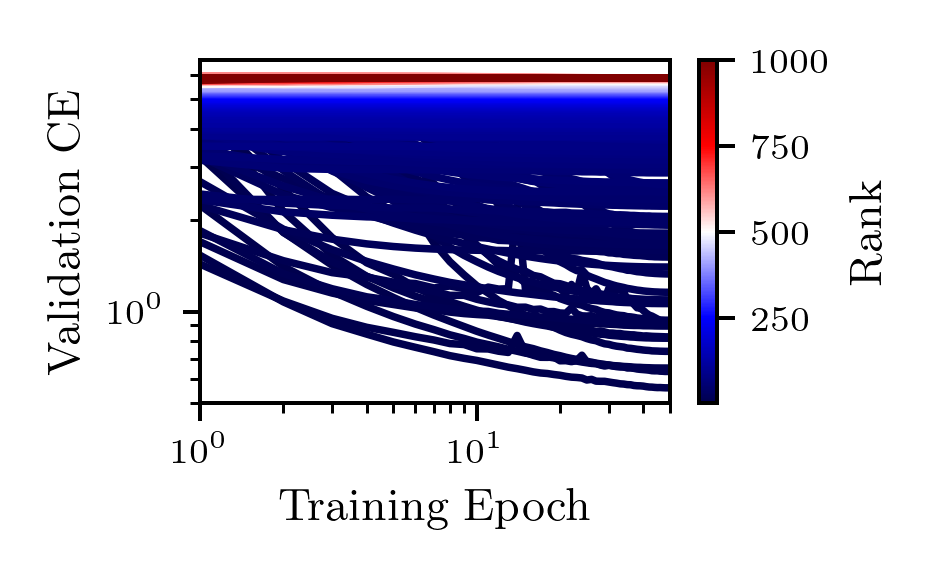}
        \caption{dionis}
        \label{fig:ranking-curves-top10-dionis}
    \end{subfigure}
    \begin{subfigure}[b]{0.24\textwidth}
        \centering
        \includegraphics[width=\textwidth]{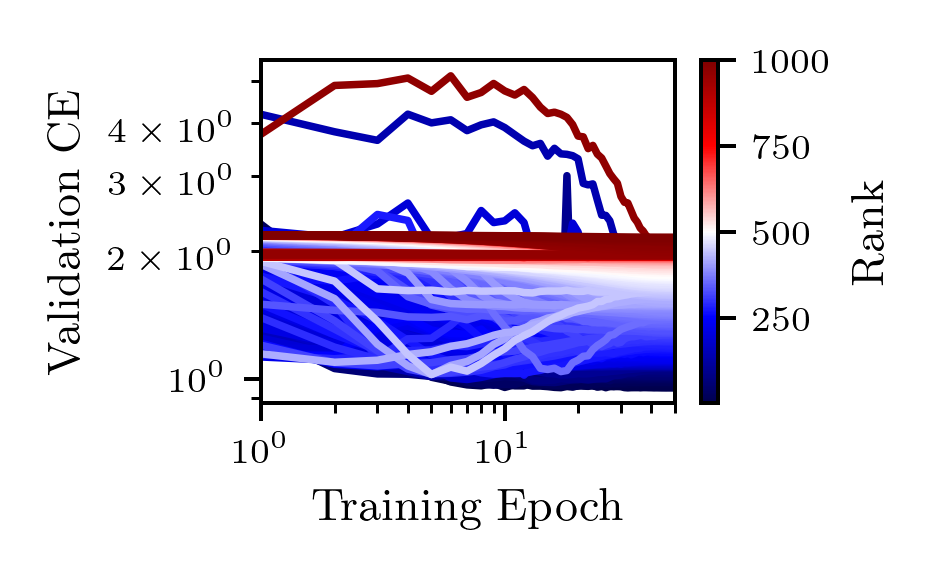}
        \caption{fabert}
        \label{fig:ranking-curves-top10-fabert}
    \end{subfigure}
    \begin{subfigure}[b]{0.24\textwidth}
        \centering
        \includegraphics[width=\textwidth]{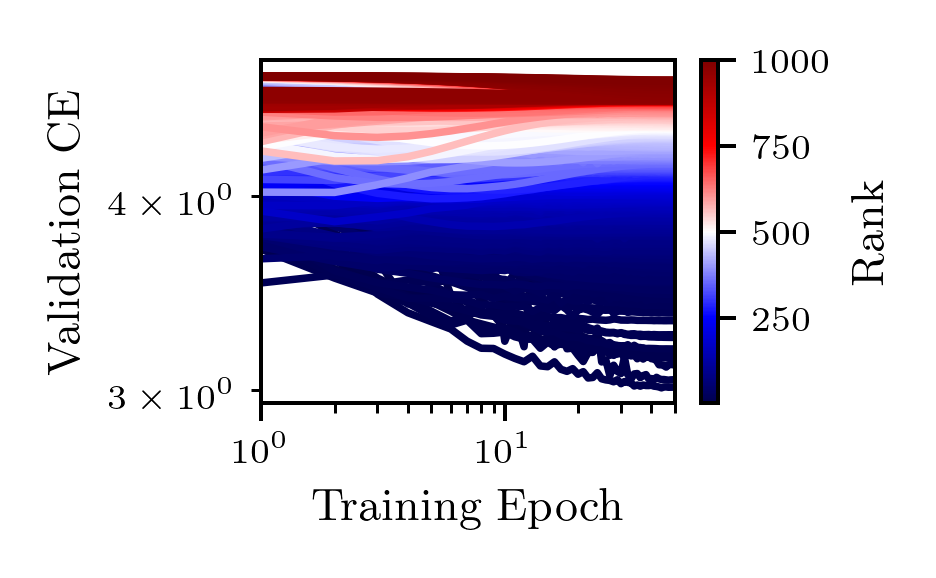}
        \caption{helena}
        \label{fig:ranking-curves-top10-helena}
    \end{subfigure}
    \begin{subfigure}[b]{0.24\textwidth}
        \centering
        \includegraphics[width=\textwidth]{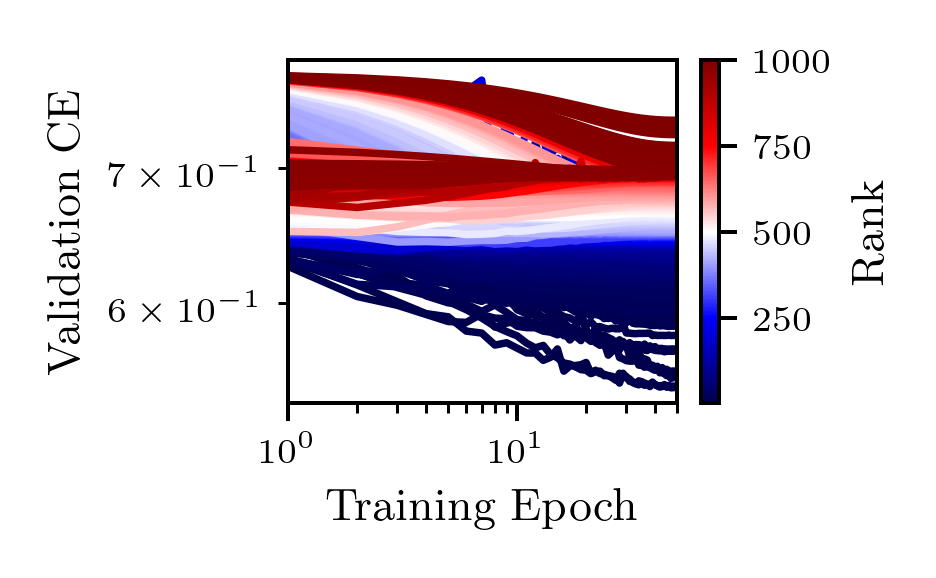}
        \caption{higgs}
        \label{fig:ranking-curves-top10-higgs}
    \end{subfigure}
    \begin{subfigure}[b]{0.24\textwidth}
        \centering
        \includegraphics[width=\textwidth]{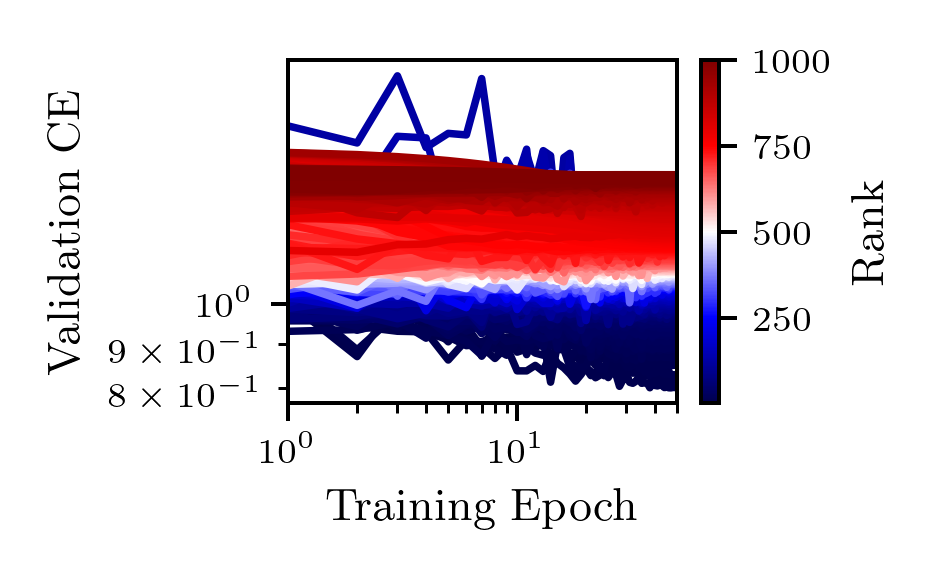}
        \caption{jannis}
        \label{fig:ranking-curves-top10-jannis}
    \end{subfigure}
    \begin{subfigure}[b]{0.24\textwidth}
        \centering
        \includegraphics[width=\textwidth]{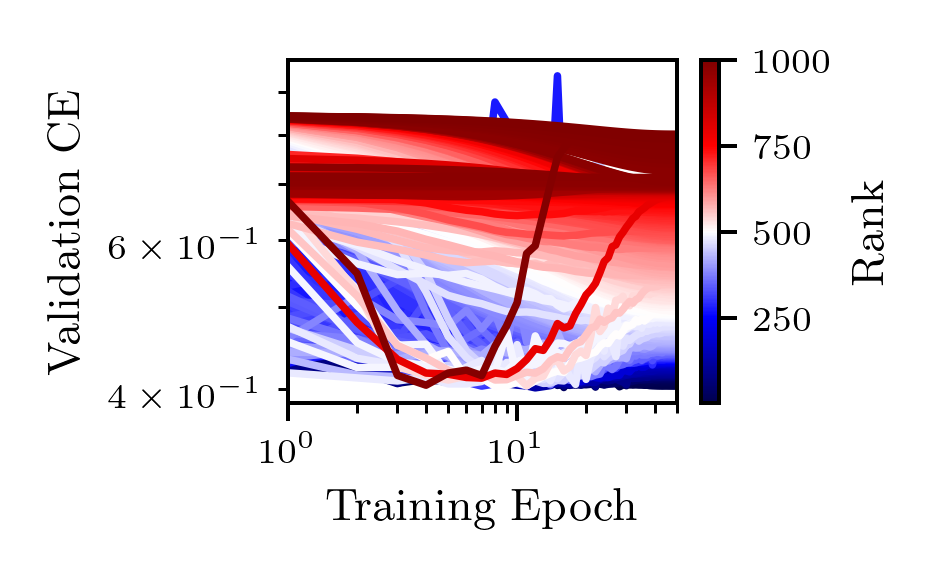}
        \caption{jasmine}
        \label{fig:ranking-curves-top10-jasmine}
    \end{subfigure}
    \begin{subfigure}[b]{0.24\textwidth}
        \centering
        \includegraphics[width=\textwidth]{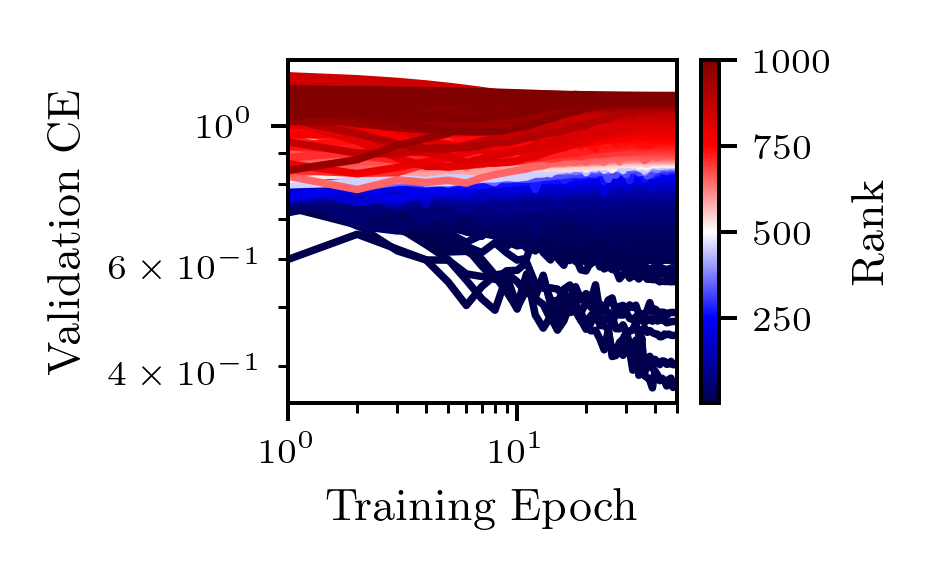}
        \caption{jungle chess 2pcs raw endgame complete}
        \label{fig:ranking-curves-top10-jungle_chess_2pcs_raw_endgame_complete}
    \end{subfigure}
    \begin{subfigure}[b]{0.24\textwidth}
        \centering
        \includegraphics[width=\textwidth]{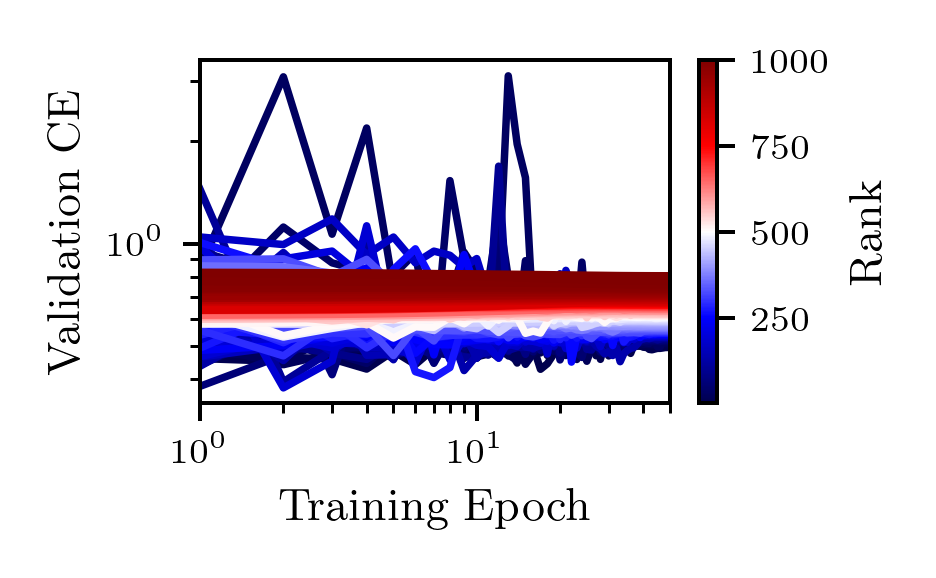}
        \caption{kc1}
        \label{fig:ranking-curves-top10-kc1}
    \end{subfigure}
    \begin{subfigure}[b]{0.24\textwidth}
        \centering
        \includegraphics[width=\textwidth]{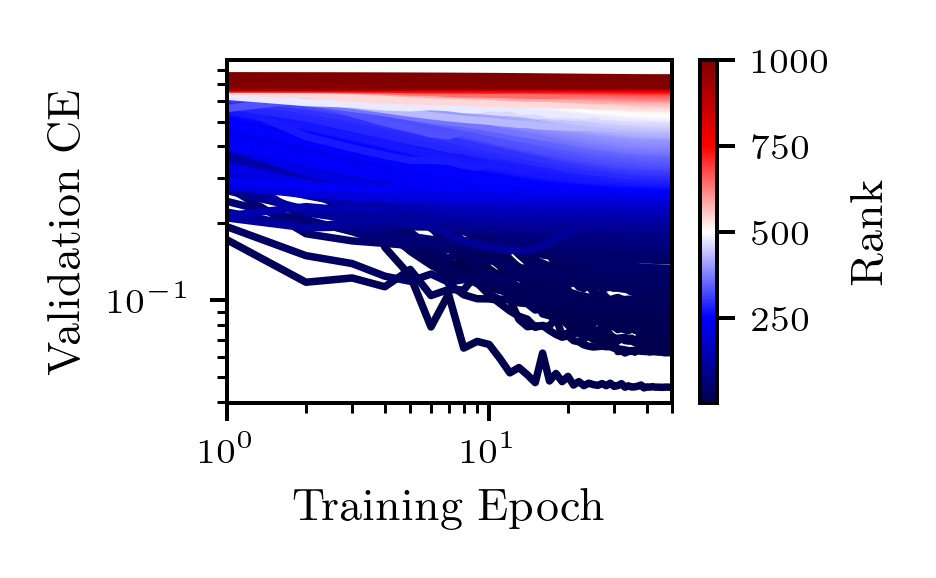}
        \caption{kr vs kp}
        \label{fig:ranking-curves-top10-kr-vs-kp}
    \end{subfigure}
    
    \caption{Learning curves of 1000 randomly sampled models from HPOBench colored by the rank at maximum fidelity ({\bf First Part}).}
    \label{fig:all-ranking-curves-lcbench}
\end{figure}

\begin{figure}[!h]
    \centering
    
    \begin{subfigure}[b]{0.24\textwidth}
        \centering
        \includegraphics[width=\textwidth]{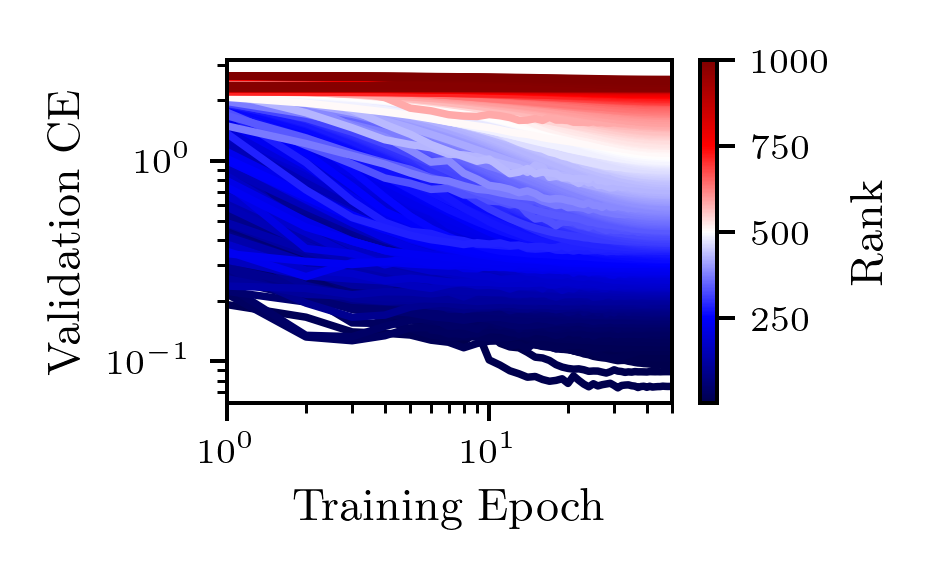}
        \caption{mfeat factors}
        \label{fig:ranking-curves-top10-mfeat-factors}
    \end{subfigure}
    \begin{subfigure}[b]{0.24\textwidth}
        \centering
        \includegraphics[width=\textwidth]{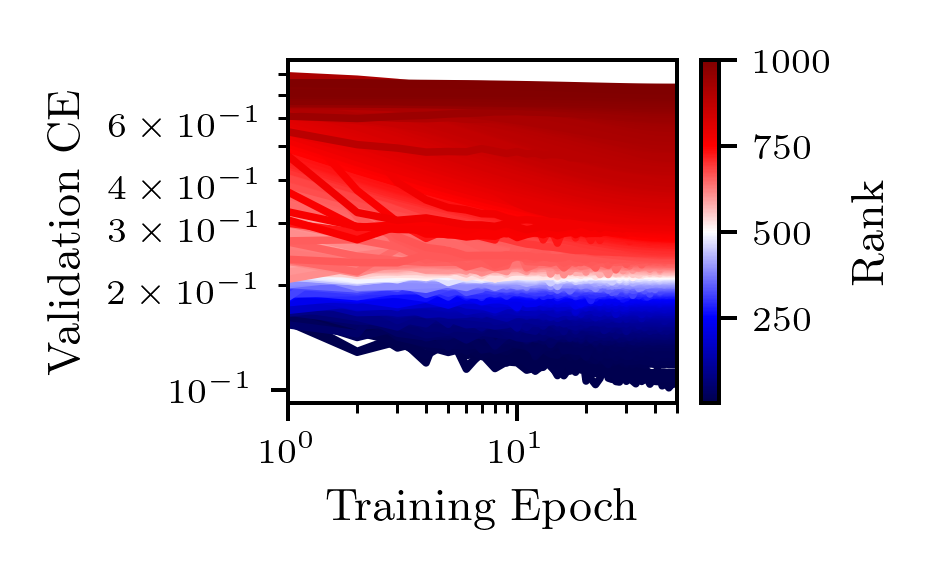}
        \caption{nomao}
        \label{fig:ranking-curves-top10-nomao}
    \end{subfigure}
    \begin{subfigure}[b]{0.24\textwidth}
        \centering
        \includegraphics[width=\textwidth]{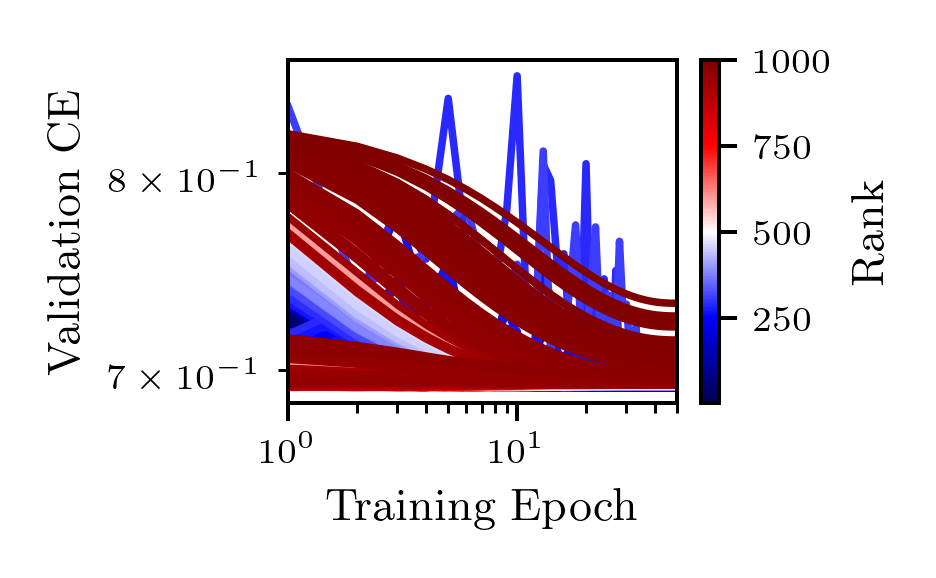}
        \caption{numerai28.6}
        \label{fig:ranking-curves-top10-numerai28.6}
    \end{subfigure}
    \begin{subfigure}[b]{0.24\textwidth}
        \centering
        \includegraphics[width=\textwidth]{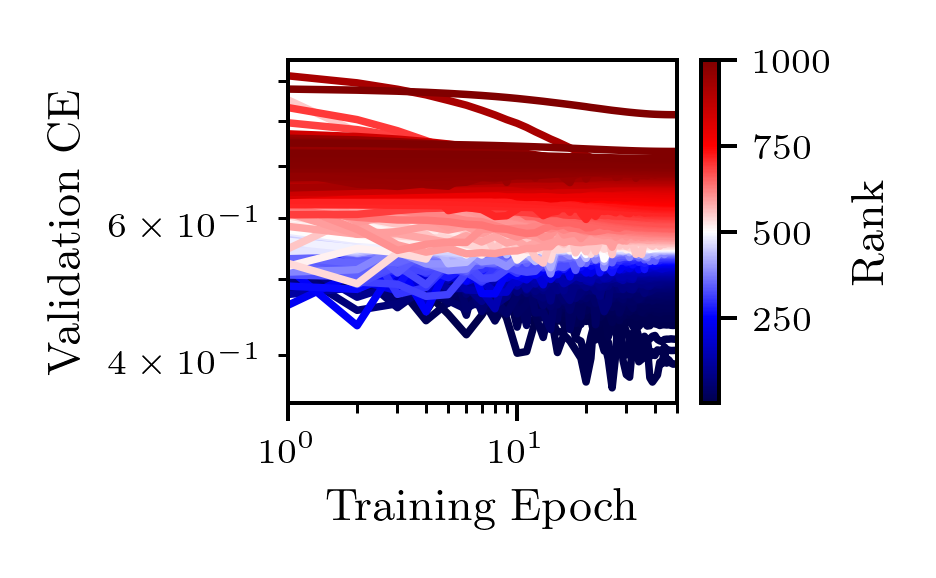}
        \caption{phoneme}
        \label{fig:ranking-curves-top10-phoneme}
    \end{subfigure}
    \begin{subfigure}[b]{0.24\textwidth}
        \centering
        \includegraphics[width=\textwidth]{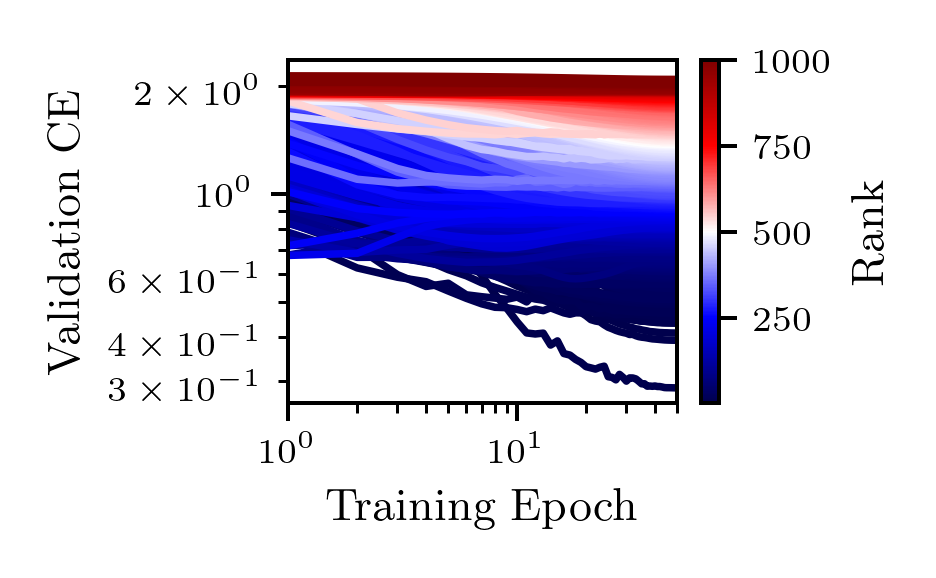}
        \caption{segment}
        \label{fig:ranking-curves-top10-segment}
    \end{subfigure}
    \begin{subfigure}[b]{0.24\textwidth}
        \centering
        \includegraphics[width=\textwidth]{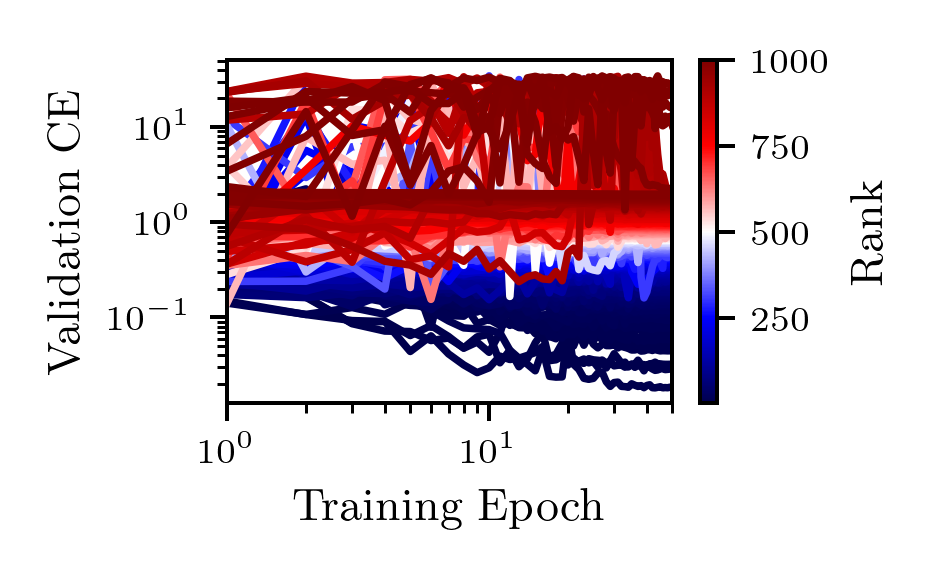}
        \caption{shuttle}
        \label{fig:ranking-curves-top10-shuttle}
    \end{subfigure}
    \begin{subfigure}[b]{0.24\textwidth}
        \centering
        \includegraphics[width=\textwidth]{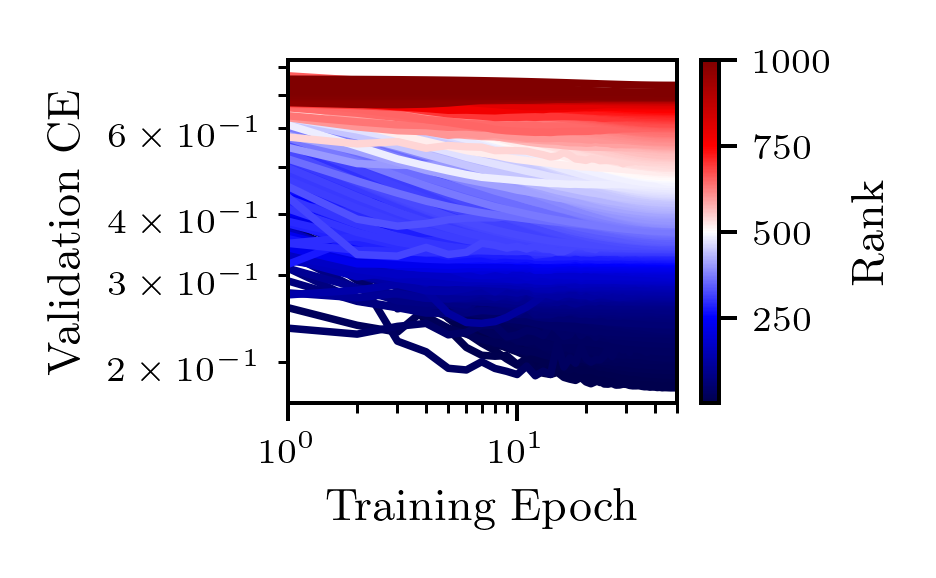}
        \caption{sylvine}
        \label{fig:ranking-curves-top10-sylvine}
    \end{subfigure}
    \begin{subfigure}[b]{0.24\textwidth}
        \centering
        \includegraphics[width=\textwidth]{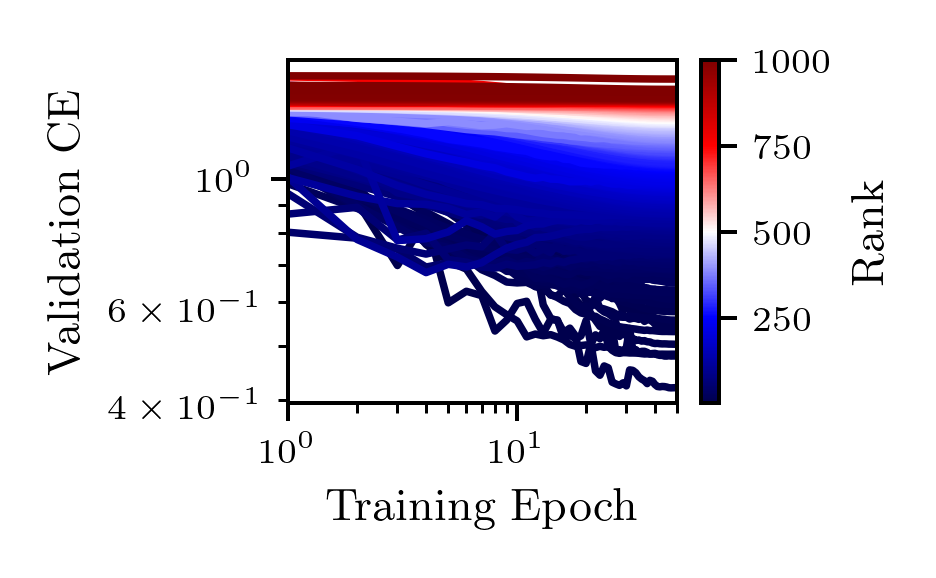}
        \caption{vehicle}
        \label{fig:ranking-curves-top10-vehicle}
    \end{subfigure}
    \begin{subfigure}[b]{0.24\textwidth}
        \centering
        \includegraphics[width=\textwidth]{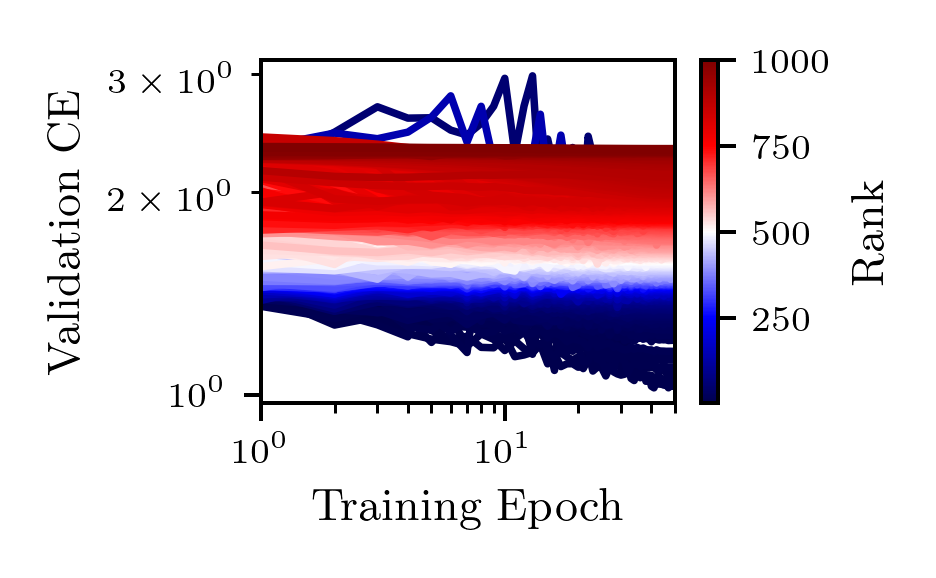}
        \caption{volkert}
        \label{fig:ranking-curves-top10-volkert}
    \end{subfigure}
    
    \caption{Learning curves of 1000 randomly sampled models from LCBench colored by the rank at maximum fidelity ({\bf Second Part}).}
    \label{fig:all-ranking-curves-lcbench-second-part}
\end{figure}

\begin{figure}[!h]
    \centering
    
    \begin{subfigure}[b]{0.24\textwidth}
        \centering
        \includegraphics[width=\textwidth]{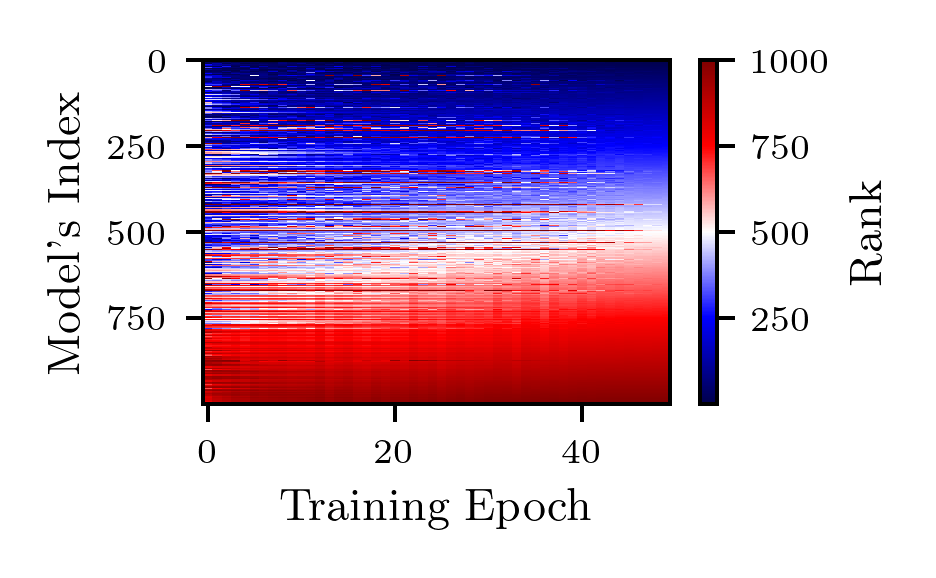}
        \caption{APSFailure}
        \label{fig:heatmap-top10-APSFailure}
    \end{subfigure}
    \begin{subfigure}[b]{0.24\textwidth}
        \centering
        \includegraphics[width=\textwidth]{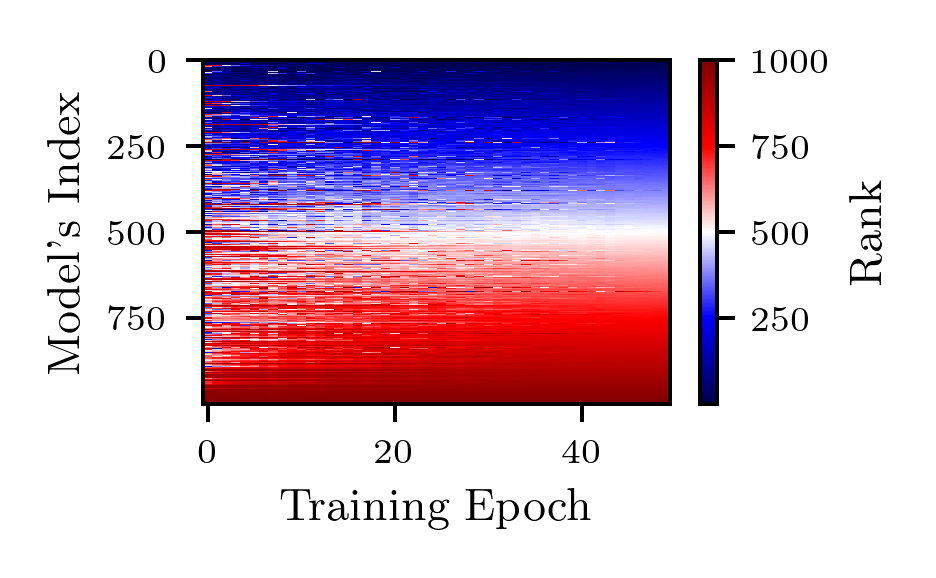}
        \caption{Amazon employee access}
        \label{fig:heatmap-top10-Amazon_employee_access}
    \end{subfigure}
    \begin{subfigure}[b]{0.24\textwidth}
        \centering
        \includegraphics[width=\textwidth]{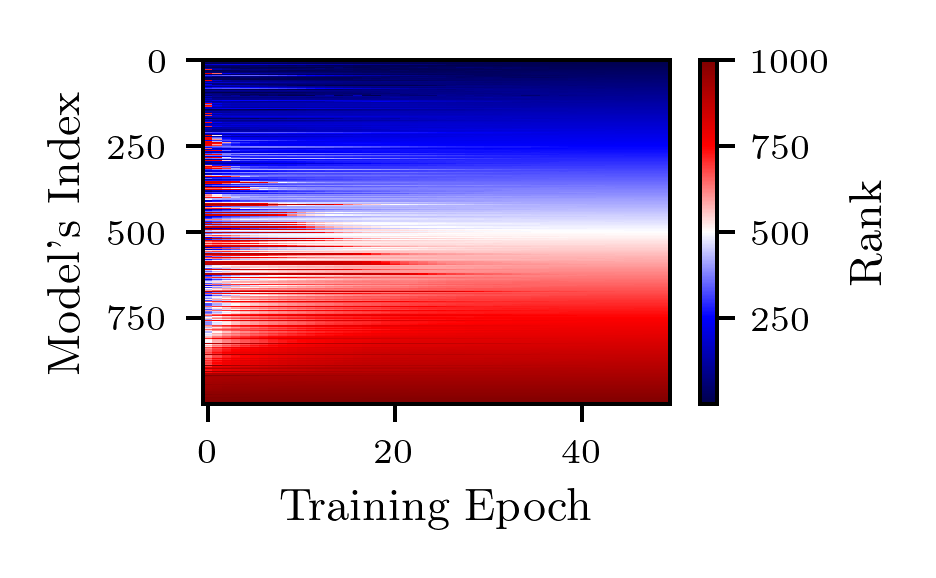}
        \caption{Australian}
        \label{fig:heatmap-top10-Australian}
    \end{subfigure}
    \begin{subfigure}[b]{0.24\textwidth}
        \centering
        \includegraphics[width=\textwidth]{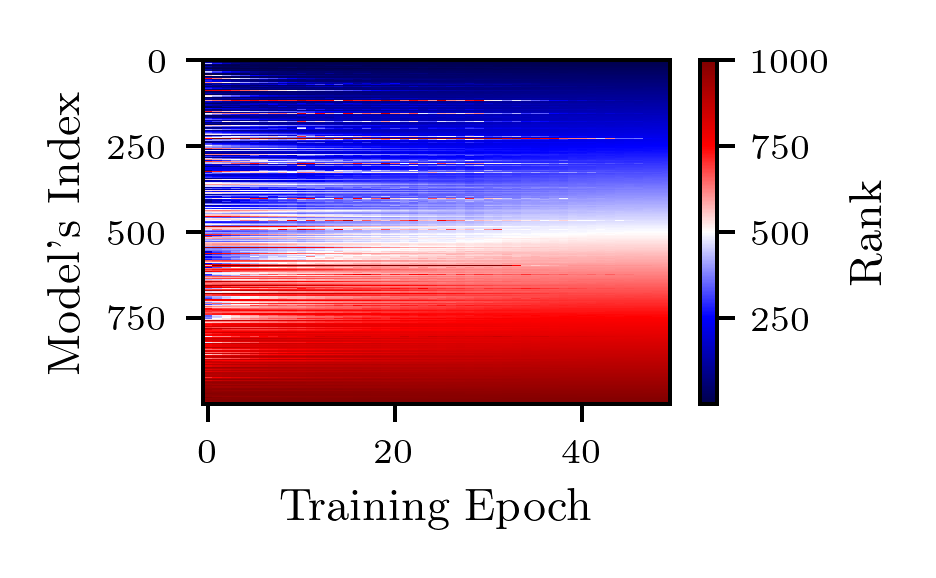}
        \caption{Fashion MNIST}
        \label{fig:heatmap-top10-Fashion-MNIST}
    \end{subfigure}
    \begin{subfigure}[b]{0.24\textwidth}
        \centering
        \includegraphics[width=\textwidth]{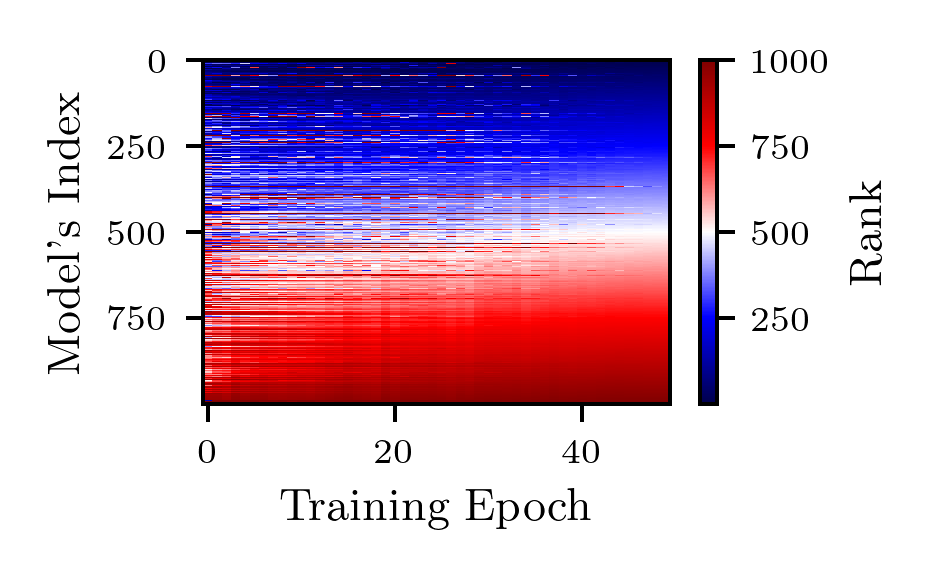}
        \caption{KDDCup09 appetency}
        \label{fig:heatmap-top10-KDDCup09_appetency}
    \end{subfigure}
    \begin{subfigure}[b]{0.24\textwidth}
        \centering
        \includegraphics[width=\textwidth]{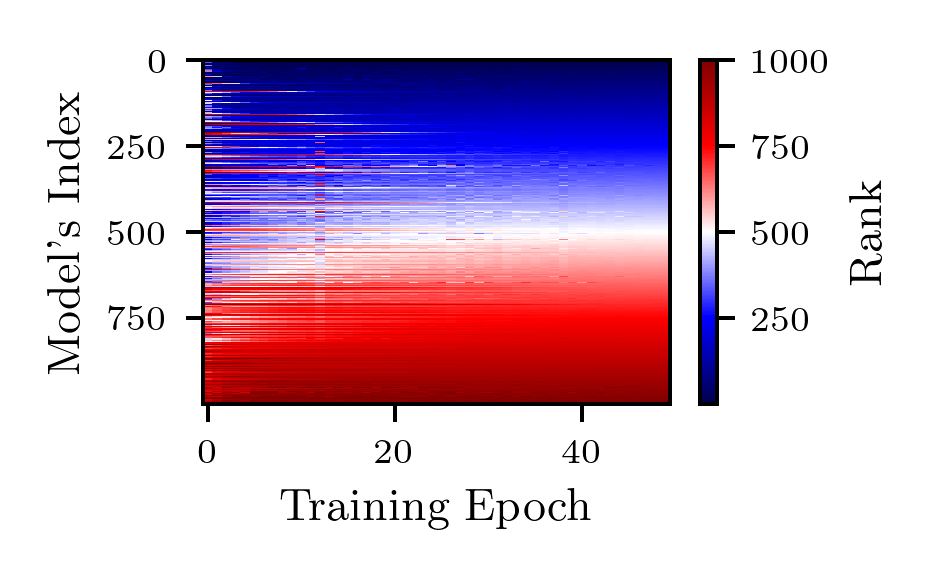}
        \caption{MiniBooNE}
        \label{fig:heatmap-top10-MiniBooNE}
    \end{subfigure}
    \begin{subfigure}[b]{0.24\textwidth}
        \centering
        \includegraphics[width=\textwidth]{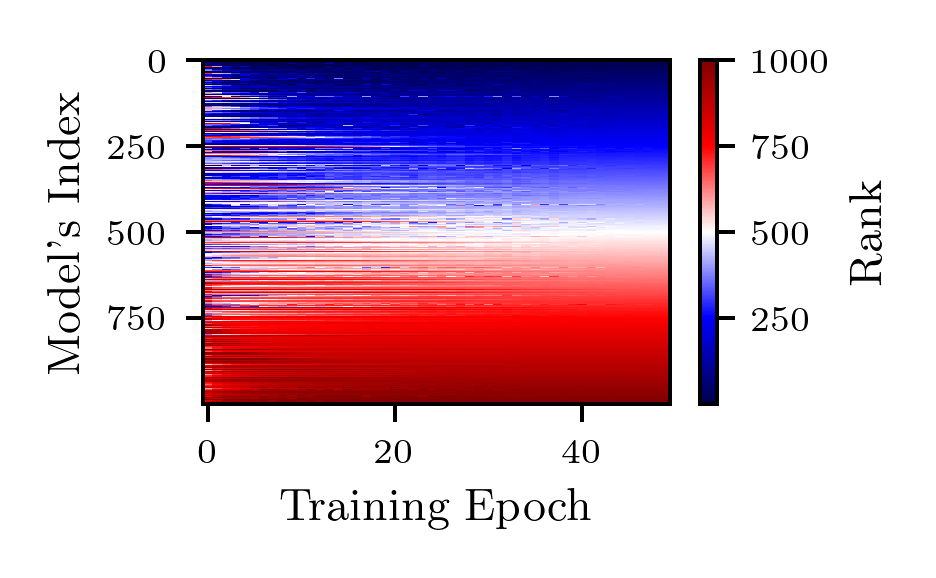}
        \caption{adult}
        \label{fig:heatmap-top10-adult}
    \end{subfigure}
    \begin{subfigure}[b]{0.24\textwidth}
        \centering
        \includegraphics[width=\textwidth]{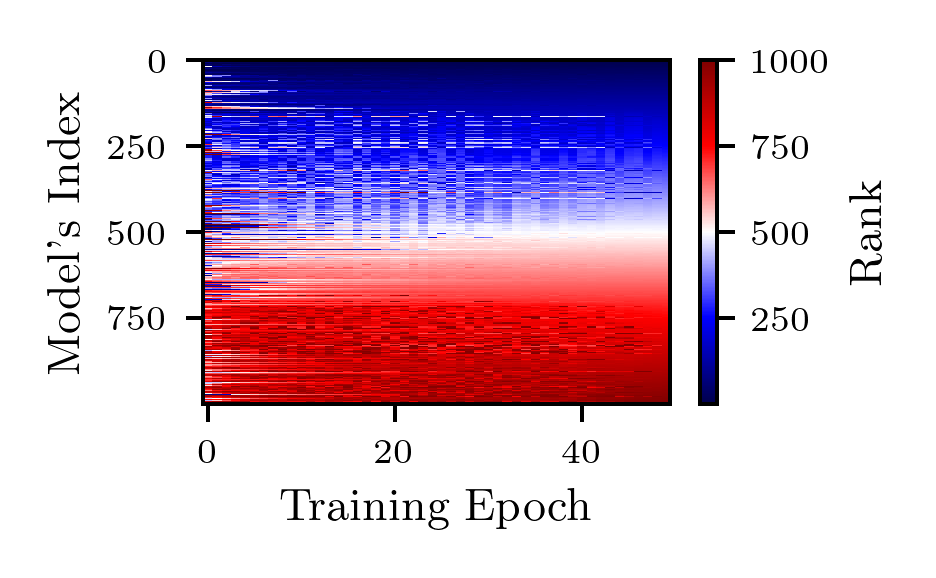}
        \caption{airlines}
        \label{fig:heatmap-top10-airlines}
    \end{subfigure}
    \begin{subfigure}[b]{0.24\textwidth}
        \centering
        \includegraphics[width=\textwidth]{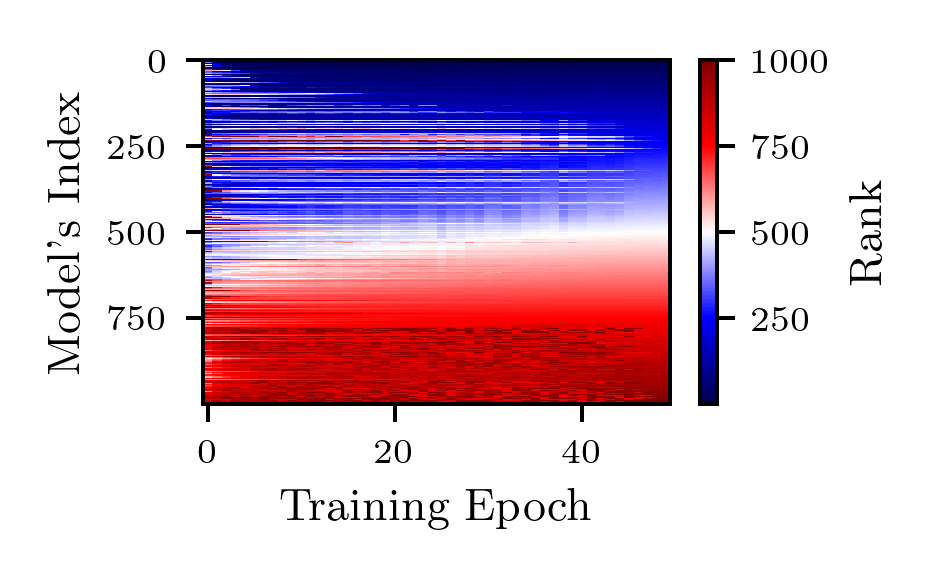}
        \caption{albert}
        \label{fig:heatmap-top10-albert}
    \end{subfigure}
    \begin{subfigure}[b]{0.24\textwidth}
        \centering
        \includegraphics[width=\textwidth]{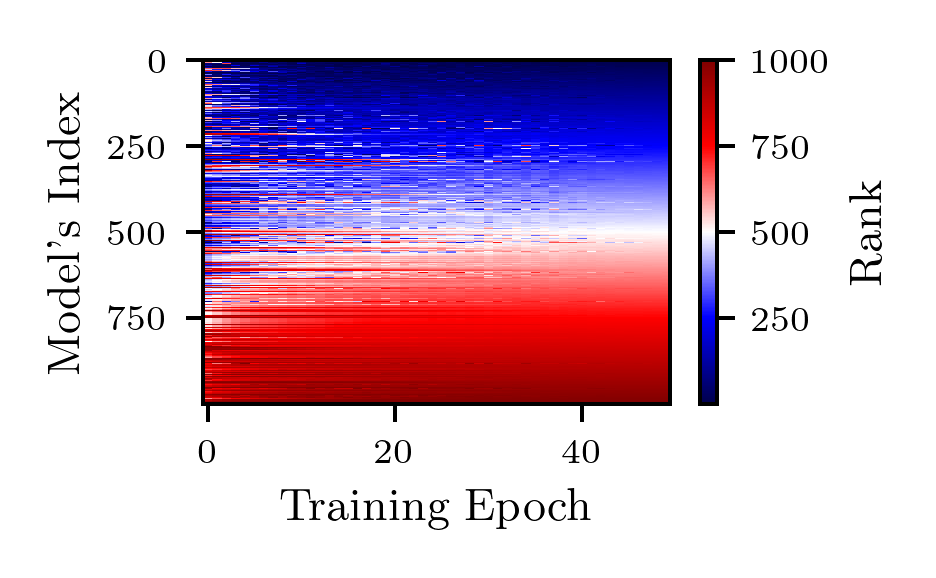}
        \caption{bank marketing}
        \label{fig:heatmap-top10-bank-marketing}
    \end{subfigure}
    \begin{subfigure}[b]{0.24\textwidth}
        \centering
        \includegraphics[width=\textwidth]{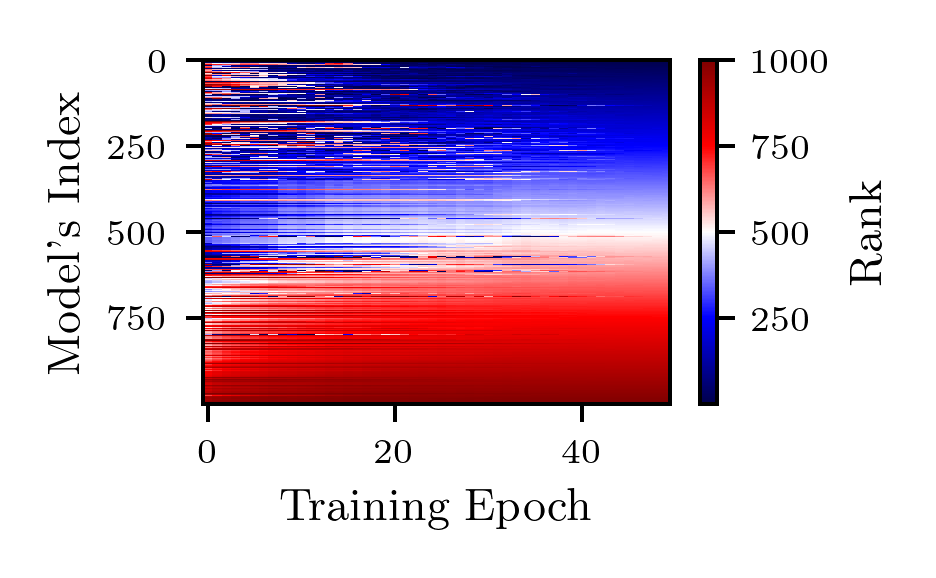}
        \caption{blood transfusion}
        \label{fig:heatmap-top10-blood-transfusion-service-center}
    \end{subfigure}
    \begin{subfigure}[b]{0.24\textwidth}
        \centering
        \includegraphics[width=\textwidth]{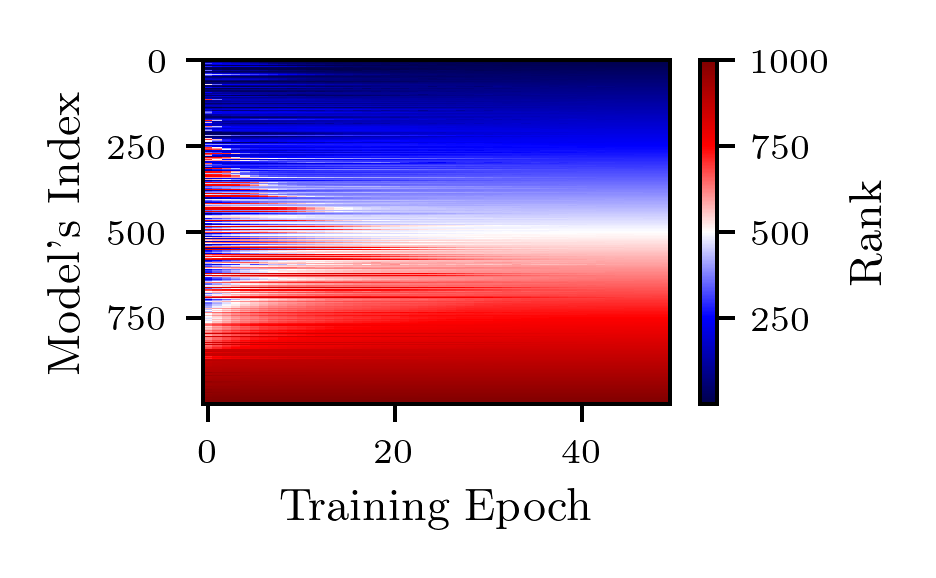}
        \caption{car}
        \label{fig:heatmap-top10-car}
    \end{subfigure}
    \begin{subfigure}[b]{0.24\textwidth}
        \centering
        \includegraphics[width=\textwidth]{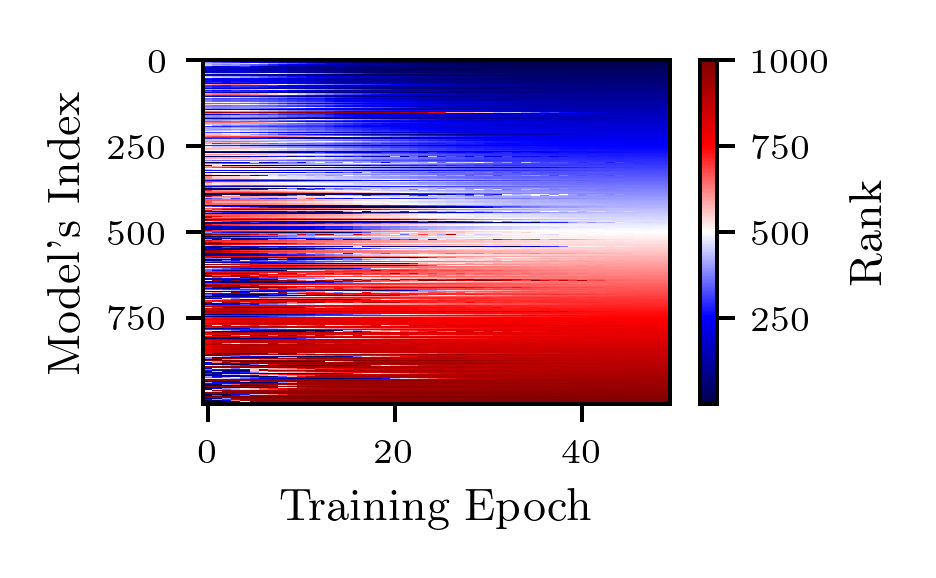}
        \caption{christine}
        \label{fig:heatmap-top10-christine}
    \end{subfigure}
    \begin{subfigure}[b]{0.24\textwidth}
        \centering
        \includegraphics[width=\textwidth]{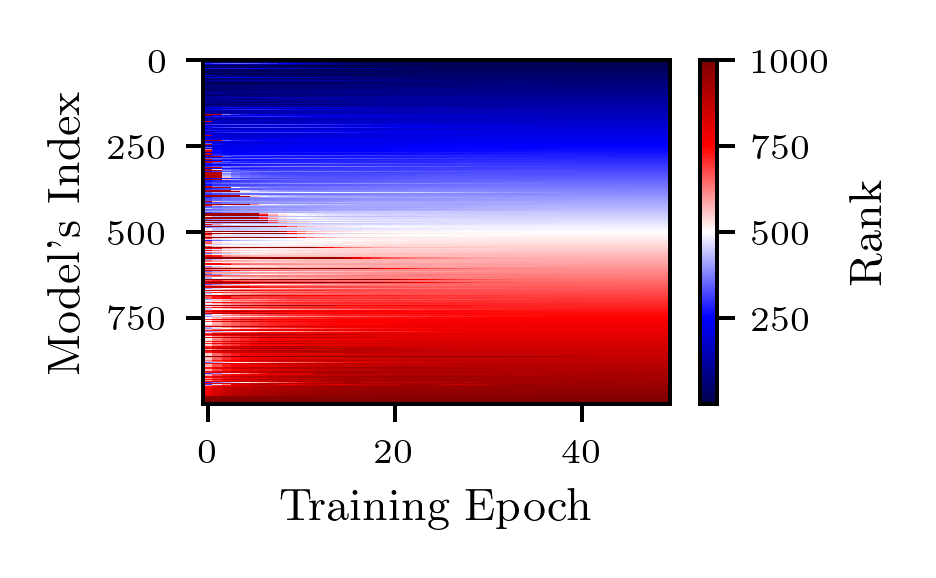}
        \caption{cnae 9}
        \label{fig:heatmap-top10-cnae-9}
    \end{subfigure}
    \begin{subfigure}[b]{0.24\textwidth}
        \centering
        \includegraphics[width=\textwidth]{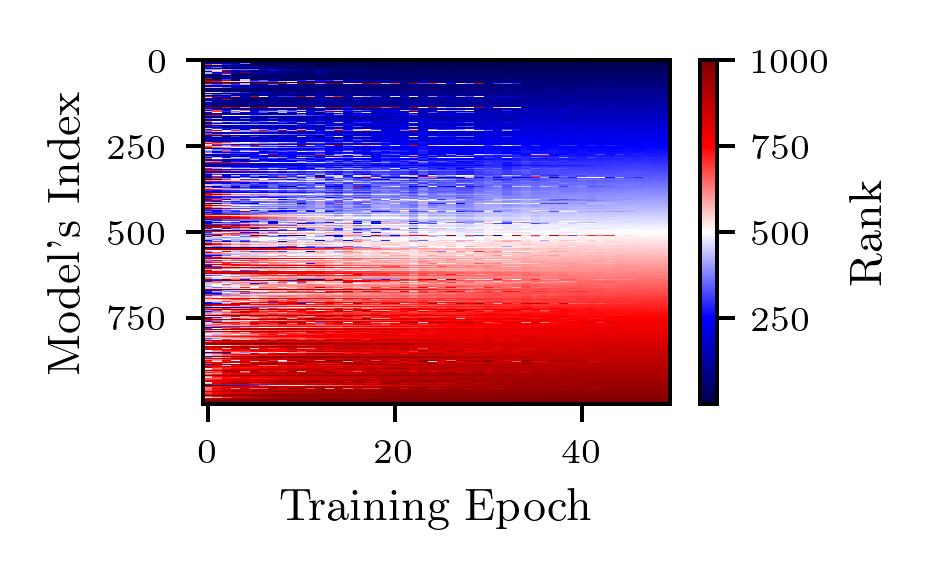}
        \caption{connect 4}
        \label{fig:heatmap-top10-connect-4}
    \end{subfigure}
    \begin{subfigure}[b]{0.24\textwidth}
        \centering
        \includegraphics[width=\textwidth]{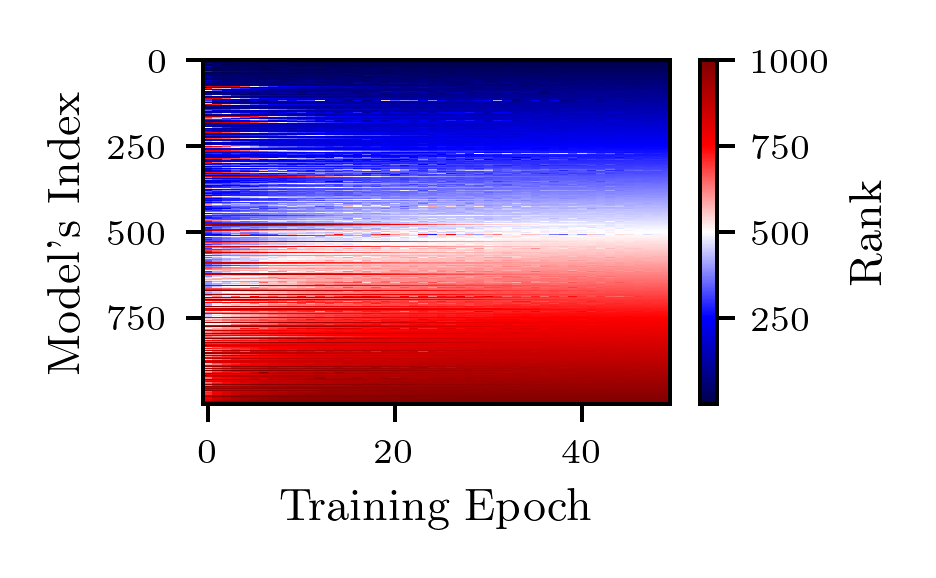}
        \caption{covertype}
        \label{fig:heatmap-top10-covertype}
    \end{subfigure}
    \begin{subfigure}[b]{0.24\textwidth}
        \centering
        \includegraphics[width=\textwidth]{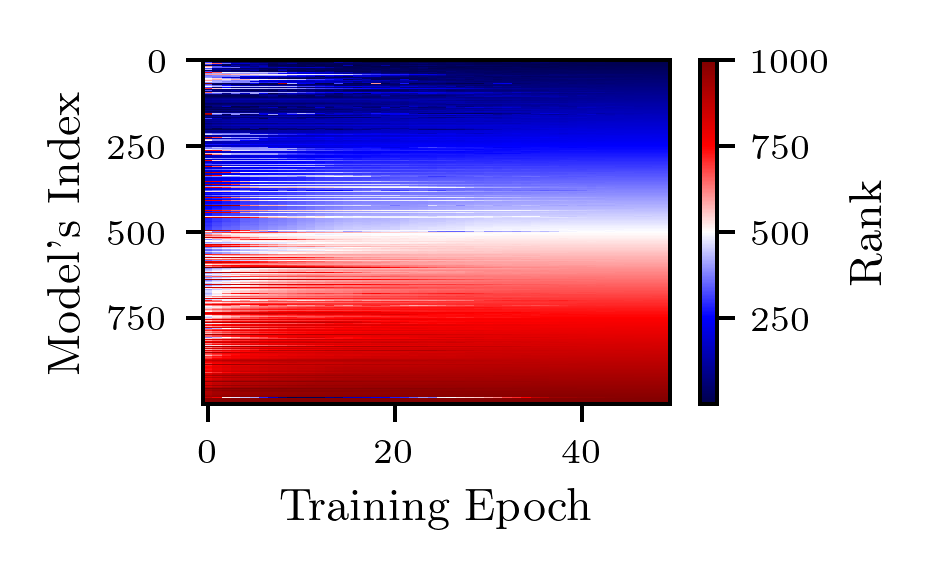}
        \caption{credit g}
        \label{fig:heatmap-top10-credit-g}
    \end{subfigure}
    \begin{subfigure}[b]{0.24\textwidth}
        \centering
        \includegraphics[width=\textwidth]{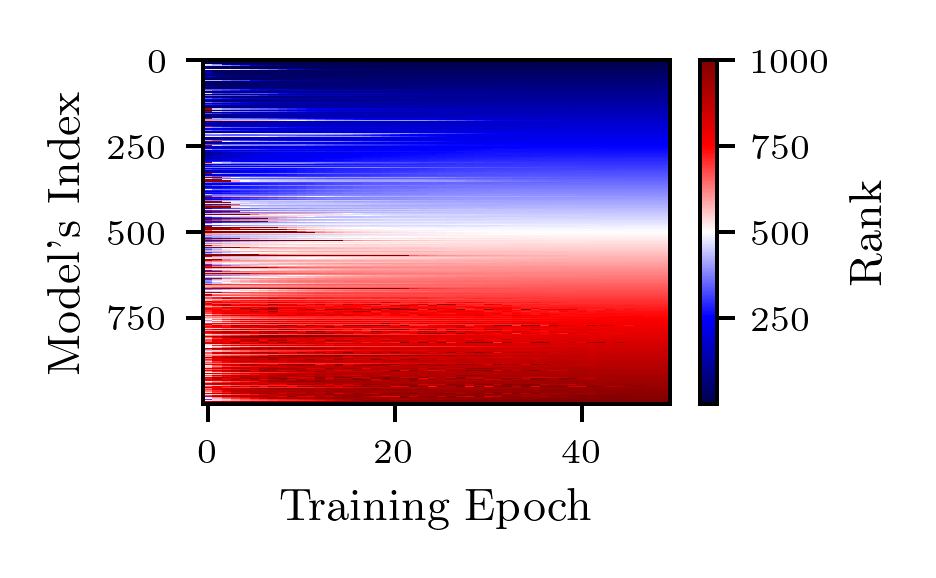}
        \caption{dionis}
        \label{fig:heatmap-top10-dionis}
    \end{subfigure}
    \begin{subfigure}[b]{0.24\textwidth}
        \centering
        \includegraphics[width=\textwidth]{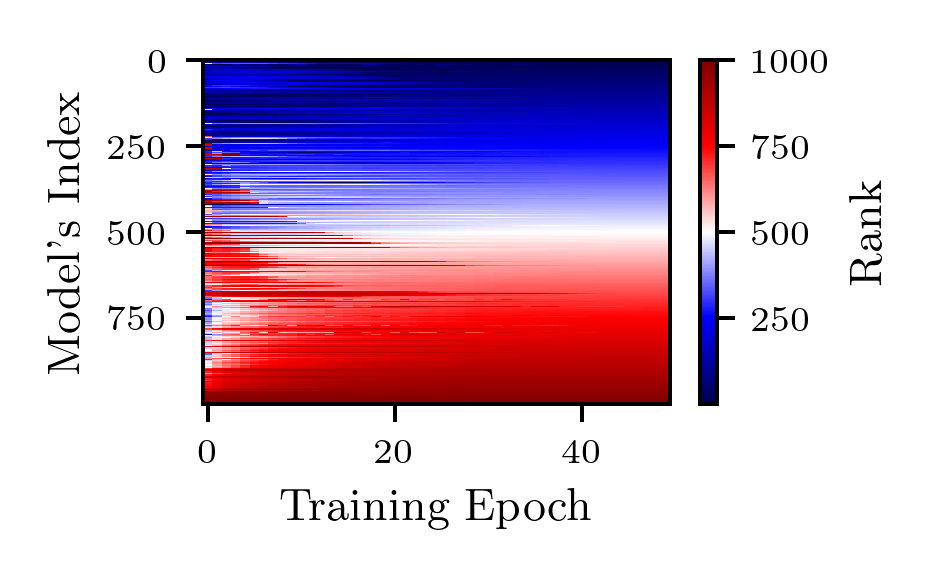}
        \caption{fabert}
        \label{fig:heatmap-top10-fabert}
    \end{subfigure}
    \begin{subfigure}[b]{0.24\textwidth}
        \centering
        \includegraphics[width=\textwidth]{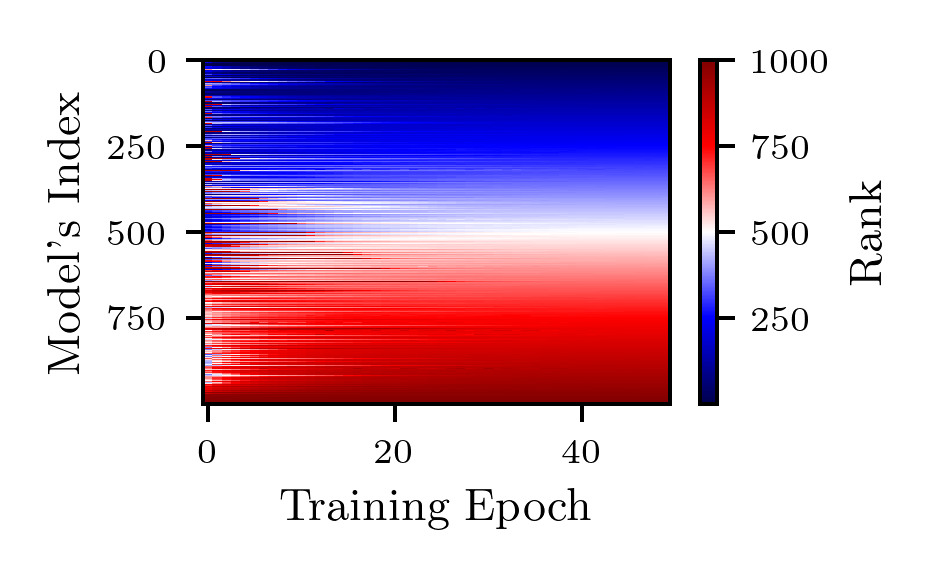}
        \caption{helena}
        \label{fig:heatmap-top10-helena}
    \end{subfigure}
    \begin{subfigure}[b]{0.24\textwidth}
        \centering
        \includegraphics[width=\textwidth]{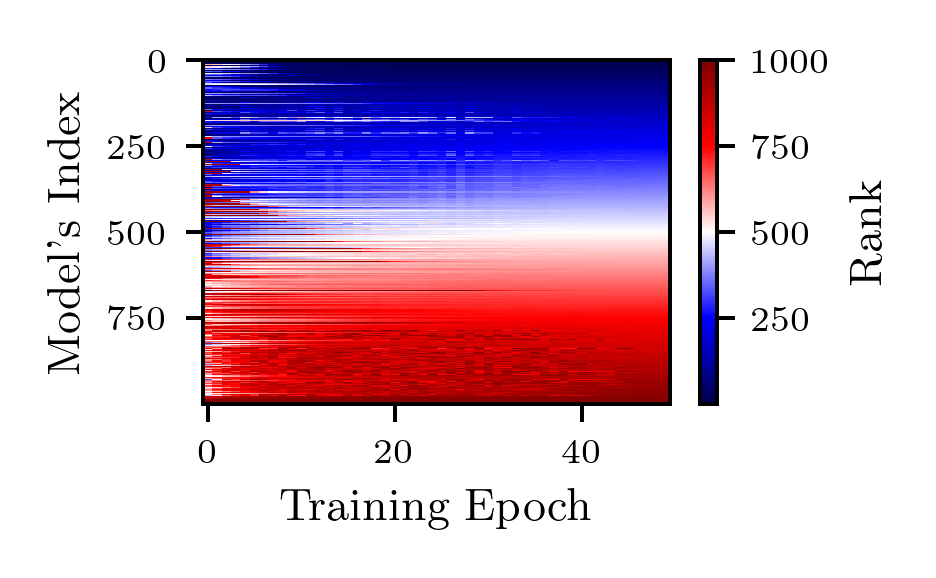}
        \caption{higgs}
        \label{fig:heatmap-top10-higgs}
    \end{subfigure}
    \begin{subfigure}[b]{0.24\textwidth}
        \centering
        \includegraphics[width=\textwidth]{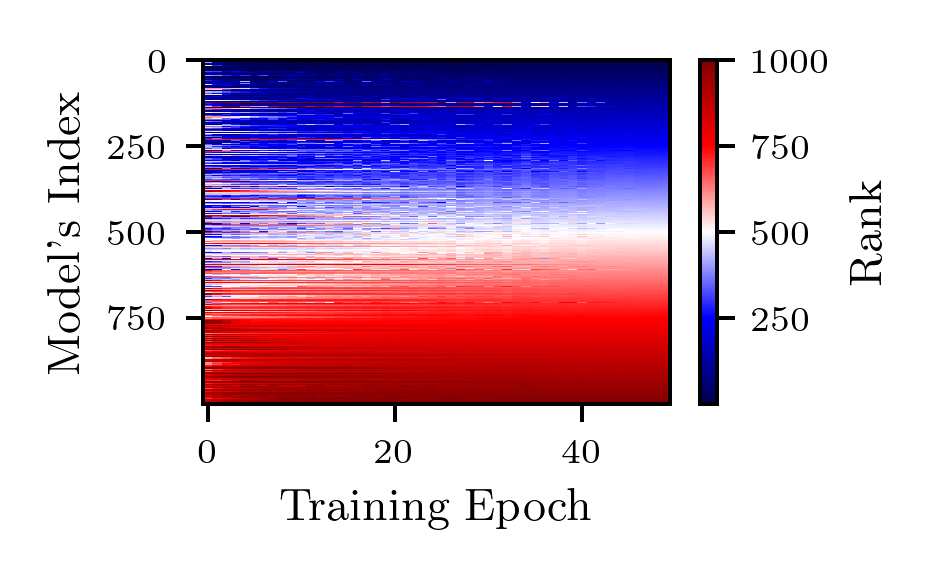}
        \caption{jannis}
        \label{fig:heatmap-top10-jannis}
    \end{subfigure}
    \begin{subfigure}[b]{0.24\textwidth}
        \centering
        \includegraphics[width=\textwidth]{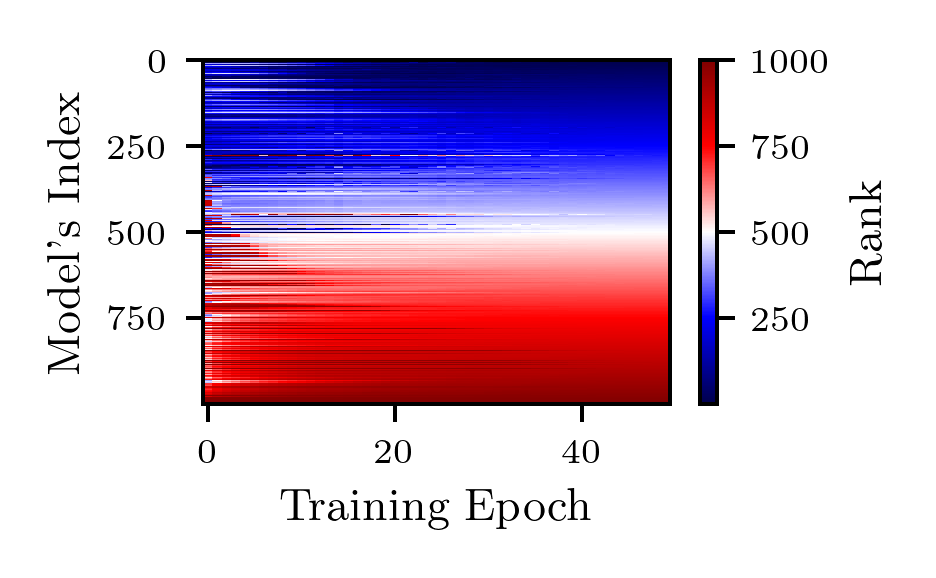}
        \caption{jasmine}
        \label{fig:heatmap-top10-jasmine}
    \end{subfigure}
    \begin{subfigure}[b]{0.24\textwidth}
        \centering
        \includegraphics[width=\textwidth]{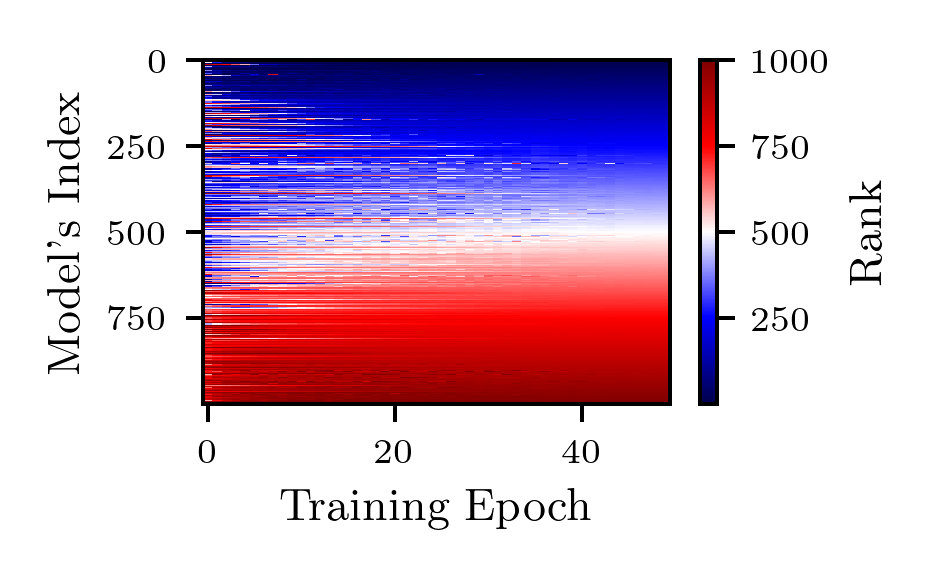}
        \caption{jungle chess 2pcs raw endgame complete}
        \label{fig:heatmap-top10-jungle_chess_2pcs_raw_endgame_complete}
    \end{subfigure}
    \begin{subfigure}[b]{0.24\textwidth}
        \centering
        \includegraphics[width=\textwidth]{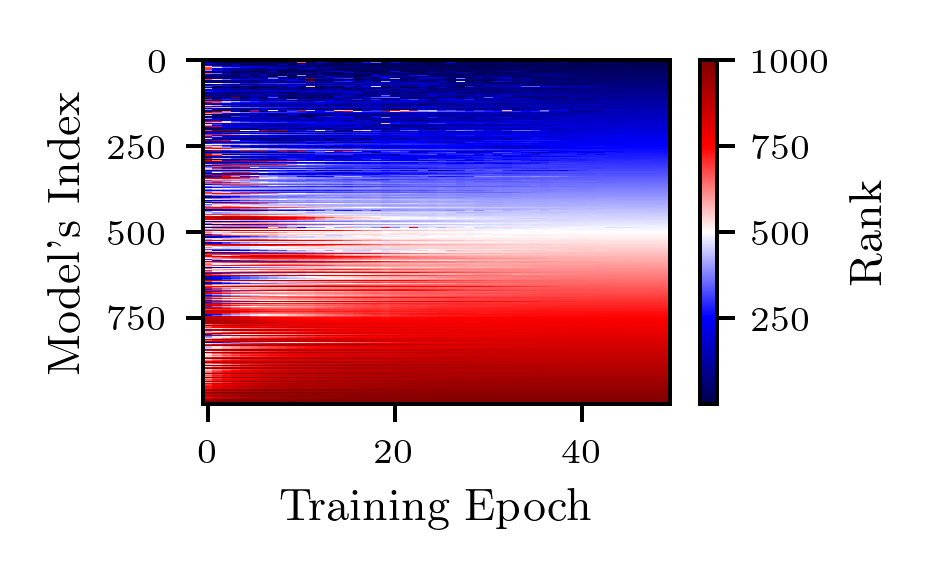}
        \caption{kc1}
        \label{fig:heatmap-top10-kc1}
    \end{subfigure}
    \begin{subfigure}[b]{0.24\textwidth}
        \centering
        \includegraphics[width=\textwidth]{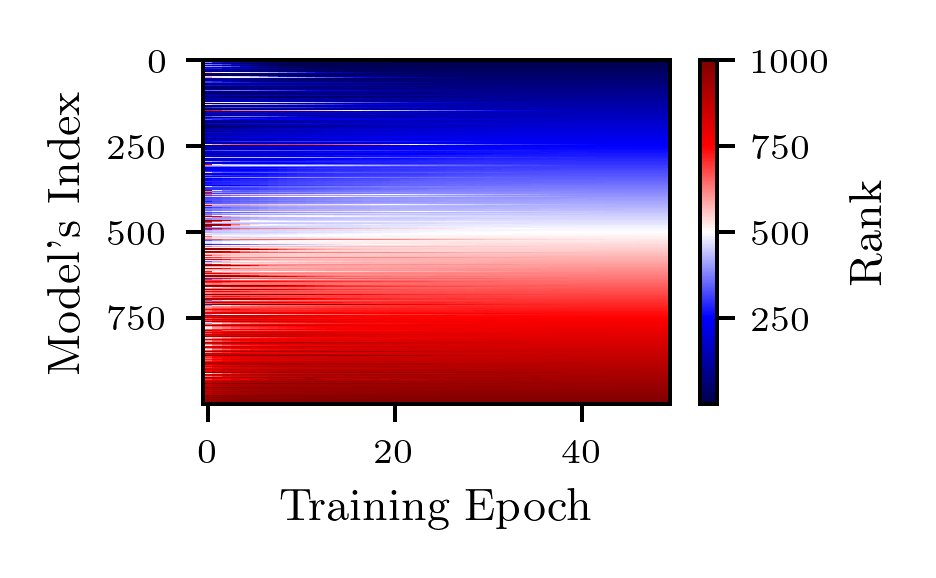}
        \caption{kr vs kp}
        \label{fig:heatmap-top10-kr-vs-kp}
    \end{subfigure}
  
    \caption{Evolution of the ranking from minimal to maximal fidelity on 1,000 randomly sampled models from the search space of LCBench ({\bf First Part}).}
    \label{fig:all-heatmap-lcbench}
\end{figure}

\begin{figure}[!h]
    \centering

    \begin{subfigure}[b]{0.24\textwidth}
        \centering
        \includegraphics[width=\textwidth]{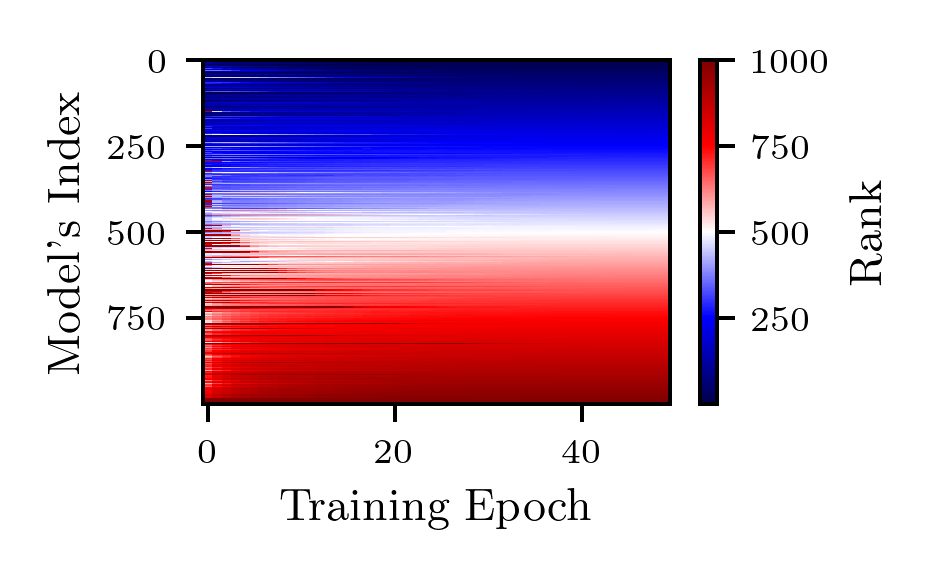}
        \caption{mfeat factors}
        \label{fig:heatmap-top10-mfeat-factors}
    \end{subfigure}
    \begin{subfigure}[b]{0.24\textwidth}
        \centering
        \includegraphics[width=\textwidth]{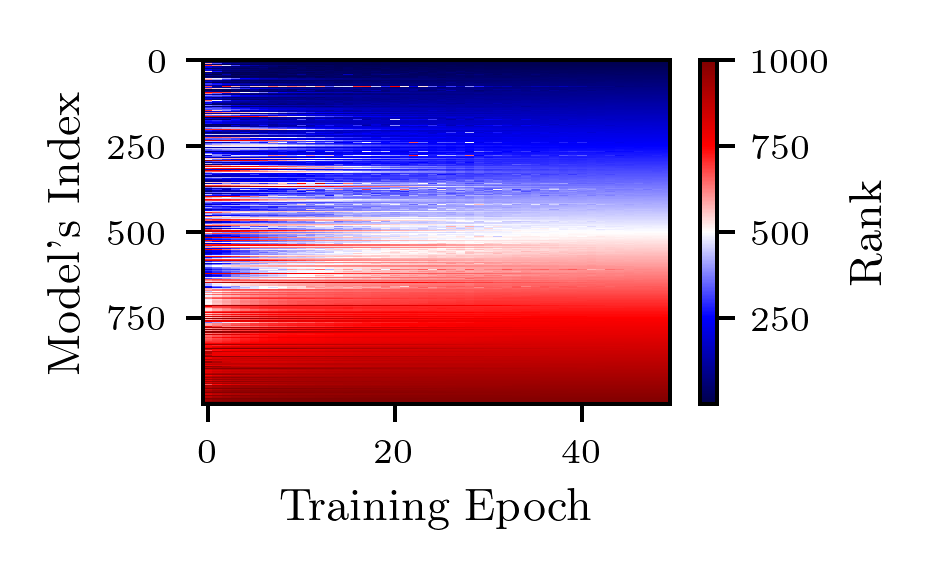}
        \caption{nomao}
        \label{fig:heatmap-top10-nomao}
    \end{subfigure}
    \begin{subfigure}[b]{0.24\textwidth}
        \centering
        \includegraphics[width=\textwidth]{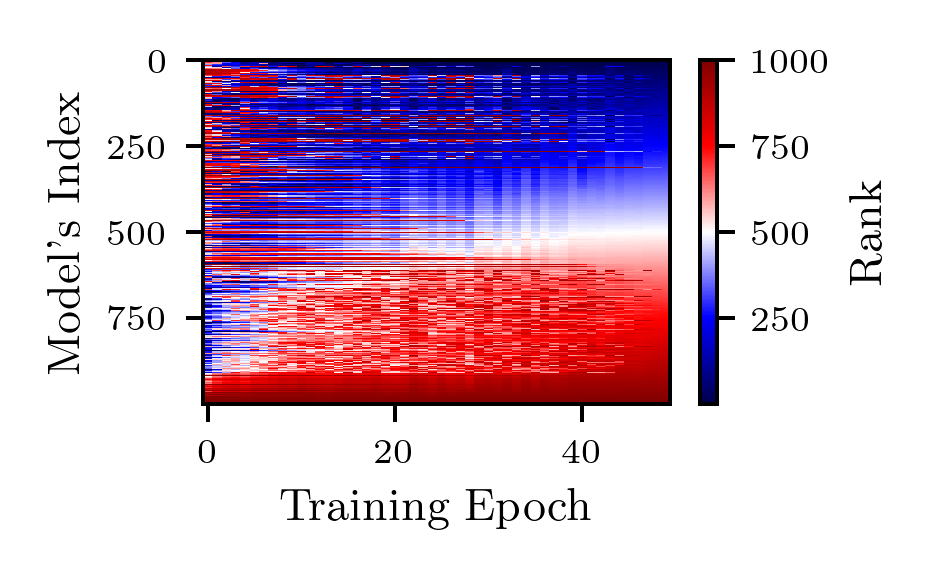}
        \caption{numerai28.6}
        \label{fig:heatmap-top10-numerai28.6}
    \end{subfigure}
    \begin{subfigure}[b]{0.24\textwidth}
        \centering
        \includegraphics[width=\textwidth]{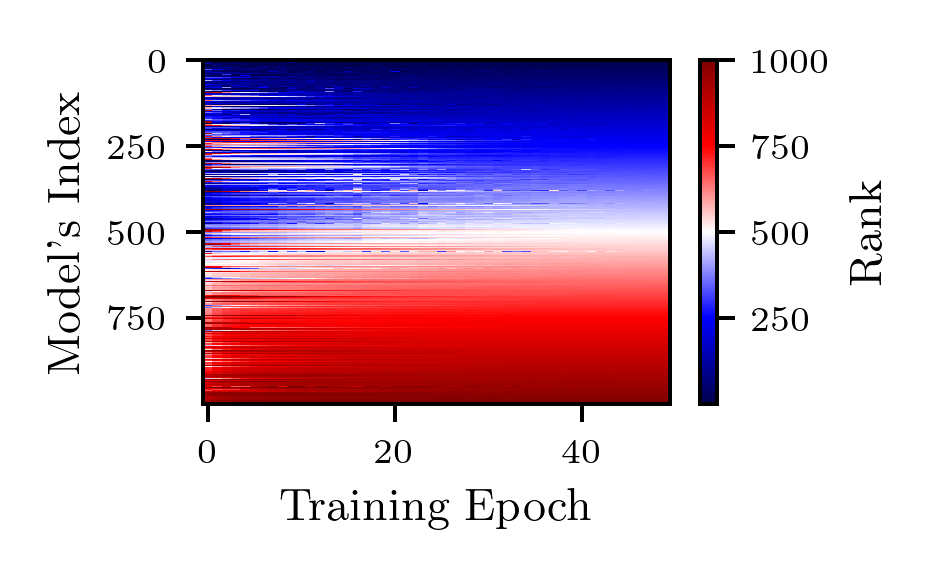}
        \caption{phoneme}
        \label{fig:heatmap-top10-phoneme}
    \end{subfigure}
    \begin{subfigure}[b]{0.24\textwidth}
        \centering
        \includegraphics[width=\textwidth]{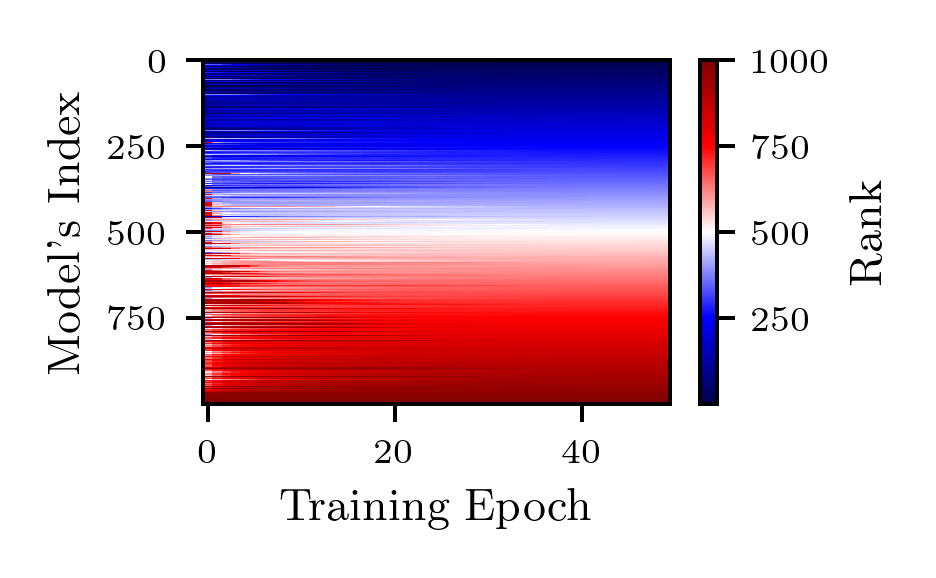}
        \caption{segment}
        \label{fig:heatmap-top10-segment}
    \end{subfigure}
    \begin{subfigure}[b]{0.24\textwidth}
        \centering
        \includegraphics[width=\textwidth]{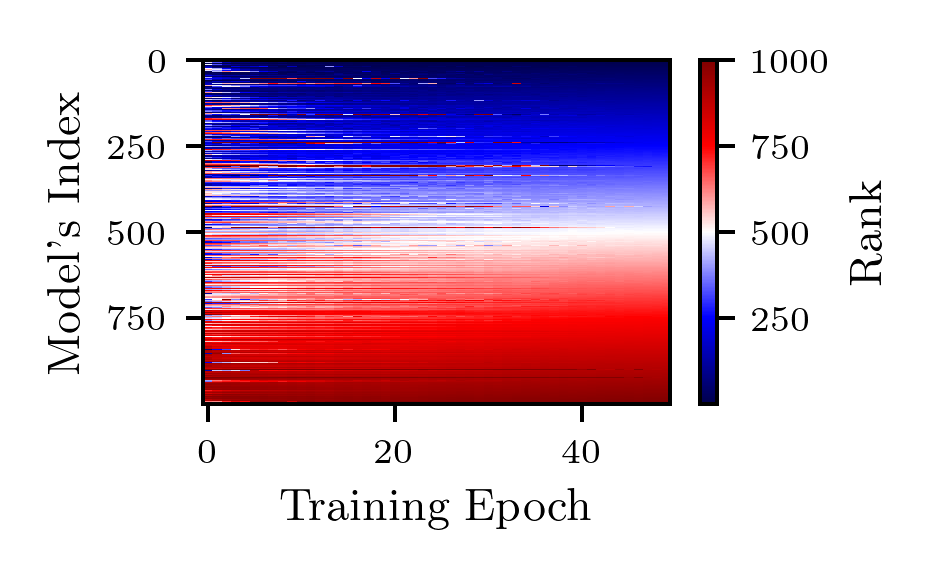}
        \caption{shuttle}
        \label{fig:heatmap-top10-shuttle}
    \end{subfigure}
    \begin{subfigure}[b]{0.24\textwidth}
        \centering
        \includegraphics[width=\textwidth]{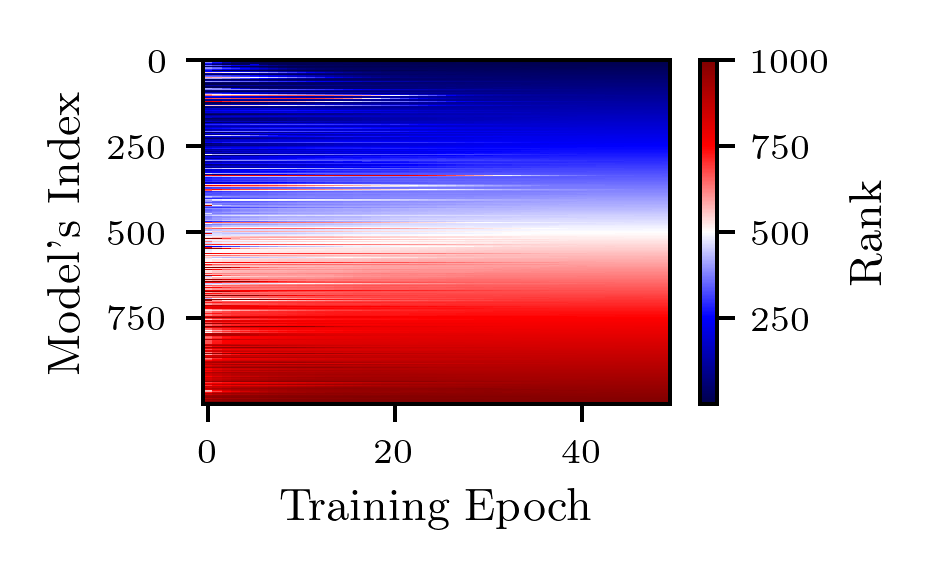}
        \caption{sylvine}
        \label{fig:heatmap-top10-sylvine}
    \end{subfigure}
    \begin{subfigure}[b]{0.24\textwidth}
        \centering
        \includegraphics[width=\textwidth]{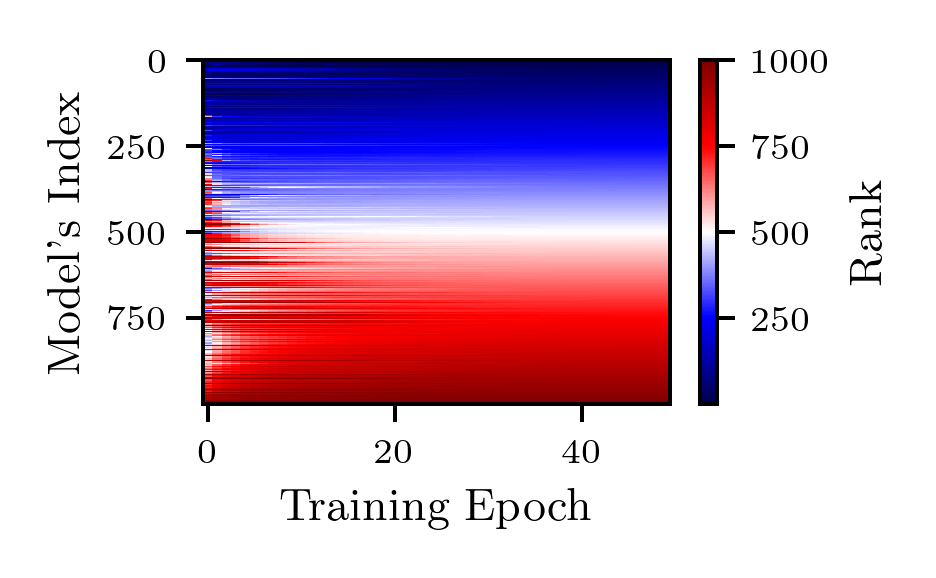}
        \caption{vehicle}
        \label{fig:heatmap-top10-vehicle}
    \end{subfigure}
    \begin{subfigure}[b]{0.24\textwidth}
        \centering
        \includegraphics[width=\textwidth]{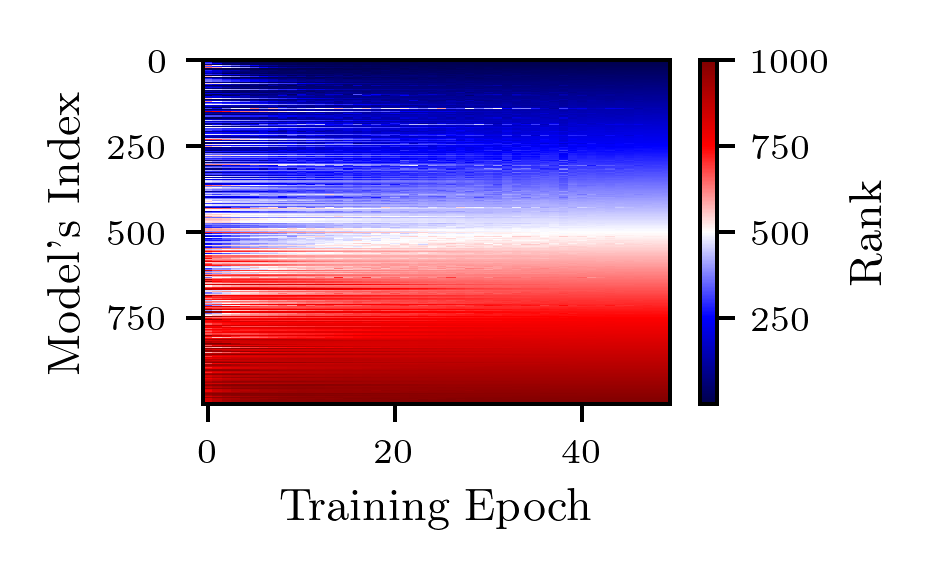}
        \caption{volkert}
        \label{fig:heatmap-top10-volkert}
    \end{subfigure}
    
    \caption{Evolution of the ranking from minimal to maximal fidelity on 1,000 randomly sampled models from the search space of LCBench ({\bf Second Part}).}
    \label{fig:all-heatmap-lcbench-second-part}
\end{figure}

Similarly to what is done on HPOBench in the previous section we perform the same visualization now on LCBench~\cite{klein_tabular_2019}. We present 1,000 randomly selected learning curves colored by their final ranking in Figure~\ref{fig:all-ranking-curves-lcbench},\ref{fig:all-ranking-curves-lcbench-second-part} and the corresponding heatmaps in Figure~\ref{fig:all-heatmap-lcbench},\ref{fig:all-heatmap-lcbench-second-part}. In such visualization, we can identify clearly two groups among learning curves (good and bad candidates). Also we can notice that it is possible to identify good candidates only based on the first epoch.

\section{Visualization of Learning Curves from YAHPO}~\label{sec:appendix-viz-learning-curves-YAHPO}

\begin{figure}[!h]
    \centering

    \begin{subfigure}[b]{0.24\textwidth}
        \centering
        \includegraphics[width=\textwidth]{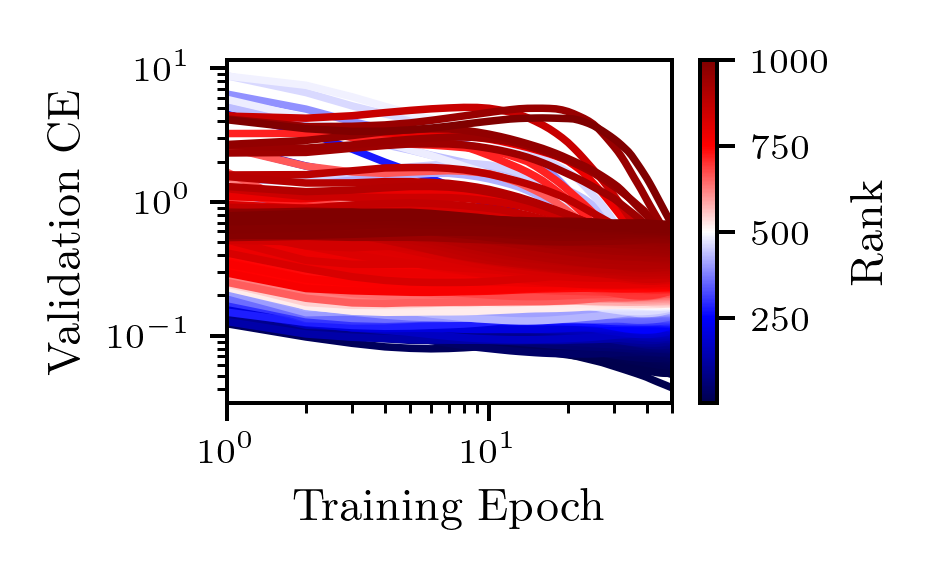}
        \caption{APSFailure}
        \label{fig:ranking-curves-top10-APSFailure}
    \end{subfigure}
    \begin{subfigure}[b]{0.24\textwidth}
        \centering
        \includegraphics[width=\textwidth]{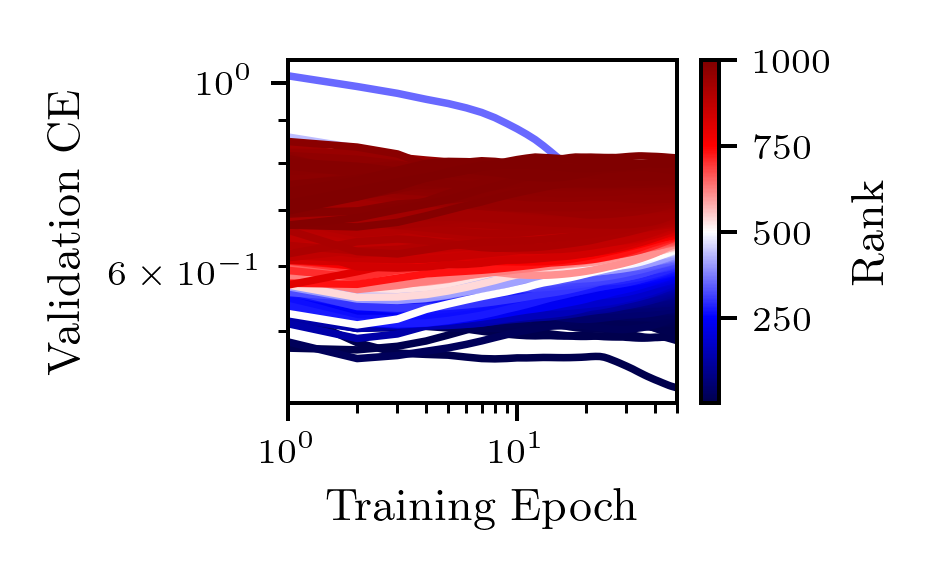}
        \caption{Amazon employee access}
        \label{fig:ranking-curves-top10-Amazon_employee_access}
    \end{subfigure}
    \begin{subfigure}[b]{0.24\textwidth}
        \centering
        \includegraphics[width=\textwidth]{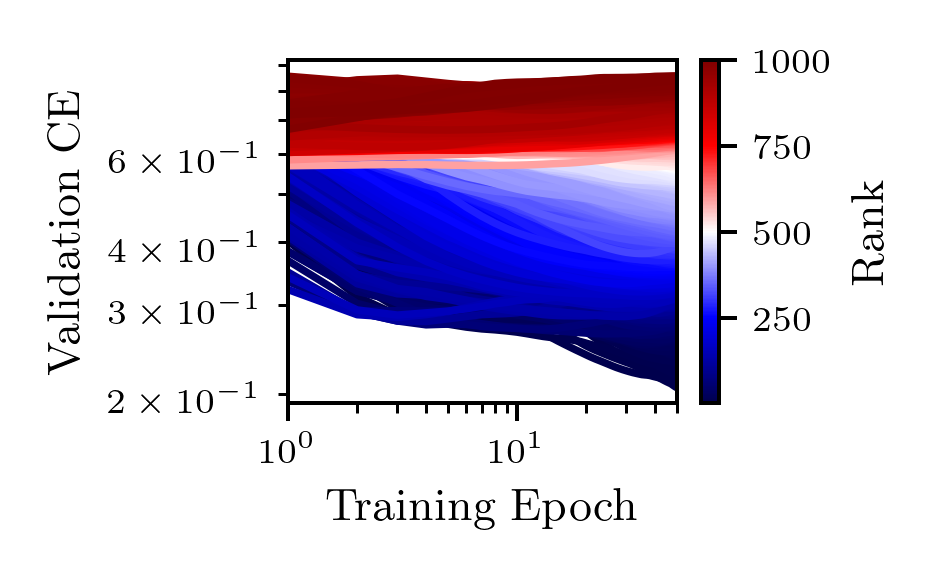}
        \caption{Australian}
        \label{fig:ranking-curves-top10-Australian}
    \end{subfigure}
    \begin{subfigure}[b]{0.24\textwidth}
        \centering
        \includegraphics[width=\textwidth]{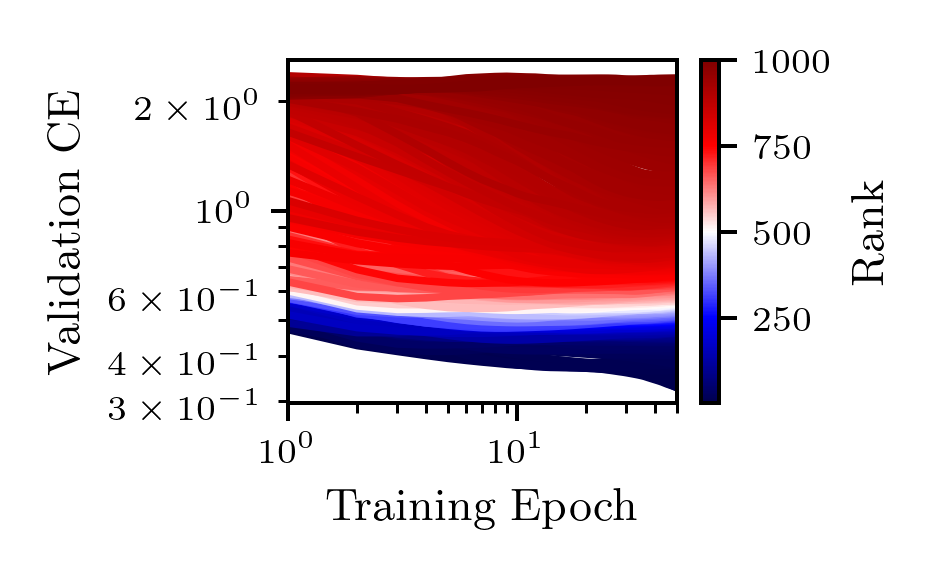}
        \caption{Fashion MNIST}
        \label{fig:ranking-curves-top10-Fashion-MNIST}
    \end{subfigure}
    \begin{subfigure}[b]{0.24\textwidth}
        \centering
        \includegraphics[width=\textwidth]{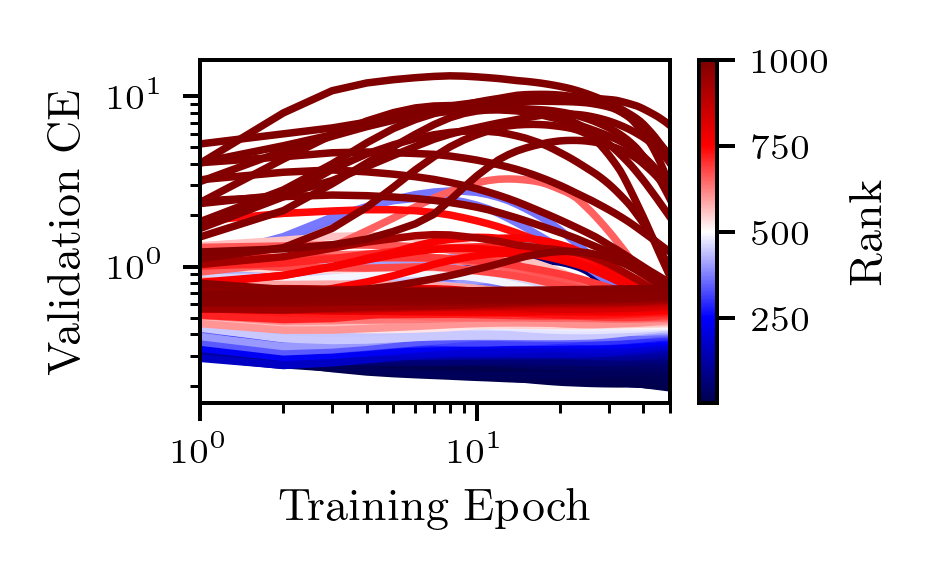}
        \caption{KDDCup09 appetency}
        \label{fig:ranking-curves-top10-KDDCup09_appetency}
    \end{subfigure}
    \begin{subfigure}[b]{0.24\textwidth}
        \centering
        \includegraphics[width=\textwidth]{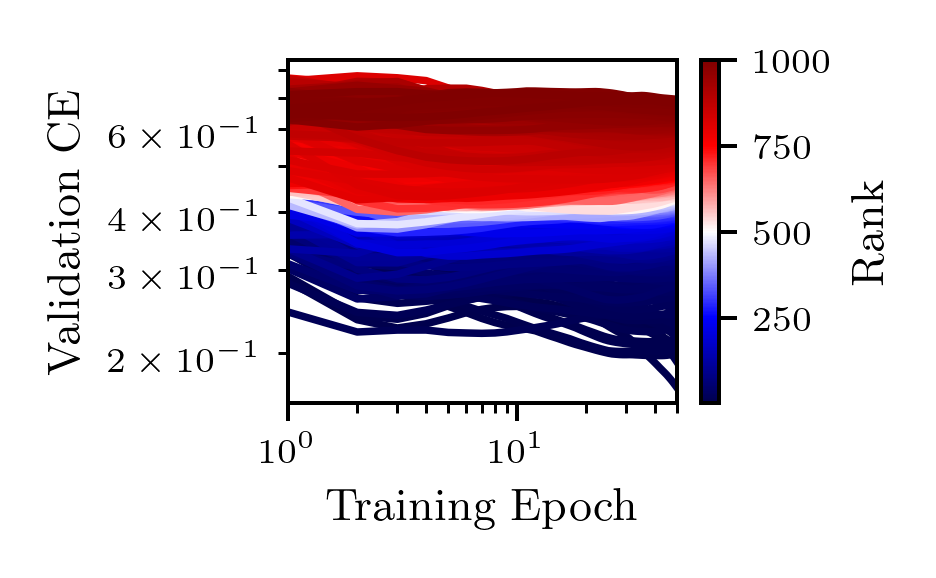}
        \caption{MiniBooNE}
        \label{fig:ranking-curves-top10-MiniBooNE}
    \end{subfigure}
    \begin{subfigure}[b]{0.24\textwidth}
        \centering
        \includegraphics[width=\textwidth]{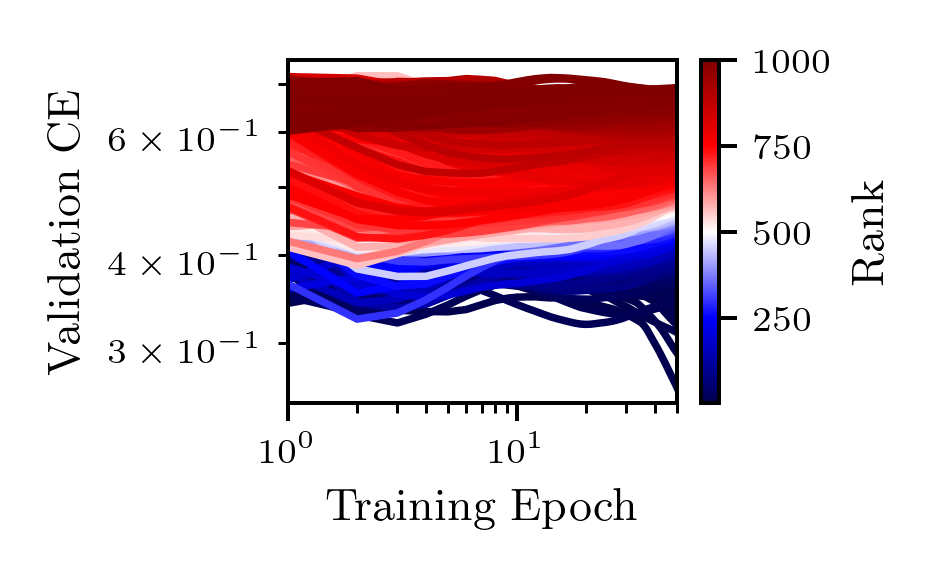}
        \caption{adult}
        \label{fig:ranking-curves-top10-adult}
    \end{subfigure}
    \begin{subfigure}[b]{0.24\textwidth}
        \centering
        \includegraphics[width=\textwidth]{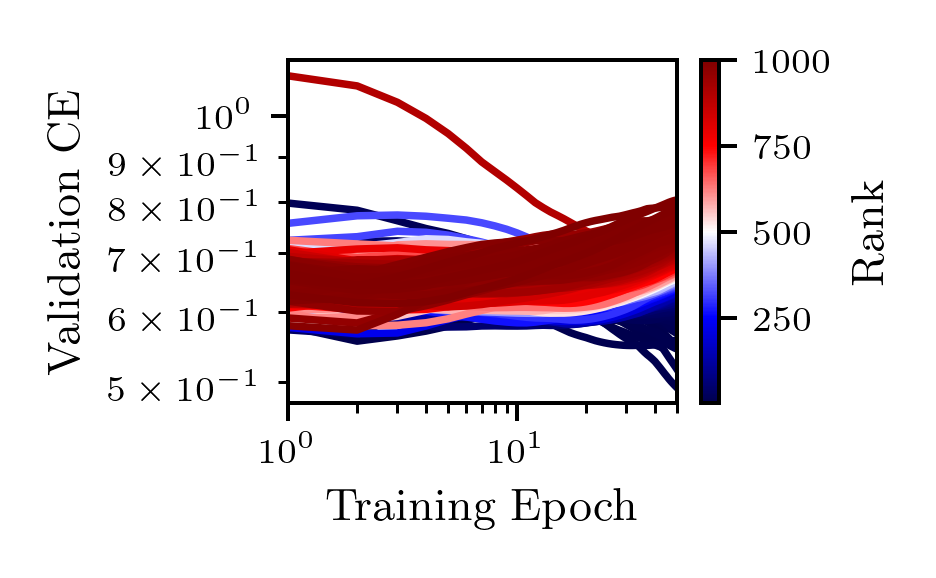}
        \caption{airlines}
        \label{fig:ranking-curves-top10-airlines}
    \end{subfigure}
    \begin{subfigure}[b]{0.24\textwidth}
        \centering
        \includegraphics[width=\textwidth]{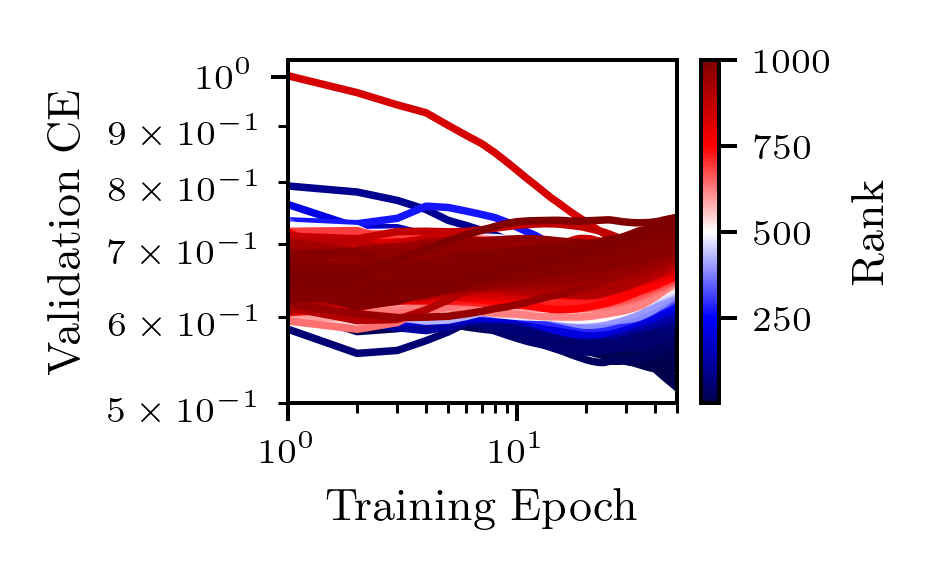}
        \caption{albert}
        \label{fig:ranking-curves-top10-albert}
    \end{subfigure}
    \begin{subfigure}[b]{0.24\textwidth}
        \centering
        \includegraphics[width=\textwidth]{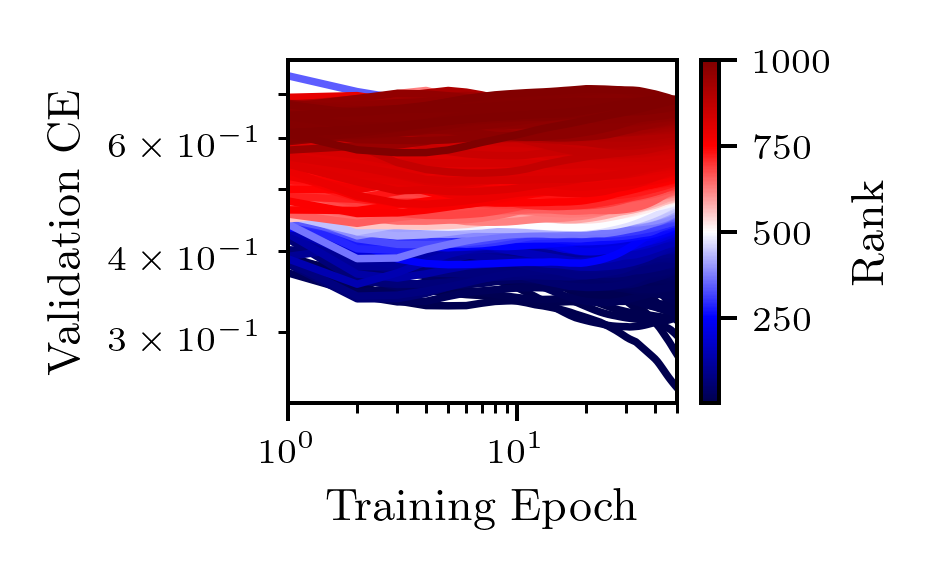}
        \caption{bank marketing}
        \label{fig:ranking-curves-top10-bank-marketing}
    \end{subfigure}
    \begin{subfigure}[b]{0.24\textwidth}
        \centering
        \includegraphics[width=\textwidth]{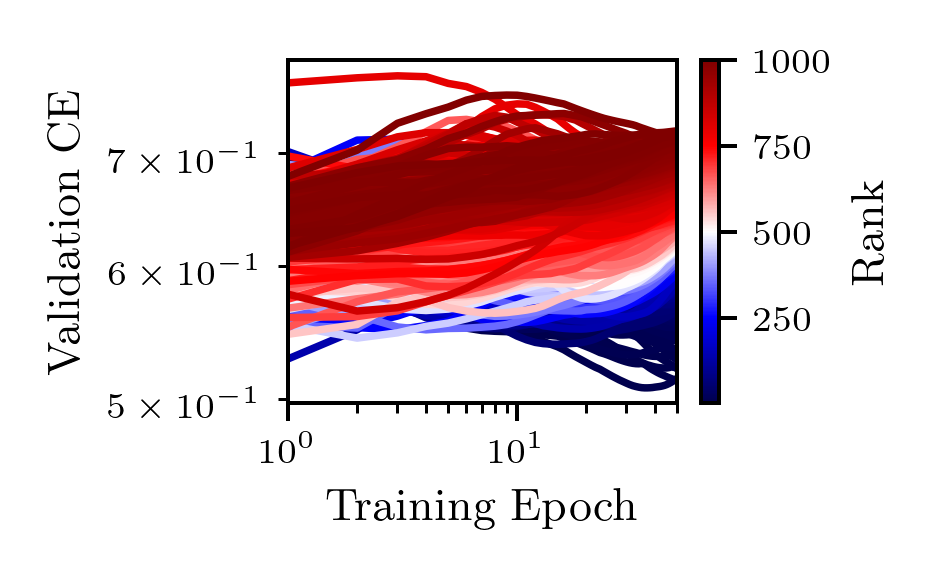}
        \caption{blood transfusion}
        \label{fig:ranking-curves-top10-blood-transfusion-service-center}
    \end{subfigure}
    \begin{subfigure}[b]{0.24\textwidth}
        \centering
        \includegraphics[width=\textwidth]{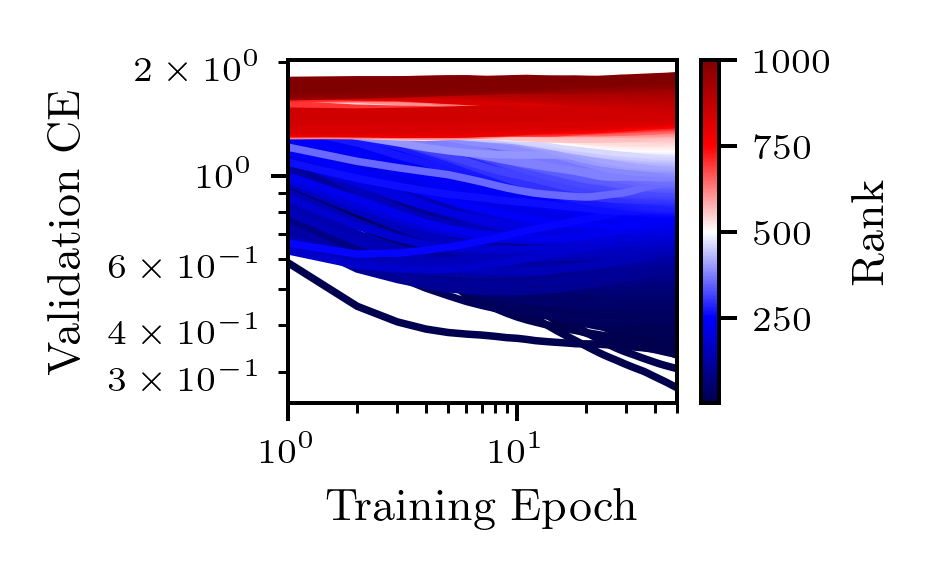}
        \caption{car}
        \label{fig:ranking-curves-top10-car}
    \end{subfigure}
    \begin{subfigure}[b]{0.24\textwidth}
        \centering
        \includegraphics[width=\textwidth]{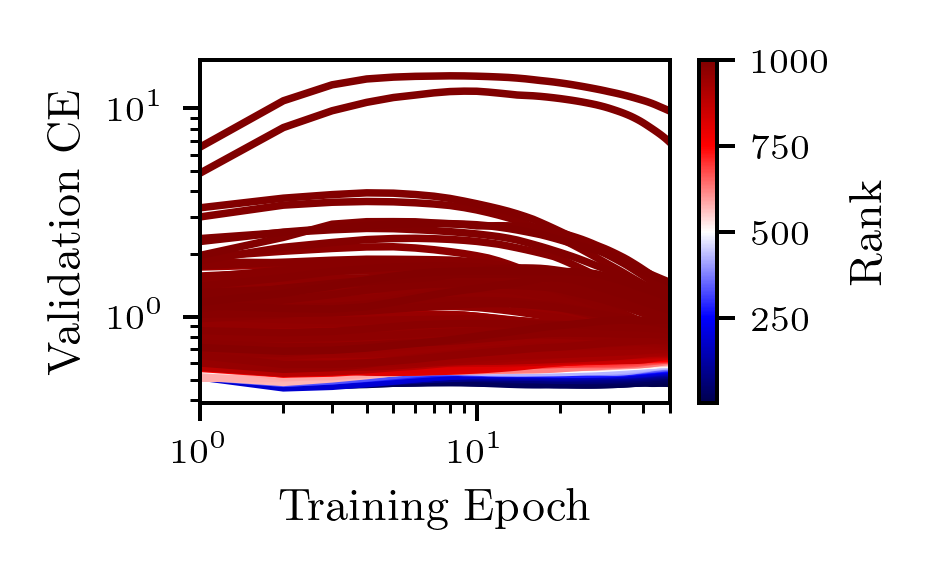}
        \caption{christine}
        \label{fig:ranking-curves-top10-christine}
    \end{subfigure}
    \begin{subfigure}[b]{0.24\textwidth}
        \centering
        \includegraphics[width=\textwidth]{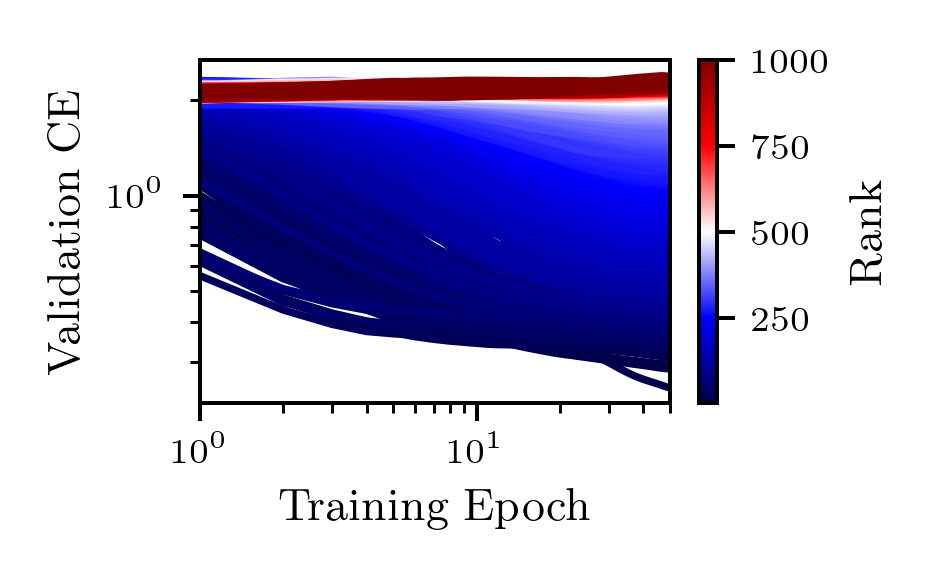}
        \caption{cnae 9}
        \label{fig:ranking-curves-top10-cnae-9}
    \end{subfigure}
    \begin{subfigure}[b]{0.24\textwidth}
        \centering
        \includegraphics[width=\textwidth]{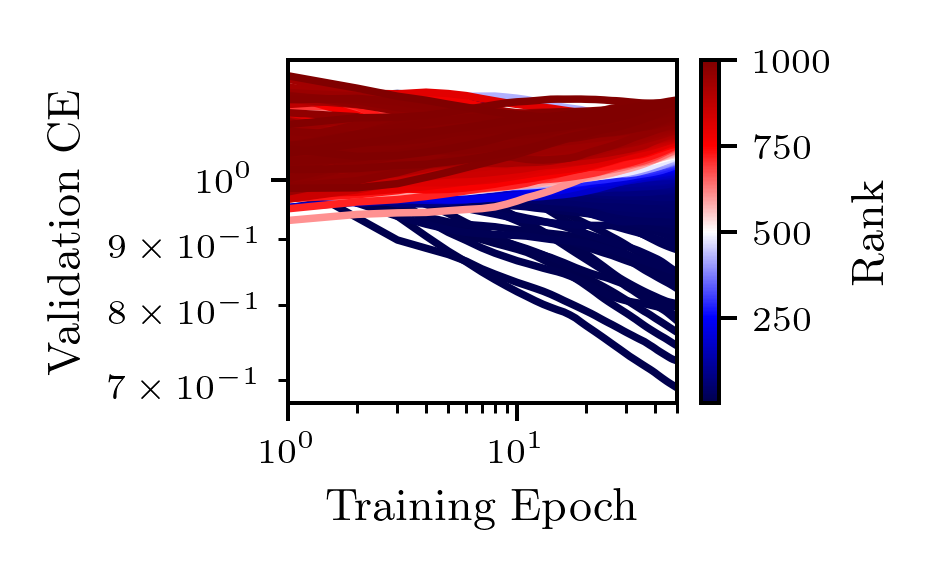}
        \caption{connect 4}
        \label{fig:ranking-curves-top10-connect-4}
    \end{subfigure}
    \begin{subfigure}[b]{0.24\textwidth}
        \centering
        \includegraphics[width=\textwidth]{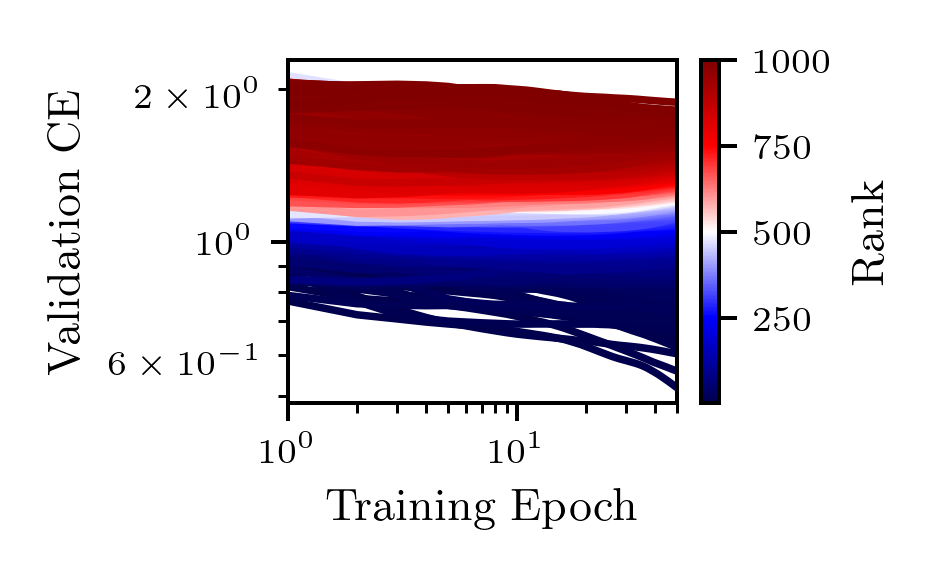}
        \caption{covertype}
        \label{fig:ranking-curves-top10-covertype}
    \end{subfigure}
    \begin{subfigure}[b]{0.24\textwidth}
        \centering
        \includegraphics[width=\textwidth]{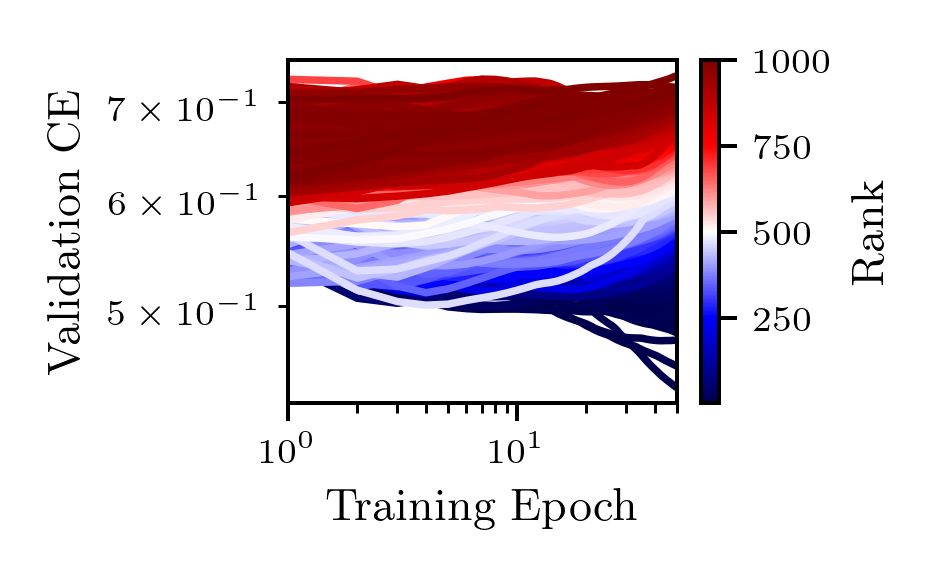}
        \caption{credit g}
        \label{fig:ranking-curves-top10-credit-g}
    \end{subfigure}
    \begin{subfigure}[b]{0.24\textwidth}
        \centering
        \includegraphics[width=\textwidth]{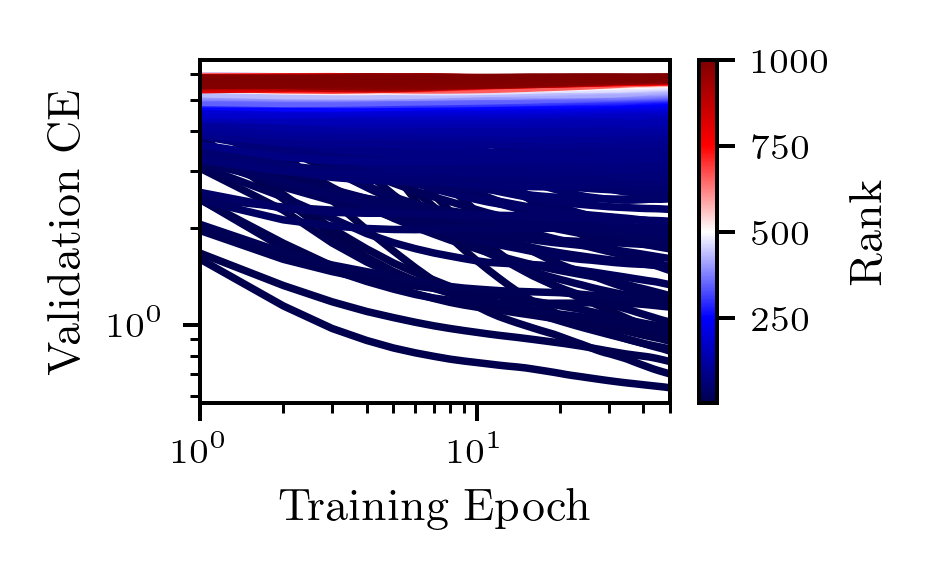}
        \caption{dionis}
        \label{fig:ranking-curves-top10-dionis}
    \end{subfigure}
    \begin{subfigure}[b]{0.24\textwidth}
        \centering
        \includegraphics[width=\textwidth]{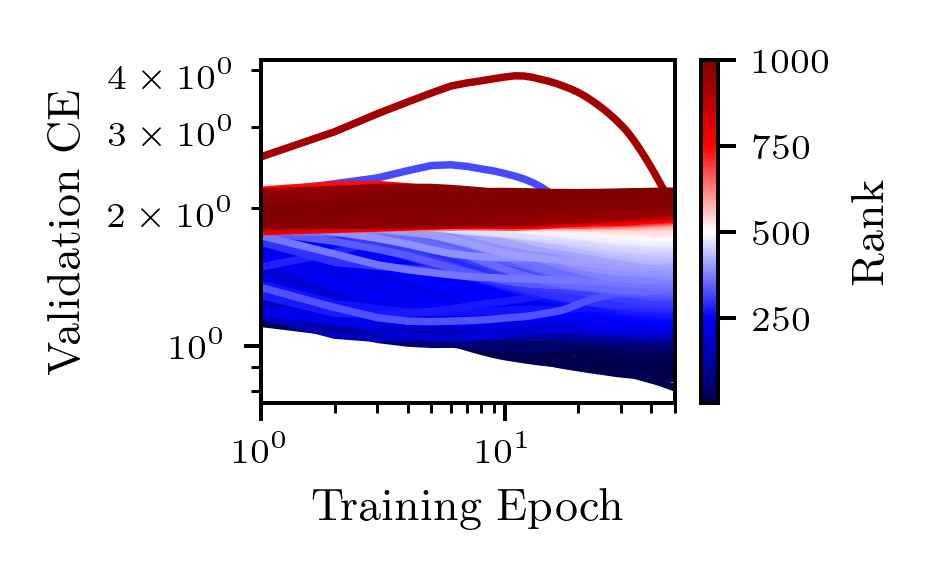}
        \caption{fabert}
        \label{fig:ranking-curves-top10-fabert}
    \end{subfigure}
    \begin{subfigure}[b]{0.24\textwidth}
        \centering
        \includegraphics[width=\textwidth]{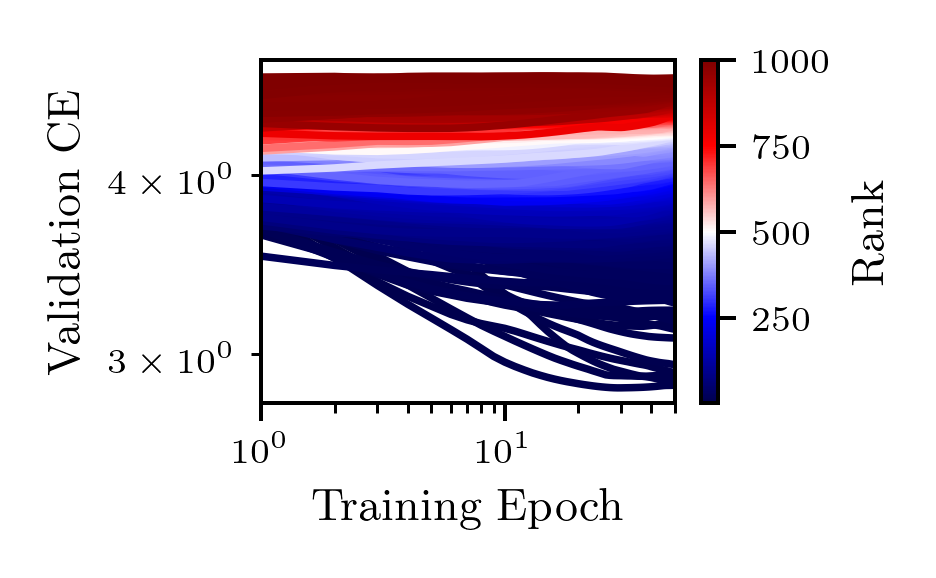}
        \caption{helena}
        \label{fig:ranking-curves-top10-helena}
    \end{subfigure}
    \begin{subfigure}[b]{0.24\textwidth}
        \centering
        \includegraphics[width=\textwidth]{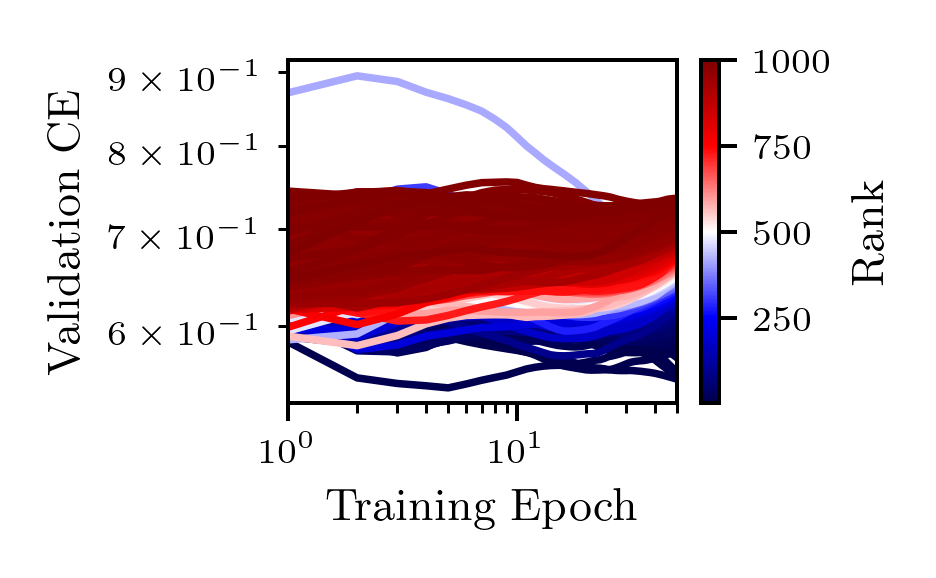}
        \caption{higgs}
        \label{fig:ranking-curves-top10-higgs}
    \end{subfigure}
    \begin{subfigure}[b]{0.24\textwidth}
        \centering
        \includegraphics[width=\textwidth]{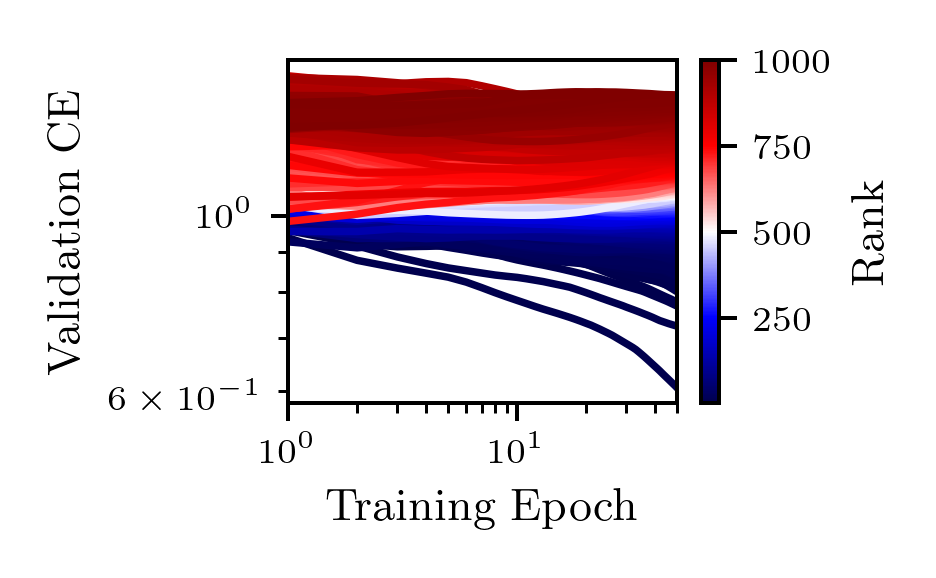}
        \caption{jannis}
        \label{fig:ranking-curves-top10-jannis}
    \end{subfigure}
    \begin{subfigure}[b]{0.24\textwidth}
        \centering
        \includegraphics[width=\textwidth]{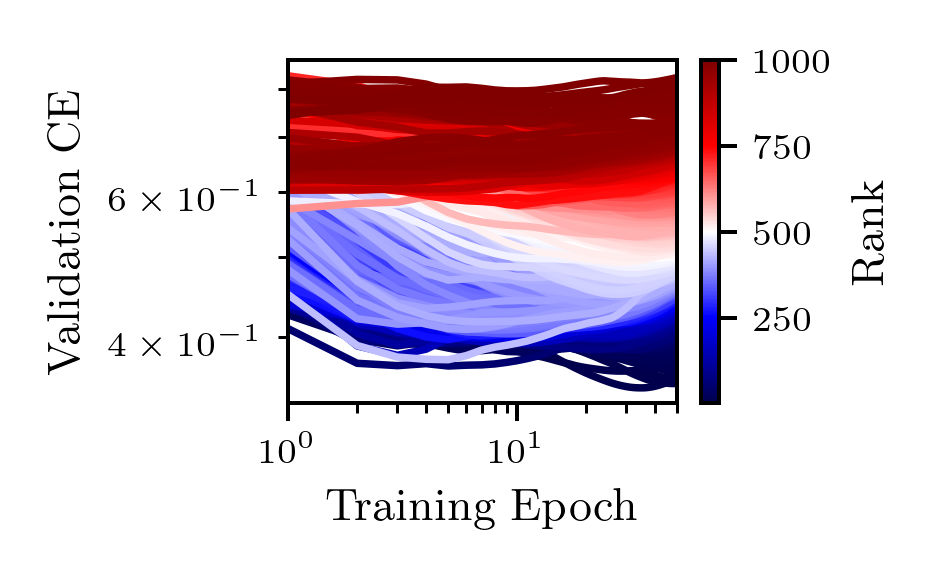}
        \caption{jasmine}
        \label{fig:ranking-curves-top10-jasmine}
    \end{subfigure}
    \begin{subfigure}[b]{0.24\textwidth}
        \centering
        \includegraphics[width=\textwidth]{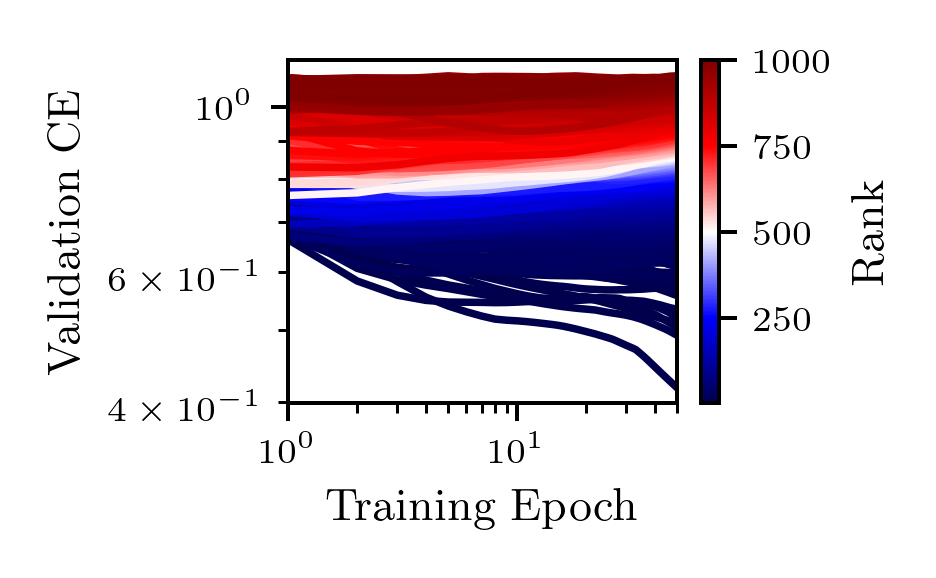}
        \caption{jungle chess 2pcs raw endgame complete}
        \label{fig:ranking-curves-top10-jungle_chess_2pcs_raw_endgame_complete}
    \end{subfigure}
    \begin{subfigure}[b]{0.24\textwidth}
        \centering
        \includegraphics[width=\textwidth]{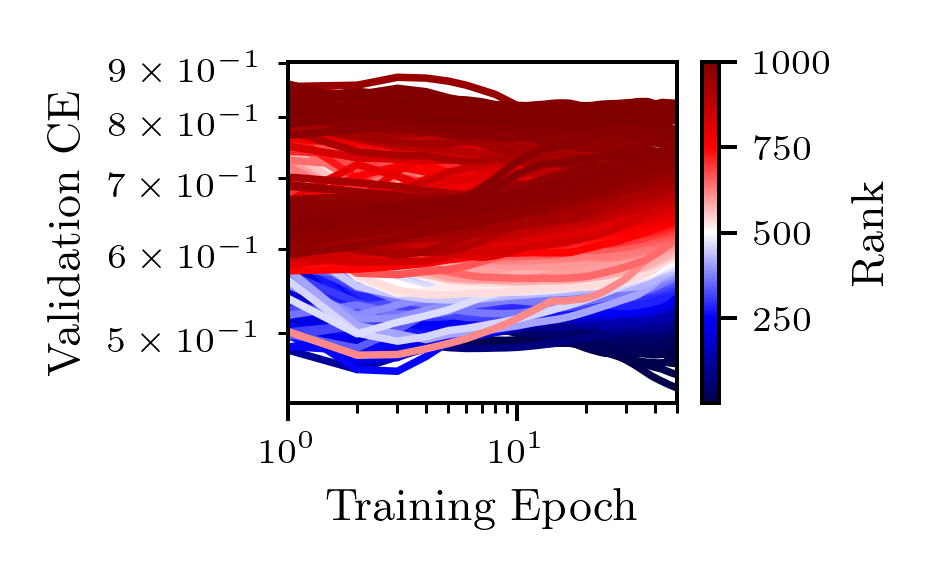}
        \caption{kc1}
        \label{fig:ranking-curves-top10-kc1}
    \end{subfigure}
    \begin{subfigure}[b]{0.24\textwidth}
        \centering
        \includegraphics[width=\textwidth]{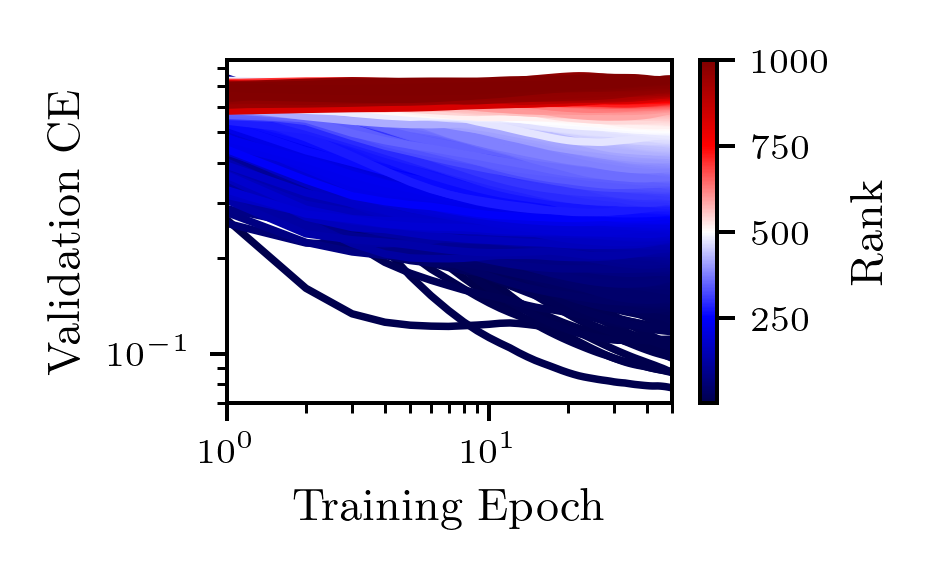}
        \caption{kr vs kp}
        \label{fig:ranking-curves-top10-kr-vs-kp}
    \end{subfigure}
    
    \caption{Learning curves of the 1000 randomly sampled models from LCBench now in YAHPO colored by the rank at maximum fidelity ({\bf First Part}).}
    \label{fig:all-ranking-curves-yahpo}
\end{figure}

\begin{figure}
    \centering
    
    \begin{subfigure}[b]{0.24\textwidth}
        \centering
        \includegraphics[width=\textwidth]{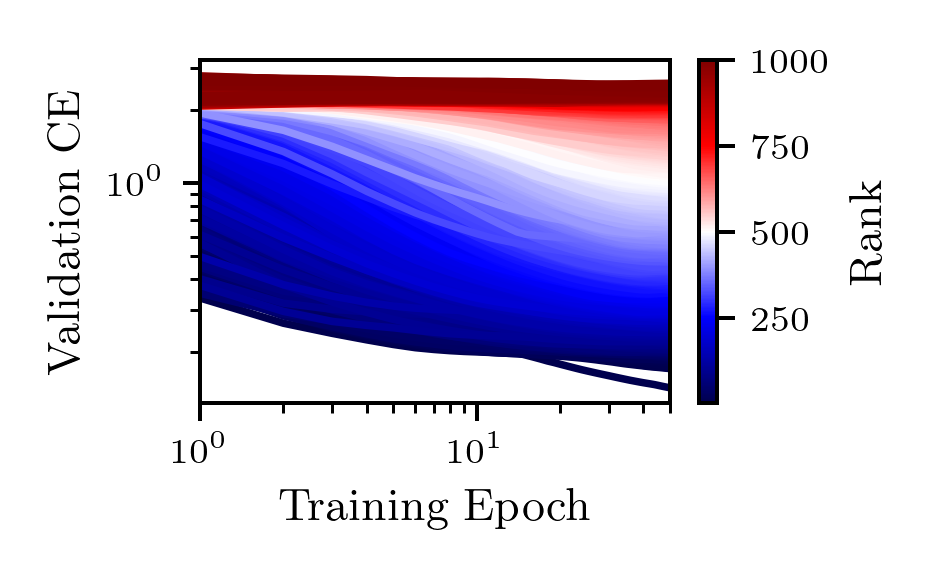}
        \caption{mfeat factors}
        \label{fig:ranking-curves-top10-mfeat-factors}
    \end{subfigure}
    \begin{subfigure}[b]{0.24\textwidth}
        \centering
        \includegraphics[width=\textwidth]{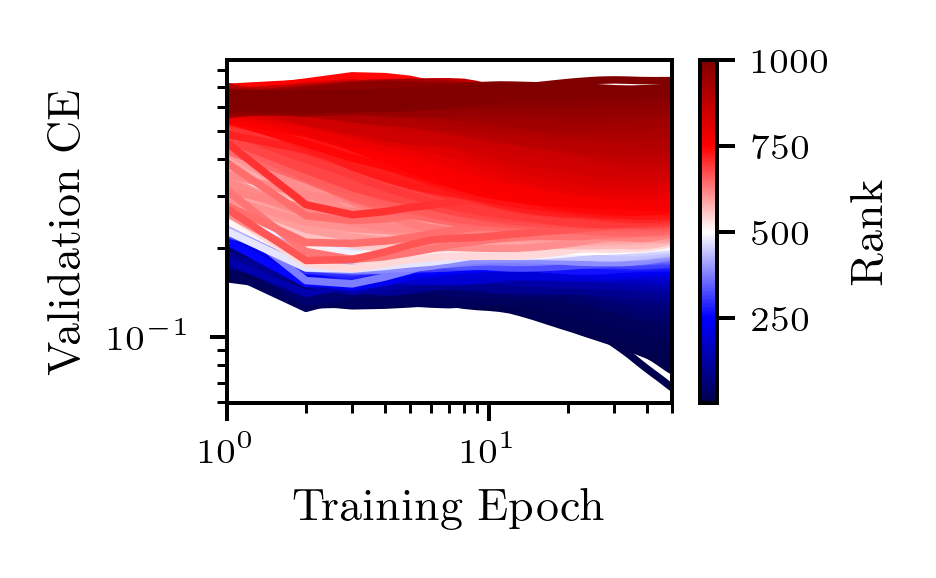}
        \caption{nomao}
        \label{fig:ranking-curves-top10-nomao}
    \end{subfigure}
    \begin{subfigure}[b]{0.24\textwidth}
        \centering
        \includegraphics[width=\textwidth,height=0.618\textwidth]{example-image}
        \caption{numerai28.6}
        \label{fig:ranking-curves-top10-numerai28.6}
    \end{subfigure}
    \begin{subfigure}[b]{0.24\textwidth}
        \centering
        \includegraphics[width=\textwidth]{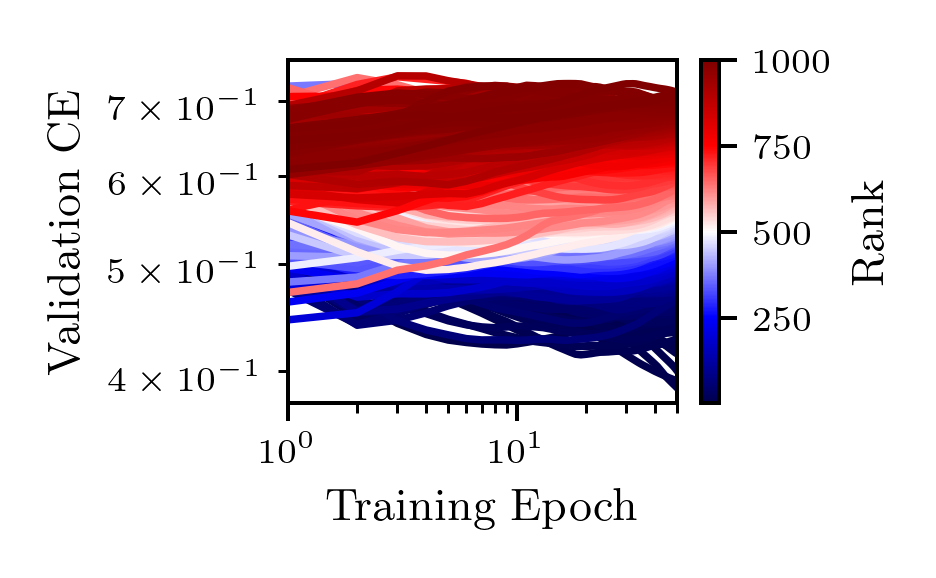}
        \caption{phoneme}
        \label{fig:ranking-curves-top10-phoneme}
    \end{subfigure}
    \begin{subfigure}[b]{0.24\textwidth}
        \centering
        \includegraphics[width=\textwidth]{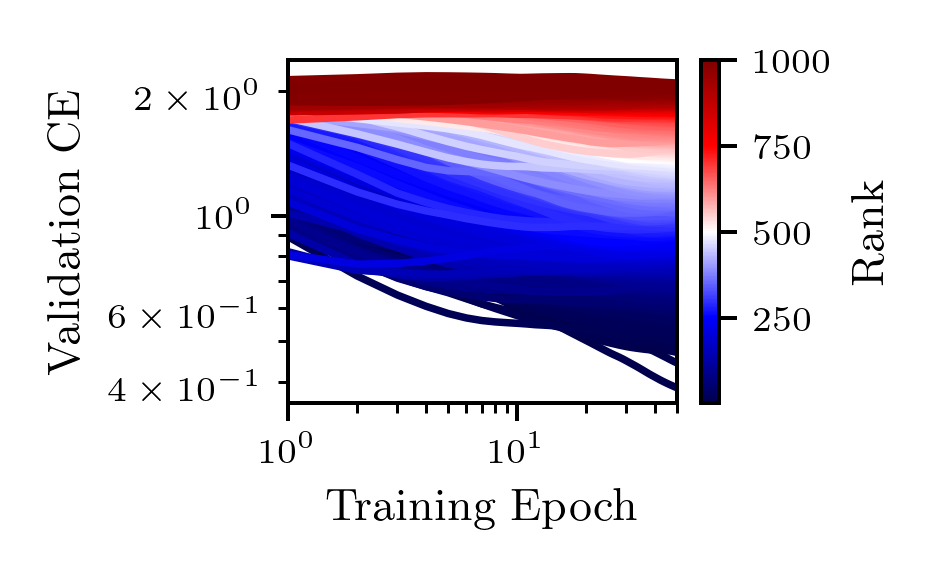}
        \caption{segment}
        \label{fig:ranking-curves-top10-segment}
    \end{subfigure}
    \begin{subfigure}[b]{0.24\textwidth}
        \centering
        \includegraphics[width=\textwidth]{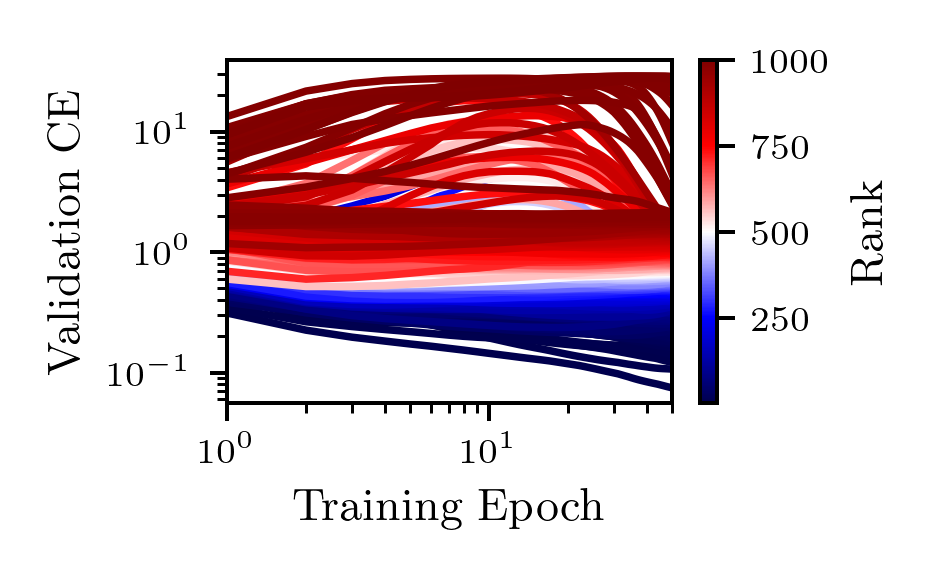}
        \caption{shuttle}
        \label{fig:ranking-curves-top10-shuttle}
    \end{subfigure}
    \begin{subfigure}[b]{0.24\textwidth}
        \centering
        \includegraphics[width=\textwidth]{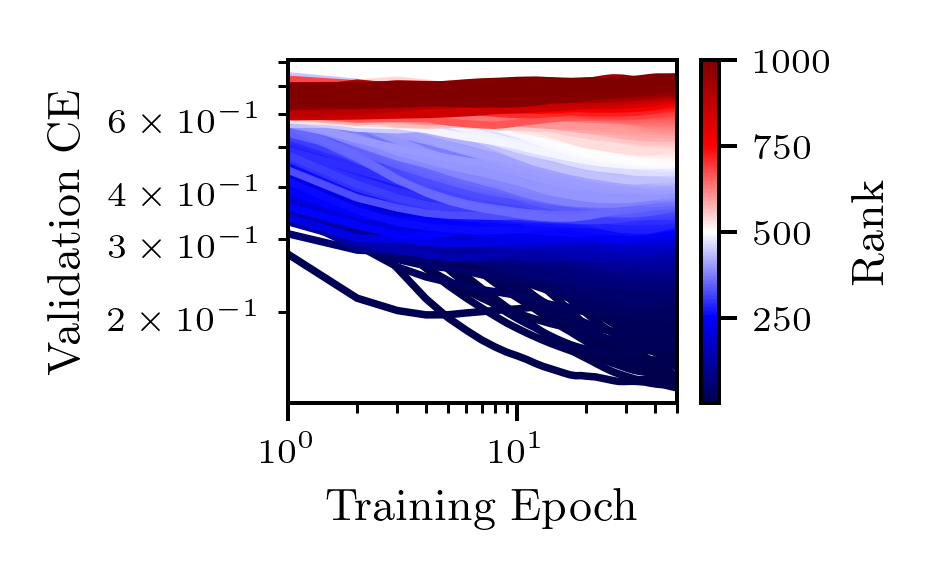}
        \caption{sylvine}
        \label{fig:ranking-curves-top10-sylvine}
    \end{subfigure}
    \begin{subfigure}[b]{0.24\textwidth}
        \centering
        \includegraphics[width=\textwidth]{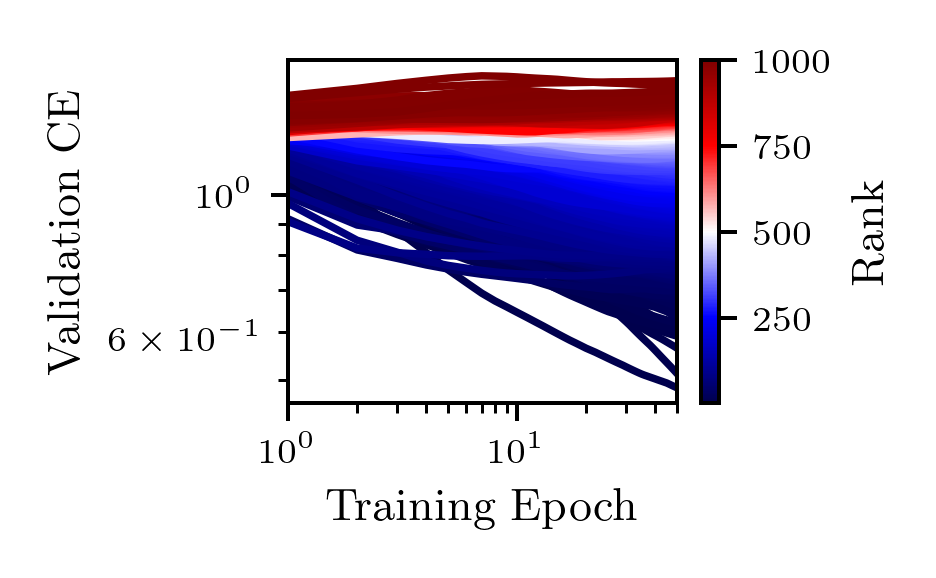}
        \caption{vehicle}
        \label{fig:ranking-curves-top10-vehicle}
    \end{subfigure}
    \begin{subfigure}[b]{0.24\textwidth}
        \centering
        \includegraphics[width=\textwidth]{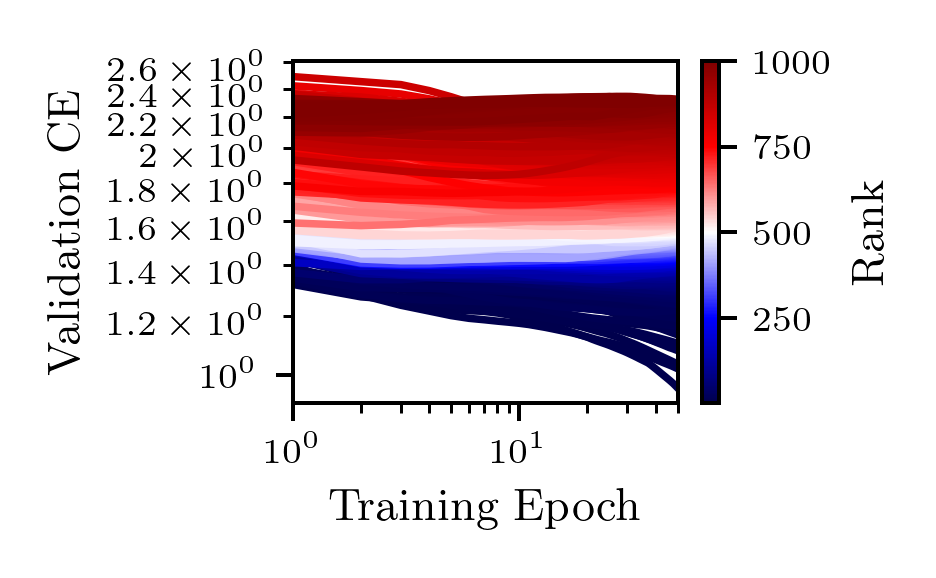}
        \caption{volkert}
        \label{fig:ranking-curves-top10-volkert}
    \end{subfigure}
    
    \caption{Learning curves of the 1000 randomly sampled models from LCBench now in YAHPO colored by the rank at maximum fidelity ({\bf Second Part}).}
    \label{fig:all-ranking-curves-yahpo-second-part}
\end{figure}

\begin{figure}
    \centering
    
    \begin{subfigure}[b]{0.24\textwidth}
        \centering
        \includegraphics[width=\textwidth]{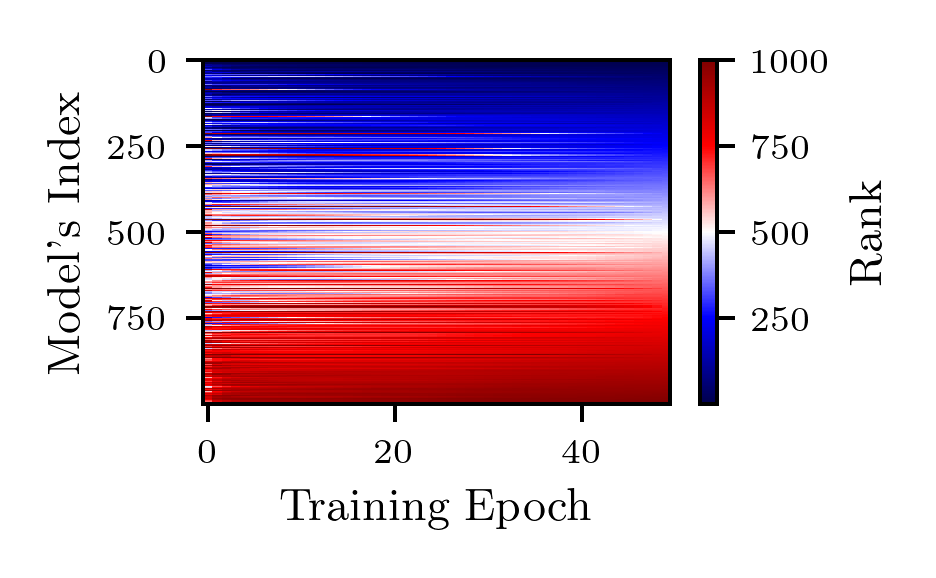}
        \caption{APSFailure}
        \label{fig:heatmap-top10-APSFailure}
    \end{subfigure}
    \begin{subfigure}[b]{0.24\textwidth}
        \centering
        \includegraphics[width=\textwidth]{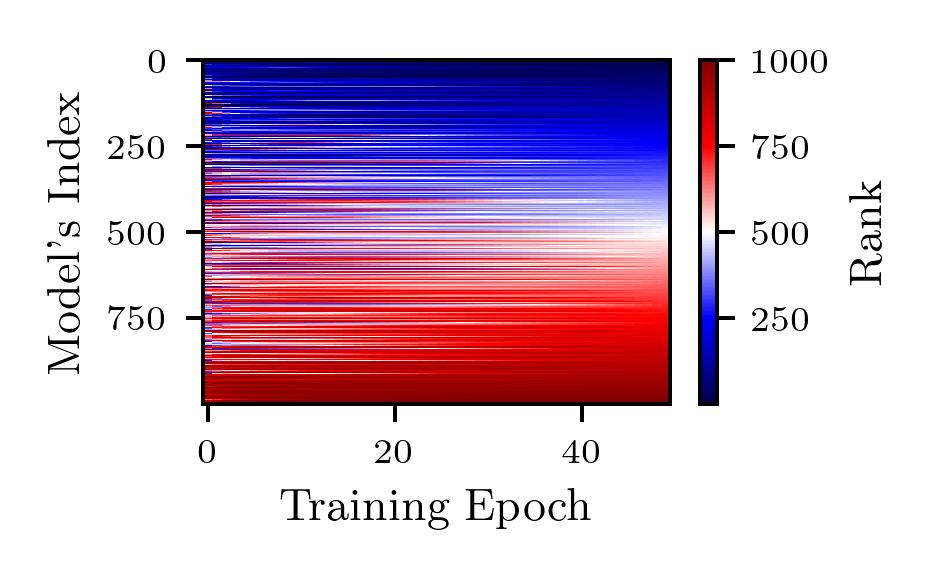}
        \caption{Amazon employee access}
        \label{fig:heatmap-top10-Amazon_employee_access}
    \end{subfigure}
    \begin{subfigure}[b]{0.24\textwidth}
        \centering
        \includegraphics[width=\textwidth]{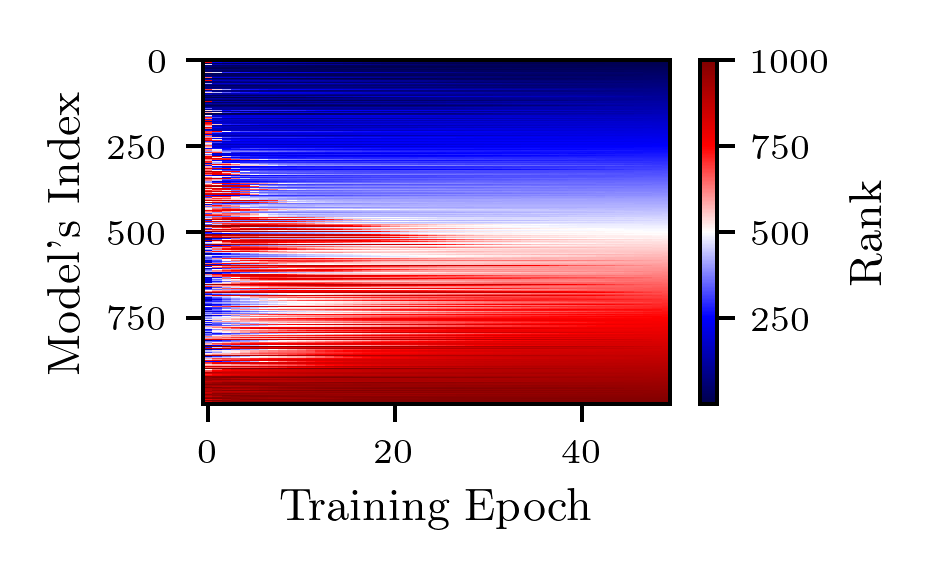}
        \caption{Australian}
        \label{fig:heatmap-top10-Australian}
    \end{subfigure}
    \begin{subfigure}[b]{0.24\textwidth}
        \centering
        \includegraphics[width=\textwidth]{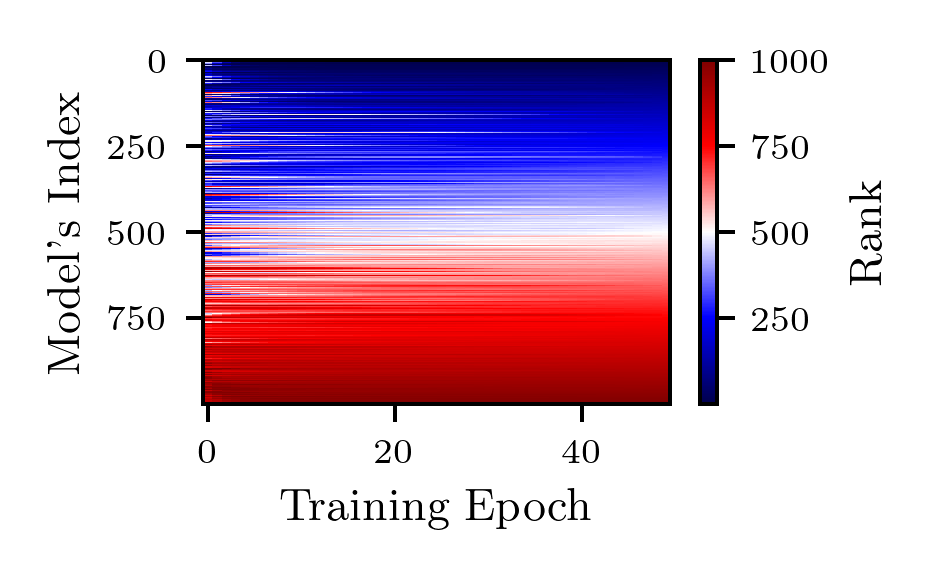}
        \caption{Fashion MNIST}
        \label{fig:heatmap-top10-Fashion-MNIST}
    \end{subfigure}
    \begin{subfigure}[b]{0.24\textwidth}
        \centering
        \includegraphics[width=\textwidth]{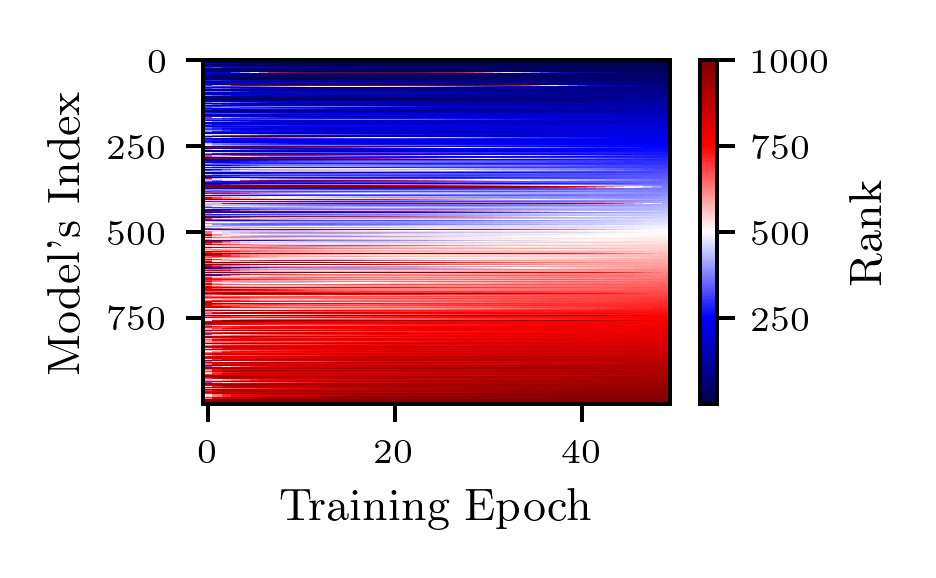}
        \caption{KDDCup09 appetency}
        \label{fig:heatmap-top10-KDDCup09_appetency}
    \end{subfigure}
    \begin{subfigure}[b]{0.24\textwidth}
        \centering
        \includegraphics[width=\textwidth]{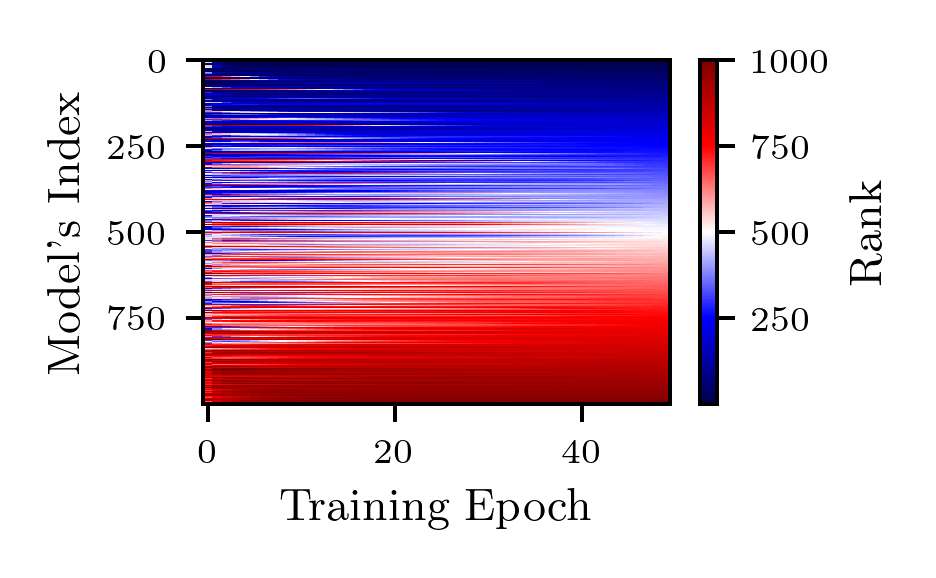}
        \caption{MiniBooNE}
        \label{fig:heatmap-top10-MiniBooNE}
    \end{subfigure}
    \begin{subfigure}[b]{0.24\textwidth}
        \centering
        \includegraphics[width=\textwidth]{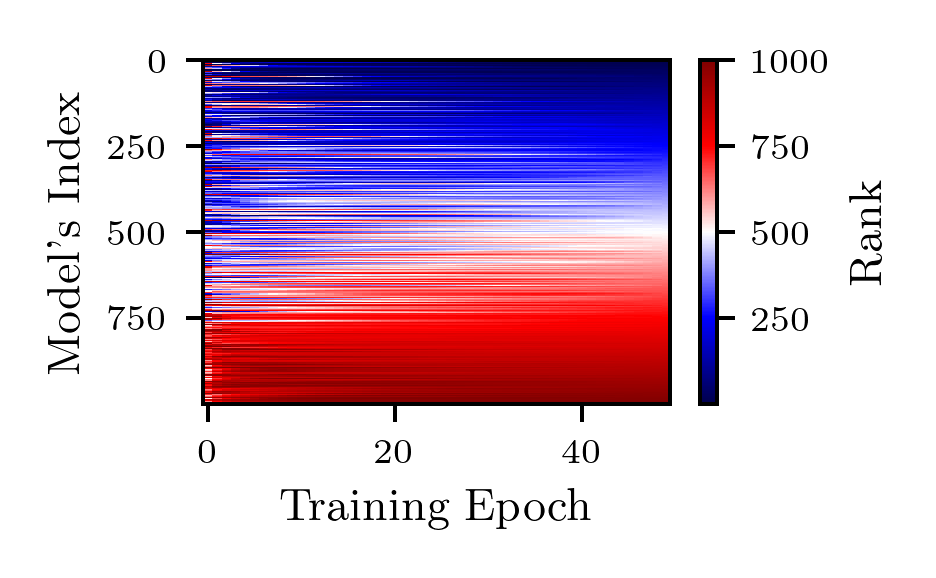}
        \caption{adult}
        \label{fig:heatmap-top10-adult}
    \end{subfigure}
    \begin{subfigure}[b]{0.24\textwidth}
        \centering
        \includegraphics[width=\textwidth]{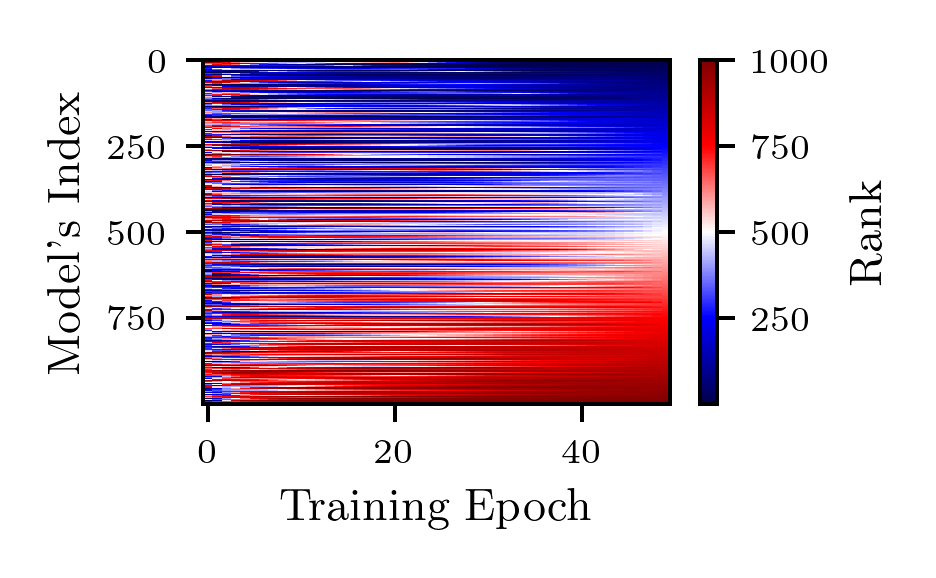}
        \caption{airlines}
        \label{fig:heatmap-top10-airlines}
    \end{subfigure}
    \begin{subfigure}[b]{0.24\textwidth}
        \centering
        \includegraphics[width=\textwidth]{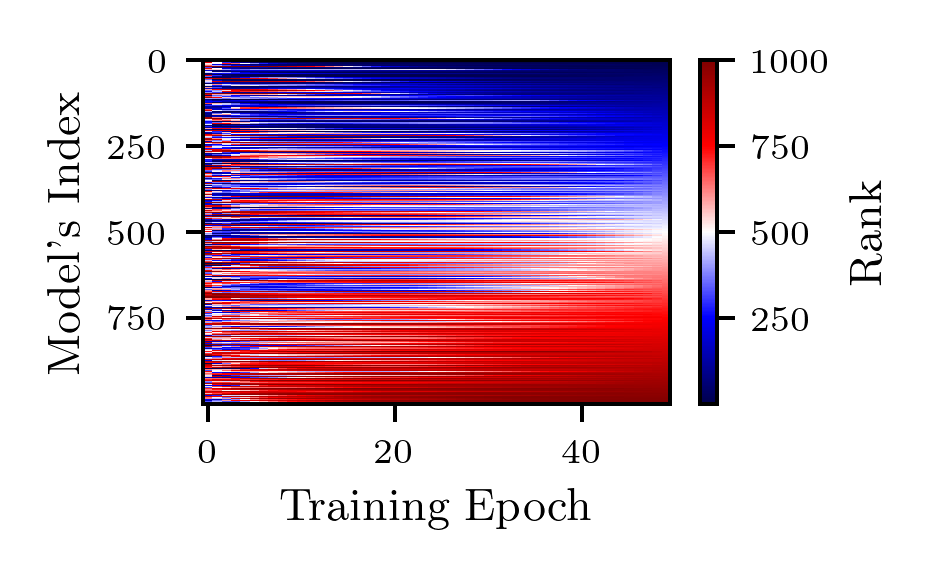}
        \caption{albert}
        \label{fig:heatmap-top10-albert}
    \end{subfigure}
    \begin{subfigure}[b]{0.24\textwidth}
        \centering
        \includegraphics[width=\textwidth]{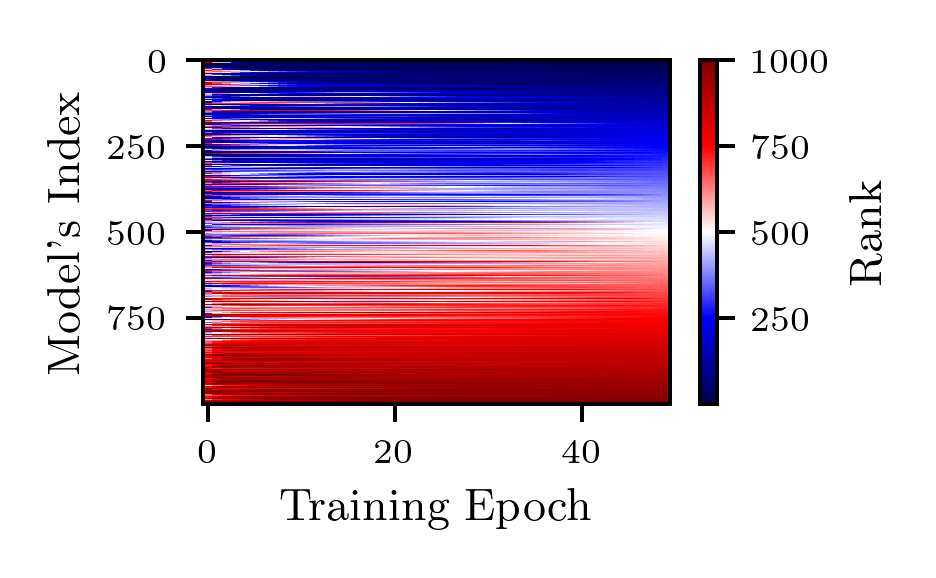}
        \caption{bank marketing}
        \label{fig:heatmap-top10-bank-marketing}
    \end{subfigure}
    \begin{subfigure}[b]{0.24\textwidth}
        \centering
        \includegraphics[width=\textwidth]{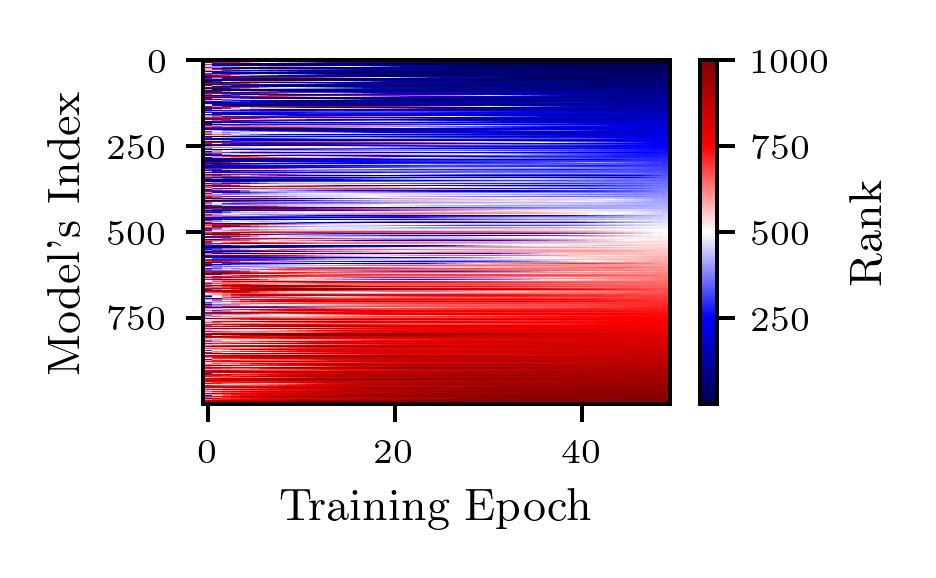}
        \caption{blood transfusion}
        \label{fig:heatmap-top10-blood-transfusion-service-center}
    \end{subfigure}
    \begin{subfigure}[b]{0.24\textwidth}
        \centering
        \includegraphics[width=\textwidth]{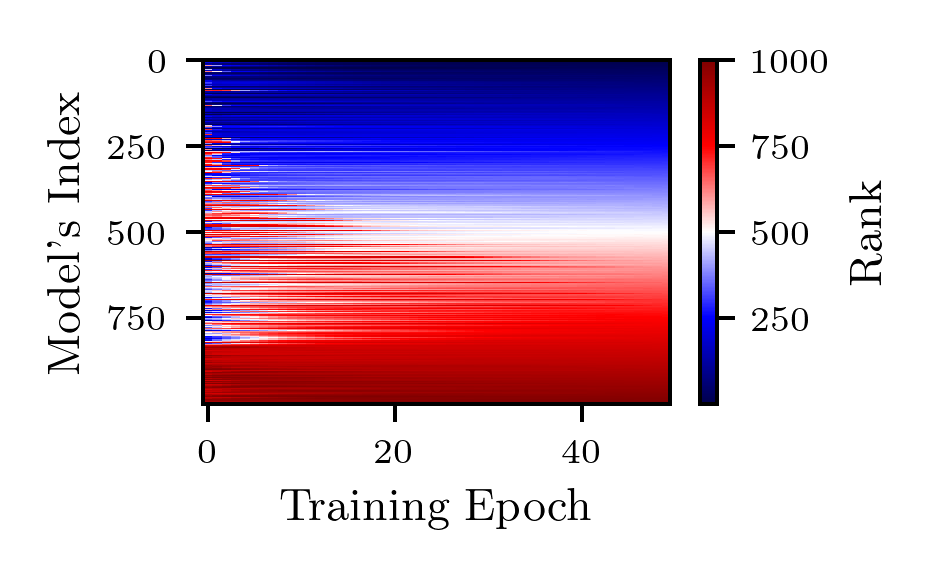}
        \caption{car}
        \label{fig:heatmap-top10-car}
    \end{subfigure}
    \begin{subfigure}[b]{0.24\textwidth}
        \centering
        \includegraphics[width=\textwidth]{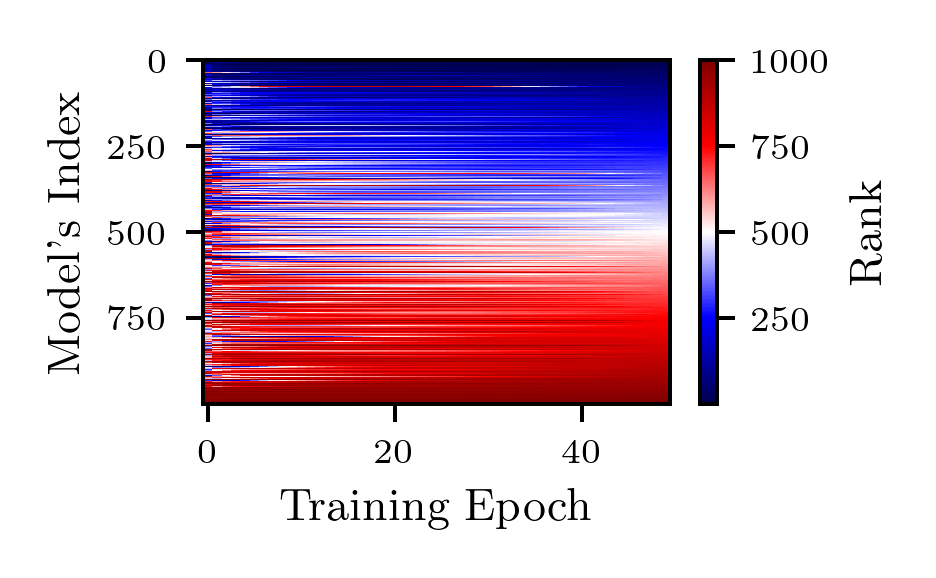}
        \caption{christine}
        \label{fig:heatmap-top10-christine}
    \end{subfigure}
    \begin{subfigure}[b]{0.24\textwidth}
        \centering
        \includegraphics[width=\textwidth]{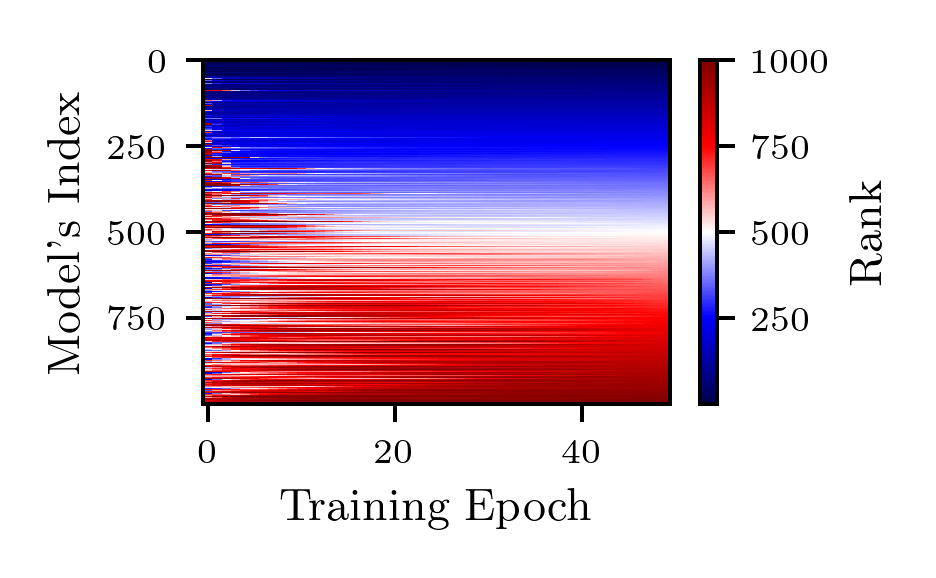}
        \caption{cnae 9}
        \label{fig:heatmap-top10-cnae-9}
    \end{subfigure}
    \begin{subfigure}[b]{0.24\textwidth}
        \centering
        \includegraphics[width=\textwidth]{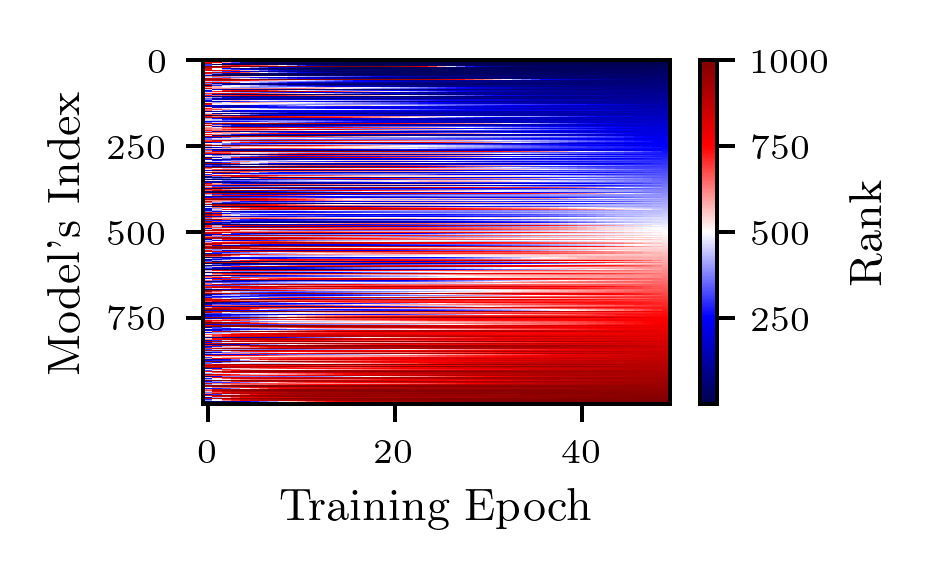}
        \caption{connect 4}
        \label{fig:heatmap-top10-connect-4}
    \end{subfigure}
    \begin{subfigure}[b]{0.24\textwidth}
        \centering
        \includegraphics[width=\textwidth]{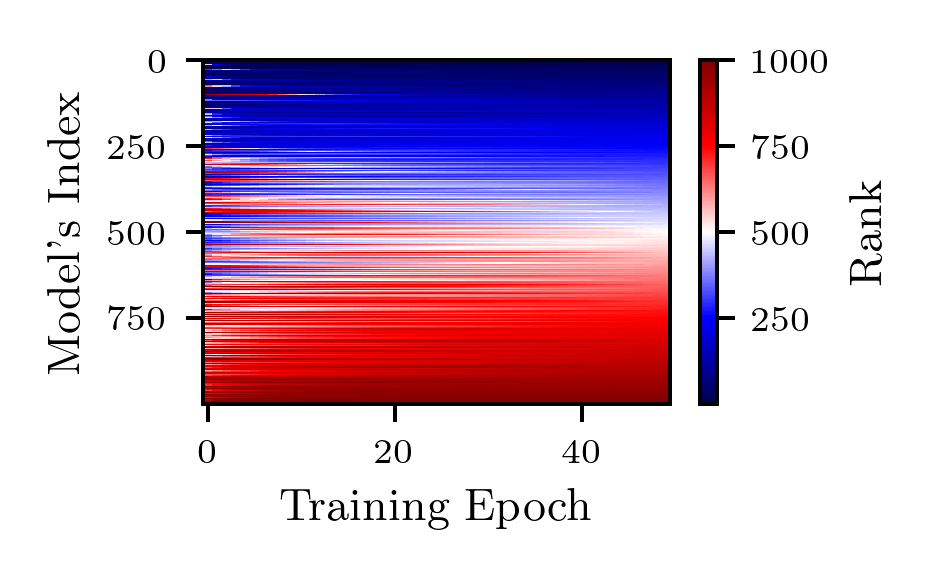}
        \caption{covertype}
        \label{fig:heatmap-top10-covertype}
    \end{subfigure}
    \begin{subfigure}[b]{0.24\textwidth}
        \centering
        \includegraphics[width=\textwidth]{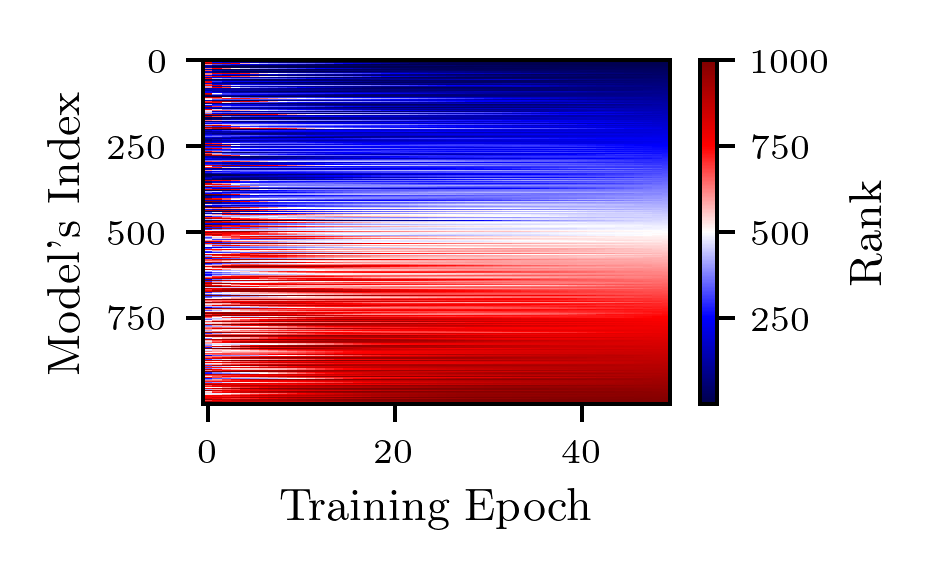}
        \caption{credit g}
        \label{fig:heatmap-top10-credit-g}
    \end{subfigure}
    \begin{subfigure}[b]{0.24\textwidth}
        \centering
        \includegraphics[width=\textwidth]{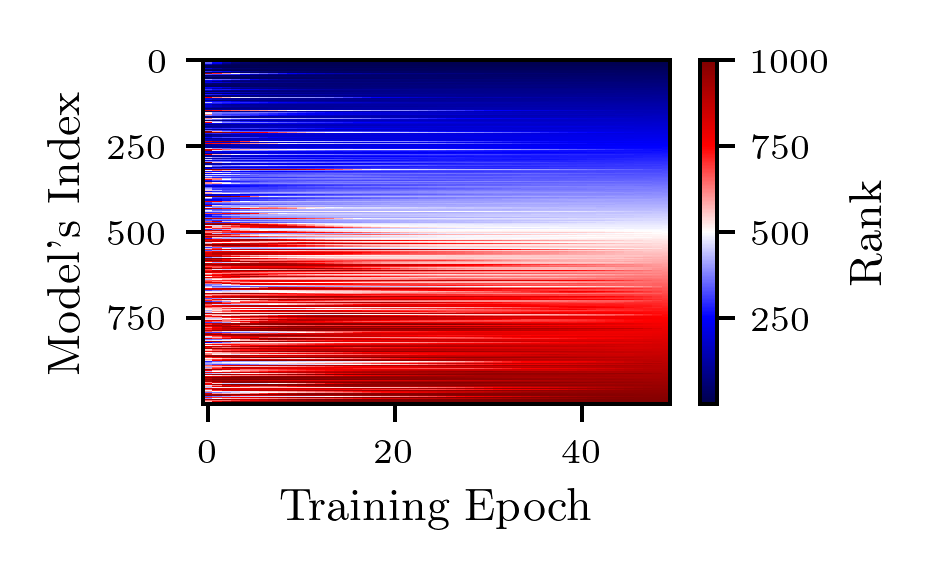}
        \caption{dionis}
        \label{fig:heatmap-top10-dionis}
    \end{subfigure}
    \begin{subfigure}[b]{0.24\textwidth}
        \centering
        \includegraphics[width=\textwidth]{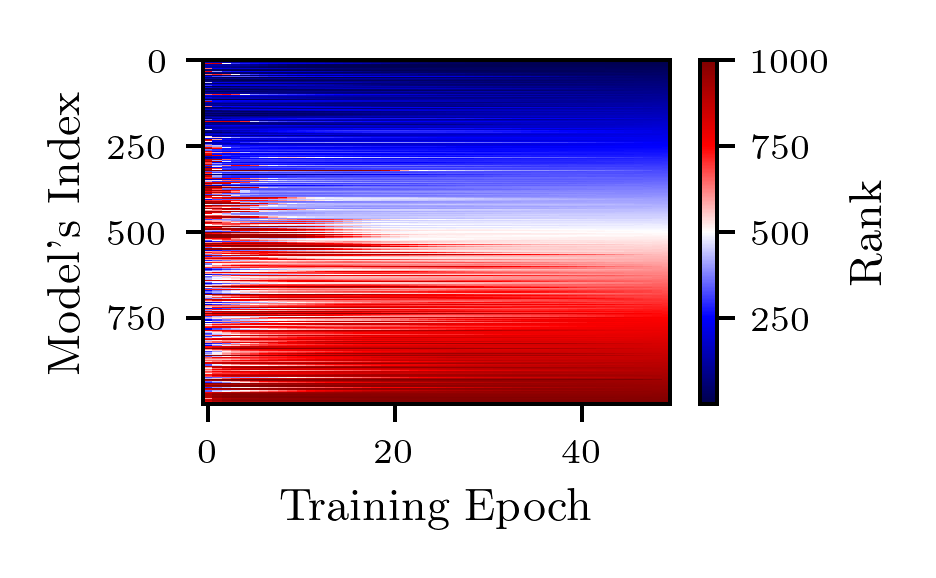}
        \caption{fabert}
        \label{fig:heatmap-top10-fabert}
    \end{subfigure}
    \begin{subfigure}[b]{0.24\textwidth}
        \centering
        \includegraphics[width=\textwidth]{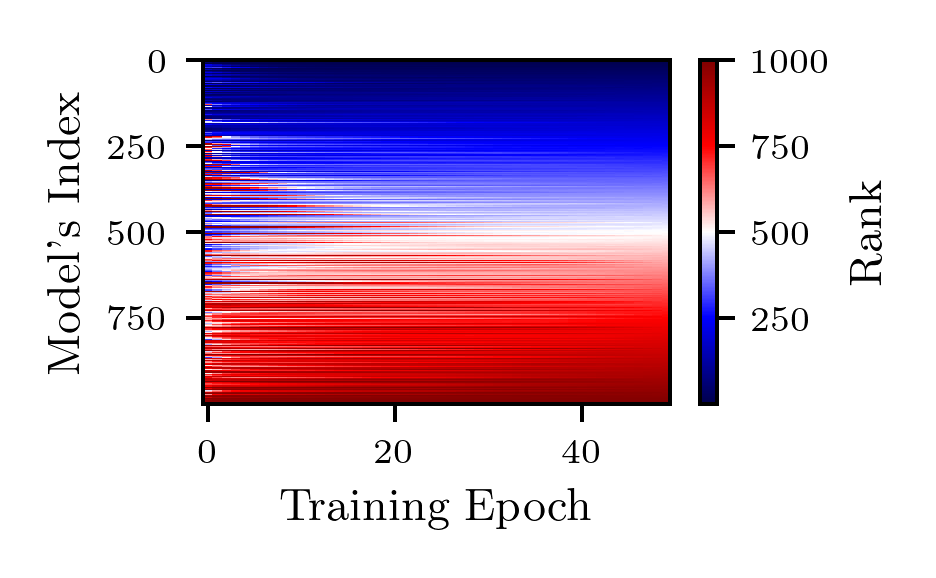}
        \caption{helena}
        \label{fig:heatmap-top10-helena}
    \end{subfigure}
    \begin{subfigure}[b]{0.24\textwidth}
        \centering
        \includegraphics[width=\textwidth]{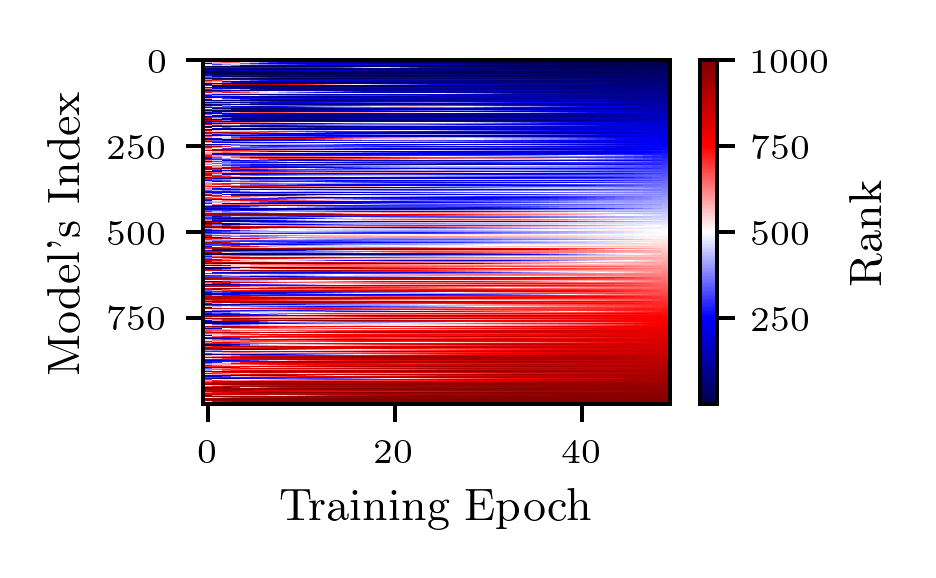}
        \caption{higgs}
        \label{fig:heatmap-top10-higgs}
    \end{subfigure}
    \begin{subfigure}[b]{0.24\textwidth}
        \centering
        \includegraphics[width=\textwidth]{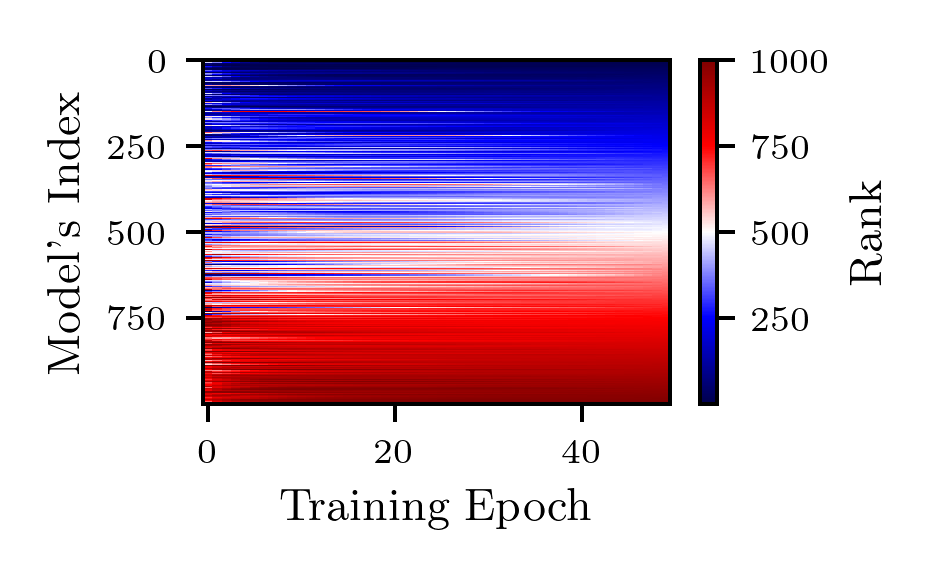}
        \caption{jannis}
        \label{fig:heatmap-top10-jannis}
    \end{subfigure}
    \begin{subfigure}[b]{0.24\textwidth}
        \centering
        \includegraphics[width=\textwidth]{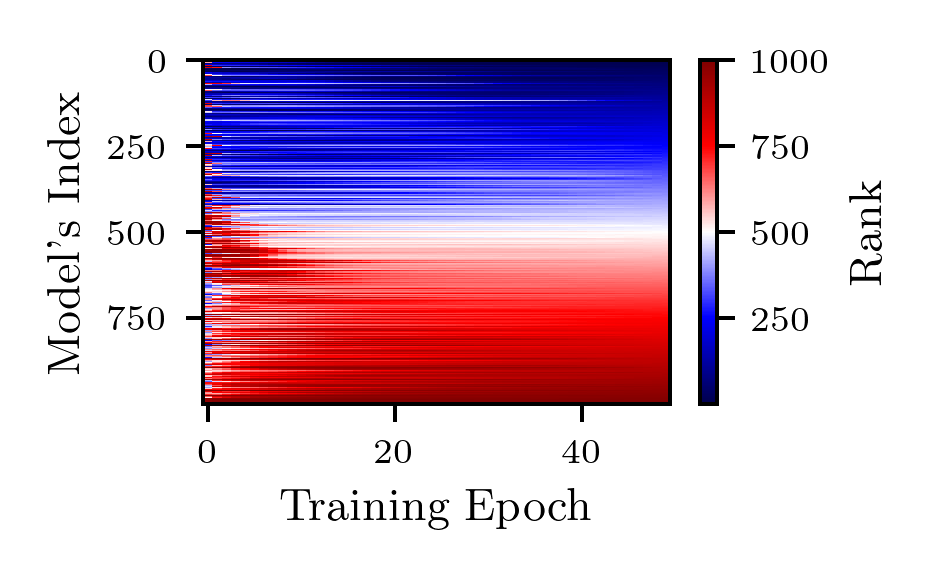}
        \caption{jasmine}
        \label{fig:heatmap-top10-jasmine}
    \end{subfigure}
    \begin{subfigure}[b]{0.24\textwidth}
        \centering
        \includegraphics[width=\textwidth]{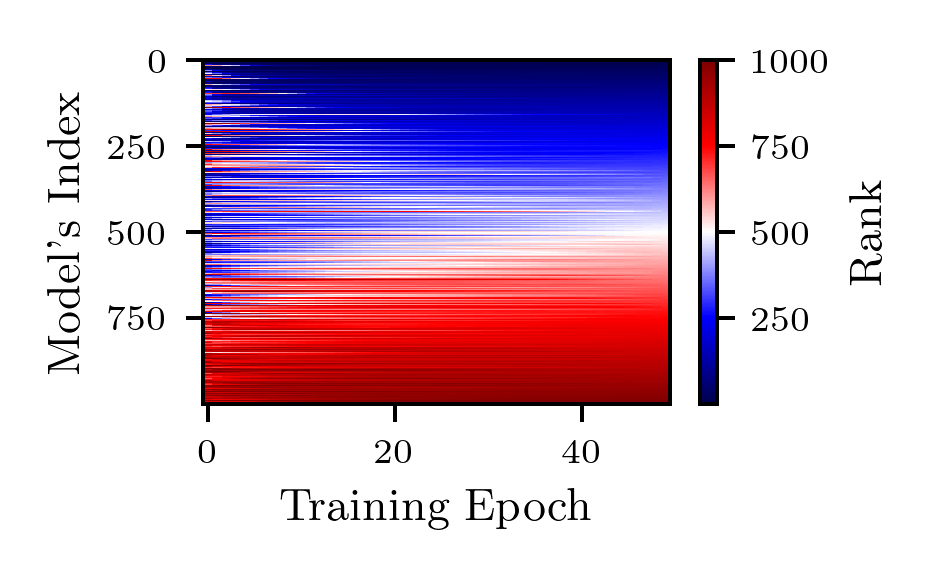}
        \caption{jungle chess 2pcs raw endgame complete}
        \label{fig:heatmap-top10-jungle_chess_2pcs_raw_endgame_complete}
    \end{subfigure}
    \begin{subfigure}[b]{0.24\textwidth}
        \centering
        \includegraphics[width=\textwidth]{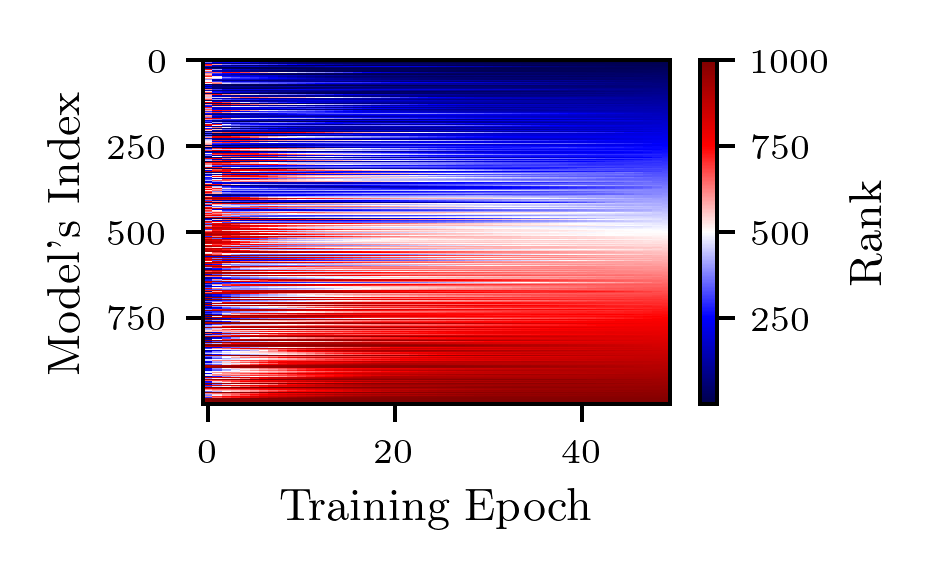}
        \caption{kc1}
        \label{fig:heatmap-top10-kc1}
    \end{subfigure}
    \begin{subfigure}[b]{0.24\textwidth}
        \centering
        \includegraphics[width=\textwidth]{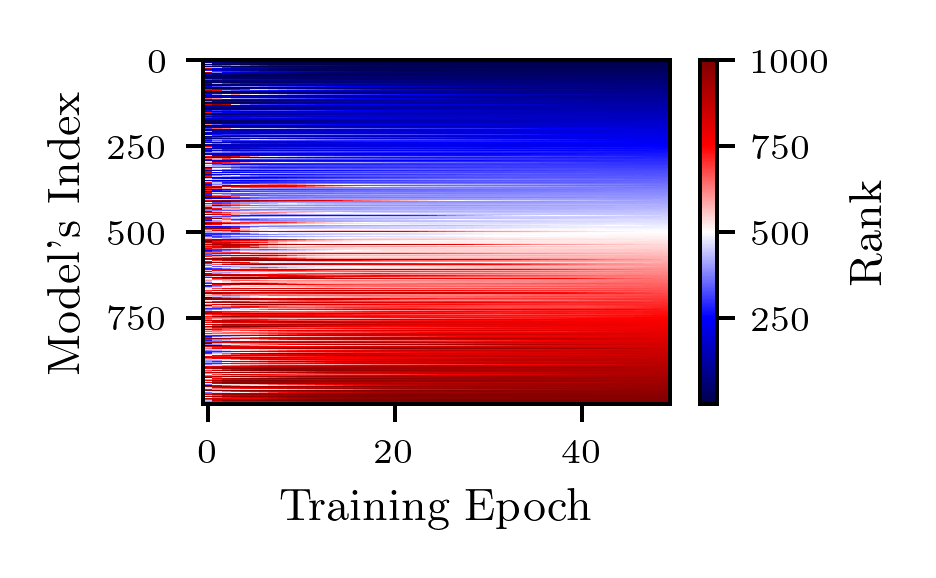}
        \caption{kr vs kp}
        \label{fig:heatmap-top10-kr-vs-kp}
    \end{subfigure}
  
    \caption{Evolution of the ranking from minimal to maximal fidelity on 1,000 randomly sampled models from the search space of YAHPO ({\bf First Part}).}
    \label{fig:all-heatmap-yahpo}
\end{figure}

\begin{figure}
    \centering

    \begin{subfigure}[b]{0.24\textwidth}
        \centering
        \includegraphics[width=\textwidth]{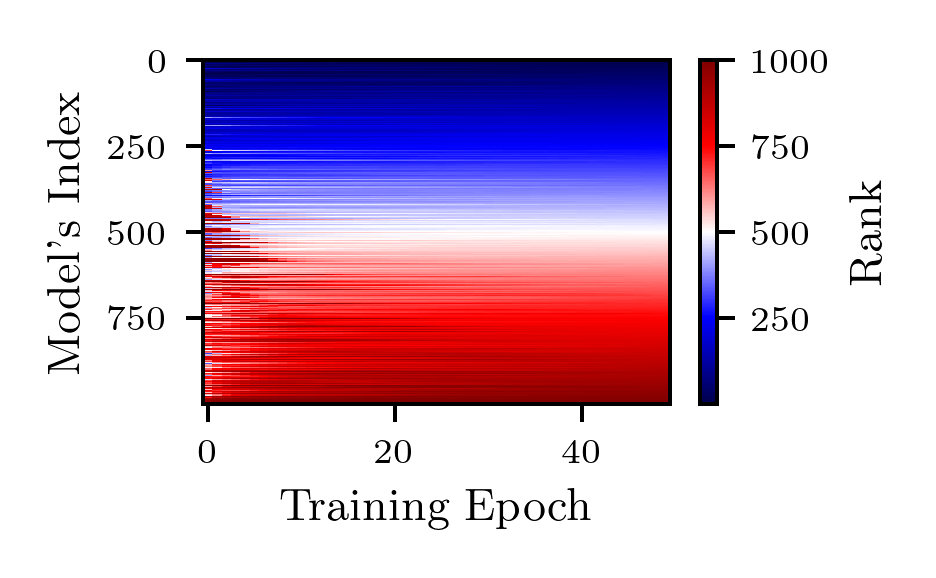}
        \caption{mfeat factors}
        \label{fig:heatmap-top10-mfeat-factors}
    \end{subfigure}
    \begin{subfigure}[b]{0.24\textwidth}
        \centering
        \includegraphics[width=\textwidth]{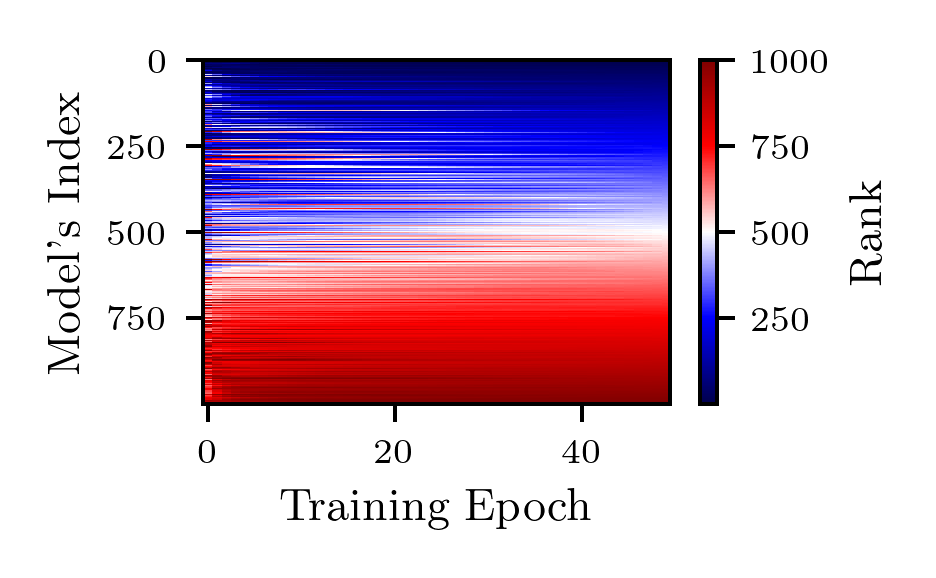}
        \caption{nomao}
        \label{fig:heatmap-top10-nomao}
    \end{subfigure}
    \begin{subfigure}[b]{0.24\textwidth}
        \centering
        \includegraphics[width=\textwidth,height=0.618\textwidth]{example-image}
        \caption{numerai28.6}
        \label{fig:heatmap-top10-numerai28.6}
    \end{subfigure}
    \begin{subfigure}[b]{0.24\textwidth}
        \centering
        \includegraphics[width=\textwidth]{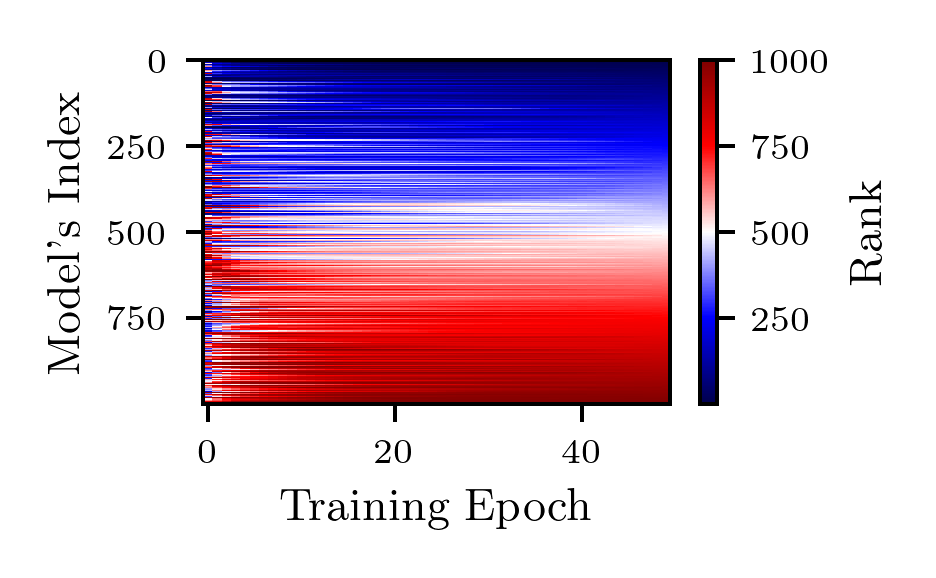}
        \caption{phoneme}
        \label{fig:heatmap-top10-phoneme}
    \end{subfigure}
    \begin{subfigure}[b]{0.24\textwidth}
        \centering
        \includegraphics[width=\textwidth]{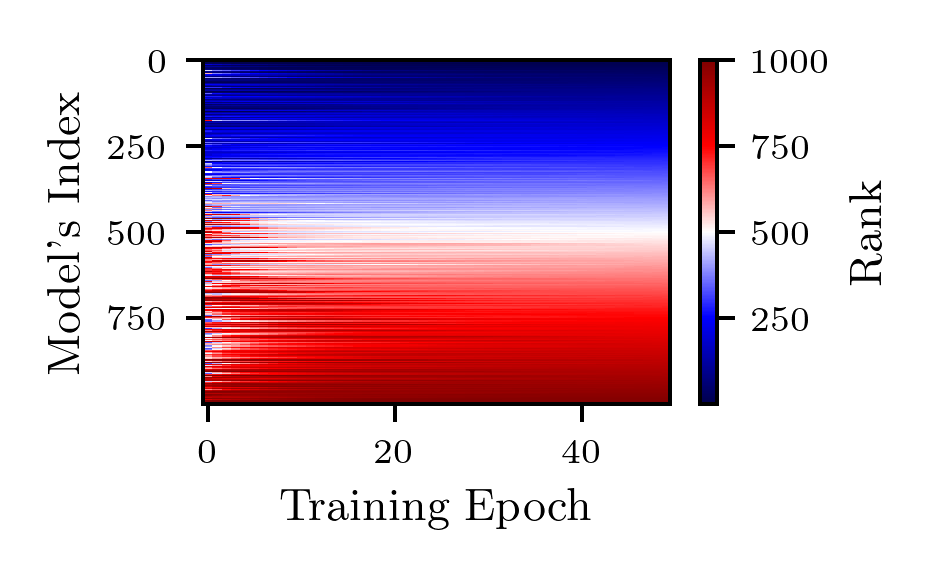}
        \caption{segment}
        \label{fig:heatmap-top10-segment}
    \end{subfigure}
    \begin{subfigure}[b]{0.24\textwidth}
        \centering
        \includegraphics[width=\textwidth]{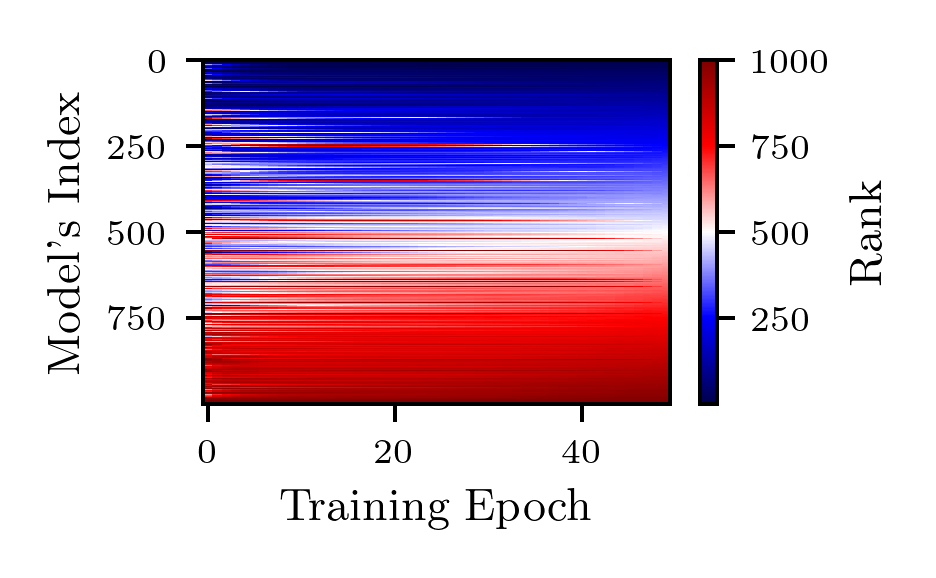}
        \caption{shuttle}
        \label{fig:heatmap-top10-shuttle}
    \end{subfigure}
    \begin{subfigure}[b]{0.24\textwidth}
        \centering
        \includegraphics[width=\textwidth]{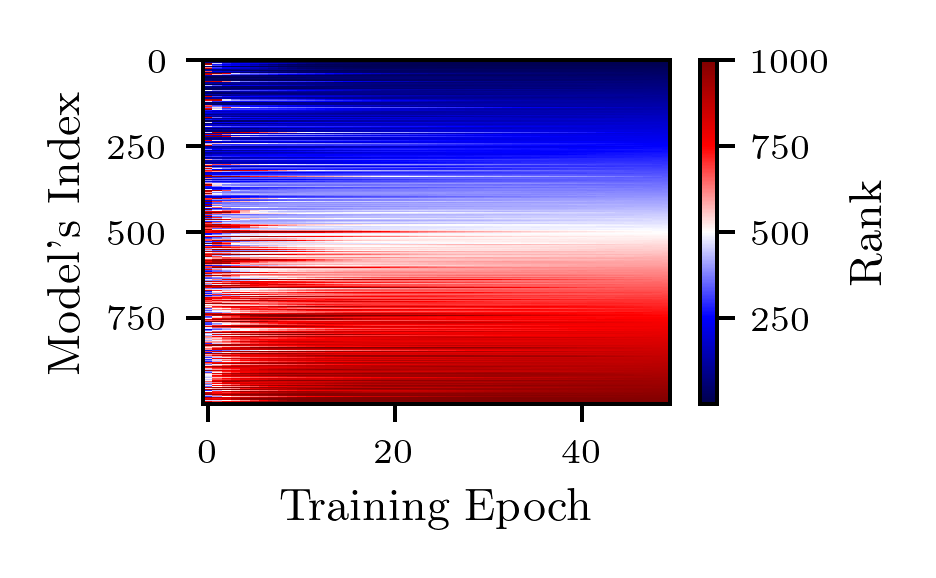}
        \caption{sylvine}
        \label{fig:heatmap-top10-sylvine}
    \end{subfigure}
    \begin{subfigure}[b]{0.24\textwidth}
        \centering
        \includegraphics[width=\textwidth]{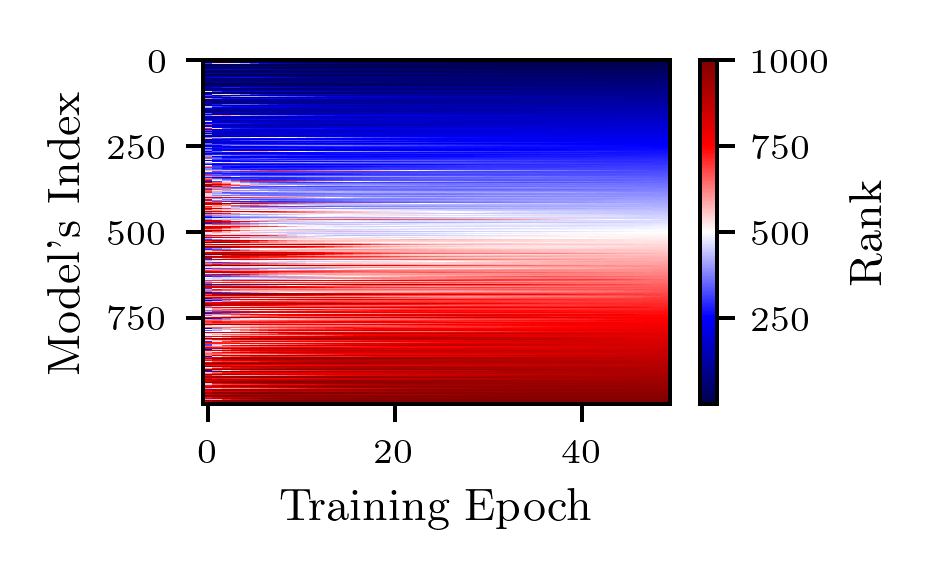}
        \caption{vehicle}
        \label{fig:heatmap-top10-vehicle}
    \end{subfigure}
    \begin{subfigure}[b]{0.24\textwidth}
        \centering
        \includegraphics[width=\textwidth]{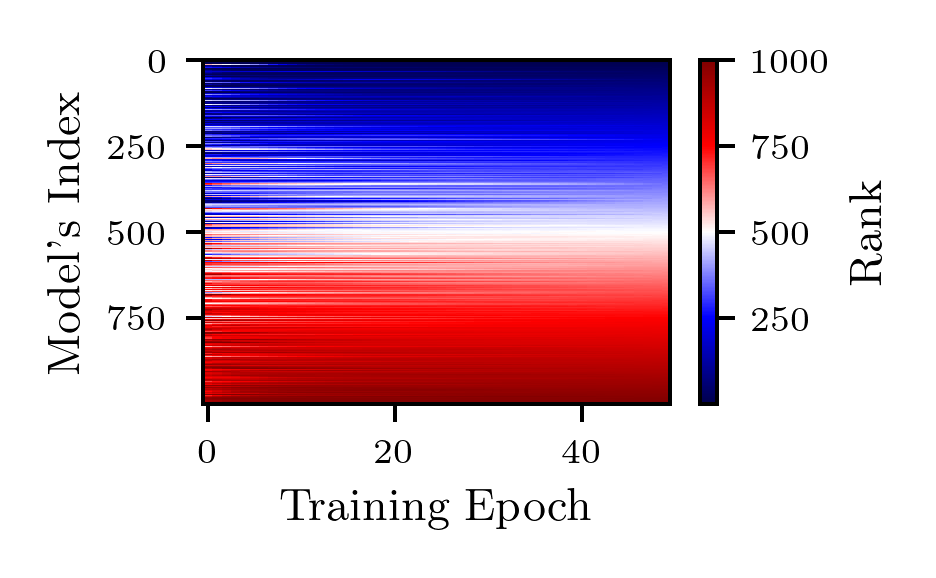}
        \caption{volkert}
        \label{fig:heatmap-top10-volkert}
    \end{subfigure}
    
    \caption{Evolution of the ranking from minimal to maximal fidelity on 1,000 randomly sampled models from the search space of YAHPO ({\bf Second Part}).}
    \label{fig:all-heatmap-yahpo-second-part}
\end{figure}

\begin{figure}
    \centering

    \begin{subfigure}[b]{0.45\textwidth}
        \centering
        \includegraphics[width=\textwidth]{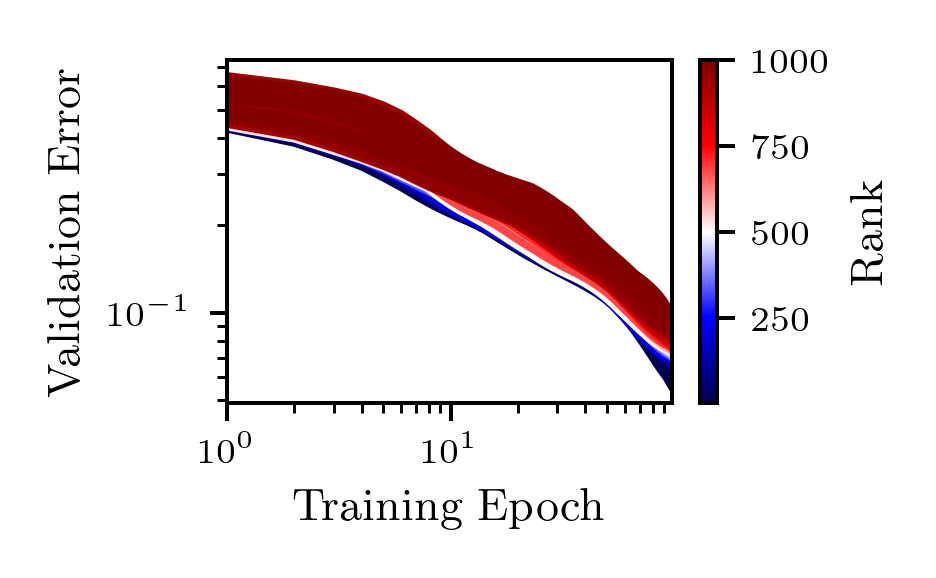}
        \caption{NB301 from YAHPO}
        \label{fig:ranking-curves-nb301-yahpo}
    \end{subfigure}
    \begin{subfigure}[b]{0.45\textwidth}
        \centering
        \includegraphics[width=\textwidth]{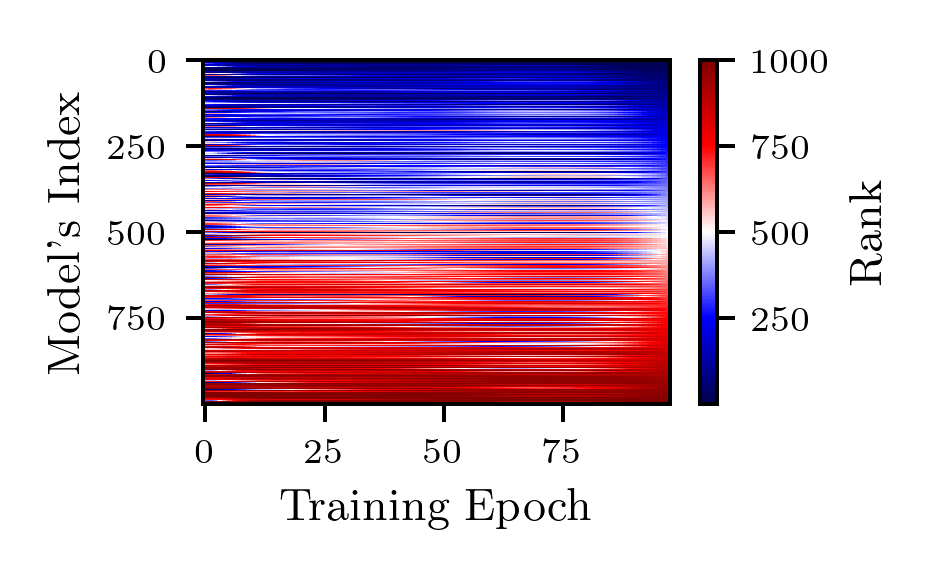}
        \caption{NB301 from YAHPO}
        \label{fig:heatmap-nb301-yahpo}
    \end{subfigure}
    
    \caption{(Left) Learning curves colored by the rank at maximum fidelity (98 epochs) and (Right) evolution of the ranking from minimal to maximal fidelity both on 1,000 randomly sampled models from the NB301 benchmark on the Cifar10 dataset from YAHPO.}
    \label{fig:nb301-cifar10-yahpo}
\end{figure}

Similarly to what is done on LCBench in the previous section we perform the same visualization now on YAHPO~\cite{pfisterer2022yahpo}. We present 1,000 randomly selected learning curves, colored by their final ranking in Figure~\ref{fig:all-ranking-curves-yahpo},\ref{fig:all-ranking-curves-yahpo-second-part} and the corresponding heatmaps in Figure~\ref{fig:all-heatmap-yahpo},\ref{fig:all-heatmap-yahpo-second-part}. In such visualization, we can identify clearly two groups among learning curves (good and bad candidates). Also we can notice that it is possible to identify good candidates only based on the first epoch. From this results, it is clear that the 1-Epoch baseline would be performing similarly well on a majority of these problems. Now, LCBench only considers tabular datasets. Therefore, within YAHPO we look at NB301 based on Cifar-10 (computer vision dataset), the only scenario (other than the LCBench scenario) using epochs as the fidelity. We present the results in Fig.~\ref{fig:nb301-cifar10-yahpo}. From the heatmap, it is clear that the best configuration is already dominating bad configurations from the first epoch. Therefore, 1-Epoch would also perform well on this benchmark.

\section{Visualization of Learning Curves from JAHS-BENCH}~\label{sec:appendix-viz-learning-curves-JAHSBENCH}

\begin{figure}
    \centering
    \begin{subfigure}[b]{0.3\textwidth}
        \centering
        \includegraphics[width=\textwidth]{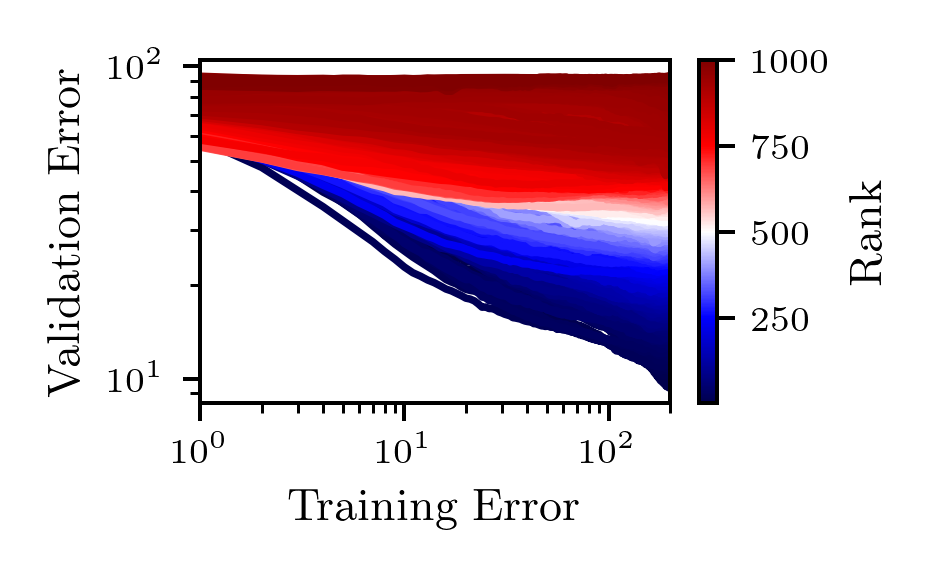}
        \caption{Cifar10}
        \label{fig:ranking-curves-cifar10-jahsbench}
    \end{subfigure}
    \begin{subfigure}[b]{0.3\textwidth}
        \centering
        \includegraphics[width=\textwidth]{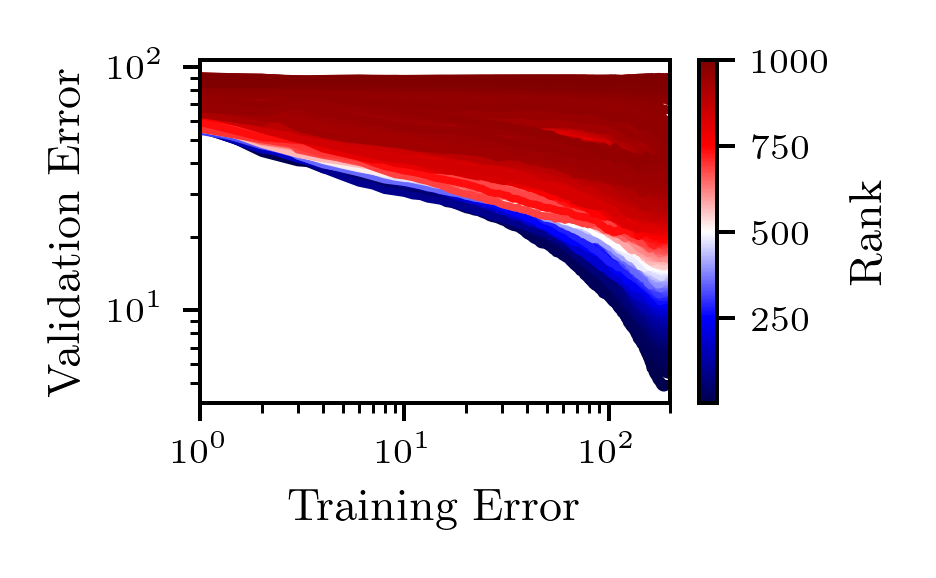}
        \caption{Colorectal Histology}
        \label{fig:ranking-curves-colorectal_histology-jahsbench}
    \end{subfigure}
    \begin{subfigure}[b]{0.3\textwidth}
        \centering
        \includegraphics[width=\textwidth]{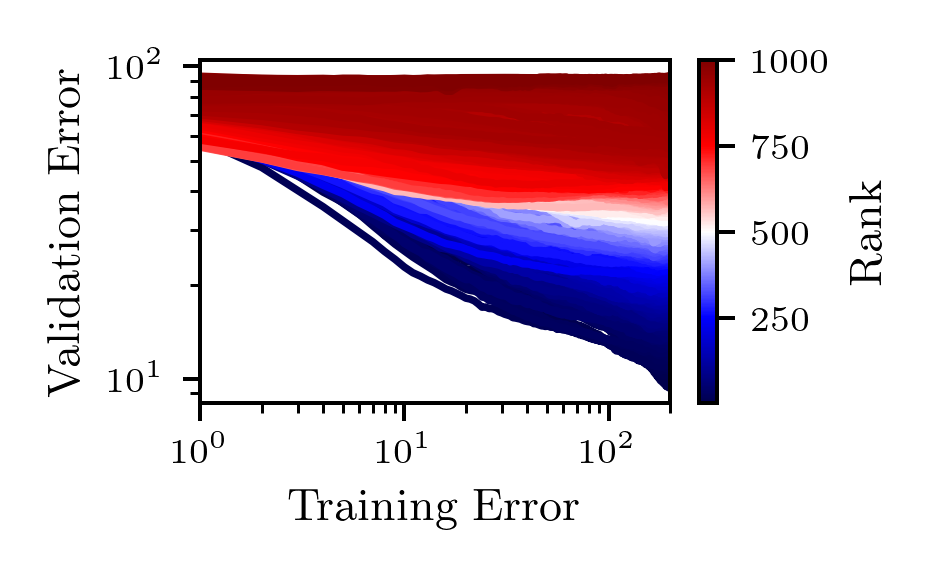}
        \caption{Fashion-MNIST}
        \label{fig:ranking-curves-fashion_mnist-jahsbench}
    \end{subfigure}
    \caption{Learning curves colored by the rank at maximum fidelity from JAHS-Bench-201.}
    \label{fig:all-ranking-curves-jahsbench}
\end{figure}

\begin{figure}
    \centering

    \begin{subfigure}[b]{0.3\textwidth}
        \centering
        \includegraphics[width=\textwidth]{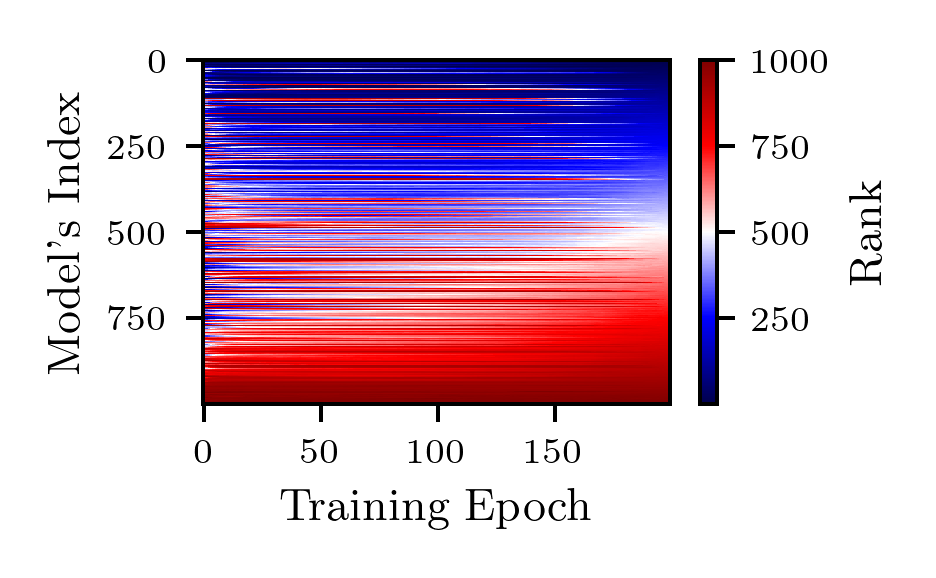}
        \caption{Cifar10}
        \label{fig:heatmap-cifar10-jahsbench}
    \end{subfigure}
    \begin{subfigure}[b]{0.3\textwidth}
        \centering
        \includegraphics[width=\textwidth]{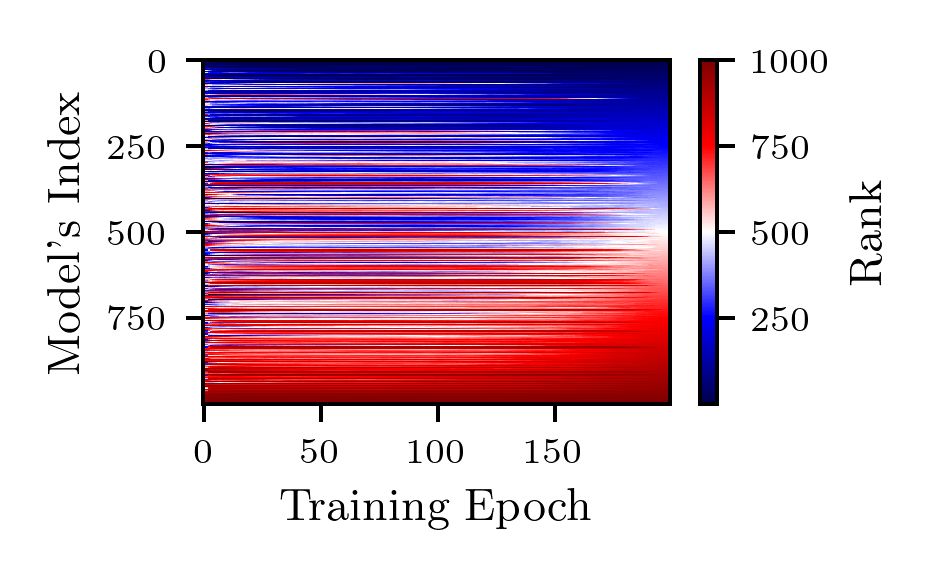}
        \caption{Colorectal Histology}
        \label{fig:heatmap-colorectal_histology-jahsbench}
    \end{subfigure}
    \begin{subfigure}[b]{0.3\textwidth}
        \centering
        \includegraphics[width=\textwidth]{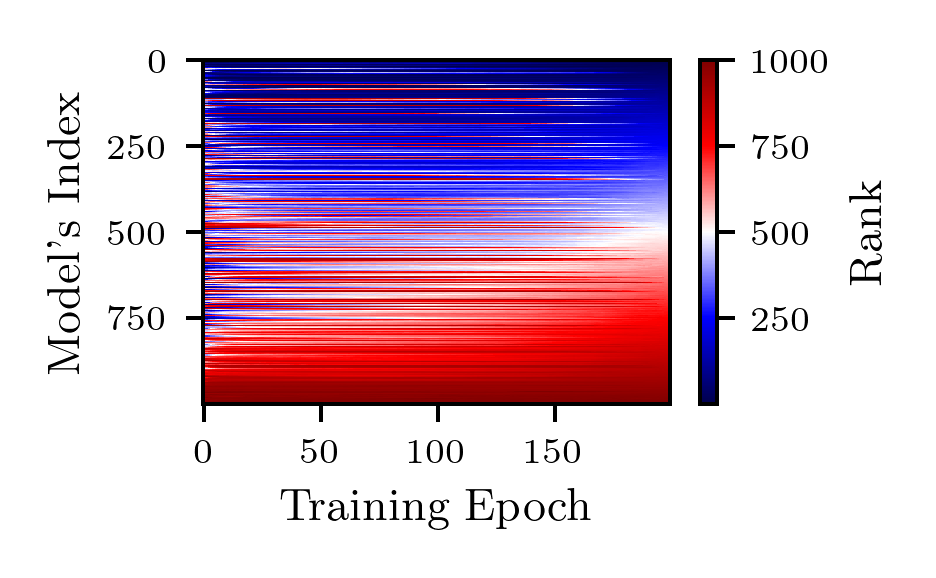}
        \caption{Fashion-MNIST}
        \label{fig:heatmap-fashion_mnist-jahsbench}
    \end{subfigure}

    \caption{Evolution of the ranking from minimal to maximal fidelity on 1,000 randomly sampled models from JAHS-Bench-201.}
    \label{fig:all-heatmap-jahsbench}
\end{figure}

Similarly to was done on HPOBench, LCBench and YAHPO in the previous section we perform the diagnostic visualization now on JAHS-Bench-201~\cite{bansal2022jahsbench}. We present 1,000 randomly selected learning curves as done before, colored by their final ranking in Figure~\ref{fig:all-ranking-curves-jahsbench} and the corresponding heatmaps in Figure~\ref{fig:all-heatmap-jahsbench}. Same conclusions as previous benchmarks can be derived. The best curves are dominant over the red curves. This can be seen well in the top row of the heatmaps Figure~\ref{fig:all-heatmap-jahsbench} which remain blue from the start to the end.

\section{Experiments on HPOBench}~\label{sec:appendix-1-epoch-hpobench}

\begin{figure}
    \centering
    \begin{subfigure}[b]{0.45\textwidth}
        \centering
        \caption{Naval Propulsion}
        \includegraphics[width=\textwidth]{figures/hpobench/navalpropulsion/pruners_navalpropulsion_RANDOM.png}
        \vspace{-0.1in}
        \label{fig:hpo-bench-stopper-comparison-navalpropulsion}
    \end{subfigure}
    \begin{subfigure}[b]{0.45\textwidth}
        \centering
        \caption{Parkinsons Telemonitoring}
        \includegraphics[width=\textwidth]{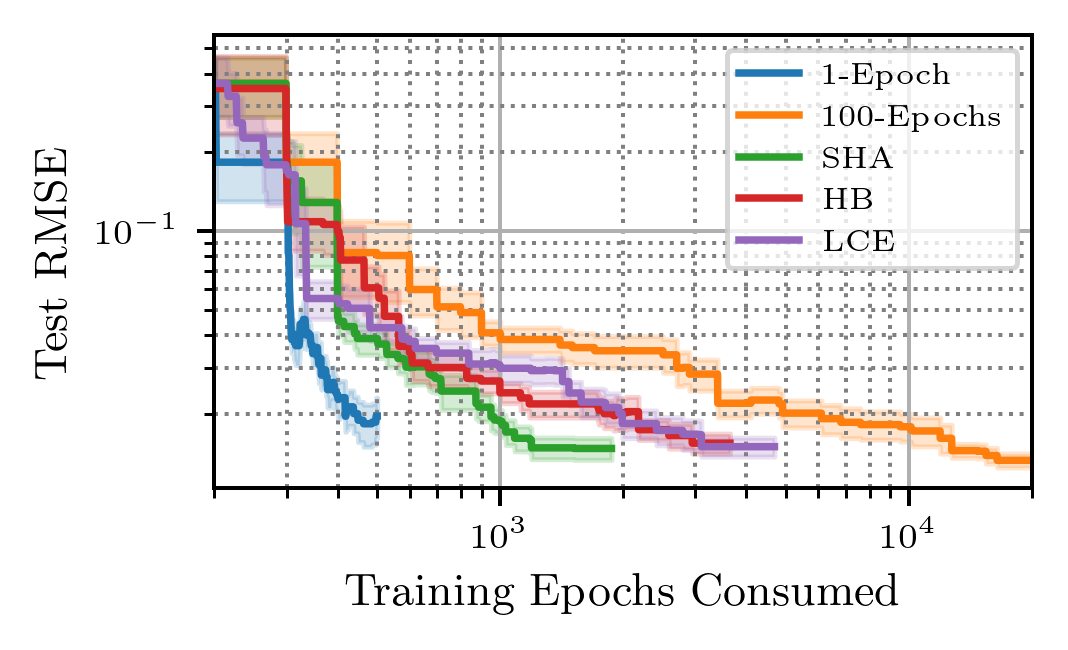}
        \vspace{-0.1in}
        \label{fig:hpo-bench-stopper-comparison-parkinsonstelemonitoring}
    \end{subfigure}
    \begin{subfigure}[b]{0.45\textwidth}
        \centering
        \caption{Protein Structure}
        \includegraphics[width=\textwidth]{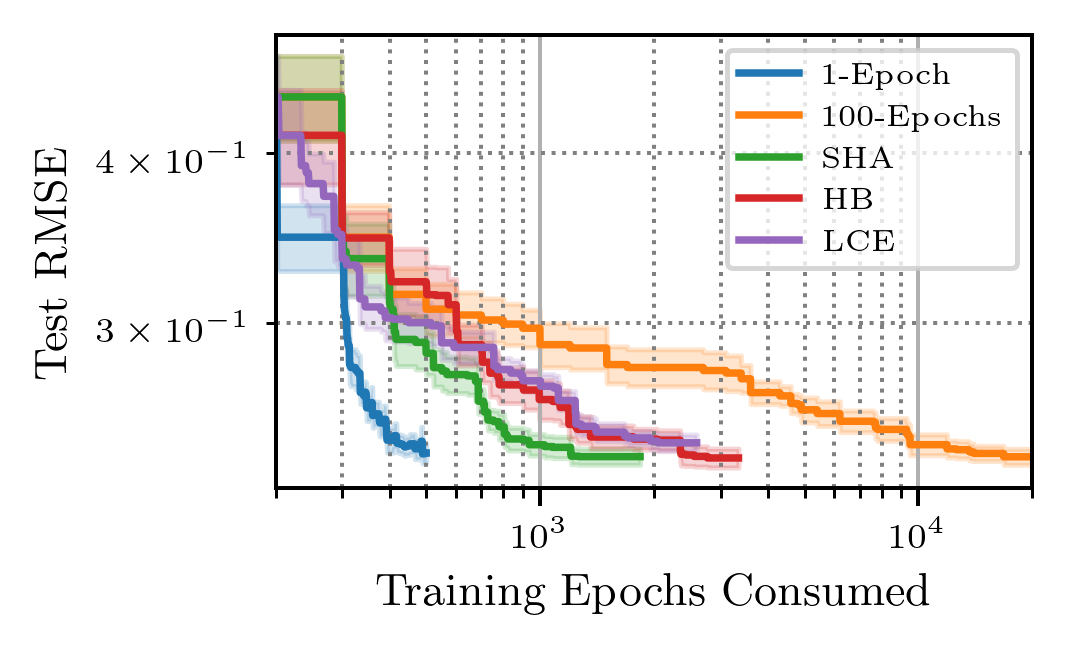}
        \vspace{-0.1in}
        \label{fig:hpo-bench-stopper-comparison-proteinstructure}
    \end{subfigure}
    \begin{subfigure}[b]{0.45\textwidth}
        \centering
        \caption{Slice Localization}
        \includegraphics[width=\textwidth]{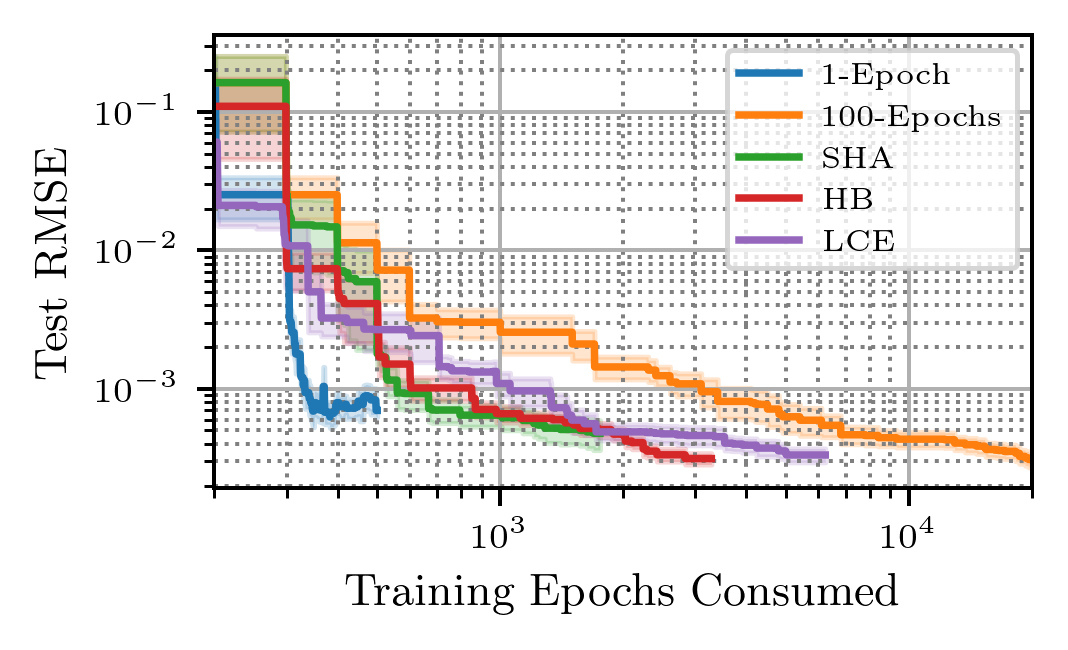}
        \vspace{-0.1in}
        \label{fig:hpo-bench-stopper-comparison-slicelocalization}
    \end{subfigure}
    \vspace{-0.1in}
    \caption{Comparing the performance (mean and standard error) of various early discarding strategies (all combined with with random search) on the HPOBench benchmark: {\color{orange}100-Epochs} and {\color{blue}1-Epoch} use a fixed number of epoch while {\color{green}Successive Halving (SHA)}, {\color{red}Hyperband (HB)} and {\color{purple}Learning Curve Extrapolation (LCE)} manage the number of epochs dynamically. Using as stopping criterion 200 iterations of the search agent, dynamic strategies consume fewer training epochs than 100-Epochs, but, 1-Epoch performs yet better than all methods. In particular, it consumes 40 times fewer epochs than 100-Epochs.}
    \label{fig:model-based-pruning-with-random-search-on-hpobench-appendix}
\end{figure}

In this section, we present the four problems of the HPOBench benchmark in Figure~\ref{fig:model-based-pruning-with-random-search-on-hpobench-appendix} to complement the Naval Propulsion problem presented in the main part of our study. As it can be seen, all problems have similar outcomes.

\section{Experiments on LCBench}~\label{sec:appendix-1-epoch-lcbench}

\begin{figure}[!h]
    \centering

    \begin{subfigure}[b]{0.24\textwidth}
        \centering
        \includegraphics[width=\textwidth]{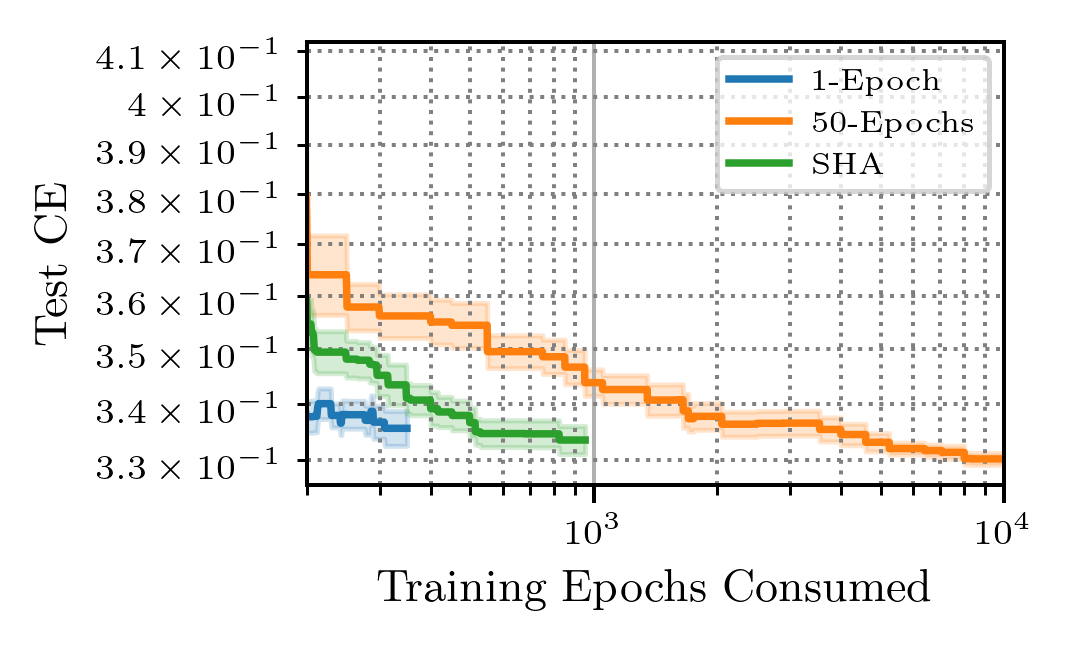}
        \caption{APSFailure}
        \label{fig:pruners-APSFailure-lcbench}
    \end{subfigure}
    \begin{subfigure}[b]{0.24\textwidth}
        \centering
        \includegraphics[width=\textwidth]{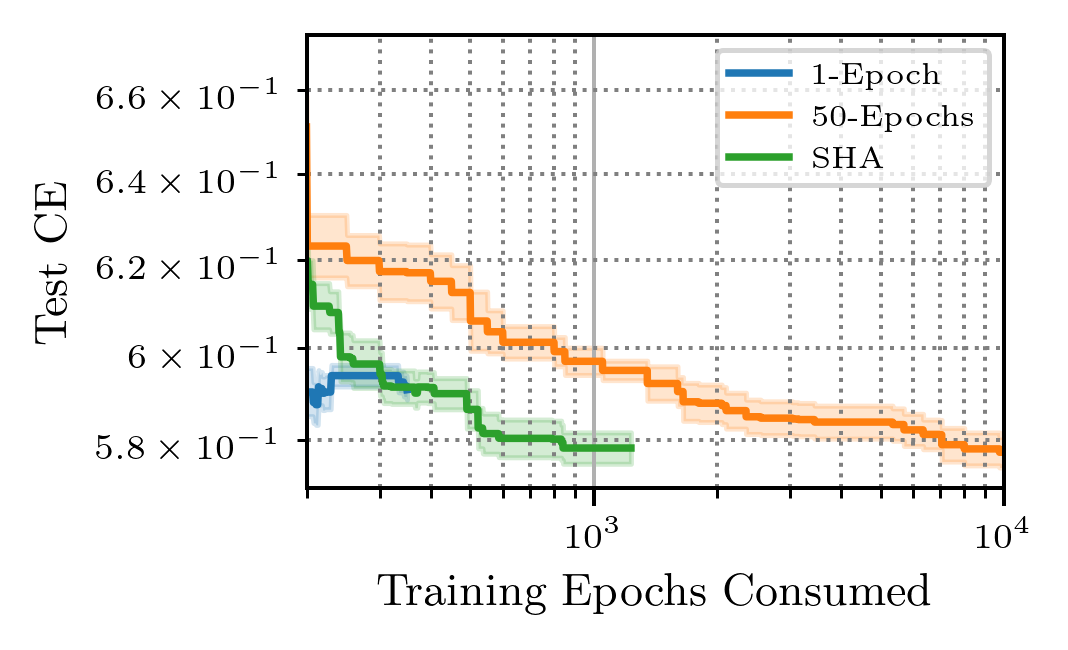}
        \caption{Amazon employee access}
        \label{fig:pruners-Amazon_employee_access-lcbench}
    \end{subfigure}
    \begin{subfigure}[b]{0.24\textwidth}
        \centering
        \includegraphics[width=\textwidth]{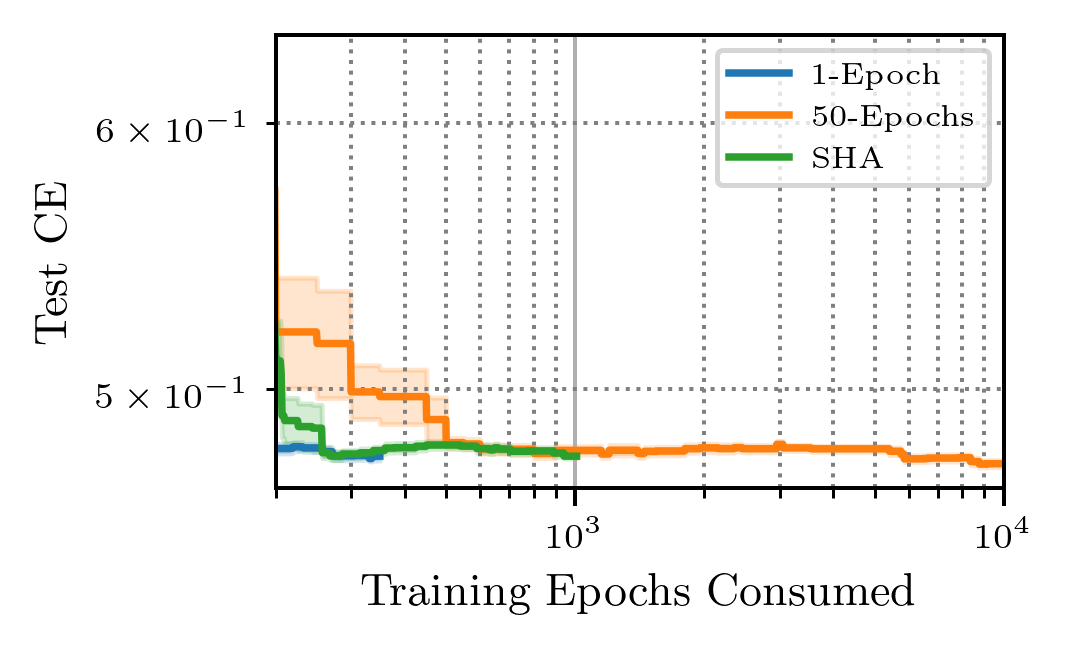}
        \caption{Australian}
        \label{fig:pruners-Australian-lcbench}
    \end{subfigure}
    \begin{subfigure}[b]{0.24\textwidth}
        \centering
        \includegraphics[width=\textwidth]{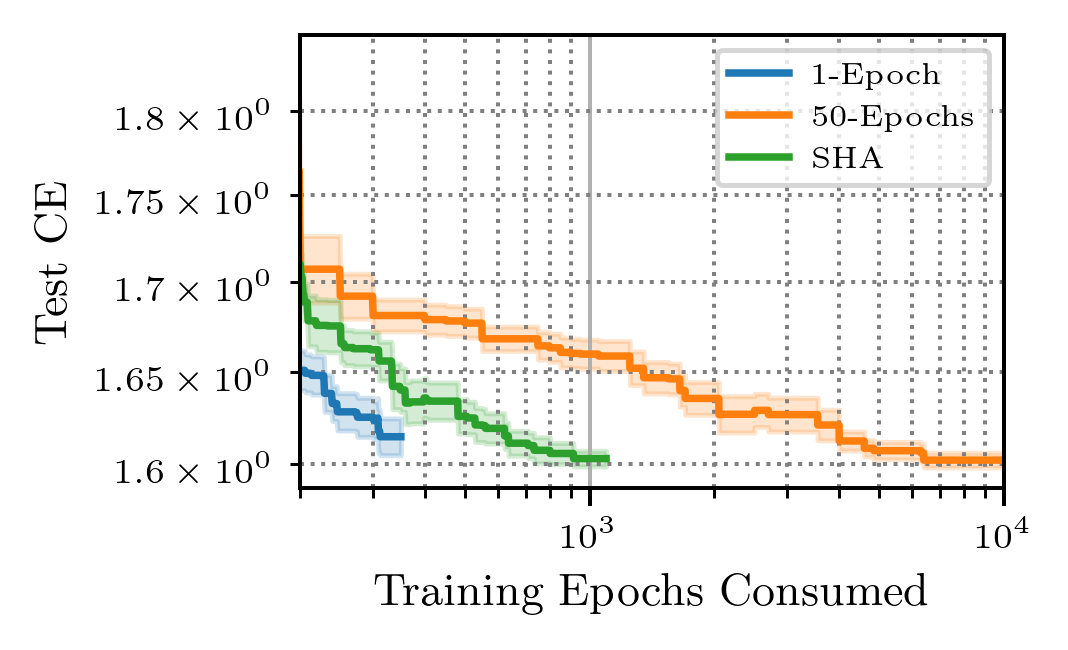}
        \caption{Fashion MNIST}
        \label{fig:pruners-Fashion-MNIST-lcbench}
    \end{subfigure}
    \begin{subfigure}[b]{0.24\textwidth}
        \centering
        \includegraphics[width=\textwidth]{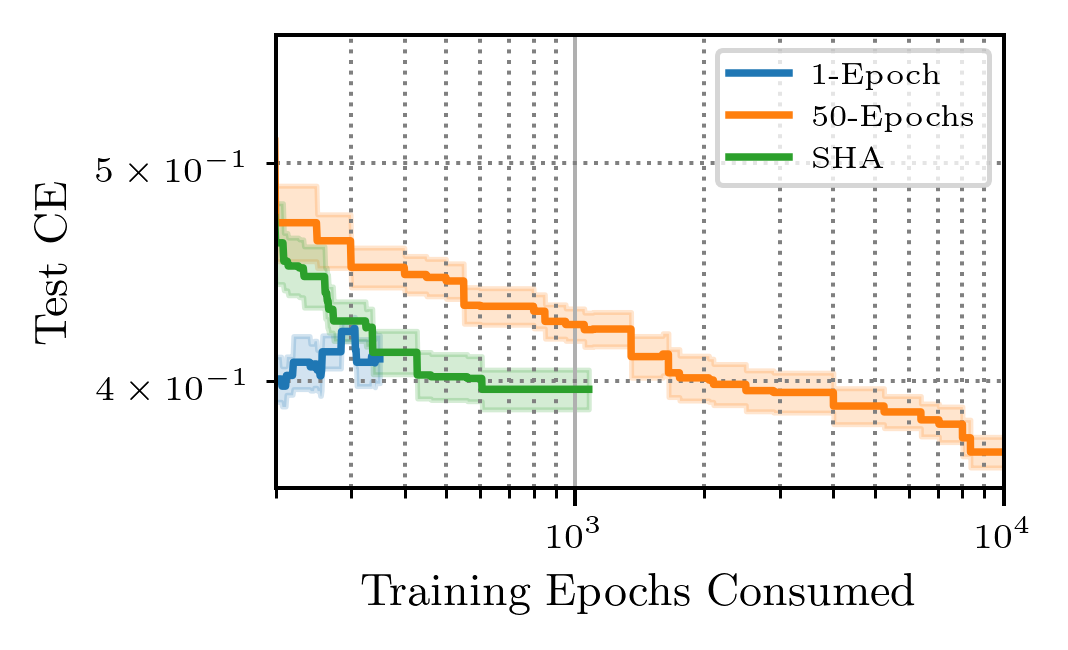}
        \caption{KDDCup09 appetency}
        \label{fig:pruners-KDDCup09_appetency-lcbench}
    \end{subfigure}
    \begin{subfigure}[b]{0.24\textwidth}
        \centering
        \includegraphics[width=\textwidth]{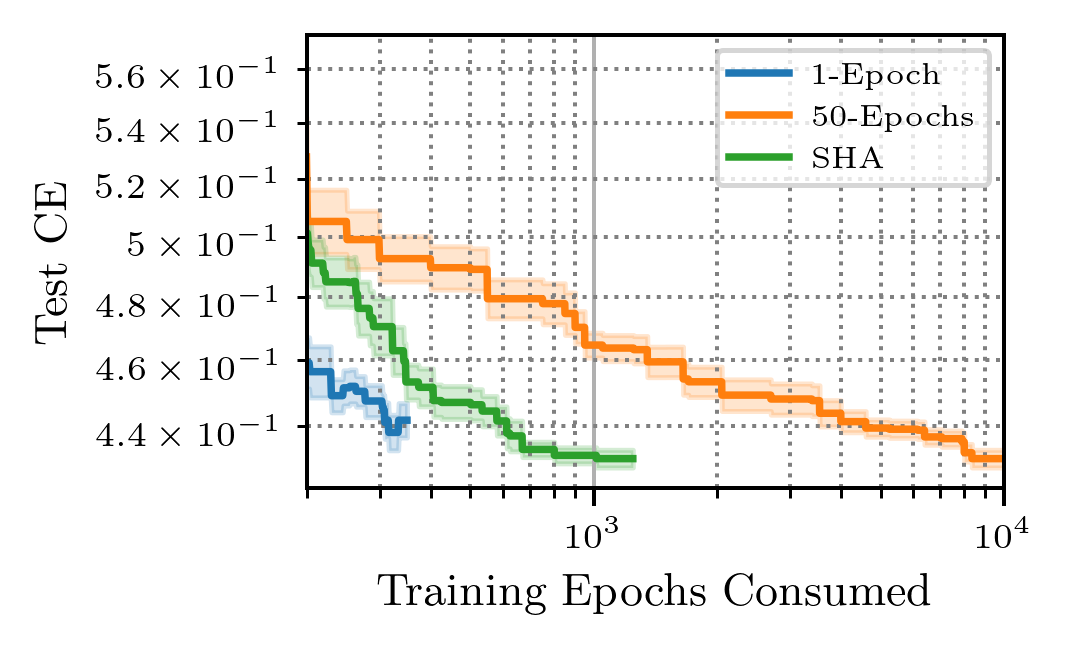}
        \caption{MiniBooNE}
        \label{fig:pruners-MiniBooNE-lcbench}
    \end{subfigure}
    \begin{subfigure}[b]{0.24\textwidth}
        \centering
        \includegraphics[width=\textwidth]{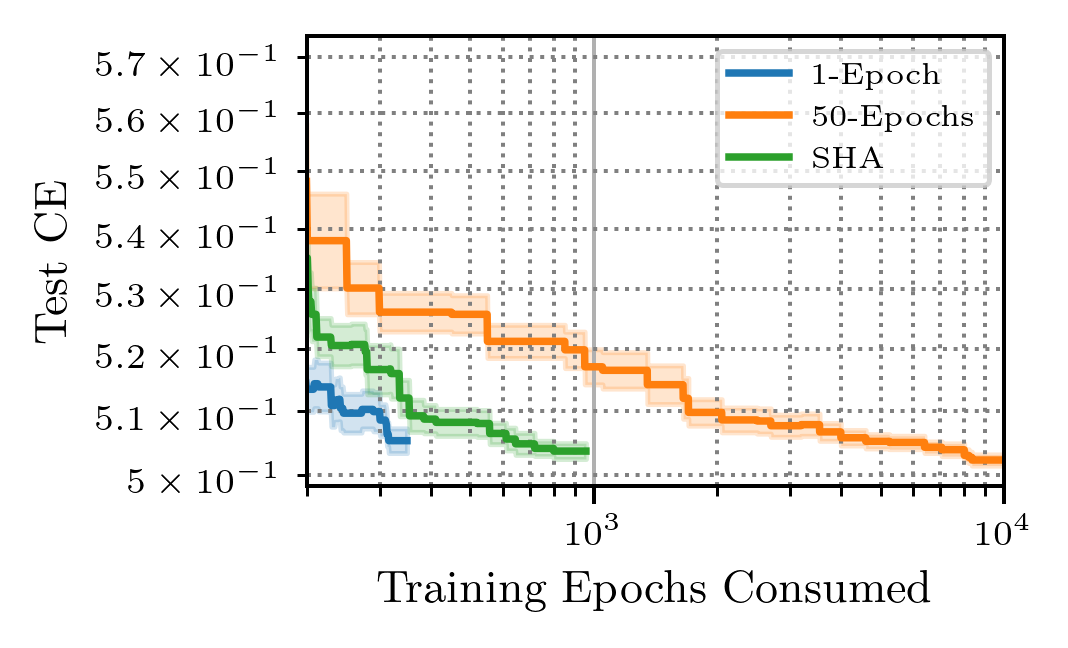}
        \caption{adult}
        \label{fig:pruners-adult-lcbench}
    \end{subfigure}
    \begin{subfigure}[b]{0.24\textwidth}
        \centering
        \includegraphics[width=\textwidth]{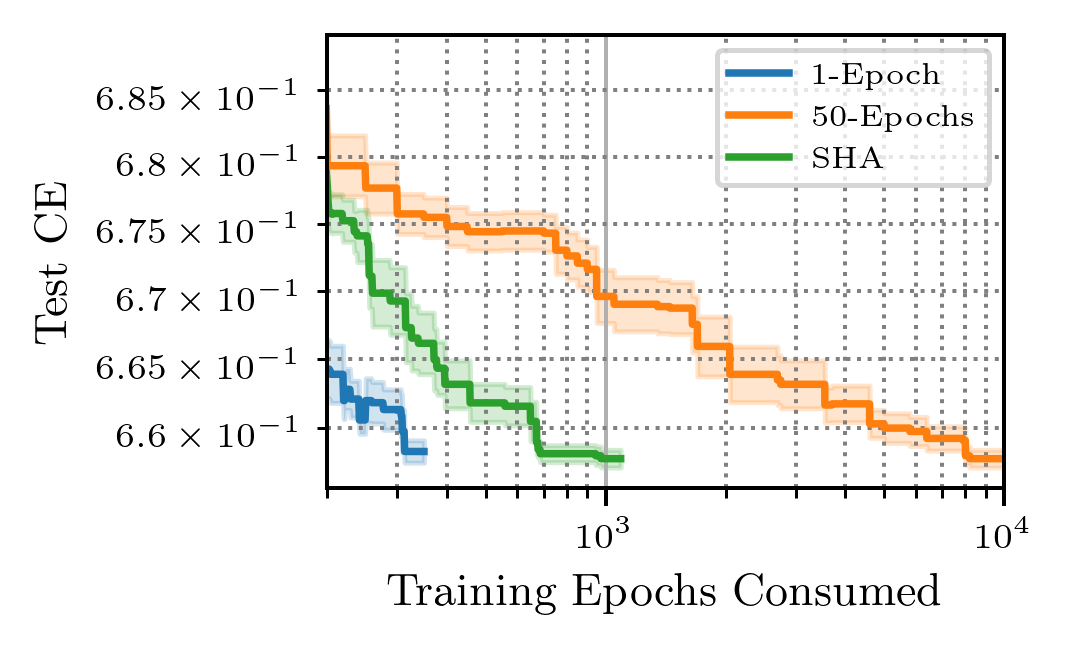}
        \caption{airlines}
        \label{fig:pruners-airlines-lcbench}
    \end{subfigure}
    \begin{subfigure}[b]{0.24\textwidth}
        \centering
        \includegraphics[width=\textwidth]{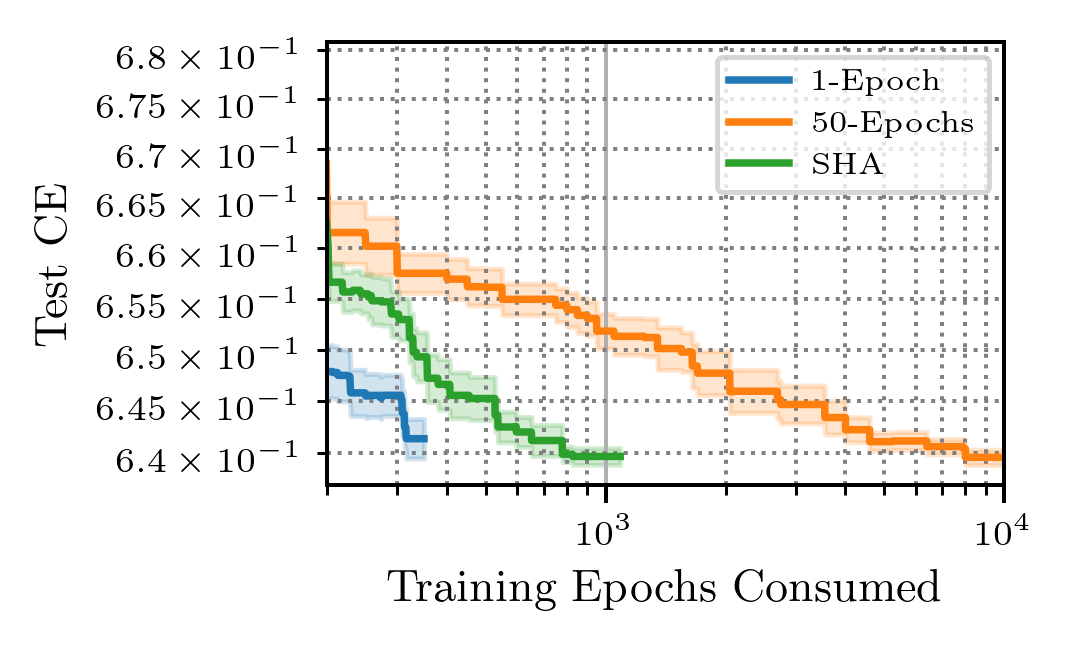}
        \caption{albert}
        \label{fig:pruners-albert-lcbench}
    \end{subfigure}
    \begin{subfigure}[b]{0.24\textwidth}
        \centering
        \includegraphics[width=\textwidth]{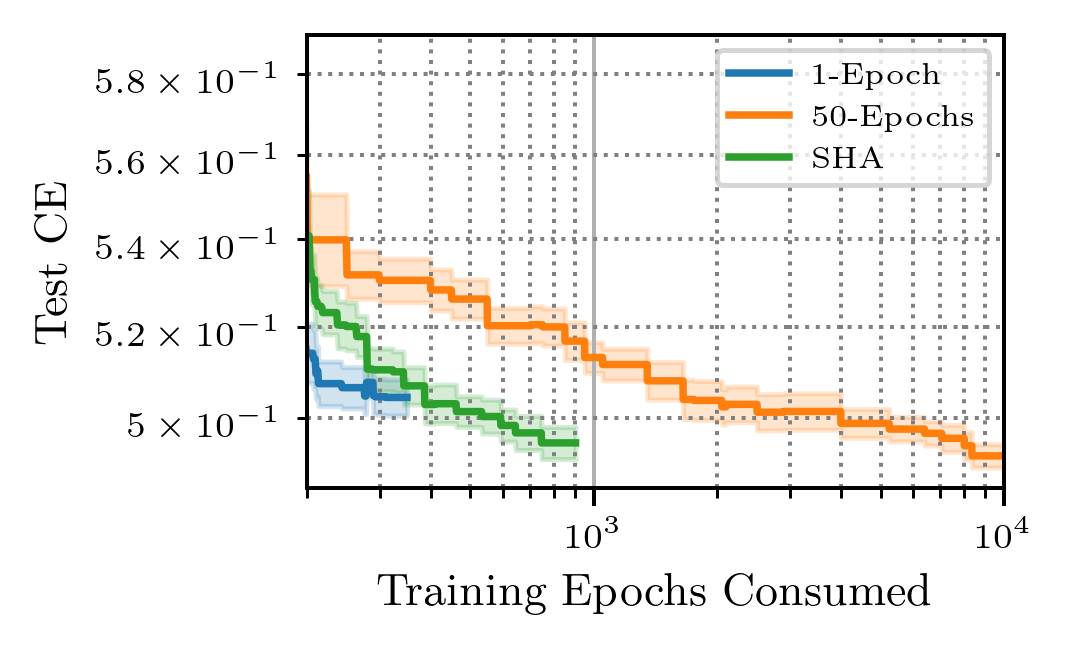}
        \caption{bank marketing}
        \label{fig:pruners-bank-marketing-lcbench}
    \end{subfigure}
    \begin{subfigure}[b]{0.24\textwidth}
        \centering
        \includegraphics[width=\textwidth]{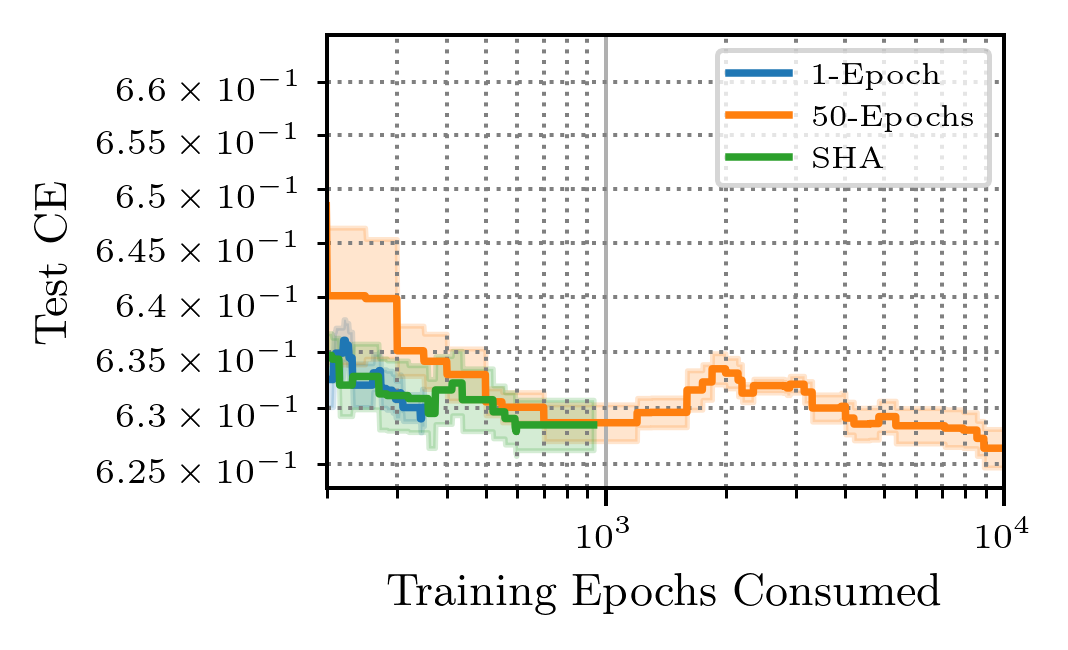}
        \caption{blood transfusion}
        \label{fig:pruners-blood-transfusion-service-center-lcbench}
    \end{subfigure}
    \begin{subfigure}[b]{0.24\textwidth}
        \centering
        \includegraphics[width=\textwidth]{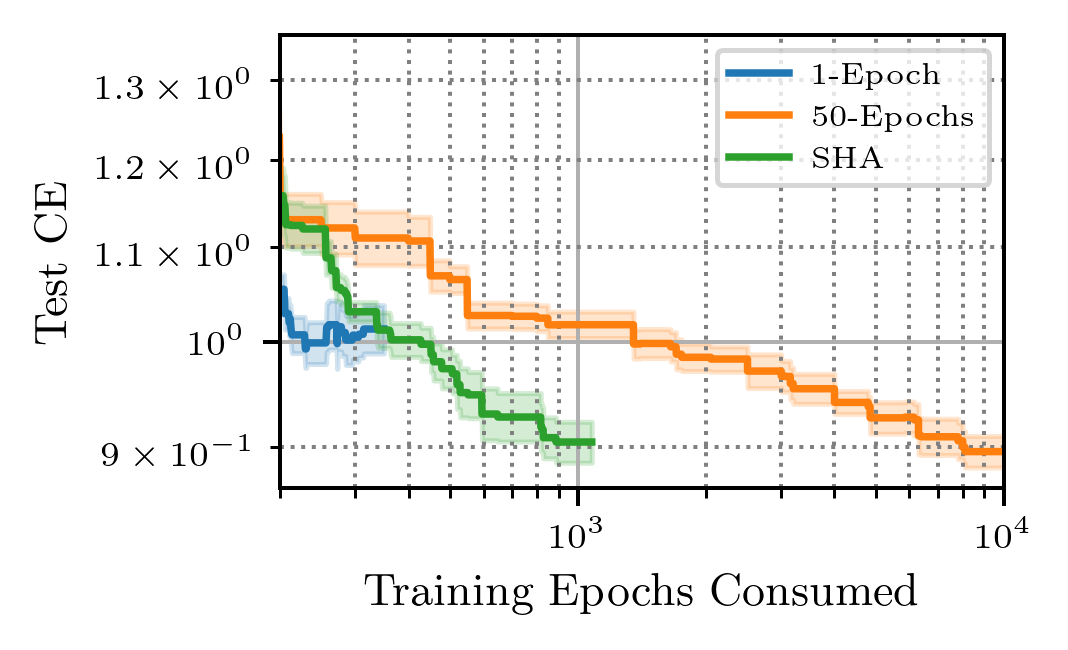}
        \caption{car}
        \label{fig:pruners-car-lcbench}
    \end{subfigure}
    \begin{subfigure}[b]{0.24\textwidth}
        \centering
        \includegraphics[width=\textwidth]{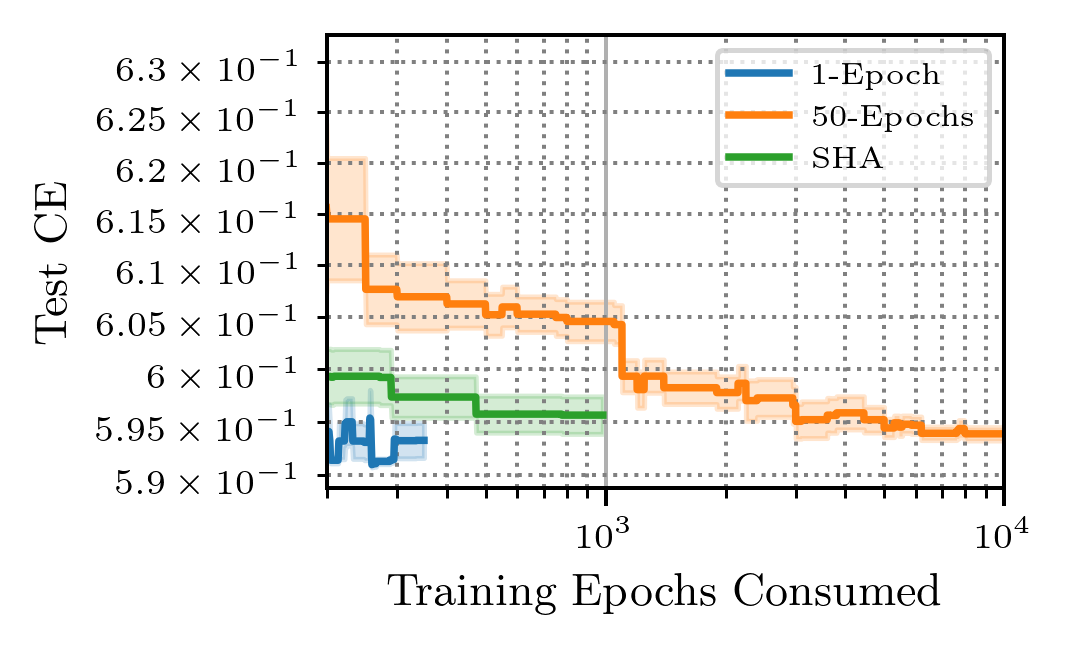}
        \caption{christine}
        \label{fig:pruners-christine-lcbench}
    \end{subfigure}
    \begin{subfigure}[b]{0.24\textwidth}
        \centering
        \includegraphics[width=\textwidth]{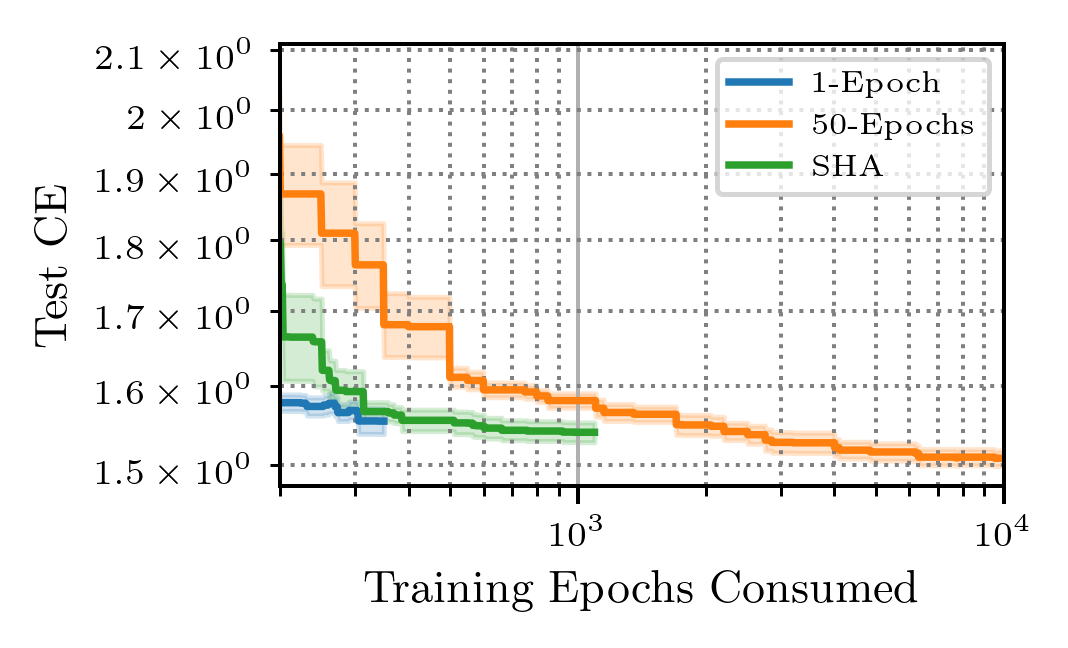}
        \caption{cnae 9}
        \label{fig:pruners-cnae-9-lcbench}
    \end{subfigure}
    \begin{subfigure}[b]{0.24\textwidth}
        \centering
        \includegraphics[width=\textwidth]{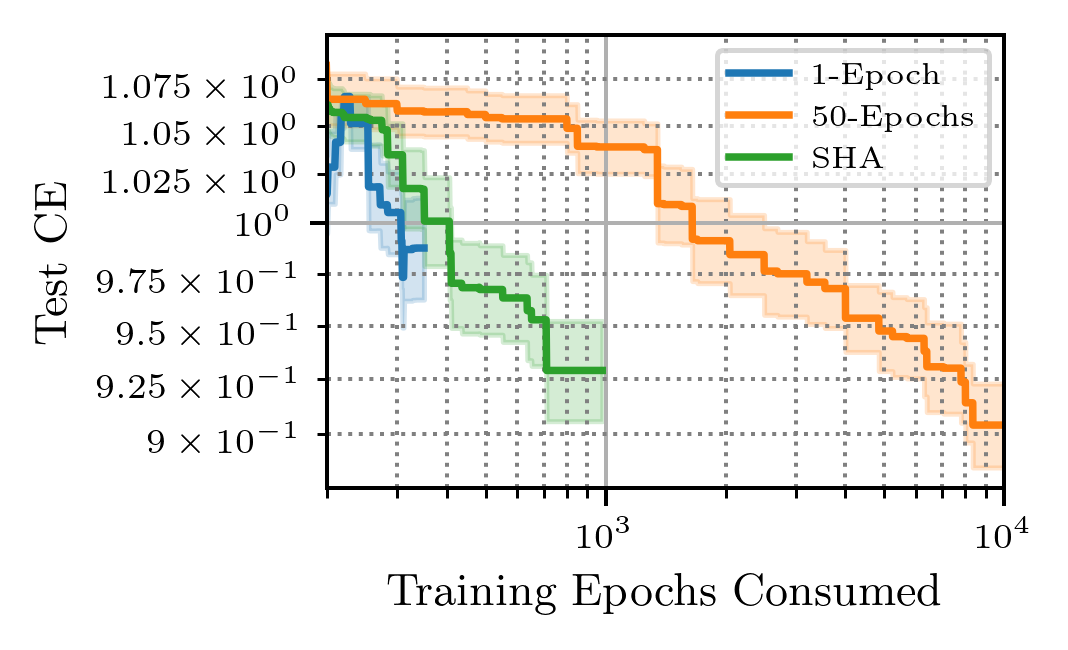}
        \caption{connect 4}
        \label{fig:pruners-connect-4-lcbench}
    \end{subfigure}
    \begin{subfigure}[b]{0.24\textwidth}
        \centering
        \includegraphics[width=\textwidth]{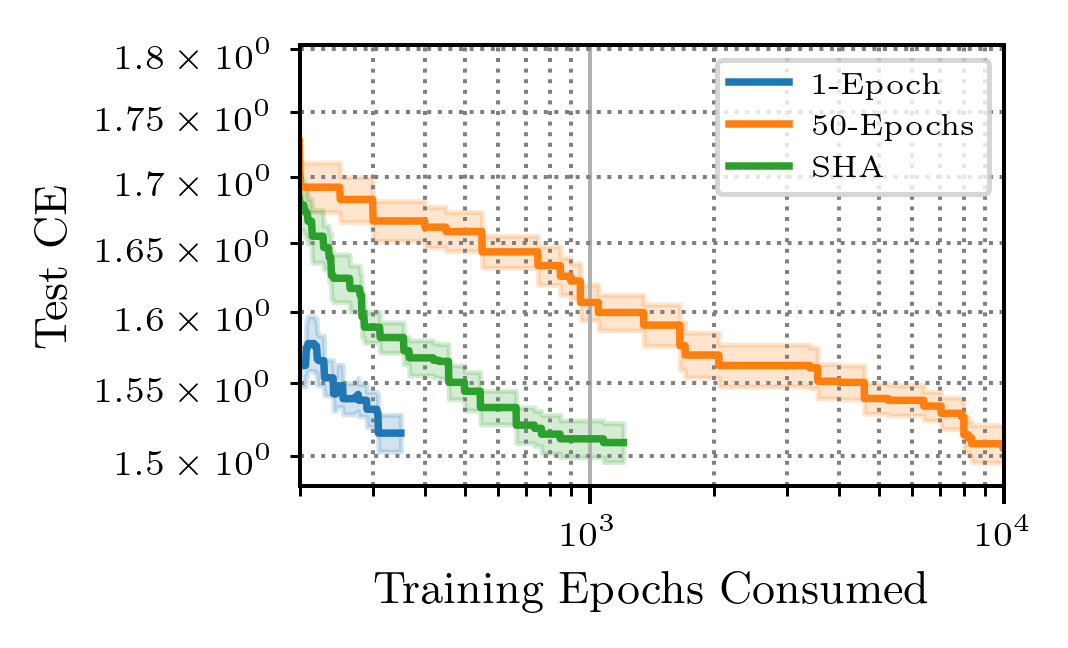}
        \caption{covertype}
        \label{fig:pruners-covertype-lcbench}
    \end{subfigure}
    \begin{subfigure}[b]{0.24\textwidth}
        \centering
        \includegraphics[width=\textwidth]{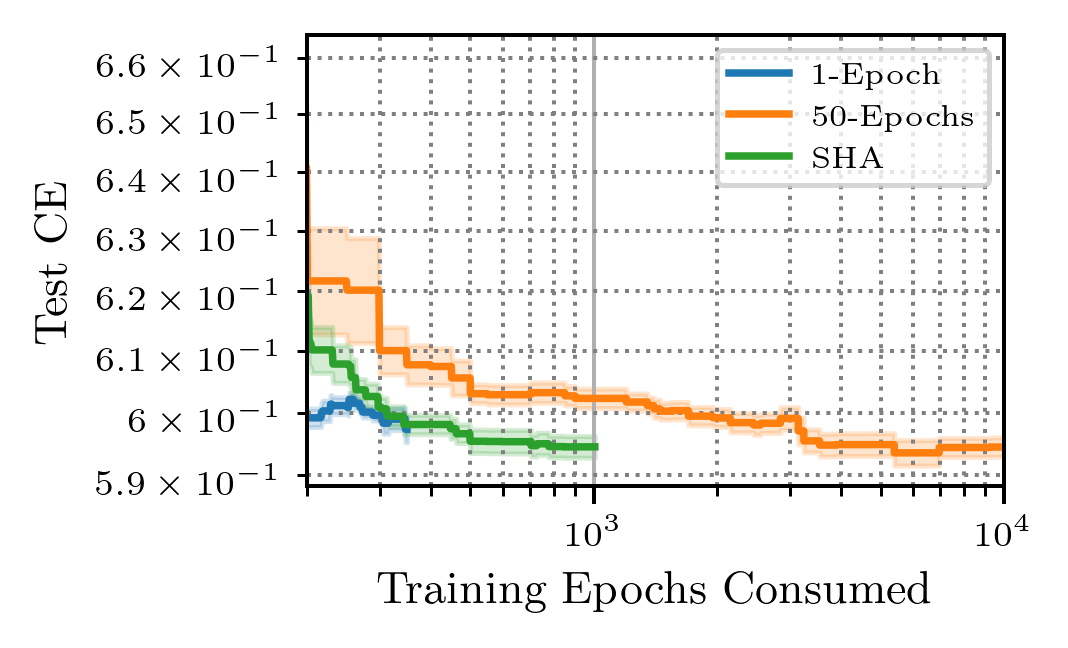}
        \caption{credit g}
        \label{fig:pruners-credit-g-lcbench}
    \end{subfigure}
    \begin{subfigure}[b]{0.24\textwidth}
        \centering
        \includegraphics[width=\textwidth]{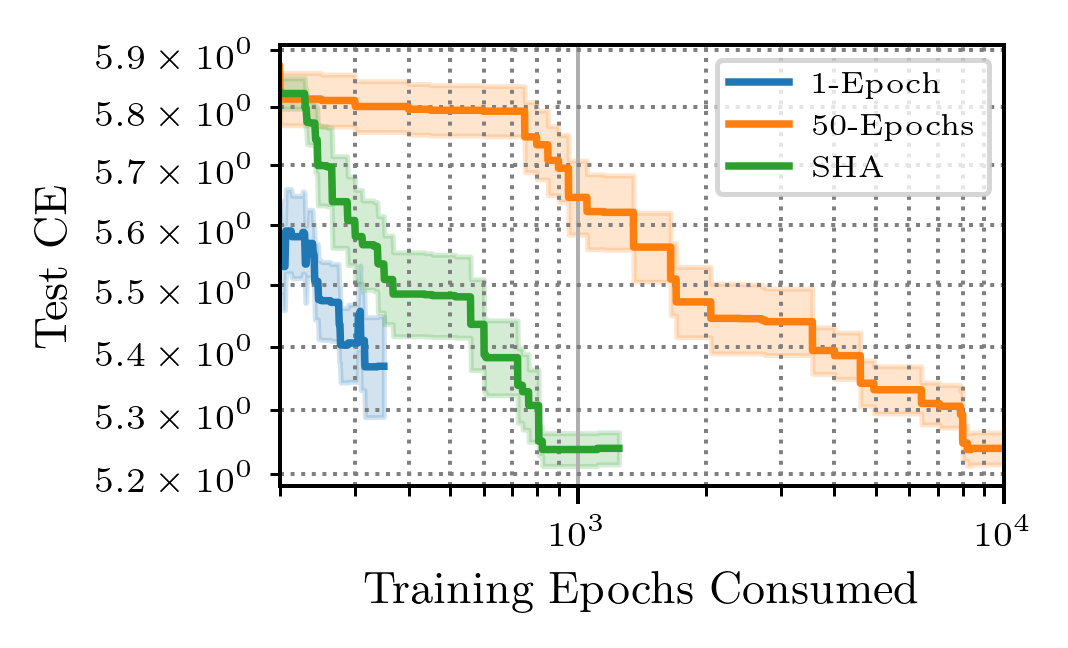}
        \caption{dionis}
        \label{fig:pruners-dionis-lcbench}
    \end{subfigure}
    \begin{subfigure}[b]{0.24\textwidth}
        \centering
        \includegraphics[width=\textwidth]{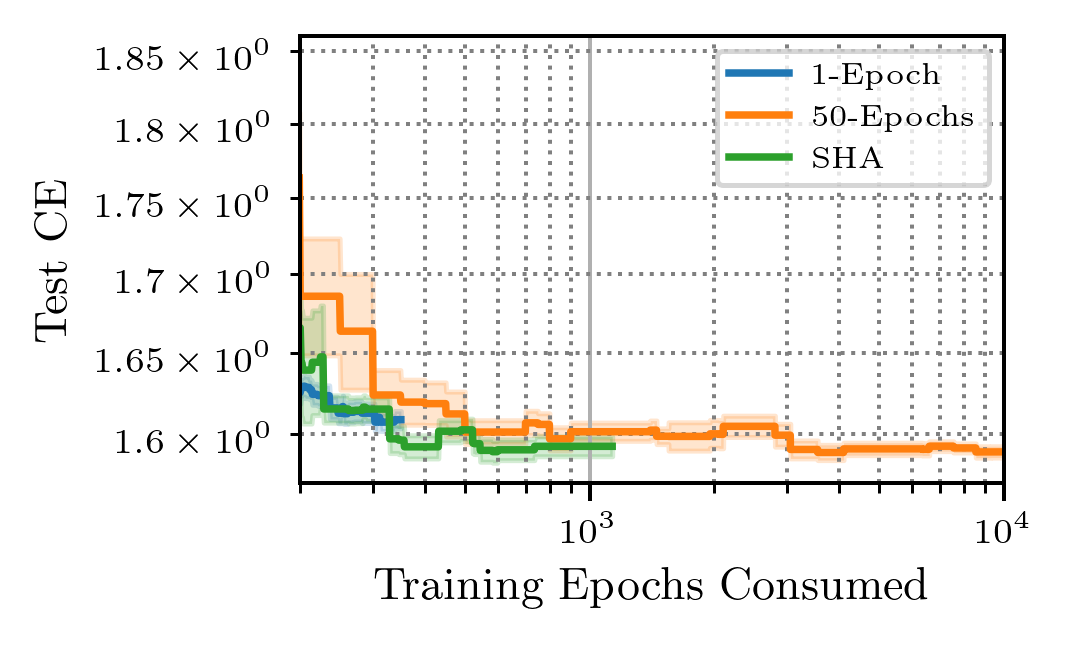}
        \caption{fabert}
        \label{fig:pruners-fabert-lcbench}
    \end{subfigure}
    \begin{subfigure}[b]{0.24\textwidth}
        \centering
        \includegraphics[width=\textwidth]{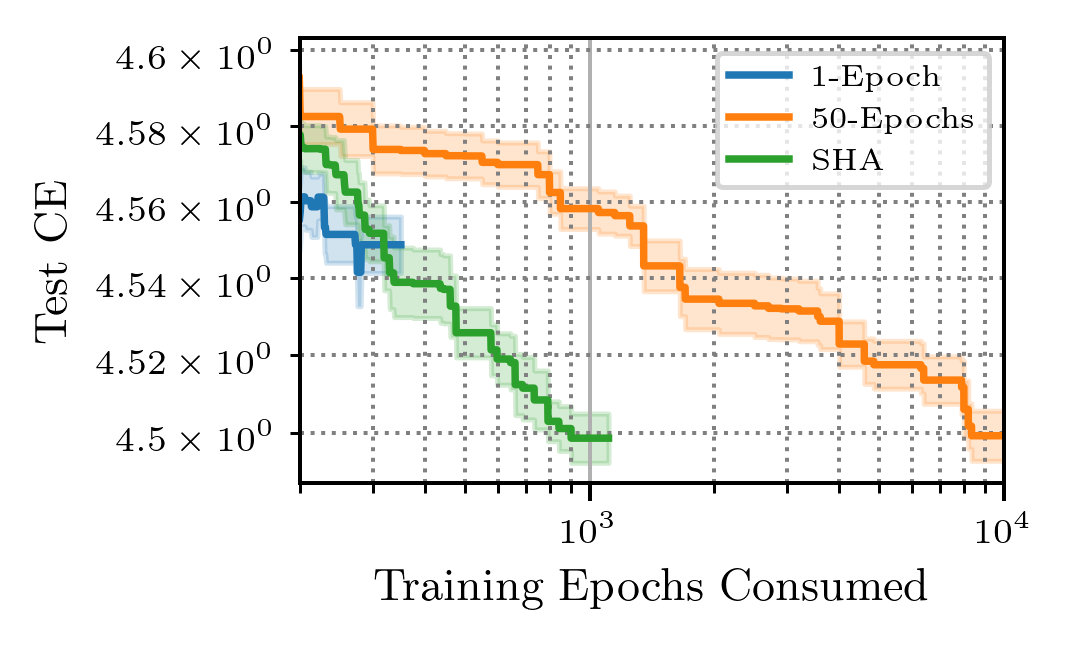}
        \caption{helena}
        \label{fig:pruners-helena-lcbench}
    \end{subfigure}
    \begin{subfigure}[b]{0.24\textwidth}
        \centering
        \includegraphics[width=\textwidth]{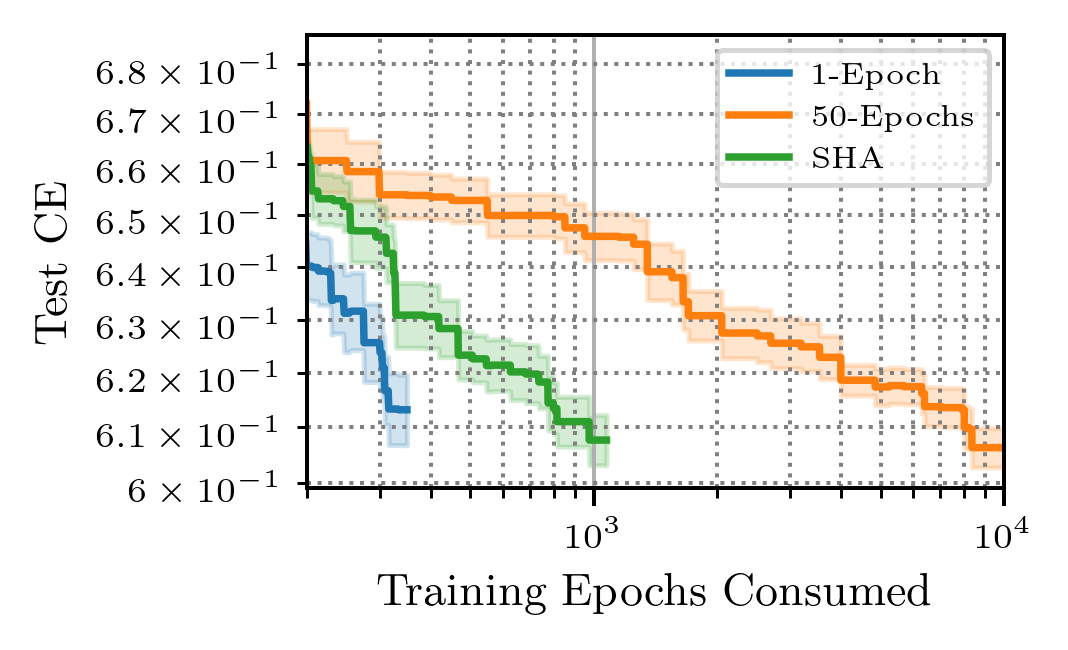}
        \caption{higgs}
        \label{fig:pruners-higgs-lcbench}
    \end{subfigure}
    \begin{subfigure}[b]{0.24\textwidth}
        \centering
        \includegraphics[width=\textwidth]{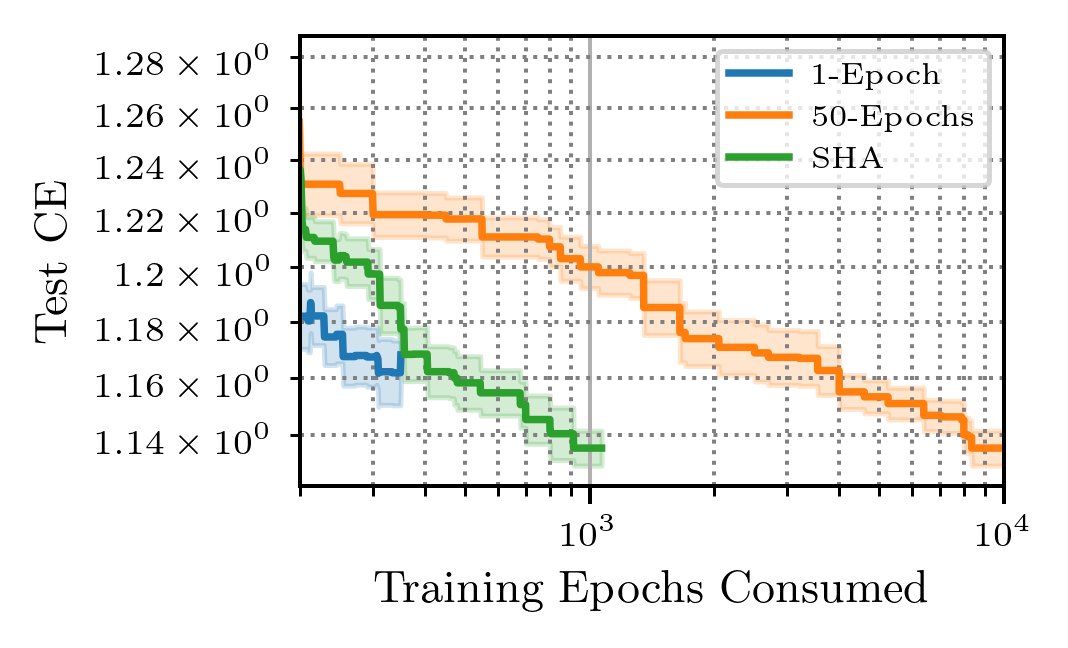}
        \caption{jannis}
        \label{fig:pruners-jannis-lcbench}
    \end{subfigure}
    \begin{subfigure}[b]{0.24\textwidth}
        \centering
        \includegraphics[width=\textwidth]{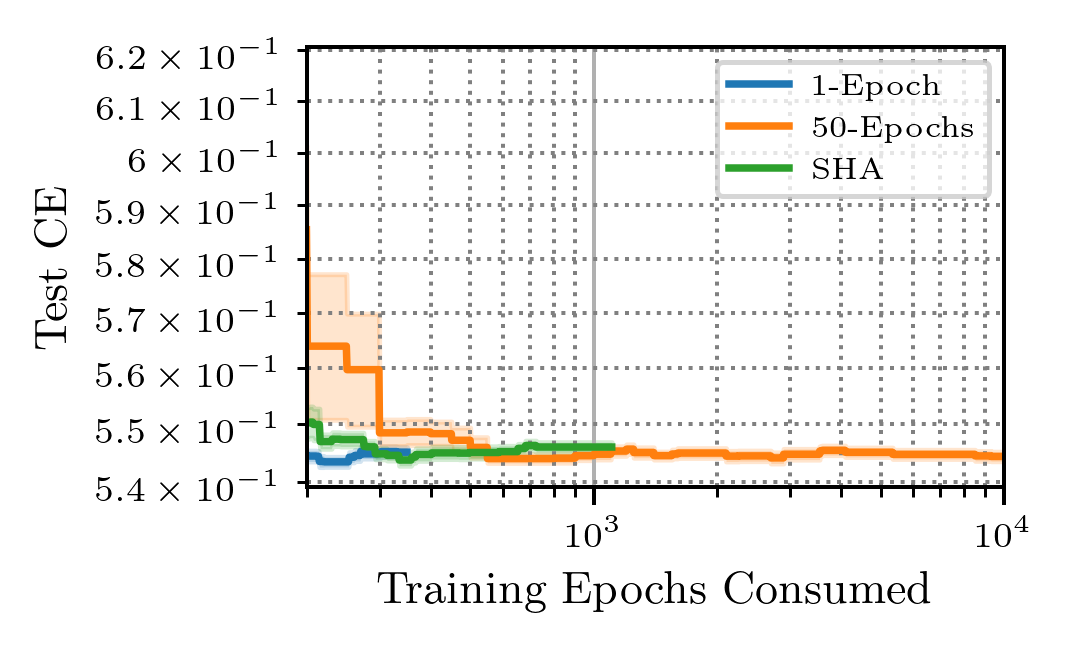}
        \caption{jasmine}
        \label{fig:pruners-jasmine-lcbench}
    \end{subfigure}
    \begin{subfigure}[b]{0.24\textwidth}
        \centering
        \includegraphics[width=\textwidth]{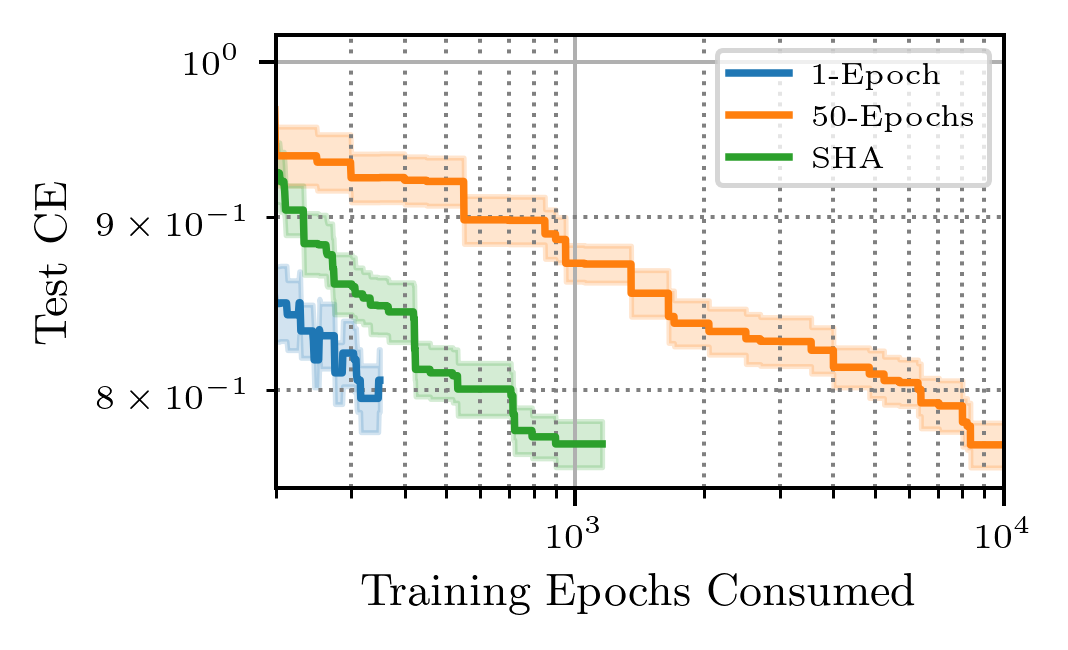}
        \caption{jungle chess 2pcs raw endgame complete}
        \label{fig:pruners-jungle_chess_2pcs_raw_endgame_complete-lcbench}
    \end{subfigure}
    \begin{subfigure}[b]{0.24\textwidth}
        \centering
        \includegraphics[width=\textwidth]{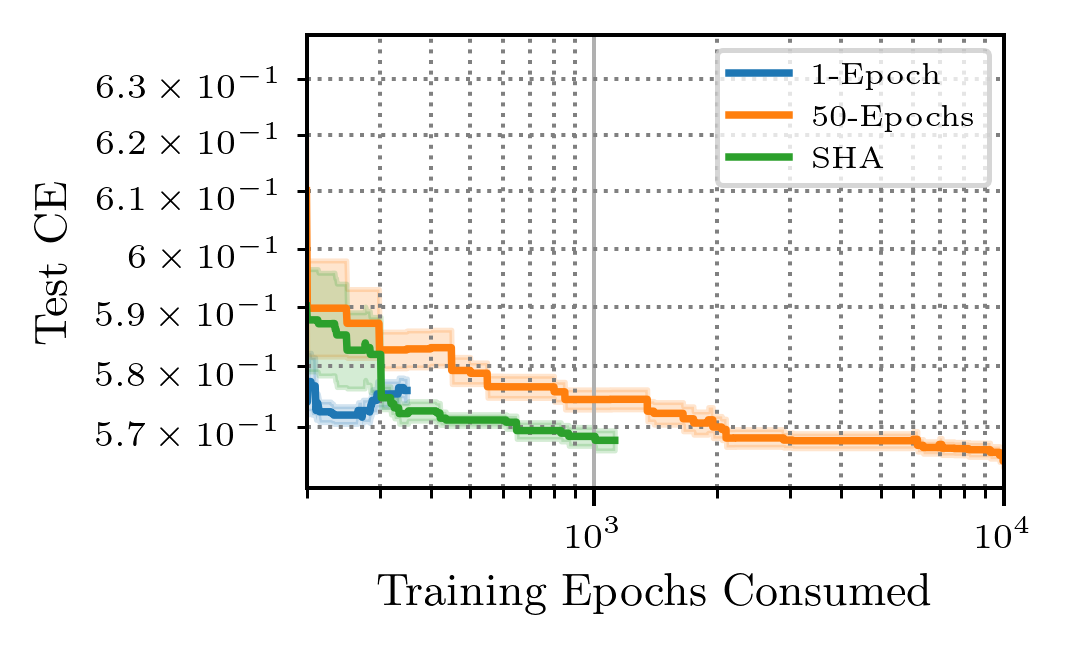}
        \caption{kc1}
        \label{fig:pruners-kc1-lcbench}
    \end{subfigure}
    \begin{subfigure}[b]{0.24\textwidth}
        \centering
        \includegraphics[width=\textwidth]{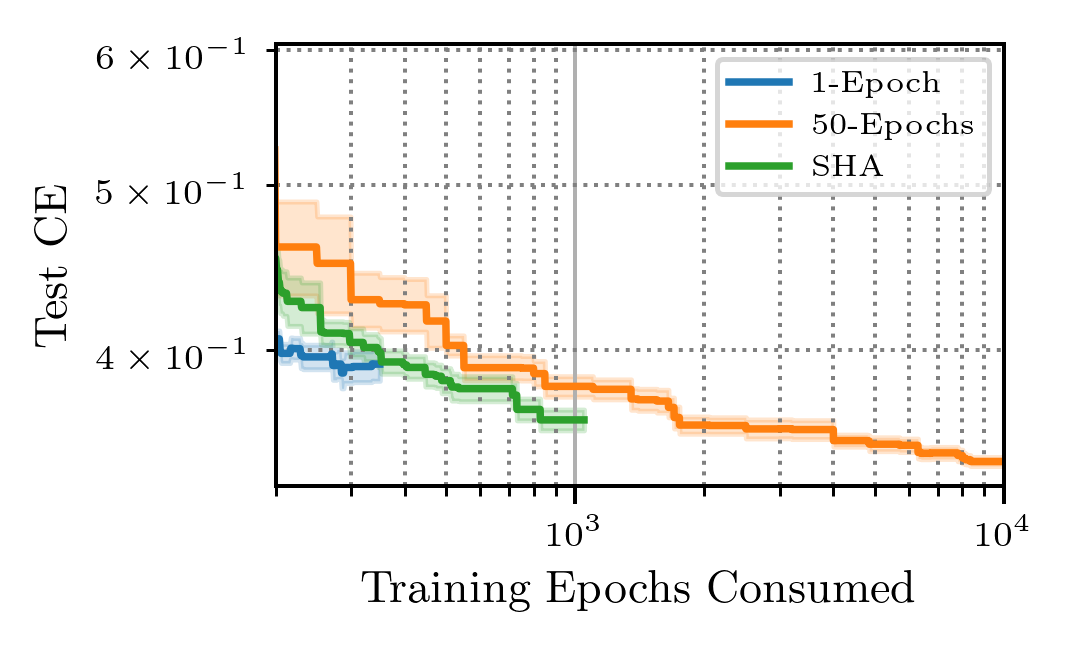}
        \caption{kr vs kp}
        \label{fig:pruners-kr-vs-kp-lcbench}
    \end{subfigure}
    
    \caption{Comparing pruners 1-Epoch, 50-Epochs and SHA on the LCBench benchmark ({\bf First Part}).}
    \label{fig:pruners-lcbench}
\end{figure}

\begin{figure}[!h]
    \centering
    
    \begin{subfigure}[b]{0.24\textwidth}
        \centering
        \includegraphics[width=\textwidth]{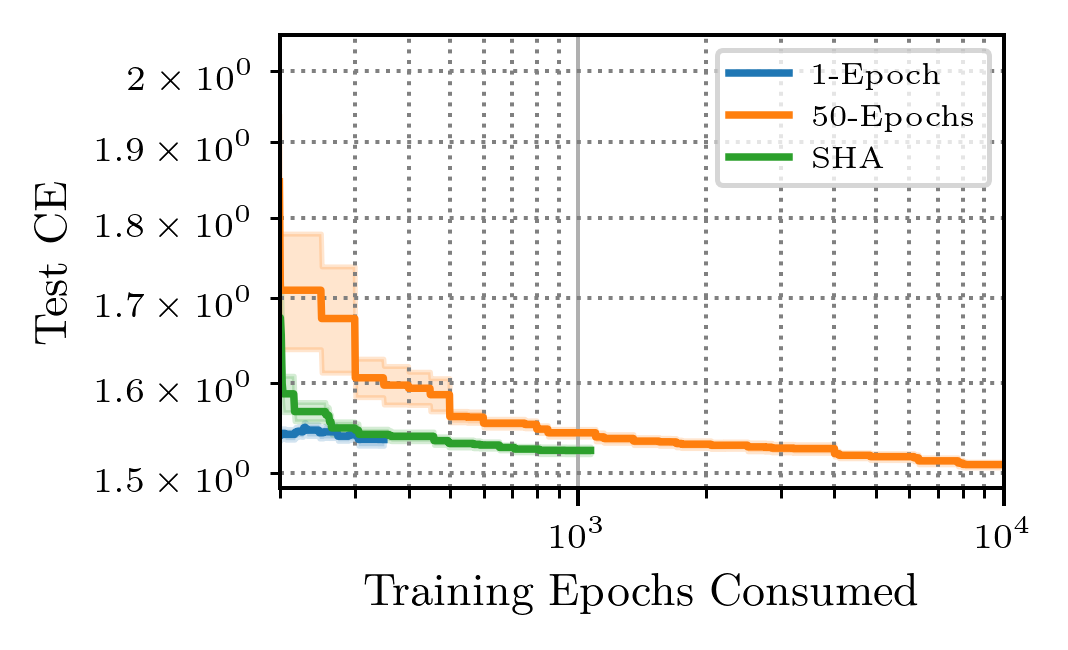}
        \caption{mfeat factors}
        \label{fig:pruners-mfeat-factors-lcbench}
    \end{subfigure}
    \begin{subfigure}[b]{0.24\textwidth}
        \centering
        \includegraphics[width=\textwidth]{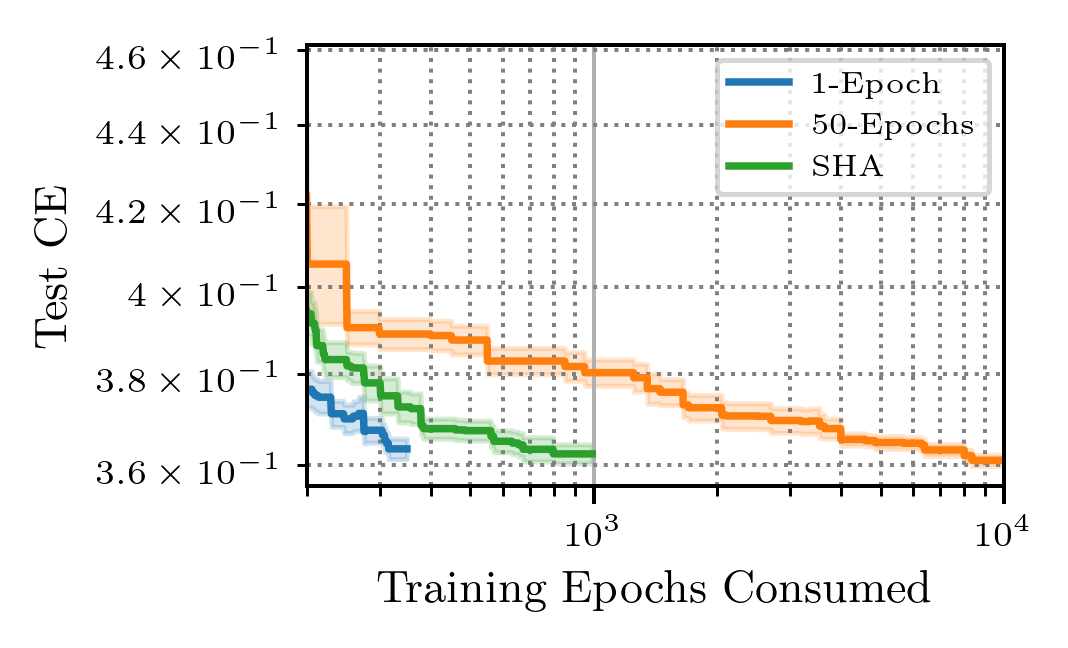}
        \caption{nomao}
        \label{fig:pruners-nomao-lcbench}
    \end{subfigure}
    \begin{subfigure}[b]{0.24\textwidth}
        \centering
        \includegraphics[width=\textwidth]{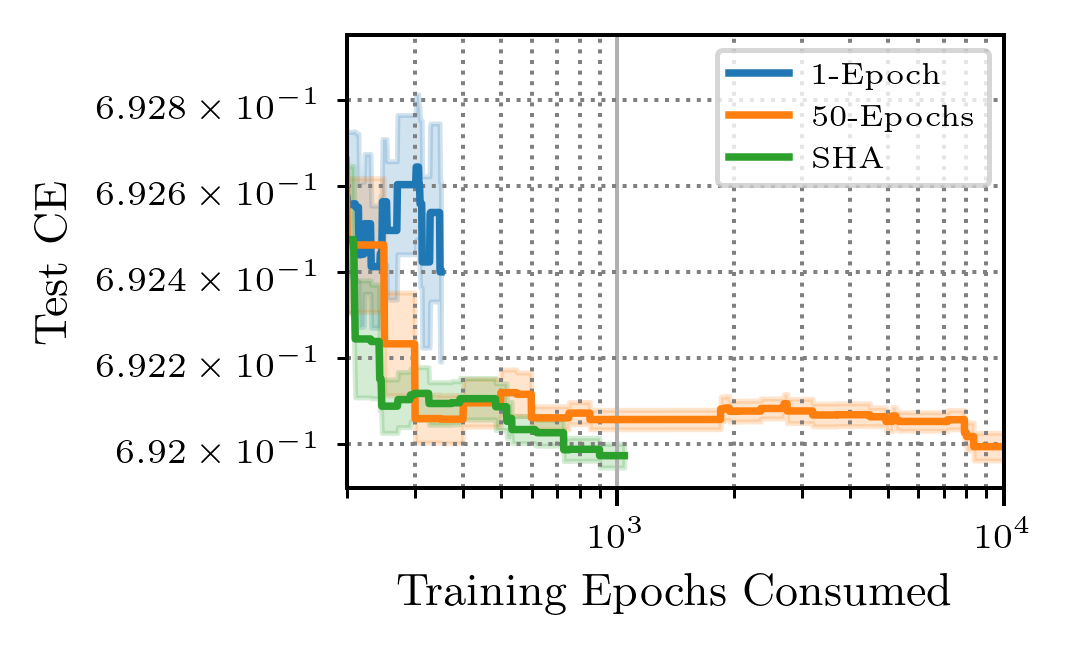}
        \caption{numerai28.6}
        \label{fig:pruners-numerai28.6-lcbench}
    \end{subfigure}
    \begin{subfigure}[b]{0.24\textwidth}
        \centering
        \includegraphics[width=\textwidth]{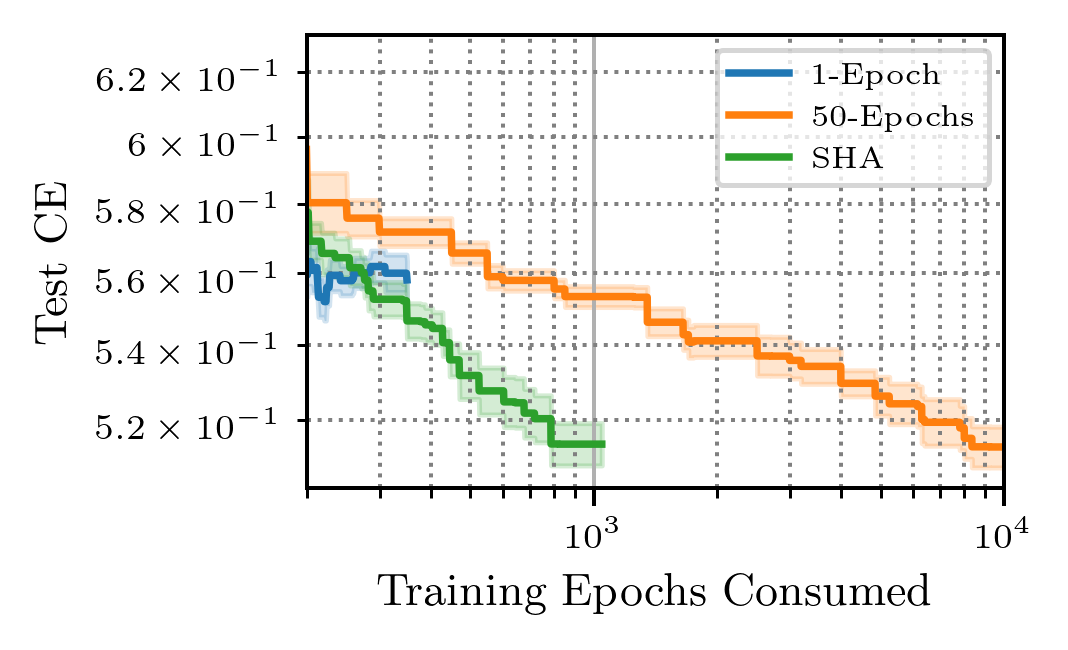}
        \caption{phoneme}
        \label{fig:pruners-phoneme-lcbench}
    \end{subfigure}
    \begin{subfigure}[b]{0.24\textwidth}
        \centering
        \includegraphics[width=\textwidth]{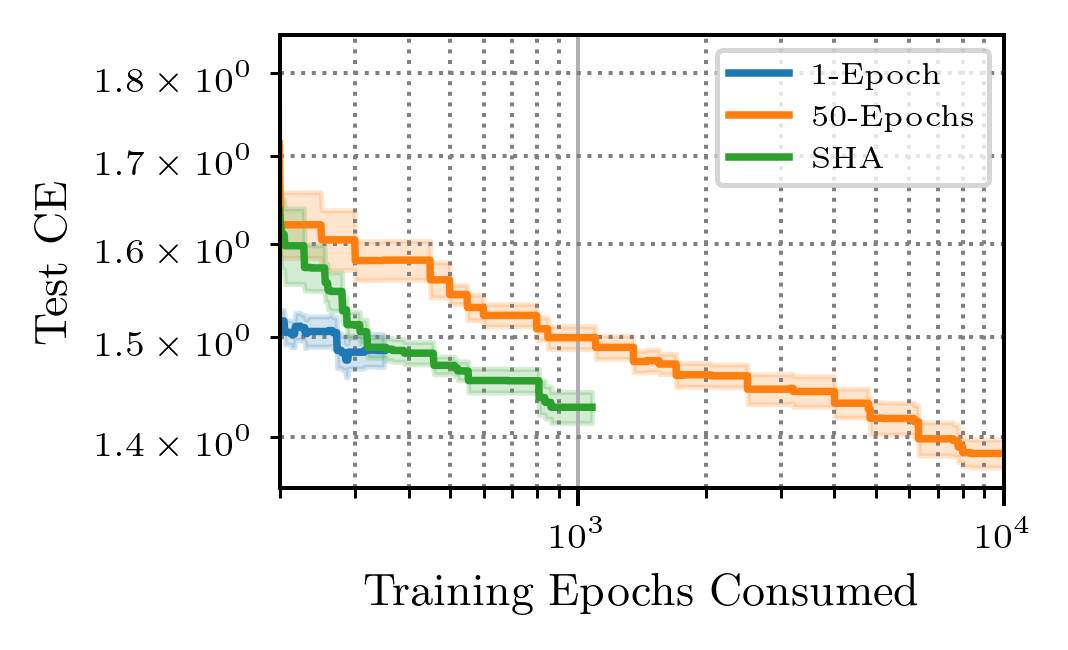}
        \caption{segment}
        \label{fig:pruners-segment-lcbench}
    \end{subfigure}
    \begin{subfigure}[b]{0.24\textwidth}
        \centering
        \includegraphics[width=\textwidth]{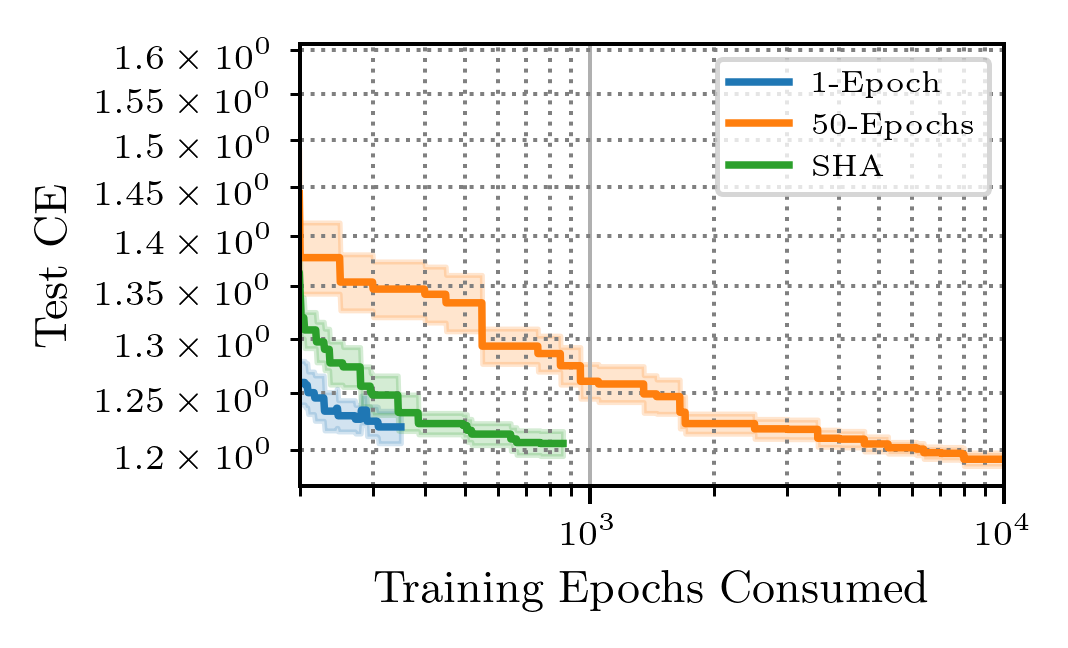}
        \caption{shuttle}
        \label{fig:pruners-shuttle-lcbench}
    \end{subfigure}
    \begin{subfigure}[b]{0.24\textwidth}
        \centering
        \includegraphics[width=\textwidth]{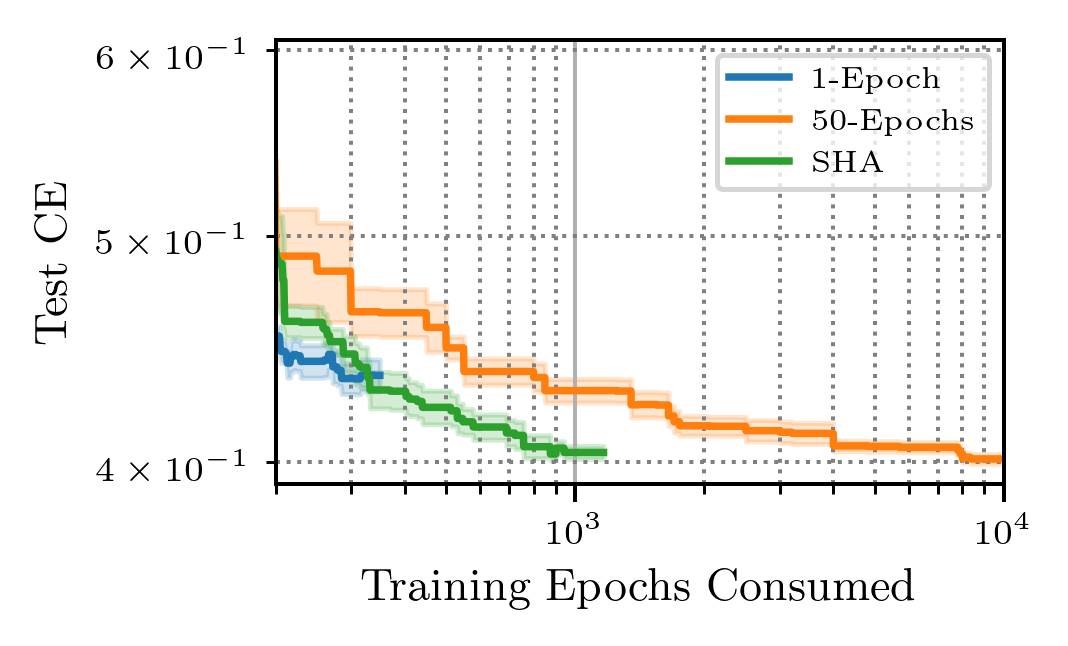}
        \caption{sylvine}
        \label{fig:pruners-sylvine-lcbench}
    \end{subfigure}
    \begin{subfigure}[b]{0.24\textwidth}
        \centering
        \includegraphics[width=\textwidth]{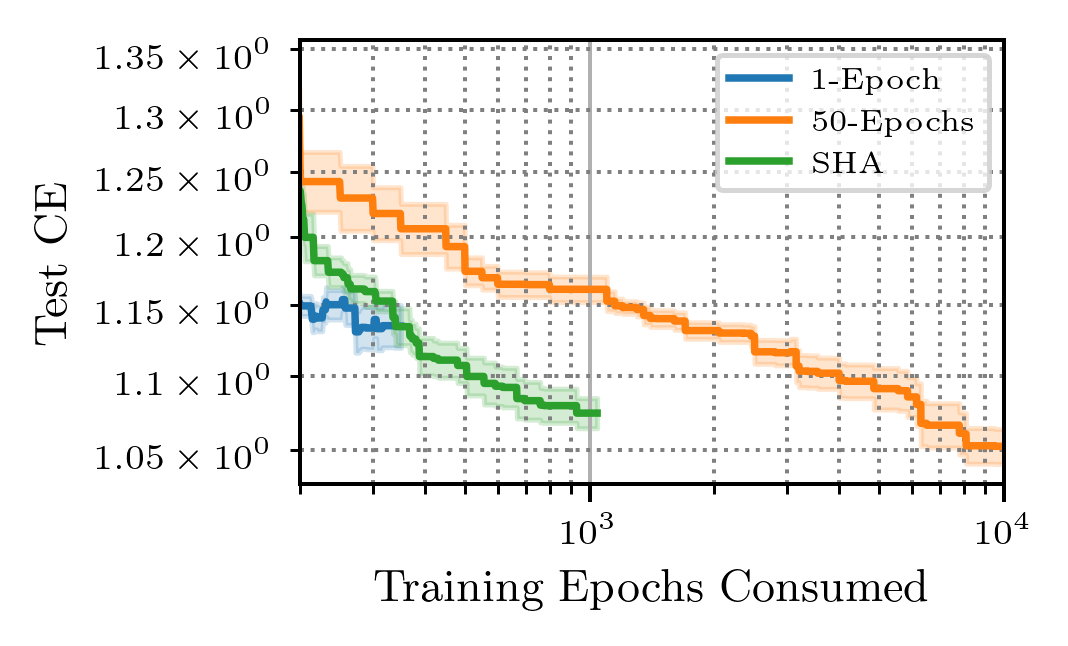}
        \caption{vehicle}
        \label{fig:pruners-vehicle-lcbench}
    \end{subfigure}
    \begin{subfigure}[b]{0.24\textwidth}
        \centering
        \includegraphics[width=\textwidth]{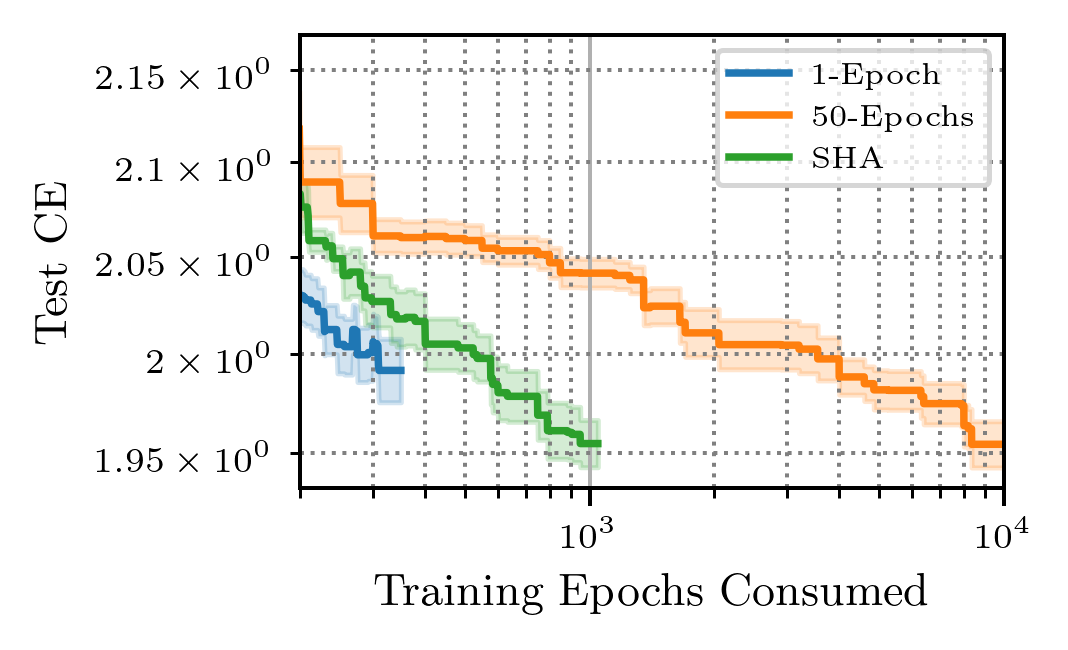}
        \caption{volkert}
        \label{fig:pruners-volkert-lcbench}
    \end{subfigure}
    
    \caption{Comparing pruners 1-Epoch, 50-Epochs and SHA on the LCBench benchmark ({\bf Second Part}).}
    \label{fig:pruners-lcbench-second-part}
\end{figure}

In this section, we compare the performance of our 1-Epoch baseline on the LCBench~\cite{klein_tabular_2019} benchmark. The experimental setting is similar to the experiments on HPOBench. A random sampling is used for the outer-loop optimization. The minimum fidelity on LCbench is 1 epoch of training, and the maximum is 50 epochs of training. The outer-loop is executed for 200 iterations. The Top-$K$ selection is run with $k=3$ for 1-Epoch. Table~\ref{tab:results-lcbench-random-search} summarizes the results of final test objective (in this case cross-entropy) as well as the final speed up in term of consumed training epochs. To be transpartent about the behaviour of the algorith we also display the search trajectories in Figure~\ref{fig:pruners-lcbench} and \ref{fig:pruners-lcbench-second-part}. As it can be observed, the test objectives found are close in general dispite a large difference in speed-up. Therefore, if the main bottleneck to execute hyperparameter optimization is compute, using 1-Epoch can be justified.

\begin{table}[!h]
\centering
\resizebox{\textwidth}{!}{%
\begin{tabular}{llllrr}
\toprule
               Dataset &     1-Epoch &         SHA &   50-Epochs &  S(1-Epochs) &  S(SHA) \\
\midrule
            APSFailure & 0.34 ± 0.00 & 0.33 ± 0.00 & 0.33 ± 0.00 &        28.71 &   12.29 \\
Amazon employee access & 0.59 ± 0.00 & 0.58 ± 0.00 & 0.58 ± 0.00 &        28.71 &   11.92 \\
            Australian & 0.48 ± 0.00 & 0.48 ± 0.00 & 0.47 ± 0.00 &        28.71 &   12.23 \\
         Fashion MNIST & 1.61 ± 0.01 & 1.60 ± 0.00 & 1.60 ± 0.00 &        28.71 &   11.10 \\
    KDDCup09 appetency & 0.41 ± 0.01 & 0.40 ± 0.01 & 0.37 ± 0.01 &        28.71 &   11.77 \\
             MiniBooNE & 0.44 ± 0.01 & 0.43 ± 0.00 & 0.43 ± 0.00 &        28.71 &   11.30 \\
                 adult & 0.51 ± 0.00 & 0.50 ± 0.00 & 0.50 ± 0.00 &        28.71 &   11.99 \\
              airlines & 0.66 ± 0.00 & 0.66 ± 0.00 & 0.66 ± 0.00 &        28.71 &   11.27 \\
                albert & 0.64 ± 0.00 & 0.64 ± 0.00 & 0.64 ± 0.00 &        28.71 &   11.59 \\
        bank marketing & 0.50 ± 0.00 & 0.49 ± 0.00 & 0.49 ± 0.00 &        28.71 &   13.31 \\
     blood transfusion & 0.63 ± 0.00 & 0.63 ± 0.00 & 0.63 ± 0.00 &        28.71 &   13.12 \\
                   car & 1.01 ± 0.02 & 0.90 ± 0.02 & 0.90 ± 0.01 &        28.71 &   11.40 \\
             christine & 0.59 ± 0.00 & 0.60 ± 0.00 & 0.59 ± 0.00 &        28.71 &   12.84 \\
                cnae 9 & 1.55 ± 0.02 & 1.54 ± 0.01 & 1.51 ± 0.01 &        28.71 &   11.85 \\
             connect 4 & 0.99 ± 0.03 & 0.93 ± 0.02 & 0.90 ± 0.02 &        28.71 &   11.66 \\
             covertype & 1.52 ± 0.01 & 1.51 ± 0.01 & 1.50 ± 0.01 &        28.71 &   12.13 \\
              credit g & 0.60 ± 0.00 & 0.59 ± 0.00 & 0.59 ± 0.00 &        28.71 &   12.28 \\
                dionis & 5.37 ± 0.08 & 5.24 ± 0.02 & 5.24 ± 0.02 &        28.71 &   11.06 \\
                fabert & 1.61 ± 0.01 & 1.59 ± 0.01 & 1.59 ± 0.00 &        28.71 &   12.54 \\
                helena & 4.55 ± 0.01 & 4.50 ± 0.01 & 4.50 ± 0.01 &        28.71 &   11.19 \\
                 higgs & 0.61 ± 0.01 & 0.61 ± 0.00 & 0.61 ± 0.00 &        28.71 &   11.10 \\
                jannis & 1.17 ± 0.01 & 1.14 ± 0.01 & 1.14 ± 0.01 &        28.71 &   11.53 \\
               jasmine & 0.55 ± 0.00 & 0.55 ± 0.00 & 0.54 ± 0.00 &        28.71 &   12.01 \\
          jungle chess & 0.81 ± 0.02 & 0.77 ± 0.01 & 0.77 ± 0.01 &        28.71 &   12.30 \\
                   kc1 & 0.58 ± 0.00 & 0.57 ± 0.00 & 0.56 ± 0.00 &        28.71 &   11.51 \\
              kr vs kp & 0.39 ± 0.01 & 0.36 ± 0.01 & 0.34 ± 0.00 &        28.71 &   11.96 \\
         mfeat factors & 1.54 ± 0.01 & 1.52 ± 0.00 & 1.51 ± 0.00 &        28.71 &   11.29 \\
                 nomao & 0.36 ± 0.00 & 0.36 ± 0.00 & 0.36 ± 0.00 &        28.71 &   12.33 \\
           numerai28.6 & 0.69 ± 0.00 & 0.69 ± 0.00 & 0.69 ± 0.00 &        28.71 &   12.37 \\
               phoneme & 0.56 ± 0.01 & 0.51 ± 0.01 & 0.51 ± 0.01 &        28.71 &   11.75 \\
               segment & 1.49 ± 0.02 & 1.43 ± 0.02 & 1.38 ± 0.01 &        28.71 &   11.20 \\
               shuttle & 1.22 ± 0.01 & 1.21 ± 0.01 & 1.19 ± 0.01 &        28.71 &   13.23 \\
               sylvine & 0.44 ± 0.01 & 0.40 ± 0.00 & 0.40 ± 0.00 &        28.71 &   11.64 \\
               vehicle & 1.13 ± 0.02 & 1.07 ± 0.01 & 1.05 ± 0.01 &        28.71 &   11.47 \\
               volkert & 1.99 ± 0.02 & 1.95 ± 0.01 & 1.95 ± 0.01 &        28.71 &   11.40 \\
\bottomrule
\end{tabular}
}
\caption{Table of results from the LCBench benchmark. The final test cross-entropy (mean ± standard error) are displayed as well as the speed up $S(.)$ for 1-Epoch and SHA with respect to using maximum fidelity (50 training epochs for this benchmark).}
\label{tab:results-lcbench-random-search}
\end{table}

\section{Details About HPOBench Search Space}

The search space for the 4 tabular benchmarks we used from HPOBench is given by~\cite{klein_tabular_2019} and detailed in table~\ref{tab:hpobench-tabular-search-space}. The backbone is a two-layers fully connected neural network with a linear output layer.

\begin{table}[!h]
\centering
\begin{tabular}{|c|c|}
\hline
\textbf{Hyperparameters} & \textbf{Choices}                            \\ \hline
Initial LR               & $\{0.0005, 0.001, 0.005, 0.01, 0.05, 0.1\}$ \\ \hline
Batch Size               & $\{8, 16, 32, 64\}$                         \\ \hline
LR Schedule              & $\{\text{cosine}, \text{fix}\}$             \\ \hline
Activation/Layer 1       & $\{\text{relu}, \text{tanh}\}$              \\ \hline
Activation/Layer 2       & $\{\text{relu}, \text{tanh}\}$              \\ \hline
Layer 1 Size             & $\{16, 32, 64, 128, 256, 512\}$             \\ \hline
Layer 2 Size             & $\{16, 32, 64, 128, 256, 512\}$             \\ \hline
Dropout/Layer 1          & $\{0.0, 0.3, 0.6\}$                         \\ \hline
Dropout/Layer 2          & $\{0.0, 0.3, 0.6\}$                         \\ \hline
\end{tabular}
\caption{Hyperparameter search space for the four tabular benchmarks from HPOBench.}
\label{tab:hpobench-tabular-search-space}
\end{table}

\section{Rober: Robust Bayesian Early Rejection}~\label{sec:appendix-rober}

In this section, we present shortly our improved version of learning curve extrapolation, we call it robust Bayesian early rejection (RoBER). The detailed algorithm of RoBER is provided in Algorithm~\ref{alg:rober-algorithm}. RoBER closely follows the proposed weighted probabilistic model (WPM) from~\cite{domhan_speeding_2015}. WPM consists in estimating the posterior distribution $P(\Theta|C)$ of a mixture distribution of parametric learning curves models, where $C=\{(z_1, y_1), ..., (z_i, y_i)\}$ is a partial learning curve of the observed scores and $\Theta$ is the concatenation of all the parameters (weights of the mixture and parameters of all learning curves models used in the mixture). Once this posterior distribution is known we can evaluate the probability of the score at the final training step $\hat{y}_\text{max}$ to be less than $y^*$ the best score observed so far (including all completed evaluations) which formally corresponds to $p := P(\hat{y}_\text{max} < y^* | C;\Theta)$. However, after noticing many instabilities in the optimization process of WPM, in RoBER we cut out the mixture distribution and consider only a single model. More precisely, we keep the \textit{mmf4} (Equation~\ref{eq:mmf4}) model proposed by~\cite{10.5555/645940.671380} and assessed in \cite{mohr2023lcdb} to be good for extrapolation.

\begin{equation}~\label{eq:mmf4}
   f_\text{mmf4}(z;\Theta) =  \frac{\Theta_0 \Theta_1 + \Theta_2 b(z)^{\Theta_3}}{\Theta_1 + b(z)^{\Theta_3}}
\end{equation}

Then, after observing at least four steps of the learning curve (number of parameters of the model), we estimate the model parameters $\Theta$ through non-linear least squares optimization (l. 17 in Algorithm~\ref{alg:rober-algorithm}). The continuous optimizer used is the Levenberg-Marquart algorithm~\cite{gavin2019levenberg} to which we can provide the exact Jacobian of the vector of residuals $\nabla_\Theta r(\textbf{z},\textbf{y};\Theta)$ through automatic differentiation with JAX~\cite{jax2018github} where $r$ is given by:

$$r(\textbf{z},\textbf{y};\Theta) = \textbf{y} - \mathcal{C}(\textbf{z};\Theta)$$

which combined with just-in-time compilation (JIT) allows for faster and robust estimation of the $\Theta$ parameters (previous works only used approximated gradient which makes it less stable). Now, that we have parameters correctly estimating the mean $y_i$ we are interested in incorporating uncertainty about the future and the noise between training steps. Therefore, we follow a Bayesian approach where we have $P(\Theta|C) \propto P(C|\Theta) \cdot P(\Theta)$. The prior on the parameters is given by $P(\Theta) := \mathcal{N}(\hat{\Theta}, 1)$. The likelihood is given by $P(C|\Theta) := \prod_{i=1}^n \mathcal{N}(y_i;\hat{y}_i, \sigma^2)$ where $\sigma^2 \sim \text{Exp}(1)$. Finally, the posterior distribution $P(\Theta|C)$ is not exactly computed but we can generate data (possible learning curves) from it using Monte-Carlo Markov Chain (MCMC) (line 18 in Algorithm~\ref{alg:rober-algorithm}) a common approach to generate data from an untractable densities. A candidate is then discarded only if $\tau \le p_\text{worse}$ with $\tau$ a threshold we are in practice setting quite high ($\tau=0.9$ in our experiments) to avoid discarding models with performance not certainly worse than the current best. In this sense, the method is quite conservative about model early rejecting because early steps will have high overall uncertainty.

Now that we explained our probabilistic model we give additional details on how we handle earliest steps where $z < |\Theta|$. Indeed, for these steps no least-square estimate is available has we have more degrees of freedom than the number of collected observations. We could remain in an over-parameterized regime and adopt a different optimization approach to alleviate this issue. Instead, we decide to place some safeguard to help us detect specific cases which are stagnation of the learning curve and unexpectedly bad candidates (commonly named "outliers" in data-science). The stagnation in learning curve is managed through a classic early stopping strategy (line 7 in Algorithm~\ref{alg:rober-algorithm}). Then, the outliers are detected through the simple inter-quantile range methodology, visually we look at the distribution of performances at a given step and consider them outlier when having a lower value than the lower whisker of a box-plot (lines 12-14 in Algorithm~\ref{alg:rober-algorithm}). It corresponds to discarding models with a score less than $Q_1 - 1.5 (Q_3 - Q_1)$ where $Q_1, Q_3$ are respectively the first and third quartiles of the distribution of observed scores at the given step

\begin{algorithm2e}
\small
\SetInd{0.25em}{0.5em}
\SetAlgoLined

\SetKwInOut{Input}{Inputs}
\SetKwInOut{Output}{Output}
\SetKwProg{Fn}{Function}{ is}{end}

\SetKwFunction{evaluate}{evaluate\_with\_RoBER}
\SetKwFunction{agent}{Agent}

\SetKwFor{For}{for}{do}{end}

\Fn{\evaluate{\texttt{config},$\tau,z_\text{max}$}}{
    $C \gets [~]$ \;
    \For{$z \in [|1,z_\text{max}|]$}{
        $\texttt{score} \gets$ Execute training step for \texttt{config} \;
        Append \texttt{score} to observations $C$ \;
        ~\\
        {\color{blue}\tcc{Decide if we continue the training}}
        \uIf{$z_\text{max} \leq z$ \textbf{or} not improved for $n_\text{patience}$ steps}{
            \text{Exit the loop} \;
        }
        $y^* \gets$ Return the maximum objective among all evaluated configurations\;
        $z_\text{test} \gets \text{Return step at which objective is checked}$ \;
        \uIf{$z_\text{test} = z$}{
            $Q_1 \gets \text{25th percentile of objectives at step } z$\;
            $Q_3 \gets \text{75th percentile of objectives at step } z$\;
            \uIf{$z \leq \min(z_\text{min},n_\text{parameters})$ \textbf{and} $y_z < Q_1 - 1.5 \cdot (Q_3-Q_1)$}{
                \text{Exit the loop} \;
            }
            \uElse{
            $\hat{\Theta} \gets$ Estimate model parameters with non-linear least-squares \;
            $p \gets \text{Estimate probability of being worse } P(\hat{y}_\text{max} < y^*|C;\hat{\Theta})$ \;
            \uIf{$\tau \leq p$}{
                \text{Exit the loop} \;
                }
            }
        }
    }
    \Return \text{last element in }$C$ \;
}
    \caption{Robust Bayesian Early Rejection (RoBER)}
    \label{alg:rober-algorithm}
\end{algorithm2e}

\section{Experiments with RoBER}~\label{sec:appendix-rober-robust}

\begin{figure}
    \centering
    \begin{subfigure}[b]{0.45\textwidth}
        \centering
        \includegraphics[width=\textwidth]{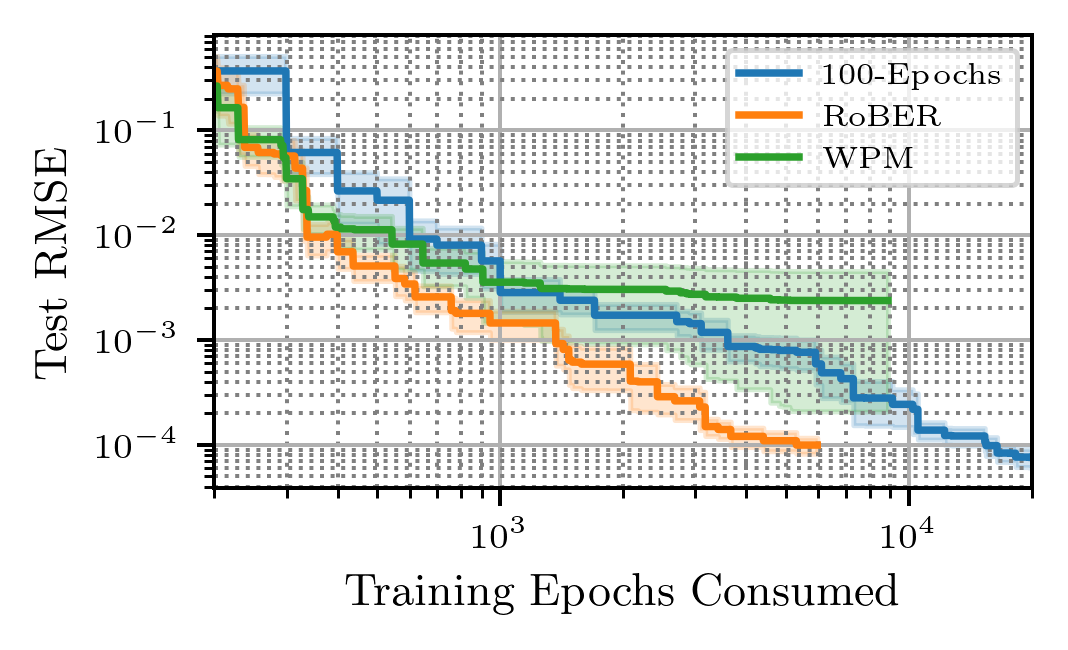}
        \caption{Naval Propulsion}
        \label{fig:rober-vs-wpm-navalpropulsion}
    \end{subfigure}
    \begin{subfigure}[b]{0.45\textwidth}
        \centering
        \includegraphics[width=\textwidth]{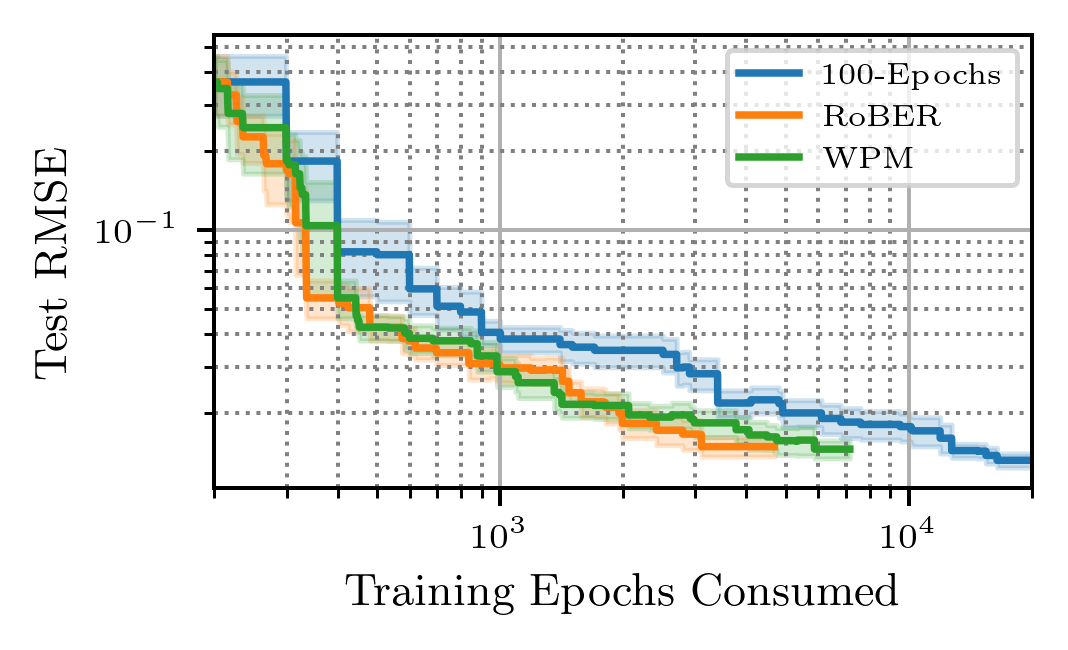}
        \caption{Parkinsons Telemonitoring}
        \label{fig:rober-vs-wpm-parkinsonstelemonitoring}
    \end{subfigure}
    \begin{subfigure}[b]{0.45\textwidth}
        \centering
        \includegraphics[width=\textwidth]{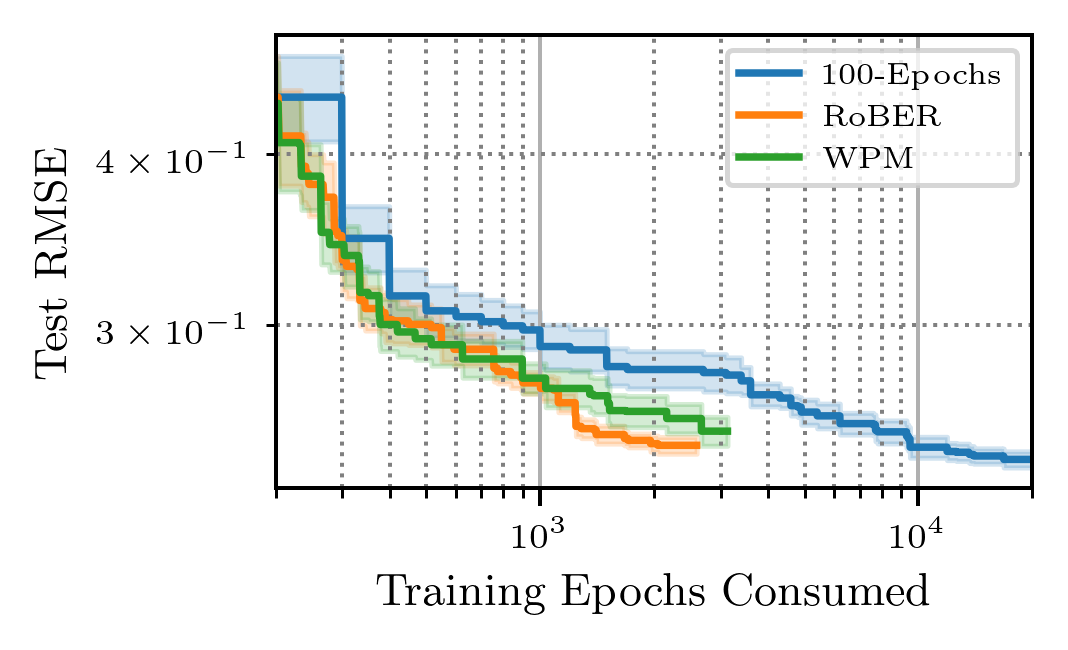}
        \caption{Protein Structure}
        \label{fig:rober-vs-wpm-proteinstructure}
    \end{subfigure}
    \begin{subfigure}[b]{0.45\textwidth}
        \centering
        \includegraphics[width=\textwidth]{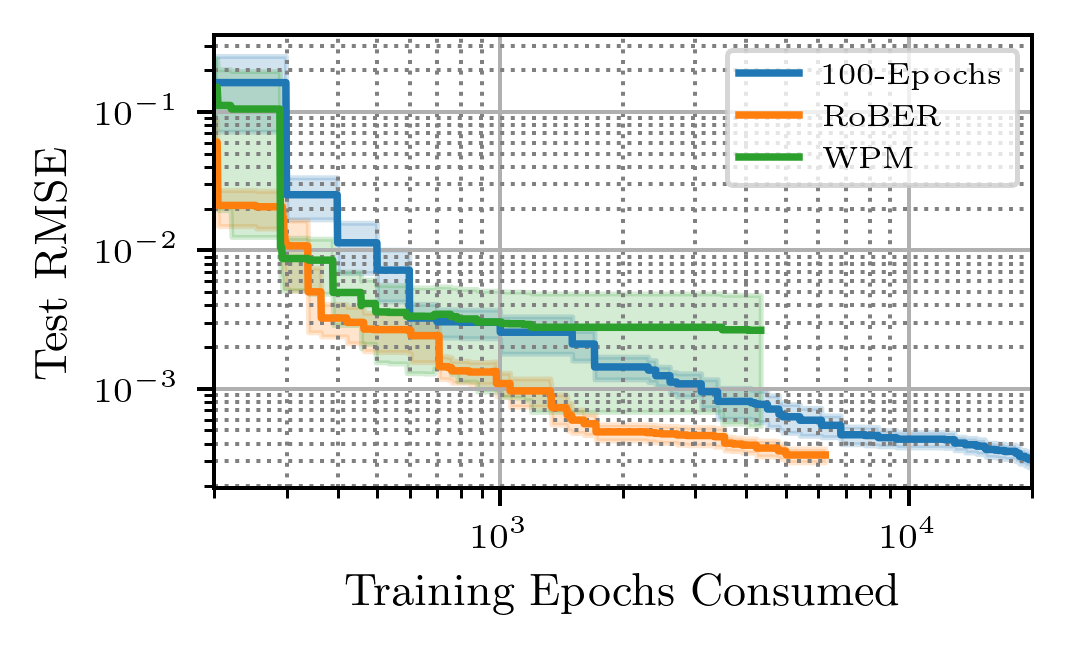}
        \caption{Slice Localization}
        \label{fig:rober-vs-wpm-slicelocalization}
    \end{subfigure}
    \caption{Comparing RoBER with Weighted Probabilistic Model (WPM) on HPOBench. RoBER is consistently better on all the problems and the WPM is something very unstable and worse than the basic 100-Epochs evaluation strategy.}
    \label{fig:all-rober-vs-wpm}
\end{figure}

In this section, we show that RoBER is more robust than previously proposed learning curve extrapolation methods. The results are presented in Figure~\ref{fig:all-rober-vs-wpm}. It can be seen that RoBER is consistent on the four problems while WPM~\cite{domhan_speeding_2015} under-perform on two of the problems \ref{fig:rober-vs-wpm-navalpropulsion} and \ref{fig:rober-vs-wpm-slicelocalization}. In addition, experimentally we RoBER is much faster to query compared to WPM which makes it more usable.

\section{More Details About our Bayesian Optimization Policy}
\label{sec:appendix-details-about-bo}

In our experiment, we also combined Bayesian optimization at the outer-optimization level with multi-fidelity agents at the inner-optimizer level. For this we use a Random-Forest surrogate model unlike very frequent Gaussian-Process based approaches or Tree-Parzen Estimation (TPE)~\cite{bergstra_algorithms_2011,bergstra_making_2013}. Our Random-Forest is slightly modified as suggested in~\cite{hutter2014algorithm} to use the "best random" split instead of the "best" split which improves significantly the estimation of epistemic uncertainty of the model (based on the Scikit-Learn~\cite{scikit-learn} implementation). Also, the objective is mapped to $[0,1]$ with a min-max standardization before applying a $\log(y_\text{min-max} + \epsilon)$ transformation on it. This equalize the errors on large and small objective values as the average error is being minimized by the learning algorithm. Then, we use the upper confidence-bound (UCB) acquisition function $\alpha(x) = \mu(x) + \kappa \cdot \sigma(x)$ with a cyclic exponential decay to manage the evolution of the exploration-exploitation parameter $\kappa$. This scheduler is able to repetitively perform exploration while having moments of strong exploitation (with a $kappa$ value cycling to converge to a value close to 0). We adopt this scheme after noticing the empirical efficiency of the TPE algorithm on HPO which has a strong exploitation scheme embedded in it.

\section{The Problem of Generalization in HPO}~\label{sec:appendix-generalization}

\begin{figure}
    \centering
    \begin{subfigure}[b]{0.45\textwidth}
        \centering
        \includegraphics[width=\textwidth]{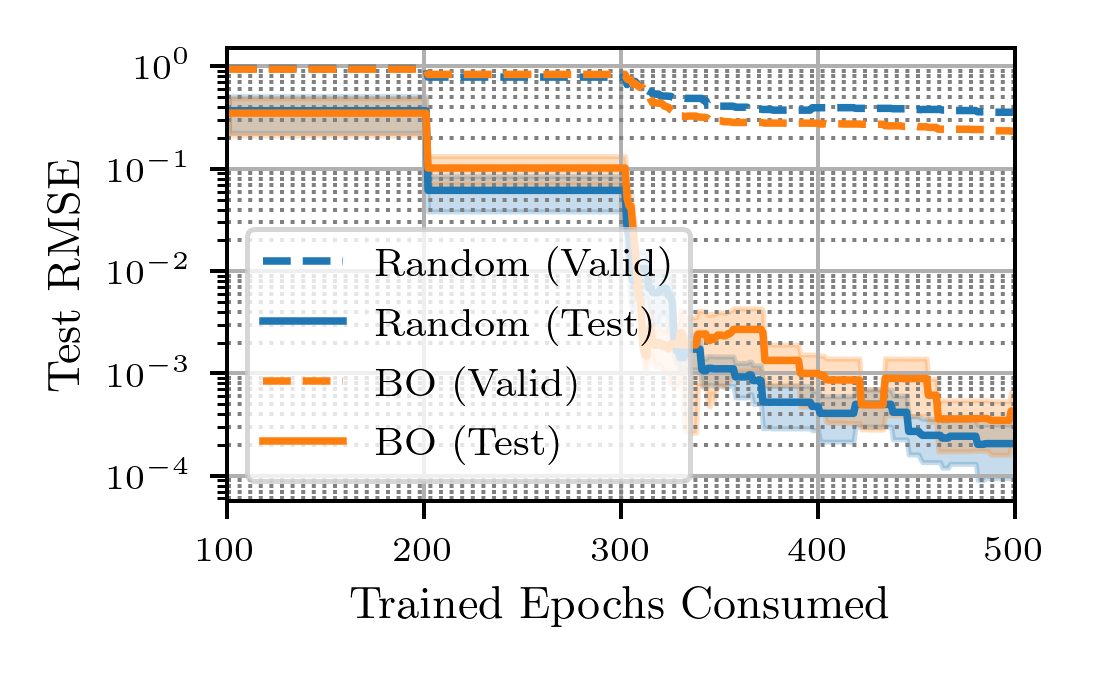}
        \caption{Naval Propulsion}
        \label{fig:all-generalization-1epoch-navalpropulsion}
    \end{subfigure}
    \begin{subfigure}[b]{0.45\textwidth}
        \centering
        \includegraphics[width=\textwidth]{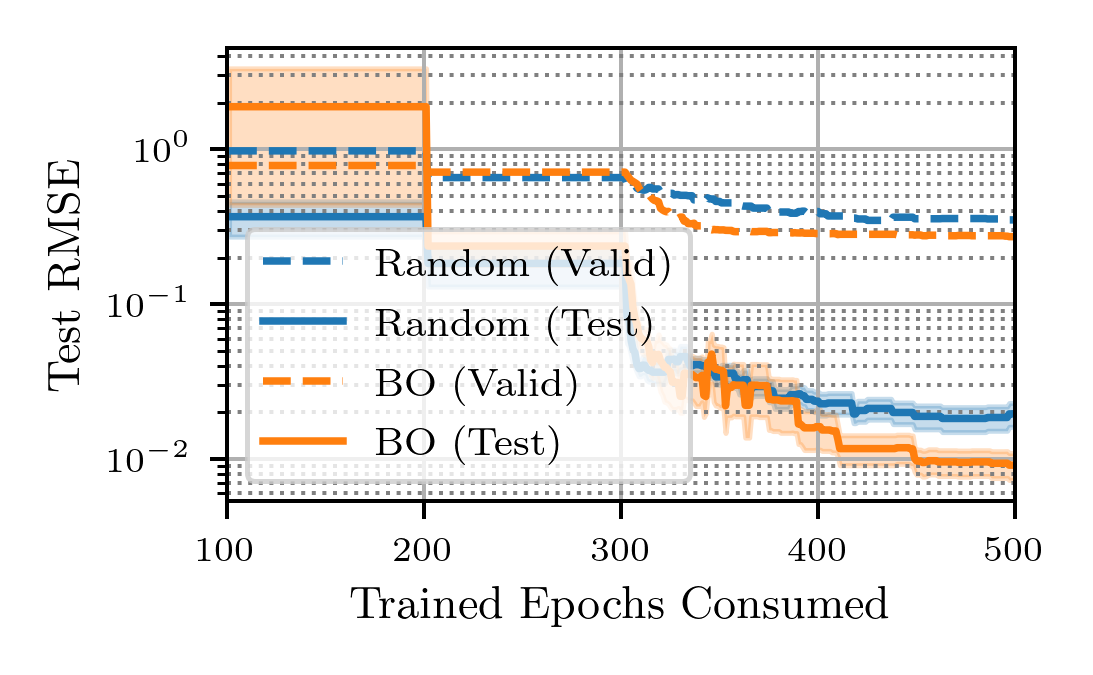}
        \caption{Parkinsons Telemonitoring}
        \label{fig:all-generalization-1epoch-parkinsonstelemonitoring}
    \end{subfigure}
    \begin{subfigure}[b]{0.45\textwidth}
        \centering
        \includegraphics[width=\textwidth]{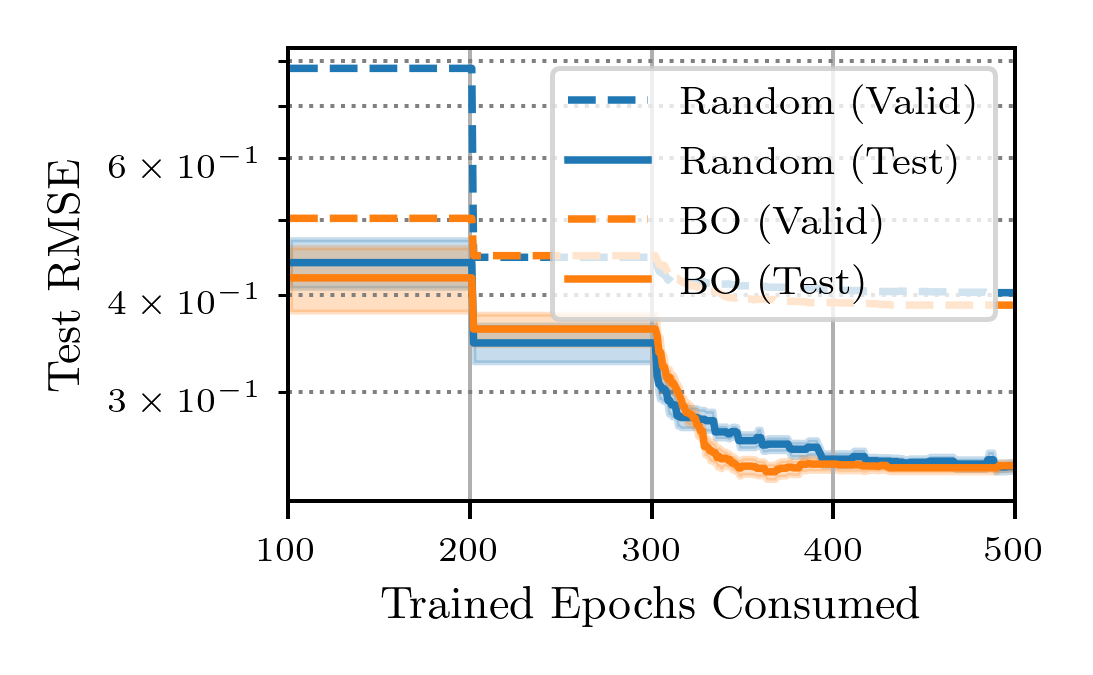}
        \caption{Protein Structure}
        \label{fig:all-generalization-1epoch-proteinstructure}
    \end{subfigure}
    \begin{subfigure}[b]{0.45\textwidth}
        \centering
        \includegraphics[width=\textwidth]{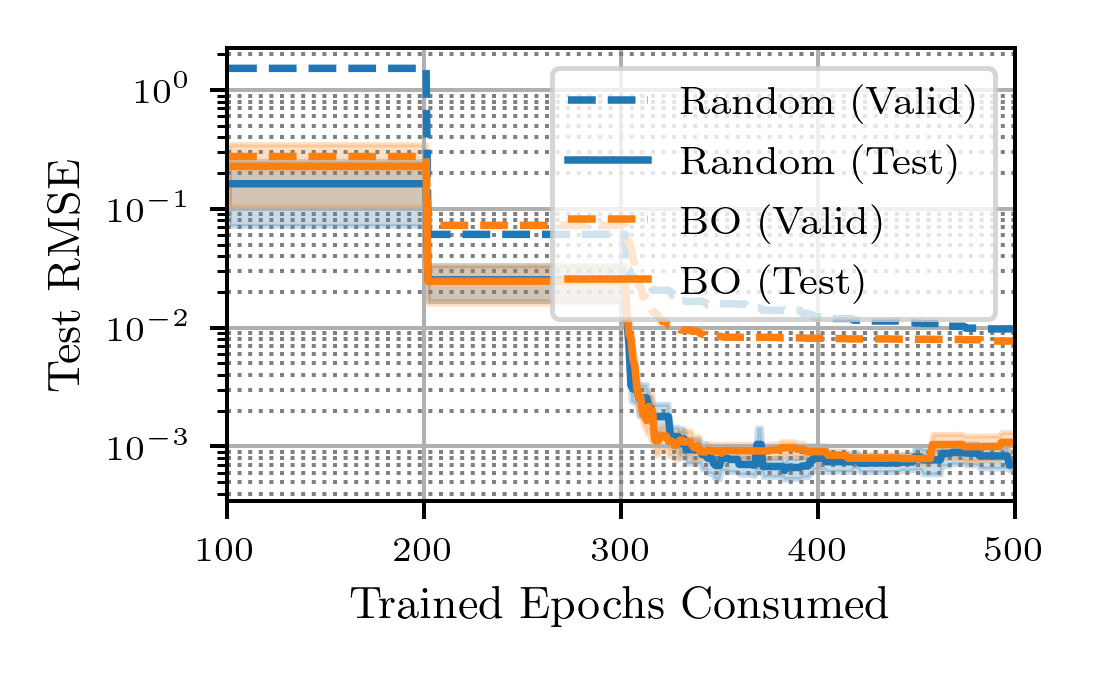}
        \caption{Slice Localization}
        \label{fig:all-generalization-1epoch-slicelocalization}
    \end{subfigure}
    \caption{Looking at the Generalization Error in HPO for the 1-Epoch multi-fidelity agent.}
    \label{fig:all-generalization-1epoch}
\end{figure}

In this section, we present the difference between the evolution of the validation error used by the optimizer during hyperparameter optimization and the generalization error (real target). As it can be observed in Figure~\ref{fig:all-generalization-1epoch} looking at the evolution of the validation error can be miss-leading of the real generalization error.

\section{Code}

The code is made publicly available with related documentation as part of an existing Hyperameter optimization Python package. Our experiments are organized as followed:
\begin{itemize}
    \item Benchmark definitions: \href{https://github.com/deephyper/benchmark}{github.com/deephyper/benchmark}
    \item Library containing the core software and algorithms: \href{https://github.com/deephyper/deephyper}{github.com/deephyper/deephyper}
    \item Organization, execution and analysis of the experiments of the paper: \href{https://github.com/deephyper/scalable-bo}{github.com/deephyper/scalable-bo}
\end{itemize}

The experiments for this paper are located in the \texttt{experiments/local/dhb/} folder of the \texttt{deephyper/scalable-bo} repository. The main entry point for the execution is \texttt{src/scalbo/scalbo/search/dbo.py}. Plots and analysis are all located in the \texttt{notebooks/learning-curves/} folder.

\end{document}